\documentclass[oneside,british,american,english,australian,12pt,a4]{book}
\usepackage{helvet}

\usepackage[T1]{fontenc}
\usepackage{xcolor}
\usepackage{array}
\usepackage{verbatim}
\usepackage{longtable}
\usepackage{calc}
\usepackage{mathrsfs}
\usepackage{mathtools}
\usepackage{bm}
\usepackage{multirow}
\usepackage{amsmath}
\usepackage{amssymb}
\usepackage{stmaryrd}
\usepackage{graphicx}
\usepackage[authoryear]{natbib}

\makeatletter

\providecommand{\tabularnewline}{\\}

\usepackage{textcomp}
\usepackage{setspace}
\usepackage{graphicx}
\usepackage{amsmath}
\usepackage{latexsym}
\usepackage{colortbl}
\usepackage{cancel}

\usepackage{helvet}
\usepackage{flushend}

\usepackage{multirow}

\usepackage{fancyhdr}
\usepackage{colortbl}
\definecolor{cobalt}{rgb}{0.0, 0.28, 0.67}
\definecolor{carmine}{rgb}{0.59, 0.0, 0.09}
\usepackage[colorlinks=true, urlcolor=carmine, citecolor=cobalt, linkcolor=carmine, pdfborder={0 0 0}]{hyperref}

\usepackage{lmodern}
\setcitestyle{round}

\usepackage{tikz}

\usepackage{epigraph,varwidth}
\usepackage{type1cm}
\usepackage{lettrine}
\definecolor{mygray}{gray}{0.6}

\usepackage{epsfig}
\usepackage{soul}
\usepackage[utf8]{inputenc}
\usepackage{microtype}
\usepackage[linesnumbered,algoruled,boxed,lined]{algorithm2e}
\usepackage{siunitx}

\usepackage{balance}

\makeatother

\usepackage{babel}
\begin{document}
\selectlanguage{english}%

\global\long\def\tr{\mathrm{tr}}%

\global\long\def\realset{\mathbb{R}}%

\global\long\def\E{\mathbb{E}}%

\global\long\def\H{\mathcal{H}}%

\global\long\def\realn{\real^{n}}%

\global\long\def\natset{\integerset}%

\global\long\def\interger{\integerset}%

\global\long\def\integerset{\mathbb{Z}}%

\global\long\def\natn{\natset^{n}}%

\global\long\def\partf{\mathcal{Z}}%

\global\long\def\rational{\mathbb{Q}}%

\global\long\def\realPlusn{\mathbb{R_{+}^{n}}}%

\global\long\def\comp{\complexset}%
 
\global\long\def\complexset{\mathbb{C}}%

\global\long\def\dataset{\mathcal{D}}%

\global\long\def\class{\mathcal{C}}%

\global\long\def\likelihood{\mathcal{L}}%

\global\long\def\normal{\mathcal{N}}%

\global\long\def\argmax#1{\underset{#1}{\text{argmax}}}%

\global\long\def\bphi{\boldsymbol{\phi}}%

\global\long\def\btheta{\boldsymbol{\theta}}%

\global\long\def\bx{\boldsymbol{x}}%

\global\long\def\by{\boldsymbol{y}}%

\global\long\def\bt{\boldsymbol{t}}%

\global\long\def\bf{\boldsymbol{f}}%

\global\long\def\bX{\boldsymbol{X}}%

\global\long\def\bY{\boldsymbol{Y}}%

\global\long\def\bp{\boldsymbol{p}}%

\global\long\def\bP{\boldsymbol{P}}%

\global\long\def\bh{\boldsymbol{h}}%

\global\long\def\bH{\boldsymbol{H}}%

\global\long\def\bv{\boldsymbol{v}}%

\global\long\def\bV{\boldsymbol{V}}%

\global\long\def\btv{\tilde{\boldsymbol{v}}}%

\global\long\def\tv{\tilde{v}}%

\global\long\def\tq{\tilde{q}}%

\global\long\def\baq{\bar{q}}%

\global\long\def\ba{\boldsymbol{a}}%

\global\long\def\bta{\tilde{\boldsymbol{a}}}%

\global\long\def\ta{\tilde{a}}%

\global\long\def\bA{\boldsymbol{A}}%

\global\long\def\bb{\boldsymbol{b}}%

\global\long\def\bB{\boldsymbol{B}}%

\global\long\def\bD{\boldsymbol{D}}%

\global\long\def\ble{\boldsymbol{e}}%

\global\long\def\bE{\boldsymbol{E}}%

\global\long\def\bI{\boldsymbol{I}}%

\global\long\def\bS{\boldsymbol{S}}%

\global\long\def\bw{\boldsymbol{w}}%

\global\long\def\bW{\boldsymbol{W}}%

\global\long\def\bL{\boldsymbol{L}}%

\global\long\def\btW{\tilde{\boldsymbol{W}}}%

\global\long\def\tW{\tilde{W}}%

\global\long\def\bz{\boldsymbol{z}}%

\global\long\def\bu{\boldsymbol{u}}%

\global\long\def\bI{\boldsymbol{I}}%

\global\long\def\bmu{\boldsymbol{\mu}}%

\global\long\def\tmu{\tilde{\mu}}%

\global\long\def\btmu{\tilde{\bmu}}%

\global\long\def\bamu{\bar{\mu}}%

\global\long\def\bbmu{\bar{\bmu}}%

\global\long\def\tlamda{\tilde{\lambda}}%

\global\long\def\balamda{\bar{\lambda}}%

\global\long\def\balpha{\boldsymbol{\alpha}}%

\global\long\def\bbeta{\boldsymbol{\beta}}%

\global\long\def\bSigma{\boldsymbol{\Sigma}}%

\global\long\def\grad{\bigtriangleup}%

\global\long\def\cov{\textnormal{cov}}%

\selectlanguage{british}%
\global\long\def\xb{\boldsymbol{x}}%
\global\long\def\wb{\boldsymbol{w}}%
\global\long\def\ub{\boldsymbol{u}}%
\global\long\def\hb{\boldsymbol{h}}%
\global\long\def\bb{\boldsymbol{b}}%
\global\long\def\zb{\boldsymbol{z}}%
\selectlanguage{australian}%

\selectlanguage{english}%
\pagenumbering{roman} 
\setcounter{page}{1}
\pagestyle{empty}

\begin{center}
\textbf{\Large{}Deep Neural Networks for Visual Reasoning}{\Large\par}
\par\end{center}

\vspace{1cm}

\vspace{1cm}

\begin{center}
by
\par\end{center}

\vspace{-3cc}

\begin{center}
\textbf{\large{}Thao Minh Le}{\large\par}
\par\end{center}

\vspace{-3cc}

\begin{center}
M.Sc.
\par\end{center}

\vfill{}

\begin{center}
\textbf{Submitted in fulfilment of the requirements for the degree
of}\\
 \textbf{Doctor of Philosophy}
\par\end{center}

\vspace{2cm}

\begin{center}
\textbf{\large{}Deakin University}{\large\par}
\par\end{center}

\vspace{-3cc}

\begin{center}
\textbf{\large{}May 2021}{\large\par}
\par\end{center}\selectlanguage{australian}%

\newpage{}

\tableofcontents{}\listoffigures
\listoftables

\newpage{}

\pagestyle{empty}

\chapter*{Abstract}

\addcontentsline{toc}{chapter}{\numberline{}{Abstract}}

\noindent

Visual perception and language understanding are - fundamental components
of human intelligence, enabling them to understand and reason about
objects and their interactions. It is crucial for machines to have
this capacity to reason using these two modalities to invent new robot-human
collaborative systems.

Recent advances in deep learning have built separate sophisticated
representations of both visual scenes and languages. However, understanding
the associations between the two modalities in a shared context for
multimodal reasoning remains a challenge. Focusing on language and
vision modalities, this thesis advances understanding of how to exploit
and use pivotal aspects of vision-and-language tasks with neural networks
to support reasoning. The contributions include: (i) effective mechanisms
for content selection and construction of temporal relations from
dynamic visual scenes in response to a linguistic query and preparing
adequate knowledge for the reasoning process (ii) new frameworks to
perform reasoning with neural networks by exploiting visual-linguistic
associations, deduced either directly from data or guided by external
priors.

In the \emph{first} work, we present a novel dual process neural architecture
resembling the dual process in the human reasoning systems for Video
Question Answering (Video QA). It is composed of a question-guided
video process module that is fast and reactive (System 1) followed
by a generic reasoning module that is slow and deliberative (System
2). The fast system is a hierarchical model that encodes visual patterns
about objects, actions and relations in space-time given the textual
cues from the question. The encoded representation is a set of high-level
visual features, which are then passed to the slow, deliberative system.
Multi-step inference is used to iteratively chain visual elements
as required by the textual elements. The system is evaluated on the
major large-scale Video QA benchmarks, demonstrating competitive results,
with large margins in the case of multi-step reasoning.

In our \emph{second} work, we delve deeper into fast, reactive system
adding introduced in the dual process neural architecture by proposing
a novel conditional computational unit to jointly encapsulate dynamic
content selection and relation construction. The unit, Conditional
Relation Network (CRN), is designed as a general-purpose reusable
neural unit taking as input a set of tensorial objects and translating
into a new set of objects that encode relations between the inputs.
Model building then becomes a simple exercise of replication, rearrangement
and stacking of these reusable units along the hierarchical structure
of video. This design thus supports high-order relational and multi-step
reasoning. We present two different resulting architectures of CRN
hierarchies for two different forms of Video QA, namely short-form
Video QA and long-form Video QA. Short-form Video QA is the most common
setting of Video QA whose video content is usually short and answers
are derived solely from the video content. Long-form Video QA (also
known as Movie QA), on the other hand, deals with longer and more
complex videos and additionally considers information from associated
channels, such as movie subtitles. Our rigorous evaluations on well-known
datasets achieved consistent improvement over state-of-the-art results,
demonstrating the effectiveness of building a general-purpose reasoning
unit on complex domains such as Video QA.

The \emph{third} work focuses on designing a neural network with strong
relational and iterative reasoning capabilities, ultimately advancing
the reasoning module (System 2) in the dual process neural architecture.
In particular, we present Language-binding Object Graph Network (LOGNet),
the first neural reasoning method with dynamic relational structures
across both visual and textual domains with applications in Image
QA. Relaxing the common assumption made by current models that the
object predicates pre-exist and stay static, passive to the reasoning
process, we propose that these dynamic predicates expand across the
modalities to include pair-wise visual-linguistic object binding.
LOGNet finds these contextualised object links, within each recurrent
reasoning step, without relying on external predicative priors. These
dynamic structures reflect the conditional dual-domain object dependency
given the evolving context of the reasoning through co-attention.
Such discovered dynamic graphs facilitate multi-step knowledge combinations
and refinements that iteratively deduce the compact representation
of the final answer. We evaluated the effectiveness of LOGNet on major
Image QA datasets, demonstrating clear advantages against other reasoning
models in performance and learning capacity when given limited amounts
of training data.

Finally, in the \emph{last} work, we address the current limitations
of most modern attention-based visual reasoning systems regarding
their unintuitive attentions. This issue comes from the lack of explicit
feedback on the binding between visual entities and their equivalent
linguistic expressions. We leverage external visual-linguistic grounded
data to improve the cross-modal binding by extracting visual-linguistic
association priors and applying this common knowledge to regulate
the attention mechanisms in Image QA models. This methodology is implemented
using the grammatical structure of the query leading to an effective
distillation learning framework. The proposed algorithm is capable
of probing attention-based reasoning models, injecting relevant associative
knowledge, and regulating the core reasoning process. This scalable
enhancement improves Image QA accuracy, fortifies robustness to linguistic
variations, and increases interpretability.

\newpage{}

\pagestyle{empty}

\chapter*{Acknowledgements}

\addcontentsline{toc}{chapter}{\numberline{}{Acknowledgements}}

First and foremost, I would like to thank my principal supervisor
A/Prof. Truyen Tran for giving me the opportunity to pursue a PhD
under his supervision. I am immensely grateful for his dedicated and
continuous support in both my academic research and personal life
in the past three years. Since the beginning of my PhD journey, A/Prof.
Truyen Tran has always helped me to develop my skills and confidence
to become an independent researcher and thinker. He has also encouraged
me to think outside of the box when it comes to my career development.
My sincere thanks also go to my co-supervisors, Alfred Deakin Prof.
Svetha Venkatesh and Dr. Vuong Le, for countless inspiring discussions,
from developing ideas for my research projects to writing rigorous
research papers. Thank you Prof. Svetha Venkatesh for organising writing
workshops to help us improve our writing skills. The success of my
research projects is largely attributed to the guidance and support
of all the advisory team members.

I was fortunate enough to be part of A2I2, the Applied Artificial
Intelligence Institute, where I had the opportunity to know and work
with fantastic people. I could not have asked for a more inspiring
research environment. Thank you to all my colleagues and friends at
A2I2 for making my PhD an unforgettable experience: Huong Ha, Dung
Nguyen, Kien Do, Tung Hoang, Hung Le, Arun Kumar, Deepthi Kuttichira,
Romero Monrais, Julian Berk, Haripriya Harikumar, Duc Nguyen, Hung
Tran, Long Dang and all others. I truly enjoyed all the group discussions,
lunchtimes and game nights that we shared. Many thanks also to Shweta
Gupta, Kim Pham and Trang Tran for their administrative support during
my studies. 

I would like to offer my special thanks to my friends, Helen Bell
and Veronnica Nagathota, for your friendship and unending support
throughout my studies in Australia. You both are significant influences
on my personal and career growth. To all my Daly crew friends, thank
you for making my life in Australia more like home and getting us
through a pandemic together: Mai Nguyen, Dominique Assaf, Kai Nguyen,
Syntyche Esenowo, Viet Nguyen, Myah Bruynen, Jono Nguyen, Mustafa
Kemal, Sam Abad.

Most of all, I would like to thank my parents, my wife Mai Nguyen
and all my family members for their unconditional love and support
which helped me overcome challenges in my life to finish this thesis.

\newpage{}

\pagestyle{empty}

\chapter*{Relevant Publications}

\addcontentsline{toc}{chapter}{\numberline{}{Relevant Publications}}

Part of this thesis has been published or documented elsewhere. The
details of these publications are as follows:

Chapter \ref{chap:DualProcess}:
\begin{itemize}
\item \textbf{Le, T. M.}, Le, V., Venkatesh, S., \& Tran, T. (2020, July).
Neural Reasoning, Fast and Slow, for Video Question Answering. In
\emph{2020 International Joint Conference on Neural Networks (IJCNN)}
(pp. 1-8), doi: \sloppy10.1109/IJCNN48605.2020.9207580. IEEE.
\end{itemize}
Chapter \ref{chap:MultimodalReasoning}:
\begin{itemize}
\item \textbf{Le, T. M.}, Le, V., Venkatesh, S., \& Tran, T. (2020). Hierarchical
Conditional Relation Networks for Video Question Answering. In \emph{Proceedings
of the IEEE/CVF Conference on Computer Vision and Pattern Recognition
(CVPR)} (pp. 9972-9981).
\item \textbf{Le, T. M.}, Le, V., Venkatesh, S., \& Tran, T. (2020). Hierarchical
Conditional Relation Networks for Multimodal Video Question Answering.
In \emph{International Journal of Computer Vision (2021), }doi: https://doi.org/10.1007/s11263-021-01514-3.
\end{itemize}
Chapter \ref{chap:RelationalVisualReasoning}:
\begin{itemize}
\item \textbf{Le, T. M.}, Le, V., Venkatesh, S., \& Tran, T. (2020). Dynamic
Language Binding in Relational Visual Reasoning. In \emph{Proceedings
of the Twenty-Ninth International Joint Conference on Artificial Intelligence
(IJCAI)} (pp. 818-824).
\end{itemize}
Chapter \ref{chap:RobustnessVR}:
\begin{itemize}
\item \textbf{Le, T. M.}, Le, V., Venkatesh, S., \& Tran, T. (2021). Linguistic
Expression Grounding as Attention Priors in Visual Question Answering.
\emph{Under submission.}
\end{itemize}
Although not the main contributor, the following collaborative studies
are done during my candidature and closely related to different parts
discussed in the thesis:
\begin{itemize}
\item Dang, L. H., \textbf{Le, T. M.}, Le, V., Tran, T. (2021). Hierarchical
Object-oriented Spatio-Temporal Reasoning for Video Question Answering.
In \emph{Proceedings of the Thirtieth International Joint Conference
on Artificial Intelligence (IJCAI)} (pp. 636-642).
\item Dang, L. H., \textbf{Le, T. M.}, Le, V., Tran, T. (2021). Object-Centric
Representation Learning for Video Question Answering. \emph{Accepted
for publication in the 2021 International Joint Conference on Neural
Networks (IJCNN'21). }arXiv preprint \emph{arXiv:2104.05166}.
\item Dang, L. H., \textbf{Le, T. M.}, Le, V., Tran, T. (2020). Object-Centric
Relational Reasoning for Video Question Answering. In \emph{2020 ECCV
2nd Workshop On Video Turing Test: Toward Human-Level Video Story
Understanding}.
\item Nguyen, T. M., Nguyen, T., \textbf{Le, T. M.}, \& Tran, T. (2020).
GEFA: Early Fusion Approach in Drug-Target Affinity Prediction. In
\emph{IEEE/ACM Transactions on Computational Biology and Bioinformatics},
vol. , no. 01, pp. 1-1, 5555. doi: 10.1109/TCBB.2021.3094217. 
\item Nguyen, T. M., Nguyen, T., \textbf{Le, T. M.}, \& Tran, T. (2020).
GEFA: Early Fusion Approach in Drug-Target Affinity Prediction. In
\emph{NeurIPS 2020 Workshop on Machine Learning for Structural Biology
(MLSB).}
\end{itemize}

\newpage{}

\pagestyle{empty}

\selectlanguage{british}%

\section*{Abbreviations}

\addcontentsline{toc}{chapter}{\numberline{}{Abbreviations}}\medskip{}

\begin{longtable}[c]{>{\raggedright}p{0.3\textwidth}>{\raggedright}p{0.65\textwidth}}
\hline 
Abbreviation & Description\tabularnewline
\hline 
\noalign{\vskip\doublerulesep}
BERT & Bidirectional Encoder Representation from Transformers\tabularnewline
\noalign{\vskip\doublerulesep}
\rowcolor[gray]{0.95}biLSTM & Bidirectional Long Short-Term Memory\tabularnewline
\noalign{\vskip\doublerulesep}
CBOW & Continuous Bag-of-Word\tabularnewline
\noalign{\vskip\doublerulesep}
\rowcolor[gray]{0.95}ClipRN & Clip-based Relational Network\tabularnewline
\noalign{\vskip\doublerulesep}
CNN & Convolution Neural Network\tabularnewline
\noalign{\vskip\doublerulesep}
\rowcolor[gray]{0.95}CRN & Conditional Relation Network\tabularnewline
\noalign{\vskip\doublerulesep}
CS & Consensus Score\tabularnewline
\noalign{\vskip\doublerulesep}
\rowcolor[gray]{0.95}GAP & Grounding-based Attention Prior\tabularnewline
\noalign{\vskip\doublerulesep}
GCN & Graph Convolution Network\tabularnewline
\noalign{\vskip\doublerulesep}
\rowcolor[gray]{0.95}GRU & Gated Recurrent Units\tabularnewline
\noalign{\vskip\doublerulesep}
HCRN & Hierarchical Conditional Relation Networks\tabularnewline
\noalign{\vskip\doublerulesep}
\rowcolor[gray]{0.95}Image QA & Image Question Answering\tabularnewline
\noalign{\vskip\doublerulesep}
LOG & Language-binding Object Graph\tabularnewline
\noalign{\vskip\doublerulesep}
\rowcolor[gray]{0.95}LOGNet & Language-binding Object Graph Networks\tabularnewline
\noalign{\vskip\doublerulesep}
LSTM & Long Short-Term Memory\tabularnewline
\noalign{\vskip\doublerulesep}
\rowcolor[gray]{0.95}MAC & Memory-Attention-Composition\tabularnewline
\noalign{\vskip\doublerulesep}
MACNet & Memory-Attention-Composition Network\tabularnewline
\noalign{\vskip\doublerulesep}
\rowcolor[gray]{0.95}MLP & Multi-layer Perceptron\tabularnewline
\noalign{\vskip\doublerulesep}
MPNN & Message Passing Neural Network\tabularnewline
\noalign{\vskip\doublerulesep}
\rowcolor[gray]{0.95}MRC & Machine Reading Comprehension\tabularnewline
\noalign{\vskip\doublerulesep}
MSE & Mean Squared Error\tabularnewline
\noalign{\vskip\doublerulesep}
\rowcolor[gray]{0.95}NMN & Neural Module Network\tabularnewline
\noalign{\vskip\doublerulesep}
NLP & Natural Language Processing\tabularnewline
\noalign{\vskip\doublerulesep}
\rowcolor[gray]{0.95}NP & Noun-phrases\tabularnewline
\noalign{\vskip\doublerulesep}
QA & Question Answering\tabularnewline
\noalign{\vskip\doublerulesep}
\rowcolor[gray]{0.95}RE & Linguistic Referring Expression\tabularnewline
\noalign{\vskip\doublerulesep}
RNN & Recurrent Neural Network\tabularnewline
\noalign{\vskip\doublerulesep}
\rowcolor[gray]{0.95}SOTA & State-of-the-art\tabularnewline
\noalign{\vskip\doublerulesep}
Text QA & Textual Question Answering\tabularnewline
\noalign{\vskip\doublerulesep}
\rowcolor[gray]{0.95}TRN & Temporal Relation Network\tabularnewline
\noalign{\vskip\doublerulesep}
UpDn & Bottom-Up Top-Down Attention\tabularnewline
\noalign{\vskip\doublerulesep}
\rowcolor[gray]{0.95}Video QA & Video Question Answering\tabularnewline
\noalign{\vskip\doublerulesep}
VP & Verb-phrases\tabularnewline
\noalign{\vskip\doublerulesep}
\rowcolor[gray]{0.95}VQA & Visual Question Answering\tabularnewline
\hline 
\noalign{\vskip\doublerulesep}
\end{longtable}

\section*{Notation}

We summarise here the notation that we use throughout this thesis.

\subsection*{Scalars, Vectors and Matrices}

Scalar values are represented by lowercase letters such as $a$. Vectors,
on the other hand, are represented by lowercase bold letters, such
as $\mathbf{a}$ and $\mathbf{w}$, or by bold Greek letters, such
as $\bm{\alpha}$ and $\bm{\beta}$. Unless otherwise stated, vectors
are represented as column matrices. 

Matrices are represented by uppercase bold letters such as $\mathbf{W}$.
An entry in a matrix is specified by its row index and column index.
For example, $w_{ij}$ refers to the entry in the $i$-th row and
$j$-th column. In the case where one letter is used multiple times
to refer to different matrices, we usually differentiate them by a
superscript that is represented by a lowercase letter. 

In regards to operations with vectors, we use $\mathbf{c}=[\mathbf{a};\mathbf{b}]$
to represent the vector concatenation operation between vector ${\bf a}\in\mathbb{R}^{d_{1}}$
and vector $\mathbf{b}\in\mathbb{R}^{d_{2}}$, where $d_{1}$ and
$d_{2}$ are lengths of vector $\mathbf{a}$ and $\mathbf{b}$, respectively.
The resultant vector $\mathbf{c}$ is a $(d_{1}+d_{2})$ dimensional
vector.

\subsection*{Sets and Sequences}

Sets or sequences/arrays are represented by uppercase bold letters
such as $\mathbf{V}$ and $\mathbf{L}$, similar to how we represent
matrices. They usually are a composition of vectors represented by
corresponding lowercase letters with indices. Occasionally, we also
use a calligraphic font to represent sets and sequences, such as $\mathcal{\boldsymbol{V}}$
and $\boldsymbol{\mathcal{E}}$. Lengths and sizes of these sets and
sequences are, on the other hand, represented by uppercase letters
such as $N$ and $T$. An item in a sequence or in a set is indexed
by a subscript represented by a lowercase letter such as $i,j,k$.
For example, a set $\mathbf{V}$ written as $\mathbf{V}=\left\{ \mathbf{v}_{i}\mid\mathbf{v}_{i}\in\mathbb{R}^{d}\right\} _{i=1}^{N}$
represents a composition of $N$ vectors $\mathbf{v}_{i}$ in $\mathbb{R}^{d}$,
for $i=1,..,N$.

\subsection*{Variables and Parameters}

Independent variables such as observed data are represented by lowercase
bold letters, such as $\mathbf{x}$, and dependent variables such
as target data are represented by lowercase letters, for example,
$y$. Learnable network parameters are usually represented by bold
Greek letters such as $\bm{\theta}$ and $\bm{\phi}$.

\subsection*{Functions and Probability Distributions}

Function are represented by either lowercase letters, as in $f(x)$,
or capitalised letters, as in $F(x)$ or $\mathcal{F}(x)$, or Greek
letters, for example, $\sigma(x)$. They are always followed by round
brackets. Composition of functions $f$ and $g$ are specified by
$f\circ g$. 

Regarding probability notation, we use $P\left(a\mid b\right)$ to
refer to the probability of event $a$ given event $b$.\selectlanguage{australian}%

\newpage\pagenumbering{arabic} 
\setcounter{page}{1}
\pagestyle{fancy}

\chapter{Introduction\label{chap:Introduction}}

\section{Motivations}

Visual perception is one of our most crucial senses, allowing us to
perceive and interact with the environment. Language is how we communicate
with each other. Humans are unique in their capability to interpret
and reason about the world using both visual and language modalities,
processing visual scenes and language seamlessly.

Having a human-level understanding of the world is the ultimate goal
of Artificial Intelligence (AI). In visual perception, we expect an
intelligent machine to be able to understand the information captured
in a static image or in dynamic scenes in a video. Powered by deep
neural networks and large-scale datasets, computer vision has made
significant progress on a wide variety of recognition tasks in recent
years. Many of them, such as object classification, object detection
and tracking, and human activity recognition, are widely integrated
into real-world applications with reliable performance. However, as
we move towards tightly coupled human-machine cooperative systems,
there is a quest for machine intelligence to be able to reason over
both vision and human language and communicate naturally with humans.
This capacity directly enables many applications, such as systems
aiding visually-impaired users, in understanding their surroundings
\citep{bigham2010vizwiz,ilievski2017generative} or visual information
search from surveillance feeds.

Based on the newly emerging generation of AI techniques based on deep
learning, a number of tasks have been proposed to assess machines
understanding of visual content either by asking them to respond to
a human query or by asking them to describe visual scenes using natural
language. We refer to such tasks as visual reasoning systems in this
thesis. Typical tasks in visual reasoning are Image/Video Captioning
\citep{chen2015microsoft}, Visual Question Answering (VQA) \citep{antol2015vqa},
Visual Dialogue \citep{das2017visual} and Visual Referring Expression
\citep{mao2016generation,hu2016natural}. Compared to traditional
computer vision tasks, such visual reasoning tasks require not only
an adequate foundation on visual perception but also demand the engagement
of more generalised problem-solving for the decision-making process
\citep{zhang2019raven}. Although there are many things in common
between the aforementioned tasks in terms of concepts and techniques,
VQA, in particular, has attracted enormous research interest because
we as humans understand the world using a process of question answering.
In addition, the arbitrariness of linguistic expressions allows us
to easily incorporate multi-disciplinary knowledge, expanding the
range of reasoning capabilities of machine intelligence. Further,
evaluating VQA systems is more intuitive than the other visual reasoning
tasks, since many questions can be answered with answers of a few
words or a set of answer choices in a multiple-choice test. Many supervised
tasks in computer vision can be reformulated as a VQA task. For example,
object classification is equivalent to answering the question \emph{``What
is present in the image?''} Action recognition task, on the other
hand, asks a machine agent to respond to the question \emph{``What
action has the person in the video performed?''} Hence, solving VQA
is one step towards building interactive AI assistants \citep{bigham2010vizwiz}.
Researchers have used VQA as a testbed, even as a visual Turing test
\citep{geman2015visual} in its early days, to measure the qualitative
performance of a computer vision system in understanding and reasoning
about natural visual scenes.

A vast number of VQA datasets have been introduced in the last couple
of years to address the visual understanding capabilities of an intelligent
system from different perspectives. For example, a large portion of
questions in VQA v1 \citep{antol2015vqa}, VQA v2 \citep{goyal2017making}
datasets and their derived datasets \citep{agrawal2018don} focus
on simple visual perception, whilst those in the CLEVR \citep{johnson2017clevr}
dataset are specifically for benchmarking machine understanding about
the compositionality of language and multi-step reasoning in a fully
controlled setting. The most recent large-scale GQA dataset \citep{hudson2019gqa}
also tends to address the multi-step reasoning capability but in real-world
scenarios with great visual and language variation. Along with introducing
those large-scale datasets, there has been a surge in methodologies
\citep{anderson2018bottom,andreas2016neural,hu2017learning,hudson2018compositional,johnson2017inferring,kim2018visual,ma2018visual}
proposed to solve VQA. 

However, most current methods focus on static images whilst answering
questions about the temporal dynamics in videos, also known as Video
Question Answering (Video QA), remains greatly challenging. In addition,
much of the existing VQA methods exploit the pattern matching capability
of generic techniques such as Recurrent Neural Networks \citep{hochreiter1997long}
and rely on attention mechanisms to exploit statistics in data rather
than constructing generic, explicit reasoning machines \citep{hudson2018compositional}.
Hence, they require large amounts of annotated training data. There
is thus a need to build more effective methods to generalise using
limited amounts of data. There have also been some efforts in addressing
Video QA \citep{jang2017tgif,gao2018motion}, but it is still unclear
how the elements across input modalities interact. 

In this thesis, we provide a family of effective neural networks to
address the reasoning capability over visual content that spans in
both space and time. We introduce two different views in approaching
VQA: one draws inspiration from the dual system of reasoning in human
reasoning \citep{evans2008dual,kahneman2011thinking}, and the other
from domain-specific perspectives. Our methods exploit the relations
between data elements to improve the robustness and generalisation
capabilities of visual reasoning systems.

\section{Aims and Scope}

\noindent This thesis focuses on how an intelligent system can learn
to reason across vision and language using neural networks. Broadly,
our objectives are:
\begin{itemize}
\item To build new neural architectures for visual reasoning embedding the
concept of dual process in human reasoning systems.
\item To leverage domain knowledge to explore properties and structures
of input modalities and their interactions to improve reasoning in
visual question answering tasks.
\end{itemize}
\noindent Specifically, we study pivotal aspects of input modalities
for visual reasoning in different settings of visual question answering
tasks:
\begin{itemize}
\item \emph{Disentangling visual pattern recognition from compositional
reasoning in Video QA}. We aim to design a framework for learning
to respond to a question about the visual content present in a short
video. The framework divides the whole computational process into
two sub-processes: one reactive visual representation process that
accommodates objects, actions and temporal relations, and the other
one that is deliberative for multi-step reasoning. The division of
labour realises a dual process for reasoning resembling the two cognitive
systems in human reasoning.
\item \emph{Deriving compositional structure of input data and relationships
between data components for Video QA}. In Video QA, information comes
not only from dynamic visual scenes and possibly from other channels
such as subtitles or audio. Hence, Video QA models need to be flexible
to adapt to changes in data modality, varying video length and question
arbitrary expressions. However, existing models are handcrafted and
not optimal for these changes. We aim to build a family of homogeneous
and flexible models that are effective for these tasks with the goal
of exploiting the compositionality of data modality and their relationships
to support the reasoning process. We build models that are highly
effective in handling different settings of Video QA, addressing the
increased complexity as input modalities grow.
\item \emph{Exploring dynamic language-driven visual relationships for visual
reasoning}. Structured representations of visual scenes are beneficial
in discovering categorical and relational information about visual
objects, facilitating visual reasoning such as answering questions
about object relations in an image. However, these object predicates
are often assumed as pre-existing, static and passive to the reasoning
process. We aim to build these structures to be dynamic and responsive
to the reasoning process. In addition, as the associations between
language and vision are crucial for answering questions, we wish to
discover the explicit connections between visual objects and linguistic
components from input data rather than treating a query as a whole,
as in existing studies.
\item \emph{Utilising language-visual grounding for robust visual reasoning}.
Much of visual reasoning methods rely on an attention mechanism to
find the cross-domain associations between linguistic components and
visual entities. However, these associations are often meaningless
and unintuitive due to the lack of supervision during training. We
aim to find the explicit connections between cross-domain components
from external sources to regulate the reasoning process. As there
are a large number of attention-based visual reasoning methods proposed,
the regularisation method is expected to be generic and easily adapted
to different variants of existing methods.
\end{itemize}

\section{Significance and Contribution}

The significance of this thesis is organised around two central lines
of work: (i) learning to reason using neural networks from the dual-system
reasoning perspective and (ii) leveraging domain knowledge to discover
the relationships of the elements living across domains from input
data, facilitating the knowledge retrieval for the reasoning process
in different applications including Image QA and Video QA. In particular,
our contributions are:
\begin{itemize}
\item Construction of a novel neural framework for learning to reason in
Video QA that creates a division of labour between two processes,
similar to the dual-process framework \citep{evans2008dual,kahneman2011thinking}
in the human cognitive system. The proposed framework contains two
interacting components: System 1 (modality-specific pattern extraction),
and System 2 (generic reasoning process with multi-step knowledge
retrieval). The effectiveness of the proposed framework is demonstrated
on SVQA \citep{song2018explore} and TGIF-QA dataset \citep{jang2017tgif}.
\item Design and validation of a new general-reusable neural unit dubbed
Conditional Relation Network (CRN) for multimodal reasoning. The CRN
takes as input a set of domain-independent tensorial objects and translates
them into a new set of objects that encode relationships between the
input objects. The generic design of CRN eases the common complex
model building process of Video QA by simple block stacking and rearrangements,
and affords flexibility in accommodating diverse input modalities
and conditioning features across both visual and linguistic domains.
The resultant network architecture, Hierarchical Conditional Relation
Networks, demonstrates its effectiveness and flexibility in tackling
Video QA task in two settings, short-form Video QA where information
for the answers is solely found in the visual content of a video,
and long-form Video QA where answers can lie in either the visual
scenes or an additional associated information channel, such as movie
subtitles. We are the first to solve short-form Video QA and long-form
Video QA with a homogeneous model. Rigorous experiments are conducted
on a variety of real-world datasets to benchmark the proposed method
against existing methods.
\item Design of a new structured model named Language-binding Object Graph
Network (LOGNet) for visual reasoning and its application on Image
QA. LOGNet takes as input an object-based representation of a visual
scene and computes the object predicates in response to a linguistic
query in a multi-step manner. It also calculates the pairwise visual-linguistic
object bindings which imply the associations of cross-domain concepts.
Such discovered dynamic structures facilitate multi-step knowledge
combination and refinements that iteratively deduce the compact representation
of the final answer. LOGNet proves its strong generalisation performance
against the state-of-the-art methods, on prominent Image QA datasets,
even when given limited amounts of training data.
\item Development of a novel framework to distil external grounding knowledge
as weakly supervision to regulate attention mechanisms in visual reasoning
models. The knowledge extraction relies on the syntactic structure
of linguistic query that allows meaningful cross-domain associations
and robust language representation. Tested on a variety of Image QA
datasets, the proposed method demonstrates its capacity to fortify
any attention-based Image QA models, both in performance and robustness
to linguistic variations.
\end{itemize}

\section{Thesis Structure}

This thesis contains eight chapters with supplementary materials in
the Appendix. The rest of the thesis is arranged as follows:
\begin{itemize}
\item Chapter \ref{chap:Background} briefly reviews the relevant background
that provides a foundation for the main contributions of this thesis.
We begin by introducing the basic deep learning models, including
Feedforward Neural Networks, Convolutional Neural Networks and Recurrent
Neural Networks. We then cover fundamental concepts in neural machine
reasoning, followed by a brief introduction about the dual process
theories in the human cognitive system and recent efforts to bring
this concept to machine intelligence. Finally, we review related works
on using deep neural networks to perform relational reasoning over
an unstructured set of objects in addition to a structured representation
of objects with graphs. 
\item Chapter \ref{chap:VisualLanguageReasoning} presents a comprehensive
literature review on visual and language reasoning, which is the main
focus of this thesis. We first introduce machine reading comprehension,
which is the first introduced form of question answering and has been
a long-standing research topic in natural language processing (NLP).
Next, we present visual and language reasoning as an extension of
machine reading comprehension in NLP to computer vision, creating
an interplay between the two domains. The rest of the chapter is dedicated
to common techniques used to perform visual and language reasoning.
\item Chapter \ref{chap:DualProcess} presents our effort to bring the concept
of dual process theories studied in human reasoning to machine reasoning,
particularly in Video QA setting. To achieve that, we propose a Clip-based
Relational Network (ClipRN) serving as System 1 that encodes visual
patterns about objects, actions and spatio-temporal relations into
a knowledge base. System 2, chosen to be a deliberative multi-step
reasoning engine, iteratively collects clues from the knowledge base
in response to the query to arrive at the answer. Experimental results
show the effectiveness of the modular design of the proposed framework
in adapting to a wide range of low-level visual processing and high-level
reasoning capabilities. It also shows that proper relational modelling
plays a crucial role in reasoning over dynamic scenes.
\item Chapter \ref{chap:MultimodalReasoning} introduces a generic computational
reasoning engine for conditional relational reasoning that can accommodate
a wide range of input types, hence supports multimodal reasoning.
We name the reasoning engine Conditional Relational Network (CRN).
To evaluate the effectiveness and universal use of the proposed CRN,
we test its application on Video QA with a family of resultant block-wise
network architectures in two common Video QA settings: short-form
Video QA where answers are reasoned solely from the visual content
of a video, and long-form Video QA where an additional associated
information channel, such as movie subtitles is presented.
\item Chapter \ref{chap:RelationalVisualReasoning} presents our approach
in bringing a structured representation of visual perception and language
to visual reasoning. We first decompose a given visual scene and linguistic
query as a composition of visual objects and primitive linguistic
components respectively. We then discover the dynamic relational structures
between objects and their adaptive connections with the linguistic
components using a series of Language-binding Object Graph (LOG) units
to retrieve adequate information to derive the answer. We carefully
justify the benefits of exploiting structured representations of data
with the application of the proposed method on Image QA, especially
when given limited amounts of training data.
\item Chapter \ref{chap:RobustnessVR} suggests a framework to utilise external
knowledge about cross-modal bindings to regulate attention mechanisms
in existing visual reasoning methods. We explain how to use the syntactic
structures of linguistic queries to obtain linguistic referring expressions
and subsequently extract meaningful linguistic-visual associations.
We then suggest the first generic dual-modality regulation mechanism
to fortify attention-based Image QA models in performance and consistency.
\item Chapter \ref{chap:Conclusion} summarises the main content of the
thesis and outlines future directions.
\end{itemize}

\newpage{}

\chapter{Background \label{chap:Background}}

The goal of this chapter is to highlight the related background of
the research objectives addressed in this thesis. As mentioned earlier,
the focus of this thesis is to apply the capabilities of deep learning
approaches in learning and reasoning; therefore, we will first go
through the basics of neural networks. Next, we will review the fundamental
concepts of learning and reasoning and how these two intelligence
faculties interact. In the later sections of this chapter, we will
introduce some generic techniques widely used in neural machine reasoning.

\section{Introduction to Neural Networks\label{sec:chap2_Introduction-to-NeuralNet}}

\subsection{Feed-forward Neural Networks\label{subsec:chap2_Feed-forward-Neural-Networks}}

\begin{figure}
\begin{centering}
\includegraphics[width=0.65\textwidth]{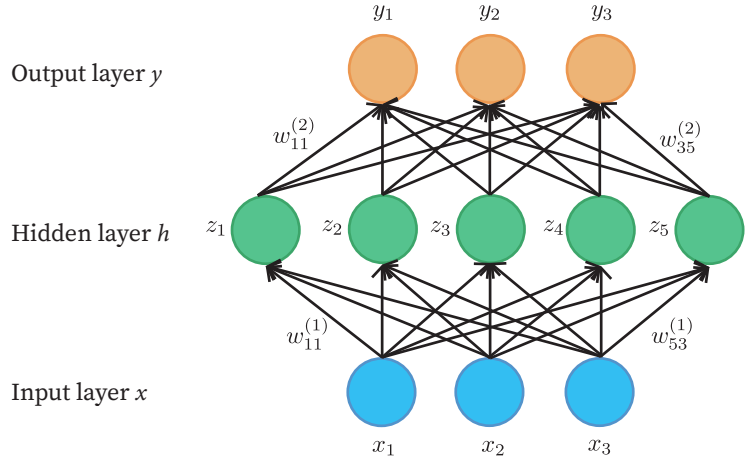}
\par\end{centering}
\caption{A multi-layer perceptron of a single hidden layer.\label{fig:chap2_A-multilayer-perceptron}}
\end{figure}

A feed-forward neural network is a type of artificial neural network
where information travels through one or multiple layers in only one
direction, from the input units to output units, without connections
in the opposite direction. The feed-forward neural network arranges
computational units in a layer-by-layer manner, and each layer is
a set of nodes having connections with nodes of the layer right before
it in the information flow. Network layers that connect to the input
layer and to the output layer are called ``hidden'' layers, and
computational units in a hidden layer are called hidden units or hidden
nodes. The multi-layer perceptron (MLP), the most common feed-forward
network, is widely used for data classification and regression tasks.
Figure \ref{fig:chap2_A-multilayer-perceptron} presents a simple
multi-layer perceptron of three layers: input, output and a single
hidden layer. Mathematically, the MLP approximates a mapping function
$y=f\left(\mathbf{x},\mathbf{\bm{\theta}}\right)$ where $\mathbf{x}$
is an input vector, $y$ is a classification label in the case of
classification tasks and a real value in the case of regression tasks,
and $\bm{\theta}$ is the network parameters. In particular, let $x_{1},x_{2},...,x_{D}$
be the input variables, the first hidden layer of the MLP computes
$M$ non-linear combinations of the input variables:
\begin{eqnarray}
a_{j} & = & \sum_{i=1}^{D}w_{ji}^{(1)}x_{i}+b_{j0}^{(1)},\\
z_{j} & = & h(a_{j}),
\end{eqnarray}
where $M$ is the number of units at the first hidden layer, and $j=1,...,M$
denotes the connections between the input layer and the first hidden
layer; $w_{ji}$ are weights of the connections, and $b_{j0}$ are
biases. $h(.)$ is a non-linear activation function. Output $a_{j}$
are called \emph{activations }and $z_{i}$ are hidden units. Similarly,
output units whose inputs are hidden units at the \emph{$\ell$-}th\emph{
}hidden layer are given by\vspace{-1cm}

\begin{eqnarray}
a_{k}^{(l)} & = & \sum_{j=1}^{M}w_{kj}^{(l)}z_{j}^{(l-1)}+b_{k0}^{(l)},\\
y_{k} & = & h(a_{k}^{(l)}),
\end{eqnarray}
where $k=1,...,K$, and $K$ is the total number of output units;
$M$ is the number of units at the $(\ell-1)$-th layer.

Regarding the non-linear activation function $h(.)$, it is usually
chosen based on the nature of data as well as the assumption of the
target variables. In practice, the most common activation functions
are\vspace{-1cm}

\begin{eqnarray}
\text{sigmoid}(z) & = & \frac{e^{z}}{1+e^{z}},\\
\text{tanh}(z) & = & \frac{e^{z}-e^{-z}}{e^{z}+e^{-z}},\\
\text{ReLU}(z) & = & \text{max}(z,0).
\end{eqnarray}

For multi-class classification problems, the activation function at
the output layer is usually chosen as the softmax function to output
probabilities of output classes:\vspace{-1cm}

\begin{eqnarray}
\text{softmax}(\mathbf{y})_{k} & = & \frac{e^{y_{k}}}{\sum_{i=1}^{K}e^{y_{i}}}\,\text{\text{for}}\,k=1,..,K\,\text{and}\,\mathbf{y}=(y_{1},...,y_{K})\in\mathbb{R}^{K}.
\end{eqnarray}

A feed-forward neural network is usually trained using an iterative
gradient-based optimisation algorithm in which it estimates the network
parameters (weights) by minimising a cost function. In practice, the
cost function is often the cross-entropy between the training data
and the model's predictions. Formally, it is described by using the
average negative log likelihood:\vspace{-1cm}

\begin{eqnarray}
\mathcal{L} & = & -\frac{1}{N}\sum_{i=1}^{N}\text{log}P(\hat{y_{i}}=y_{i}\mid x_{i}),
\end{eqnarray}
where $N$ is the total number of training instances, $x_{i}$ denotes
the \emph{i}-th input data and $y_{i}$ is its corresponding label,
and $\hat{y}_{i}$ is a predicted label at the end of the forward
propagation. In order to compute the gradients of the loss $\mathcal{L}$
with respect to the network parameters (w.r.t) $\bm{\theta}=\{(w_{ji}^{(1)},b_{j0}^{(1)}),...,(w_{kj}^{(l)},b_{k0}^{(l)})\}$,
the back-propagation algorithm \citep{rumelhart1986learning} is used.
Upon the gradients computed, an optimisation algorithm, such as stochastic
gradient decent, is used to perform network learning by updating the
network parameters:\vspace{-1cm}

\begin{eqnarray}
w_{ji} & \coloneqq & w_{ji}-\lambda\frac{\partial\mathcal{L}}{\partial w_{ji}},\\
b_{j0} & \coloneqq & b_{j0}-\lambda\frac{\partial\mathcal{L}}{\partial b_{j0}},
\end{eqnarray}
where $\lambda$ is a small learning rate.

\subsection{Convolutional Neural Networks}

\begin{figure}[t]
\centering{}\includegraphics[clip,width=0.8\columnwidth]{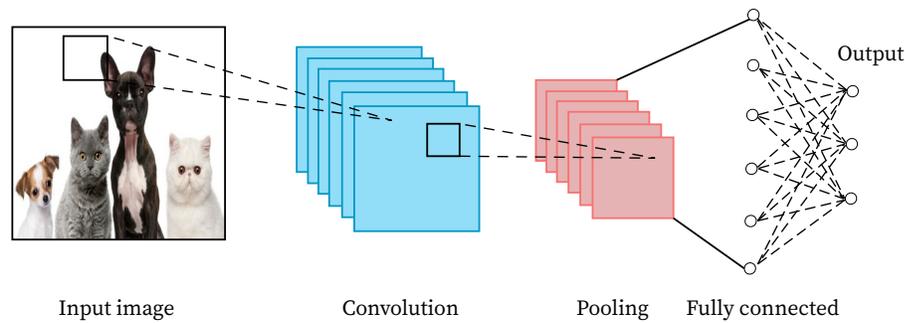}
\caption{Illustration of a simple CNN architecture. It is composed of three
types of layers: convolutional layer, pooling layer and fully-connected
layer. A convolution is to compute an activation between a small filter
and an image patch. The output of the convolutional layer is a feature
map that describes detected feature in the input image. The pooling
layer is a downsampling operation that reduces the output feature
maps of the convolutional layer. Finally, the fully connected layer
is a feed-forward network that is used as a classifier.\label{fig:chap2_cnn_architecture}}
\end{figure}

Convolutional Neural Networks (CNNs) \citep{lecun1998gradient} are
a type of feed-forward neural network for data with grid-like-topology.
Originally, CNN was neurobiologically motivated by exploring local
connectivities in the visual cortex, therefore, mainly designed for
image recognition tasks. In practice, CNN has been successfully applied
to a wide range of domains, other than just image processing, such
as natural language processing tasks \citep{sutskever2014sequence,bahdanau2014neural},
video understanding \citep{trando2015learning}, speech recognition
\citep{abdel2014convolutional}, and time-series analysis \citep{gamboa2017deep}.
In a CNN model, image data can be viewed as 2D grid data, whilst time-series
data can be viewed as 1D data taking samples at overlapping time windows.
There can be numerous variants of CNN model, however, all the CNN
models follow a common topology of three main types of layers: convolutional
layer, pooling layer, and fully-connected layer. Figure \ref{fig:chap2_cnn_architecture}
illustrates a simple CNN network architecture. Here, we explain the
intuition of the elements of a CNN network for the case of 2D image
input data.

\textbf{Convolutional layer} uses convolution operators to extract
high-level features of a given input image. Conceptually, convolution
is defined as an integral giving the amount of overlapping information
of one function when translated over another function. The convolution
of two signals $f$ and $g$ in a continuous space is given by
\begin{eqnarray}
(f*g)(t) & = & \int_{-\infty}^{\infty}f(\tau)g(t-\tau)d\tau.
\end{eqnarray}
When working with discrete data, the corresponding discrete convolution
is given by: 
\begin{eqnarray}
(f*g)(n) & = & \sum_{m=-\infty}^{\infty}f(m)g(n-m),\label{eq:chap2_1d_conv}
\end{eqnarray}
where $f$ is usually referred as input and $g$ as the filter or
kernel.

In computer vision, input data is usually a two-dimensional image.
Thus, the convolution operator in Eq. \ref{eq:chap2_1d_conv} becomes\vspace{-1cm}

\begin{eqnarray}
(F*G)(x,y) & = & \sum_{i}\sum_{j}F(i,j)G(i-x,j-y).\label{eq:chap2_2d_conv}
\end{eqnarray}

Alternatively, as convolution is communicative, we also can present
the convolution between 2-dimensional input $F$ and two-dimensional
kernel $K$ as:
\begin{eqnarray}
(G*F)(x,y) & = & \sum_{i}\sum_{j}F(i-x,j-y)G(i,j).\label{eq:chap2_conv2d_communicative}
\end{eqnarray}

\begin{figure}[t]
\centering{}\includegraphics[clip,width=0.8\columnwidth]{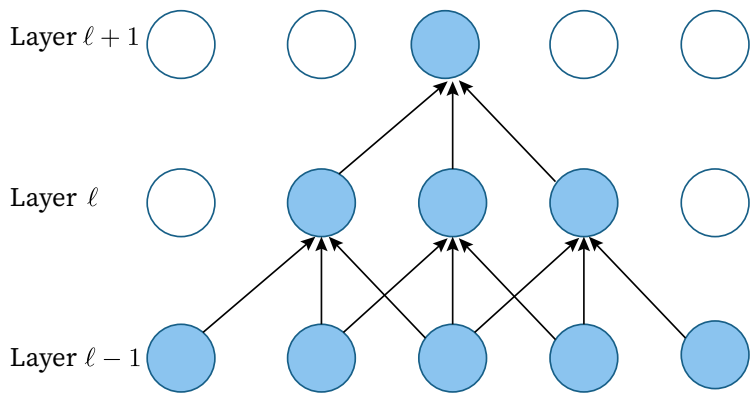}
\caption{Sparse connectivity \citep{goodfellow2016deep}. Each unit in layer
$\ell$ only is connected with a certain number of units instead of
being connected with all units present in the previous layer.\label{fig:chap2_sparse}}
\end{figure}

The output of a 2D convolution layer is often called a feature map.
A convolution is used to extract meaningful features such as object's
edges and object's shapes. In practice, people usually use filters
of small sizes such as $(3\times3\times3)$ or $(5\times5\times3)$.
Hence, it is more convenient to use the commutative form as in Eq.
\ref{eq:chap2_conv2d_communicative} during implementation as the
smaller ranges of the subscripts $(i,j)$. The benefits of using a
CNN over a MLP, explained in Sec. \ref{subsec:chap2_Feed-forward-Neural-Networks}
can be explained in two aspects: sparse connectivity and parameter
sharing \citep{goodfellow2016deep}. First, since we use a small filter
kernel to detect features of the whole image, which is usually much
bigger than the filter kernel, it helps reduce the total number of
network parameters needed and memory space compared to the traditional
MLP. Figure \ref{fig:chap2_sparse} illustrates the notion of sparse
connectivity in CNN with a filter size of $(3\times3\times3)$. Sparse
connectivity means that a computational unit in the $\ell$-th layer
has connections with only a certain number of computational units
in the $(\ell-1)$-th layer rather than having connections to all
the hidden units in the $(\ell-1)$-th layer.

Second, since we use a filter to slide over the whole input image
repeatedly, we can share the same set of filter kernel parameters
for all image patches instead of using separate weights for each image
pixel. Parameter sharing significantly reduces the number of parameters
needed compared to dense matrix multiplication used in an MLP. Parameter
sharing also explains why CNN is translation invariant which allows
it to detect objects regardless of their locations in an image.

\textbf{Pooling Layer} uses downsampling operations to reduce the
spatial size of feature maps produced by a convolutional layer. In
practice, max pooling \citep{ranzato2007sparse} and average pooling
\citep{lecun1998gradient} are popular choices. These operations replace
a rectangular neighbourhood in the feature maps with the maximum value
in the neighbourhood for max pooling or the averaged value of them
for average pooling. Applying a pooling operation significantly reduces
the number of training parameters, therefore, helps control the problem
of overfitting. Besides, the pooling operation shrinks the spatial
size of the feature maps and makes the network invariant to small
translations of the input. By having invariance to translation, CNN
can detect objects that vary in size. The ability to handle inputs
of varying size makes the pooling layer an essential component in
many tasks \citep{goodfellow2016deep}.

\textbf{Fully-Connected Layer} is basically a MLP as in Sec. \ref{subsec:chap2_Feed-forward-Neural-Networks}
where a unit in one layer connects to every unit in the previous layer.
Similar to the use of MLP for multi-class classification tasks, a
softmax layer is used as the activation function at the last layer
of the network to calculate probabilities of class labels. To put
it another way, in a CNN architecture, convolution and pooling layers
are responsible for a feature extraction phase, while the fully-connected
layer can be regarded as a classification phase. The network parameters
are trained using a back-propagation algorithm similar to the training
paradigm of a feed-forward neural network described in Sec. \ref{subsec:chap2_Feed-forward-Neural-Networks}.

Most CNN network architectures follow a general design principle of
stacking convolution layer, pooling layer and fully-connected layer
on top of each other. Depending on the nature of data and the amount
of training data, there are various CNN architectures proposed. Generally,
these CNN models can be classified into two groups: classical CNN
models which are relatively shallow (e.g. LeNet-5 \citep{lecun1998gradient},
AlexNet \citep{krizhevsky2012imagenet}, VGG16 \citep{simonyan2014very})
and modern CNN models which can be very deep with hundreds of total
layers (e.g. Inception \citep{szegedy2015going}, ResNet \citep{he2016deep}).

\subsection{Recurrent Neural Networks}

Unlike feed-forward neural networks where information only travels
in one way from input to output, a recurrent neural network (RNN)
is a type of artificial neural network that uses a feedback loop to
pass information from the output to the input at some point. In terms
of operation, a feed-forward neural network transforms a fixed-size
vector input into a fixed-size vector output, whilst an RNN is designed
to map a sequence varying in length into a different sequence also
varying in length. This makes RNNs suitable to model sequential or
time-dependent data. We will discuss two common variants of RNN in
this section: vanilla RNNs \citep{rumelhart1986learning} and Long
Short-Term Memory (LSTM) \citep{hochreiter1997long}.

\paragraph{Vanilla RNNs}

\begin{figure}
\begin{centering}
\includegraphics{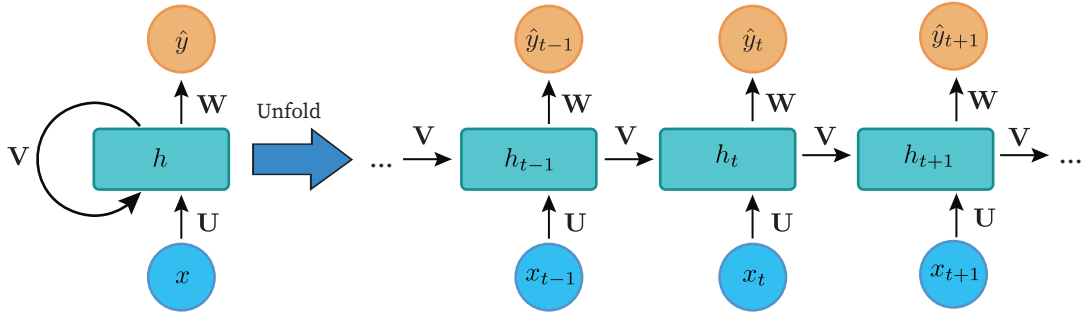}
\par\end{centering}
\caption{Illustration of a vanilla Recurrent Neural Network (left) and its
unfolded computational graph (right) \citep{fdeloche2017rnn}. At
a time step $t$, a computational unit computes a non-linear combination
of the current input $\mathbf{x}_{t}$ and the previous hidden state
$\mathbf{h}_{t-1}$ and generates an output $\mathbf{o}_{t}$. $\mathbf{W},\mathbf{U},\mathbf{V}$
are learnable weight matrices of the model.\label{fig:chap2_vanilla_rnn}}
\end{figure}

A vanilla RNN is illustrated in Fig. \ref{fig:chap2_vanilla_rnn}.
It is a series of RNN units that map a sequence of input vectors to
a sequence of output vectors. Given an input sequence $\mathbf{X}=\{\mathbf{x}_{1},\mathbf{x}_{2},...,\mathbf{x}_{T}\}$,
for each RNN unit at a time step $t$, there are three components
associated with it: input vector $\mathbf{x}_{t}$ $(\mathbf{x}_{t}\in\mathbb{R}^{N})$,
hidden state $\mathbf{h}_{t}$ $(\mathbf{h}_{t}\in\mathbb{R}^{D})$
and output vector $\mathbf{o}_{t}$ $(\mathbf{o}_{t}\in\mathbb{R}^{M})$.
The hidden state $\mathbf{h}_{t}$ is also known as ``memory'' of
an RNN as it takes into account historical information of the input
sequence. Formally, the forward pass of the computational flow happening
at time step $t$ is given by:\vspace{-1cm}

\begin{eqnarray}
\mathbf{h}_{t} & = & f(\mathbf{U}\mathbf{x}_{t}+\mathbf{V}\mathbf{h}_{t-1}+\mathbf{b}_{1}),\\
\hat{\mathbf{y}}_{t} & = & g(\mathbf{W}\mathbf{h}_{t}+\mathbf{b}_{2}),
\end{eqnarray}
where $\mathbf{b}_{1}\in\mathbb{R}^{D},\:\mathbf{b}_{2}\in\mathbb{R}^{M}$
are bias vectors, $\mathbf{U}\in\mathbb{R}^{D\times N},\:\mathbf{V}\in\mathbb{R}^{D\times D},\:\mathbf{W}\in\mathbb{R}^{M\times D}$
are learnable weight matrices, and $f\left(.\right),\:g\left(.\right)$
are non-linear activation functions. Unlike feed-forward neural networks,
the network parameters $\mathbf{U},\mathbf{V},\mathbf{W},\mathbf{b}_{1},\mathbf{b}_{2}$
in an RNN are tied across all steps. This means that all RNN units
perform the same task but taking different input information. While
$f\left(.\right)$ is usually the tanh function or the ReLU function,
$g\left(.\right)$ is chosen depending on the task. For example, in
a language model where, given a sentence, we want to predict the next
word, the RNN's $g(.)$ would be the softmax function to calculate
a vector of probabilities of words appeared in the vocabulary.

Let $\hat{\mathbf{Y}}=\left\{ \hat{\mathbf{y}}_{1},\hat{\mathbf{y}}_{2},...,\hat{\mathbf{y}}_{T}\right\} $
be the output sequence and $\mathbf{Y}=\left\{ \mathbf{y}_{1},\mathbf{y}_{2},...,\mathbf{y}_{T}\right\} $
be the ground-truth sequence, the total loss is computed as the sum
of losses over all the time steps:\vspace{-1cm}

\begin{eqnarray}
\mathcal{L}(\mathbf{Y}\mid\mathbf{X}) & = & \sum_{t=1}^{T}\mathcal{L}_{t}(\mathbf{y}_{t}\mid\mathbf{x}_{1},\mathbf{x}_{2},..,\mathbf{x}_{T})\\
 & = & -\sum_{t=1}^{T}\text{log}P(\hat{\mathbf{y}}=\mathbf{y}\mid\mathbf{x}_{1},\mathbf{x}_{2},..,\mathbf{x}_{T}).
\end{eqnarray}

Similar to feed-forward neural networks, the loss function can be
minimised by using an iterative gradient-based optimisation algorithm.
Computing the gradient for the loss $\mathcal{L}(\mathbf{Y}\mid\mathbf{X})$
is expensive as it is inherently sequential. The gradient w.r.t the
network parameters can be computed by the Back-Propagation Through
Time algorithm \citep{werbos1990backpropagation}. 

In theory, vanilla RNNs are capable of capturing the long-term dependencies
between inputs in a sequence. However, they face difficulties to carry
information from earlier steps to later ones in practice due to gradient
vanishing or gradient exploding during back propagation \citep{bengio1994learning}.
Next, we will explain who Long Short-Term Memory (LSTM) helps mitigate
this problem.

\paragraph{Long Short-Term Memory}

\begin{figure}
\begin{centering}
\includegraphics[width=1\textwidth]{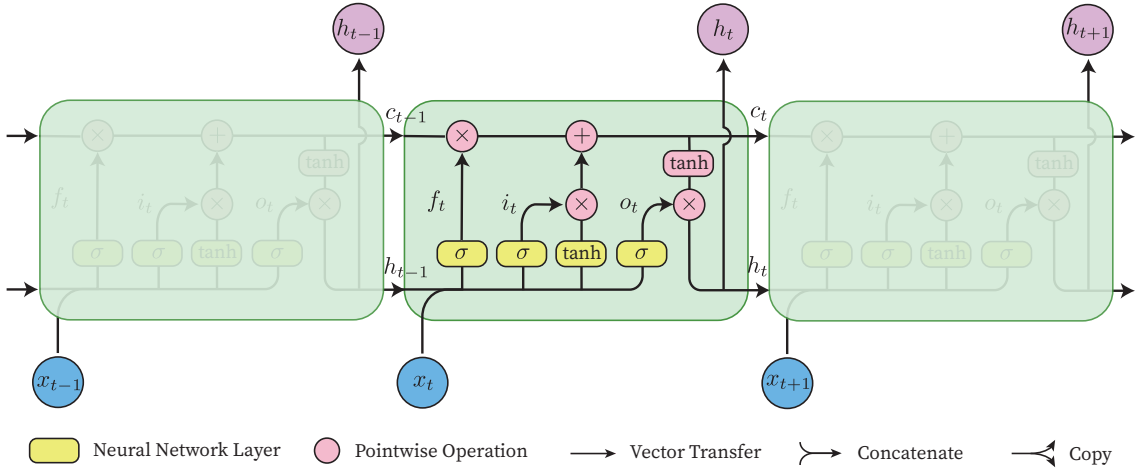}
\par\end{centering}
\caption{Illustration of Long Short-Term Memory (LSTM) \citep{olah2015understanding}.
At a time step $t$, a computational unit computes a non-linear combination
of the current input $\mathbf{x}_{t}$ and the previous hidden state
$\mathbf{h}_{t-1}$ and generates an timely output $\mathbf{c}_{t}$.\label{fig:chap2_lstm}}
\end{figure}

LSTM \citep{hochreiter1997long} is proposed to capture the long-term
dependencies in a given input sequence. Similar to a vanilla RNN,
an LSTM network is also a chain of LSTM cells (See Fig. \ref{fig:chap2_lstm}).
However, different from an RNN cell of simply having a single neural
network, an LSTM cell is a combination of different networks interacting
with each other in a special mechanism to regulate the flow of information,
hence, help pass the gradient more effectively. As a result, the LSTM
does not suffer from the gradient vanishing problem. In particular,
each LSTM cell at a time step $t$ takes as input a previous cell
state $\mathbf{c}_{t-1}\in\mathbb{R}^{h}$, a previous hidden state
$\mathbf{h}_{t-1}\in\mathbb{R}^{h}$ and current input $\mathbf{x}_{t}\in\mathbb{R}^{d}$
and returns a new cell state $\mathbf{c}_{t}\in\mathbb{R}^{h}$, a
new hidden state $\mathbf{h}_{t}\in\mathbb{R}^{h}$. The cell state
$\mathbf{c}_{t}$ acts as the ``memory'' of the network that transports
the information flow across the sequence of LSTM cells. The information
passing from one state to later ones is controlled by a forget gate
$\mathbf{f}_{t}$. Intuitively, the forget gate decides which information
from the previous step is important for the next step and throws irrelevant
information away. Formally, the forget gate $\mathbf{f}_{t}$ is given
by
\begin{eqnarray}
\mathbf{f}_{t} & = & \sigma(\mathbf{W}^{f}\mathbf{\mathbf{x}}_{t}+\mathbf{U}^{f}\mathbf{h}_{t-1}+\mathbf{b}^{f}),
\end{eqnarray}
where $\sigma$ indicates the sigmoid function, and $\mathbf{W}^{f}\in\mathbb{R}^{h\times h},\:\mathbf{U}^{f}\in\mathbb{R}^{h\times h}$
are weight matrices, $\mathbf{b}^{f}\in\mathbb{R}^{h}$ is a bias
vector.

In the next step, the network prepares a new candidate $\tilde{\mathbf{c}}_{t}$
for the cell state. At the same time, it regulates which new information
to store into the new cell state by using an input gate layer $i_{t}$:\vspace{-1cm}

\begin{eqnarray}
\tilde{\mathbf{c}}_{t} & = & \text{tanh}(\mathbf{W}^{c}\mathbf{x}_{t}+\mathbf{U}^{c}\mathbf{h}_{t-1}+\mathbf{b}^{c}),\\
\mathbf{i}_{t} & = & \sigma(\mathbf{W}^{i}\mathbf{x}_{t}+\mathbf{U}^{i}\mathbf{h}_{t-1}+\mathbf{b}^{i}).
\end{eqnarray}

Once the contribution of the previous steps to the new cell state
is calculated, we update the cell state by\vspace{-1cm}

\begin{eqnarray}
\mathbf{c}_{t} & = & \mathbf{f}_{t}*\mathbf{c}_{t-1}+\mathbf{i}_{t}*\tilde{\mathbf{c}}_{t}.
\end{eqnarray}

Finally, we compute the output hidden state $\mathbf{h}_{t}$ from
the newly acquired memory state $\mathbf{c}_{t}$ under control of
an output gate $\mathbf{o}_{t}$:\vspace{-1cm}

\begin{eqnarray}
\mathbf{o}_{t} & = & \sigma(\mathbf{W}^{o}\mathbf{x}_{t}+\mathbf{U}^{o}\mathbf{h}_{t-1}+\mathbf{b}^{o}),\\
\mathbf{h}_{t} & = & \mathbf{o}_{t}*\text{tanh}(\mathbf{c}_{t}).
\end{eqnarray}

The LSTM has been widely applied to a variety of applications including
natural language processing \citep{wang2015predicting,liu2016recurrent},
speech recognition \citep{graves2005framewise,graves2013speech} and
time series prediction \citep{schmidhuber2005evolino}. In practice,
the bidirectional Long Short-Term Memory (BiLSTM) \citep{graves2005framewise}
is often used to improve the performance of the unidirectional LSTM.
BiLSTM is an extension of the traditional unidirectional LSTM where
two independent LSTM passes are used, one moves forward through time,
and the other one moves backward through time across a given input
sequence. By doing this, information at every time step includes information
in the future and information in the past.

Other than LSTM and its variants, Gated Recurrent Units (GRU) \citep{cho2014learning}
is a common alternative RNN that also offers the ability to capture
long-term dependencies in input sequences. The GRU has demonstrated
comparable performance with LSTM in many different tasks \citep{chung2014empirical}.

\section{Neural Machine Reasoning\label{sec:chap2_Neural-Machine-Reasoning}}

The ultimate goal of AI is to have an agent that is able to interact
with humans naturally. We as humans are not only superb at learning
to differentiate our surroundings but also excellent at reasoning
about the world based on the available information. How to design
an AI agent with such capability remains largely open. This section
will review the key concepts of learning and reasoning, especially
learning to reason with neural networks.

Although there exist different formal definitions of a reasoning system
\citep{khardon1997learning,bottou2014machine}, it is generally defined
as a decision-making process that solves problems or draws conclusions
from a base representation of knowledge in response to a query. The
presentation of knowledge can be either in the form of symbolic logical
rules or knowledge graphs or high-level representation of concepts/facts
(e.g. deep features). Different reasoning systems use different inference
algorithms to manipulate information facilitating decision making.
Ideally, people expect the learning and inference process in a reasoning
system to be interpretable; hence, the resulting answer is more trustworthy. 

We evaluate our learning process in perceiving the world and reasoning
from what we have learned using a process of question answering. Therefore,
we can also use question answering as a testbed to assess the reasoning
capability of an agent. A question answering (QA) system can be easily
formed into a machine reasoning system under the above framework.
Given a linguistic query and a context either in the form of textual
content (textual QA) or visual content (Visual QA), the knowledge
base in the machine reasoning framework plays the role of the context
in a QA system whilst the query is given in the form of a natural
language question. Another reason that makes QA a suitable task for
assessing machine reasoning capability is that most supervised machine
learning tasks can be reformulated under the question answering framework
\citep{mccann2018natural}. Hence, if an agent can learn to answer
arbitrary questions, it can successfully reason on a wide range of
tasks.

Although symbolic-based machine reasoning methods have many advantages
in representing high-level abstraction, and offering interpretability,
they do not generalise well on downstream tasks such as question answering
\citet{green1970application}. On the contrary, neural networks have
demonstrated their strength in terms of empirical performance on a
wide range of tasks, including textual QA \citep{devlin2018bert}
and Visual QA \citep{antol2015vqa}. The key power of neural networks
lies in the learning capability from raw data. However, one of the
downsides of machine reasoning systems using neural networks is that
they are usually huge models containing millions of parameters and
heavily rely on a huge amount of training data, making it hard to
explain what happens under the hood. In this thesis, we will build
novel reasoning models that are more data-efficient and explore what
necessitates learning to reason with neural networks. Chapter \ref{chap:VisualLanguageReasoning}
will present in detail related studies on reasoning about vision and
language with neural networks.

\section{Dual System of Reasoning\label{sec:chap2_dual_system_reasoning}}

Researchers in psychology had realised the existence of two different
processes of thought, which is formally known as dual process theories
\citep{stanovich2000individual}. Cognitive scientists later confirmed
a similar framework in human reasoning, which contains two separate
cognitive systems with distinct evolutionary histories, and drives
the way humans think and reason \citep{evans2003two}. This framework
in the human cognitive systems is well discussed and popularised by
\citet{kahneman2011thinking} in his book \emph{Thinking Fast and
Slow}. 

The fast thinking process, also known as System 1, is associative
and domain-specific and typically operates in a parallel manner. According
to Daniel Kahneman, System 1 handles the tasks unconsciously and automatically
without much effort. On the contrary, the slow thinking process, also
known as System 2, is deliberative and domain-agnostic and operates
in a sequential manner. In other words, System 2 handles the task
consciously and effortfully under the control of mental processes
and radical thinking. Recently, the idea of bringing a dual-process
theory in human reasoning to machine intelligence has been emerged,
e.g., as discussed in the AAAI panel 2019 with the attendance of Nobel
laureate Kahneman and Turing Award winner Yoshua Bengio. In such systems,
perception tasks such as image classification, object detection and
speech recognition serve as System 1, whilst high-level tasks such
as reasoning and planning serve as System 2. 

In the past decade, machine learning techniques powered by deep learning
have made significant progress on System 1, while applying deep learning
to address System 2 tasks remains relatively new. \citet{bengio2017consciousness}
formalised a framework for System 2 under the context of a learning
agent with goals where an attention mechanism drives the access to
conscious elements. However, there are very limited theories and experiments
to back up the correctness of this speculation. 

The interactions between the two systems in mind is also well studied
by cognitive scientists \citep{kahneman2011thinking}. Most of the
time, tasks are done by System 1. When the intuitions in System 1
are not sufficient to handle the tasks, it triggers the involvement
of System 2. Inputs to System 2 are the intuitions and impressions
from System 1. The purpose of System 2 is to turn these intuitions
into beliefs through more detailed processing. This means information
processed by System 2 is prepared by System 1 with minimal modifications
or without any modifications. The interactions between the two systems
work on the principle of achieving the best performance with minimal
effort. There is a need to turn these aspects of the human mind into
a learning agent when dealing with real-world problems.

\section{Relational Reasoning\label{sec:chap2_relational_reasoning}}

Reasoning over entities and their relations plays a critical role
in a wide range of tasks \citep{lake2017building}. For example, visual
understanding requires understanding the predicates (subject, action,
object), for instance (man, ride, bike), rather than recognising the
objects represented in a visual scene independently. Hence, it is
a desire for machines to be capable to reason about co-existing entities
and their relations in a particular space, such as a scene or a document.
An entity is a real-world object or a person, or even an abstract
concept. There are two types of representation of entities: (i) \emph{set}
representation of entities where relations between elements are undefined,
and their orders do not matter (permutation invariance), and (ii)
\emph{graph} representation of entities where its topology reflects
the relations between connected entities, by either their inherent
relations or relations induced from data. 

In the following subsections, we will discuss learning to represent
and reason over these two types of data with neural networks.

\subsection{Reasoning over Unstructured Sets\label{subsec:chap2_Reasoning-over-Set}}

Compositionality is an inherent property of data in a variety of domains
in computer vision and natural language processing. Compositionality
allows us to break down complex data into a composition of primitive
components, facilitating the learning and reasoning process. For example,
in computer vision, a visual scene is often segmented into a set of
image regions or patches. In a more complex setting, dynamic scenes
in a video can be viewed as a composition of events and activities
evolving over time. In natural language processing, a sentence can
be tokenised into a set of words. Therefore, having the capacity for
reasoning over these sets and deriving the relations between their
elements are crucial for machine intelligence. 

As elements in a set are inherently unordered, operations on sets
should be permutation equivariant and capable of handling the flexible
cardinalities of input. Common set operations are sum, average and
maximum of the elements. These operations are widely used in machine
learning tasks. For example, \citet{lopez2017discovering} uses average
pooling across bags of points for causality detection. \citet{su2015multi},
on the other hand, uses average pooling to summarise information across
multiple views of an object for shape recognition. Although these
operations are simple and widely used, they process elements independently
and almost ignore the interactions between elements. Also, even though
neural networks have proven their success in a variety of machine
learning tasks, they are fundamentally based on vector inputs. Reasoning
over unstructured sets with neural networks is not a trivial extension.

An attention-based mechanism that performs weighted summation over
a set of feature vectors is a primitive operation on sets. The attention
mechanism was initially introduced in sequence-to-sequence translation
\citep{bahdanau2014neural} and later widely applied in other domains
such as computer vision \citep{xu2015show} and speech processing
\citep{chorowski2015attention}. Given input as a set of $N$ objects
$\mathbf{O}=\left\{ \mathbf{o}_{1},\mathbf{o}_{2},...,\mathbf{o}_{N}\right\} $,
where $\mathbf{o}_{i}\in\mathbb{R}^{d}$ the feature vector of the
$i^{th}$ object in the set, $d$ is feature size, an attention-based
function $h_{\bm{\theta}}\left(.\right)$ aims at mapping the input
set into a single vector $\mathbf{c}\in\mathbb{R}^{d}$:
\begin{eqnarray}
\mathbf{c} & = & h_{\bm{\theta}}\left(\left\{ \mathbf{o}_{1},\mathbf{o}_{2},..,\mathbf{o}_{N}\right\} \right),
\end{eqnarray}
where $\bm{\theta}$ is learnable parameters. In sequence-to-sequence
context, $c$ refers to a context vector that represents an input
sentence. One of the earliest implementations of the function $h_{\bm{\theta}}\left(.\right)$
is by \citet{bahdanau2014neural}: 
\begin{eqnarray}
\mathbf{c} & = & \sum_{i=1}^{N}\alpha_{i}\mathbf{o}_{i},\label{eq:chap2_bahdanau_attn}
\end{eqnarray}
where the weight $\alpha_{i}$ of the $i^{th}$ object is calculated
by\vspace{-1cm}

\begin{eqnarray}
\alpha_{i} & = & \frac{\text{exp}(\mathbf{W}^{o}\mathbf{o}_{i})}{\sum_{j=1}^{N}\text{exp}(\mathbf{W}^{o}\mathbf{o}_{j})},\label{eq:chap2_softmax}
\end{eqnarray}
with network parameters $\mathbf{W}^{o}\in\mathbb{R}^{1\times d}$. 

Recent attention-based models \citep{vaswani2017attention,lee2019set}
instead compute the attention weights based on query-key dot products
where both the key and the query are simple linear transformations
of the input set $O$:
\begin{eqnarray}
\mathbf{c} & = & \text{softmax}\left(\frac{\mathbf{Q}\mathbf{K}^{\top}}{\sqrt{d_{k}}}\right)\mathbf{V},\label{eq:chap2_dot_product_attn}
\end{eqnarray}
where $\mathbf{Q},\:\mathbf{K},\:\mathbf{V}$ denote the query, key
and value, respectively. In practice, these matrices are different
linearly projected versions of the input matrix constructed by stacking
the elements in the input set. $d_{k}$ is the size of each vector
in the query $\mathbf{K}$. The \emph{softmax} function is given by
Eq. \ref{eq:chap2_softmax}.

Note that both ways of implementing the attention mechanism in Eq.
\ref{eq:chap2_bahdanau_attn} and Eq. \ref{eq:chap2_dot_product_attn}
are permutation invariant. Later in Chapter \ref{chap:VisualLanguageReasoning},
we extensively review related studies applying attention mechanisms
to reason about visual and language, which is closely related to the
aims of this thesis.

Recently, there has been a surge of interest in using relation networks
\citep{raposo2017discovering,santoro2017simple,zaheer2017deep,lee2019set}
to explicitly capture the dependencies between elements in unstructured
sets with neural networks. Given the input set $\mathbf{O}=\left\{ \mathbf{o}_{1},\mathbf{o}_{2},...,\mathbf{o}_{N}\right\} $,
we first define a function $g_{\bm{\psi}}\left(.\right)$ with parameters
$\bm{\psi}$ operating on a factorisation (subset) $\mathbf{S}$ of
the original set $\mathbf{O}$:\vspace{-1cm}

\begin{eqnarray}
g_{\bm{\psi}}(\mathbf{S}) & \equiv & g_{\bm{\psi}}(\mathbf{o}_{1},\mathbf{o}_{2},..,\mathbf{o}_{s}),\label{eq:chap2_relation_g_psi}
\end{eqnarray}
where $\mid\mathbf{S}\mid\leq N$. When $\mid\mathbf{S}\mid\equiv N$,
$\mathbf{S}$ contains all contents of the input set $\mathbf{O}$.

A relation network is defined as a composite function $f\circ g$,
where $f$ is a mapping function that transforms the set relations
into an output prediction. As the definition of function $g_{\bm{\psi}}\left(.\right)$
in Eq.\ref{eq:chap2_relation_g_psi} is generic, it can operate on
objects directly or subsets of arbitrary cardinalities. Depending
on task-specific properties, low-order or high-order relations may
impose a prior for the reasoning process. For example, \citet{santoro2017simple}
uses pair-wise relations between objects to solve different relational
reasoning tasks such as Image QA, text-based QA and dynamic physical
systems. In particular, the pair-wise-based relation network is given
by\vspace{-1cm}

\begin{eqnarray}
f\circ g\left(\mathbf{O}\right) & = & f_{\bm{\phi}}\left(a\left(g_{\bm{\psi}}\left(\mathbf{o}_{1},\mathbf{o}_{2}\right),g_{\bm{\psi}}\left(\mathbf{o}_{1},\mathbf{o}_{3}\right),...,g_{\bm{\psi}}\left(\mathbf{o}_{N-1},\mathbf{o}_{N}\right)\right)\right),
\end{eqnarray}
where $f_{\bm{\phi}}\left(.\right)$ and $g_{\bm{\psi}}\left(.\right)$
are multi-layer perceptrons (MLPs); $a\left(.\right)$ is an aggregation
function that incorporates permutation invariance. In practice, $a\left(.\right)$
is simply the summation function. Hence, the final form of the relation
network is
\begin{eqnarray}
f\circ g\left(\mathbf{O}\right) & = & f_{\bm{\phi}}\left(\sum_{i,j}g_{\bm{\psi}}\left(\mathbf{o}_{i},\mathbf{o}_{j}\right)\right),i=1,..,N;j=1,..,N.\label{eq:chap2_relation_net_santoro}
\end{eqnarray}

\citet{zhou2018temporal}, on the other hand, utilises multi-scale
relations between frames at different time steps for activity recognition
tasks. This study proves the significance of combining low-order relations
and high-order relations in temporal relational reasoning. 

In a broader setting, \citet{zaheer2017deep} provides an in-depth
theoretical analysis and a generic deep neural network framework for
input sets, formally called DeepSets, by stacking multiple relation
network layers in Eq. \ref{eq:chap2_relation_net_santoro} on top
of each other. This is a natural property of operations on sets as
a composition of permutation equivariant functions is also permutation
invariance. DeepSets allows relation networks to take advantage of
deeper networks that are successfully proven to have better generalisation
than shallow networks across various tasks in computer vision and
natural language processing \citep{he2016deep,vaswani2017attention}. 

In this thesis, we formulate a visual reasoning problem, particularly
in VQA settings, as a set of linguistic words in interaction with
another set of visual objects. We extensively use relation networks
to learn to reason about the relations between elements in visual
scenes, considering the linguistic query as additional meta-information.
In Chapter \ref{chap:DualProcess}, we summarise dynamic visual information
in a video by using a temporal attention mechanism at the clip-level
followed by a query-induced relation network at the video level. In
Chapter \ref{chap:MultimodalReasoning}, we build a novel hierarchical
relation network that is flexible in input modalities and size, allowing
to explicitly model complex interactions of cross-domain inputs.

\subsection{Reasoning over Graphs}

A graph is a set of entities, represented as \emph{nodes,} with explicit
pairwise relations between the elements, represented as \emph{edges}.
Data representation using graphs is ubiquitous across various domains,
such as visual scene graphs, social networks, molecular structure.
Thus, having an accurate graph-based learning system would have a
significant impact on real-world applications. For example, in chemistry,
if a machine agent can effectively exploit the relationships between
the atoms in a graph-based compound structure, it can identify the
bioactivity required for drug discovery. Another example is in social
networks in which properly modelling user interactions in the large-scale
graph will allow us to connect to relevant people sharing the same
interest. In this subsection, we are interested in extending neural
networks to perform inference on graph-based data.

\paragraph{Graph Neural Networks (GNN)}

The idea of using GNN for graph inference is first introduced by \citet{scarselli2008graph},
where it aims at learning hidden embeddings at nodes by aggregating
the information within a neighbourhood of each node. Because graphs
can come in different forms, there have been various variants of the
GNN framework. Here, we describe the GNN framework based on the formulation
by \citet{gilmer2017neural} for simple undirected graphs in which
the authors refer it to \emph{message passing neural network} (MPNN).

\begin{figure*}
\begin{centering}
\begin{minipage}[t]{0.95\textwidth}%
\begin{center}
\begin{minipage}[t]{0.44\textwidth}%
\begin{center}
\includegraphics[width=0.9\columnwidth]{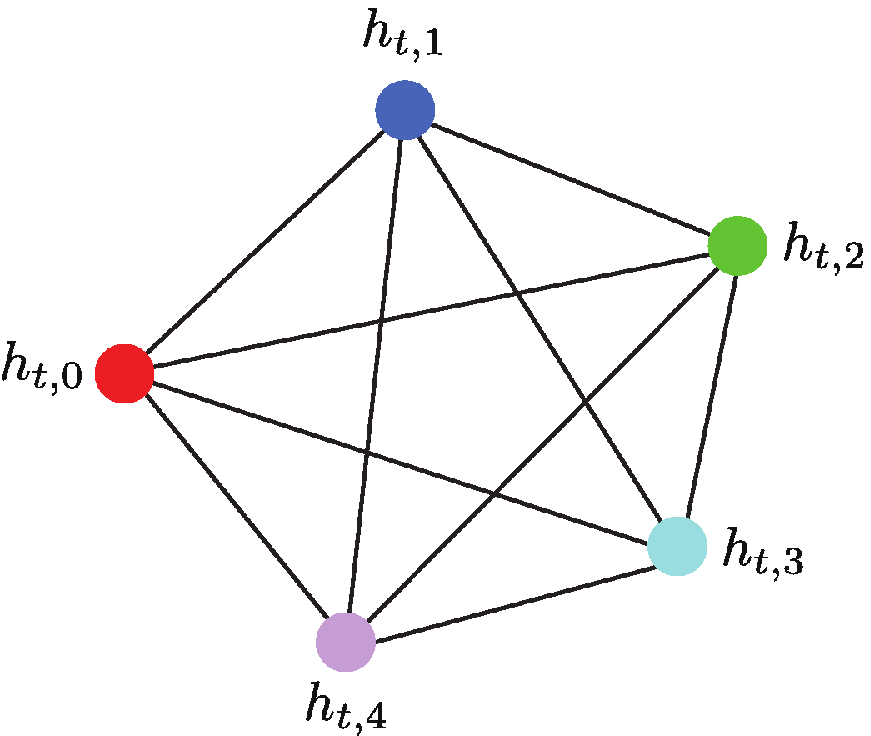}\vspace{-1em}
\par\end{center}
\begin{center}
(a) Node representations at time $t$
\par\end{center}%
\end{minipage}\quad{}%
\begin{minipage}[t]{0.42\textwidth}%
\begin{center}
\includegraphics[width=1\columnwidth]{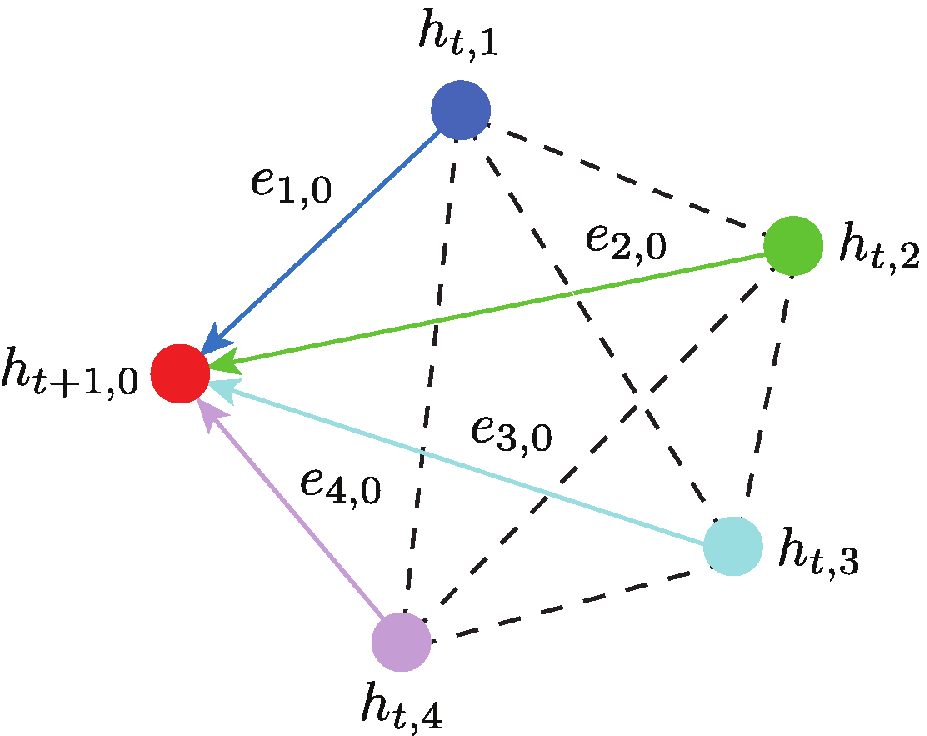}\vspace{-1em}
\par\end{center}
\begin{center}
(b) Message passing phase
\par\end{center}%
\end{minipage}
\par\end{center}%
\end{minipage}\medskip{}
\par\end{centering}
\caption{Illustration of the message passing phase in the graph neural network
at a time step $t$ on a fully connected graph of five nodes (with
identity indicated by colours). $\mathbf{h}_{t,i}$ and $\mathbf{e}_{ij}$
are hidden states at the nodes and edge features, respectively. Left
half: node representations at time $t$. Right half: the hidden state
$\mathbf{h}_{t+1,i}$ at node $i$ is refined by using its state at
time $t$ and messages provided by its neighbours and edge features
(if present).\label{fig:graph_neural_net_message_passing}}
\end{figure*}

Consider a graph $\mathcal{G}=(\mathcal{\bm{\mathcal{V}}},\bm{\mathcal{E}})$
as a composition of vertices (also nodes) $\bm{\mathcal{V}}$ with
edges $\bm{\mathcal{E}}$ denoting the associations between nodes.
Here, each node $i$ is characterised by a feature vector $\mathbf{x}_{i}$
and the edge attribute between node $i$ and node $j$ is denoted
as $\mathbf{e}_{ij}$. The MPNN framework is composed of two phases:
a message passing phase and a readout phase. The message passing phase
is an iterative propagation process of $T$ time steps to refine node
features that are done by a message function $M_{t}\left(.\right)$
and a node update function $U_{t}\left(.\right)$ at every time step
$t$. Mathematically, the operations of the message passing phase
are as follows:\vspace{-1cm}

\begin{eqnarray}
\mathbf{m}_{t+1,i} & = & \sum_{j\in N(i)}M_{t}\left(\mathbf{h}_{t,i},\mathbf{h}_{t,j},\mathbf{e}_{ij}\right),\label{eq:chap2_gnn_message_func}\\
\mathbf{h}_{t+1,i} & = & U_{t}\left(\mathbf{h}_{t,i},\mathbf{m}_{t+1,i}\right),\label{eq:chap2_gnn_node_update_func}
\end{eqnarray}
where $\mathbf{m}_{t+1,i}$ is the message for representation refinement
of hidden state $\mathbf{h}_{t+1,i}$ at node $i$ at time step $t+1$,
and $N(i)$ is the neighbours of node $i$ defined by the graph $\mathcal{G}$.
The initial hidden state is assigned as the initial feature on node
$\mathbf{h}_{0,i}=\mathbf{x}_{i}$. We illustrate the message passing
phase of the MPNN framework in Fig. \ref{fig:graph_neural_net_message_passing}.

The readout phase is a mapping function $R\left(\left\{ .\right\} \right)$
taking as input the final hidden states of the nodes to transform
into a feature vector for prediction\vspace{-1cm}

\begin{eqnarray}
\hat{\mathbf{y}} & = & R\left(\left\{ \mathbf{h}_{T,i}\mid i\in\mathcal{\bm{V}}\right\} \right),
\end{eqnarray}
where $R\left(\left\{ .\right\} \right)$ is a function on set; hence,
it should satisfy the set function's properties as mentioned in Sec.
\ref{subsec:chap2_Reasoning-over-Set} in terms of permutation invariance.

In regards to training the MPNN framework, as the message functions
$M_{t}(.)$, the update functions $U_{t}(.)$ and the readout function
$R(\{.\})$ are neural networks, it can be trained end-to-end with
the back-propagation algorithm similar to other neural networks on
regular Euclidean data introduced in Sec. \ref{sec:chap2_Introduction-to-NeuralNet}.

\paragraph{Graph Convolutional Networks (GCN)}

There is a special interest in generalising the notion of a \emph{convolutional}
\emph{network} to structured data since convolutional networks are
the most successful deep learning technique for grid data in computer
vision and related fields. The family of neural network models that
perform graph convolutions are called graph convolution networks (GCN).
Technically, GCN aim at replacing the message function $M_{t}$ (Eq.
\ref{eq:chap2_gnn_message_func}) and the node update function (Eq.
\ref{eq:chap2_gnn_node_update_func}) in the graph neural network
framework described above with a graph-based convolution operation.
In general, there are two different approaches to define a graph-based
convolution operation: spectral-based GCN and spatial-based GCN. The
spectral-based convolution operation is defined as a parameterised
filtering operation in the Fourier domain based on the spectral graph
theory in graph signal processing. The spatial-based GCN, on the other
hand, defines convolution operations operating directly on graph data
based on the spatial connections between nodes. In Chapter \ref{chap:RelationalVisualReasoning},
we particularly use the spectral-based GCN to refine the representations
of visual objects, presuming a visual scene is represented as a scene
graph of visual objects. We argue that both the graph representation
of visual objects and their interactions with the linguistic elements
in the query set are rather dynamic and required a multi-step inference
as the reasoning proceeds. Here, we explain more detail on the\emph{
spectral-based GCN} approaches.

Recall that the given graph $\mathcal{G}=(\mathcal{\bm{V}},\mathcal{\bm{E}},\bm{A})$
is an undirected graph of $N$ nodes, $\bm{A}\in\mathbb{R}^{N\times N}$
is the adjacency matrix, studies in graph theory \citep{chung1997spectral}
proved that we could represent the graph with its mathematical representation
via the normalised graph Laplacian matrix $\mathbf{L}$:\vspace{-1cm}

\begin{eqnarray}
\mathbf{L} & = & \mathbf{I}_{N}-\mathbf{D}^{-\frac{1}{2}}\mathbf{A}\mathbf{D}^{-\frac{1}{2}},\label{eq:chap2_laplacian_matrix}
\end{eqnarray}
where $\mathbf{I}_{N}$ is the identity matrix, $\mathbf{D}\in\mathbb{R}^{N\times N}$
is the diagonal degree matrix with $\mathbf{D}_{ii}=\sum_{j}\mathbf{A}_{ij}$.
As \textbf{$\mathbf{L}$} is a real symmetric positive semidefinite
matrix, it can be factored as a bilinear form $\mathbf{L}=\mathbf{U}\mathbf{\bm{\varLambda}}\mathbf{U}^{\top}$,
where $\mathbf{U}\in\mathbb{R}^{N\times N}$ is the unitary matrix
of its eigenvectors $\left\{ \mathbf{u}_{i}\right\} _{i=1}^{N}$ and
$\mathbf{\bm{\varLambda}}\in\mathbb{R}^{N\times N}$ is a diagonal
matrix of the associated eigenvalues $\left\{ \lambda_{i}\right\} _{i=1}^{N}$.

Considering a signal on node (node feature) $\mathbf{x}\in\mathbb{R}^{N}$
(each node is characterised by a scalar) and a filter $g_{\bm{\theta}}\left(\bm{\varLambda}\right)=\text{diag}(\bm{\theta})$
with parameter $\bm{\theta}\in\mathbb{R}^{N}$, the convolution operation
on graph is given by\vspace{-1cm}

\begin{eqnarray}
g_{\bm{\theta}}\star\mathbf{x} & = & g_{\bm{\theta}}\left(\mathbf{L}\right)\mathbf{x}=g_{\bm{\theta}}\left(\mathbf{U}\bm{\varLambda}\mathbf{U}^{\top}\right)\mathbf{x}=\mathbf{U}g_{\bm{\theta}}\left(\bm{\varLambda}\right)\mathbf{U}^{\top}\mathbf{x}.\label{eq:chap2_graph_conv_operation}
\end{eqnarray}
All spectral-based GCN methods follow this formula. The computation
of Eq. \ref{eq:chap2_graph_conv_operation} is expensive as the multiplication
with the unitary matrix \textbf{$\mathbf{U}$} has complexity $\mathcal{O}(N^{2})$.
This is more costly than the traditional convolution operator on grid
data, which takes only $\mathcal{O}(N)$. Therefore, many works in
the literature have attempted to offer different choices of the filter
$g_{\bm{\theta}}\left(\bm{\varLambda}\right)$ to reduce the complexity
of the graph convolution operation. 

\citet{defferrard2016convolutional} approximates $g_{\bm{\theta}}\left(\bm{\varLambda}\right)$
by truncated expansions of the Chebyshev polynomial of order $\left(K-1\right)$
of the diagonal matrix of eigenvalues $\bm{\varLambda}$ as suggested
by \citet{hammond2011wavelets}:\vspace{-1cm}

\begin{eqnarray}
g_{\bm{\theta}}\left(\bm{\varLambda}\right) & \approx & \sum_{k=0}^{K-1}\bm{\theta}_{k}T_{k}\left(\tilde{\bm{\varLambda}}\right),
\end{eqnarray}
where $\tilde{\bm{\varLambda}}=\frac{2}{\lambda_{\text{max}}}\bm{\varLambda}-\mathbf{I}_{N}$
is a diagonal matrix of eigenvalues that is scaled to be in $\left[-1,1\right]$,
and $\lambda_{\text{max}}=\text{max}\left(\left\{ \lambda_{i}\right\} _{i=1}^{N}\right)$.
$k$ indicates the Chebyshev polynomial's order. The parameter $\bm{\theta}\in\mathbb{R}^{K}$
is now Chebyshev coefficients and $T_{k}(x)=2xT_{k-1}(x)-T_{k-2}(x)$,
with $T_{0}(x)=1$ and $T_{1}(x)=x$. The graph convolution operation
in Eq. \ref{eq:chap2_graph_conv_operation} can be now written as\vspace{-1cm}

\begin{eqnarray}
g_{\bm{\theta}}\star\mathbf{x} & \approx & \sum_{k=0}^{K-1}\bm{\theta}_{k}T_{k}(\tilde{\mathbf{L}})\mathbf{x},\label{eq:chap2_graph_conv_chebyshev}
\end{eqnarray}
where $\tilde{\mathbf{L}}=\frac{2}{\lambda_{\text{max}}}\mathbf{L}-\mathbf{I}_{N},$
$\mathbf{L}$ is given by Eq. \ref{eq:chap2_laplacian_matrix}, $T_{k}\left(\tilde{\mathbf{L}}\right)\in\mathbb{R}^{N\times N}$.
This approximated graph convolution operation costs $\mathcal{O}(K\mid\bm{\mathcal{E}}\mid)\ll\mathcal{O}(N^{2}),$
where $\mid\mathcal{E\mid}$is the number of edges in the given graph
$\mathcal{G}$.

\citet{kipf2016semi} introduces a multi-layer graph convolution network
based on a first-order approximation of the graph convolution operation
in Eq. \ref{eq:chap2_graph_conv_chebyshev}. Assuming that $\lambda_{\text{max}}\approx2$,
Eq. \ref{eq:chap2_graph_conv_chebyshev} becomes\vspace{-1cm}

\begin{eqnarray}
g_{\bm{\theta}}\star\mathbf{x} & \approx & \bm{\theta}_{0}\mathbf{x}+\bm{\theta}_{1}(\mathbf{L}-\mathbf{I}_{N})\mathbf{x}=\bm{\theta}_{0}\mathbf{x}-\bm{\theta}_{1}\mathbf{D}^{-\frac{1}{2}}\mathbf{A}\mathbf{D}^{-\frac{1}{2}}\mathbf{x},\label{eq:chap2_first_order_graph_conv}
\end{eqnarray}
where $\bm{\theta}_{0}$ and $\bm{\theta}_{1}$ are parameters. In
order to avoid overfitting problem when training deep graph convolution
networks in practice, we can constrain $\bm{\theta}_{0}$ and $\bm{\theta}_{1}$
as a single parameter $\bm{\theta}$. This results to a simplified
formulation of Eq. \ref{eq:chap2_first_order_graph_conv}:\vspace{-1cm}

\begin{eqnarray}
g_{\bm{\theta}}\star\mathbf{x} & \approx & \bm{\theta}\left(\mathbf{I}_{N}+\mathbf{D}^{-\frac{1}{2}}\mathbf{A}\mathbf{D}^{-\frac{1}{2}}\right)\mathbf{x}.\label{eq:chap2_grap_conv_param_constraint}
\end{eqnarray}

\citet{kipf2016semi} rewrites Eq. \ref{eq:chap2_grap_conv_param_constraint}
in the form of a compositional layer and further applies a normalisation
trick to formulate it as feature aggregation at nodes, similar to
the definition of spatial-based graph convolution network:
\begin{eqnarray}
\mathbf{H} & = & \tilde{\mathbf{D}}^{-\frac{1}{2}}\tilde{\mathbf{A}}\tilde{\mathbf{D}}^{-\frac{1}{2}}\mathbf{X}\bm{\Theta},
\end{eqnarray}
where $\tilde{\mathbf{A}}=\mathbf{A}+\mathbf{I}_{N},\tilde{\mathbf{D}}_{ii}=\sum_{j}\tilde{\mathbf{A}}_{ij}$.
$\mathbf{X}\in\mathbb{R}^{N\times d_{1}}$ are the input matrix obtained
by stacking node feature vectors together, and $d_{1}$ is the length
of an input feature vector. The resultant feature map $\mathbf{H}\in\mathbb{R}^{N\times d_{2}}$
is the convolved signal matrix, $d_{2}$ is the size of an output
feature vector, $\bm{\Theta}\in\mathbb{R}^{d_{1}\times d_{2}}$.

Spectral-based GCN has been successfully applied to a variety of domain
such as human behaviour understanding \citep{yan2018spatial}, recommendation
system \citep{ying2018graph}, and video understanding \citep{wang2018videos}.

\section{Closing Remarks}

We have briefly reviewed basic forms of neural network that are related
to the content of this thesis. We also provided an overview of how
to train those neural networks with back-propagation and gradient
descent. To draw a broad picture of how to reason with neural networks,
we have further introduced generic neural reasoning techniques that
will be used or extended later in this thesis in the context of vision
and language reasoning. The following chapter will provide more detailed
explanations of how neural networks can reason across visual and language
in a wide range of configurations and applications.

\newpage{}

\chapter{Visual and Language Reasoning \label{chap:VisualLanguageReasoning}}

The research problem of this thesis presents one of the most exciting
venues for neural networks because it sits at the intersection between
four distinct domains: machine learning, reasoning, computer vision
and natural language processing (NLP). This chapter aims to introduce
the concepts used in visual and language reasoning, and provide a
comprehensive review of existing studies on this topic and closely
related fields.

This chapter is organised into four sections. The first section is
on neural machine reading comprehension, one of the long pursued NLP
research topics. We then present visual and language reasoning tasks
as extensions of the machine reading comprehension in NLP for computer
vision. In the later sections, we review existing methods on visual
reasoning with increasing complexity in visual content, going from
static images to more complex temporal dynamics in short videos and
finally, long-term temporal dependencies in movies.

\section{Neural Machine Reading Comprehension\label{sec:chap3_Neural-Machine-Reading-Comp}}

Machine reading comprehension (MRC), also known as textual question
answering (Text QA), is a task to test if a machine agent understands
natural language by asking it to respond to a natural language question
given context information in a piece of text. Early MRC systems based
on hand-crafted rules and features did not scale well in real-world
applications. Deep neural networks with the capability to accommodate
large training data have significantly advanced MRC. We refer to the
MRC systems applying deep learning techniques as neural MRC in the
rest of this thesis. 

\begin{figure}
\begin{centering}
\medskip{}
\includegraphics[width=0.8\textwidth]{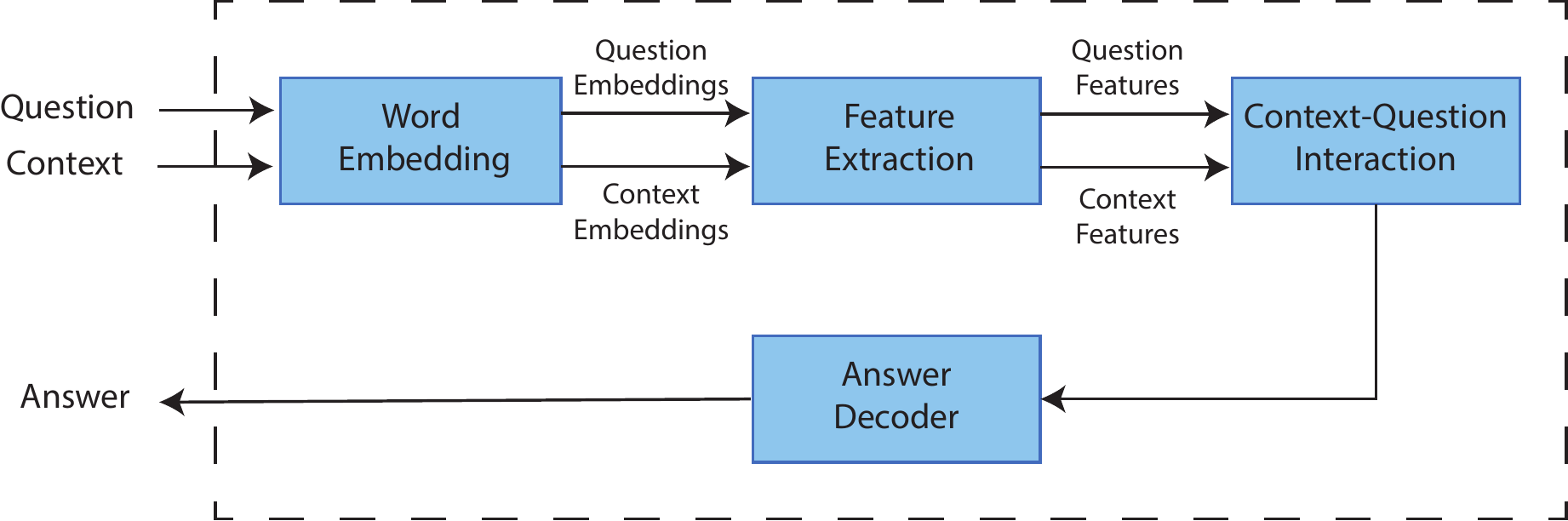}
\par\end{centering}
\caption{General architecture of a Machine Reading Comprehension (MRC) system.\label{fig:chap3_General-MRC}}

\end{figure}

Figure \ref{fig:chap3_General-MRC} presents the general architecture
of a neural MRC system. Inputs to the MRC are a question and context,
both are in a textual form, and the output is an answer to the question.
A typical design of an MRC system contains four key components: word
embedding, feature extraction, context-question interaction, and answer
decoder. Thanks to the arbitrary nature of the question, the MRC system
can come in different forms of expected answer forms, resulting in
different answer decoder's designs. In general, MRC systems can be
categorised into different tasks depending on the expected answer
forms: open-ended answering, multi-choice answering, span extraction.
We will detail each of these tasks later in Section \ref{subsec:chap3_Answer-Decoder}.

\subsection{Word Embedding\label{subsec:chap3_Word-Embedding}}

Since machines cannot read and understand natural language as humans
do, we have to numerically represent text inputs before sending it
to an MRC system. Word embeddings are numerical representations of
textual words in which they should capture as much of the lexical
meaning of the words as possible. In NLP, words are usually encoded
into a fixed-length vector space due to their convenience to analyse
their relationships. Different choices in word embeddings can lead
to significant differences in performance \citep{dhingra2017comparative}.
Hence, word embeddings have become a dominant area of research in
NLP. In general, word representation techniques can be categorised
into two main approaches: traditional word representation (context-free
word representation) and pre-trained contextualised word representation.

\subsubsection{Traditional Word Representation}

\paragraph{One-hot Encodings:}

This is the most straightforward way to represent a word as a fix-length
vector. Given the vocabulary of $n$ words, a word is represented
as an $n$-dimensional spare vector whose values are mostly ``0''
except a single ``1'' at the position that the word appears in the
vocabulary. Although it is convenient and straightforward to use,
one-hot word representation is sub-optimal as it does not reflect
the linguistic similarities across words. For example, we would expect
the representation for the word ``boy'' to be more related to the
representation for ``girl'' in the vector space than for ``tree''.
However, one-hot encodings are equally distant between any two words
in the vocabulary in terms of lexical representation. In addition,
one-hot encodings also cause computational burden when the vector
sizes increase as the vocabulary size grows. Training such high dimensional
and sparse vectors given a limited amount of training data is also
problematic. 

\paragraph{Word2Vec:}

A large body of studies had conducted based on the distribution hypothesis
\citep{firth1957synopsis} to integrate lexical meanings into word
representation by modelling the relationship between a target word
and its surrounding words in a given context. The output distributed
word representation helps address the shortcomings of conventional
word representation, for example, one-hot encodings. With the development
of deep neural networks along with more powerful hardware, training
language models \citep{bengio2003neural,mnih2007three,mnih2008scalable,mnih2013learning,collobert2008unified}
on a massive amount of data have considerably improved the distributed
word representation. Word2Vec \citep{mikolov2013distributed} is one
of the most successful methods in this line of work. It has been widely
applied to different NLP downstream tasks, such as machine translation
\citep{mikolov2013exploiting}, text classification \citep{lai2015recurrent}.

Word2Vec offers two models: the Continuous Bag-of-Word (CBOW) model
and the Skip-gram model. These models are algorithmically similar.
Whilst the CBOW model tries to model a target word's representation
conditioned on its context words, the Skip-gram model's objective
is to learn the representation of surrounding context words from the
given target word. Formally, considering a sequence $\{w_{i}\}_{i=1}^{T}$
of $T$ words, let $w_{t}$ be the target word, the objective of the
CBOW model is to maximise the log probability:\vspace{-1cm}

\begin{eqnarray}
\mathcal{L} & = & \frac{1}{T}\sum_{t=1}^{T}\sum_{-c\leq j\leq c,j\neq0}\text{log}P(w_{t}\mid w_{t+1}),
\end{eqnarray}
where $c$ is the context window size. The basic form of $P(w_{t}\mid w_{t+1})$
is the softmax function\vspace{-1cm}

\begin{eqnarray}
P(w_{t} & \mid & w_{t+1})=\frac{\text{exp}(\text{sim}(w_{t},w_{t+1}))}{\sum_{w^{\prime}\in\mathbf{V}}\text{exp}(\text{sim}(w^{\prime},w_{t+1}))},
\end{eqnarray}
where $w^{\prime}$ is a word in the vocabulary $\mathbf{V}$, $sim(w_{t},w_{t+1})$
denotes the similarity between the target word $w_{t}$ and its context
words $w_{t+1}$.

The roles of $w_{t}$ and $w_{t+1}$ are swapped in the case of the
Skip-gram model.

\paragraph{GloVE:}

Even though the word representation produced by Word2Vec offers richer
representations of words in terms of semantics than one-hot embeddings,
it only uses the local information within a context window scanning
over the corpus but ignores the global statistic information. GloVE
\citep{pennington2014glove} is proposed to address this shortcoming.
In particular, the GloVE first constructs a global co-occurrence matrix
$\mathbf{X}_{ij}$ which represents how often the target word $i$
appears in the context given by word $j$. The relationship between
the target word and the context word is approximated by:\vspace{-1cm}

\begin{eqnarray}
\mathbf{w}_{i}^{\top}\mathbf{w}_{j}+b_{i}+b_{j} & = & \text{log}(\mathbf{X}_{ij}),
\end{eqnarray}
where $\mathbf{w}_{i},\mathbf{w}_{j}$ are vector representation of
the target word $i$ and the context word $j$, respectively; $b_{i}$
and $b_{j}$ are scalar bias. Finally, we minimise a reconstruction
loss to factorise the co-occurrence matrix, which yields a lower-dimensional
matrix containing vector representations of words in the corpus. The
reconstruction loss is given by:\vspace{-1cm}

\begin{eqnarray}
J & = & \sum_{i,j=1}^{\mathbf{\mid V\mid}}f(\mathbf{X}_{ij})(\mathbf{w}_{i}^{\top}\mathbf{w}_{j}+b_{i}+b_{j}-\text{log}(\mathbf{X}_{ij})),
\end{eqnarray}
where $\mid\mathbf{V}\mid$ is the size of the vocabulary. $f(x)$
is a weight function to prevent common word pairs from dominating
the learning process. In practice, $f(x)$ is chosen as:\vspace{-1cm}

\begin{eqnarray}
f(x) & = & \begin{cases}
(x/x_{\text{max}})^{\alpha} & \text{if}\,x<x_{\text{max}}\\
1 & \text{otherwise,}
\end{cases}
\end{eqnarray}
where $\alpha$ is a scalar value.

\subsubsection{Pre-trained Contextualised Word Representation}

Even though the traditional word representation approaches are widely
used in NLP systems, they only offer a global representation for each
word in a corpus. Ideally, people expect a representation of a word
to be tailored to their context in a specific sentence or paragraph.
For example, considering the word \emph{``bank''} in two following
sentences, \emph{``I need to go to the bank''} and \emph{``I stroll
along the river bank''}. It carries different meanings in these two
examples, one refers to a financial institute, and the other implies
a land near a river. Contextualised word representation approaches
embed tokens in consideration of the other tokens in a sentence, allowing
them to capture more precise semantic meanings of words across diverse
linguistic contexts. Even though contextualised word representation
techniques are relatively new, they have surpassed all the traditional
methods in a wide range of NLP tasks \citep{devlin2018bert} with
large margins. In practice, contextual word embeddings are learned
by a language model which is trained on large-scale corpora in an
unsupervised manner.

\paragraph{ELMo:}

The ELMo model \citep{peters2018deep} extracts context-dependent
word representation based on a bidirectional language model. A forward
$L$-layer LSTM and a backward $L$-layer LSTM are used to run over
a sequence of $N$ tokens in opposite directions. The obtained contextualised
representations at each position in the sentence from the two LSTM
passes' layers are further concatenated to output $N$ hidden representations
for $N$ given tokens. Eventually, for each token in the sentence,
the bidirectional language model produces $(L+1)$-layer representations,
including the global context-independent representation obtained via
a traditional word embedding method.

To use the ELMo model in downstream tasks such as question answering,
sentiment analysis, we simply squeeze the $(L+1)$-layer representations
into a single vector for each token $k$:\vspace{-1cm}

\begin{eqnarray}
\text{ELMo}_{k}^{task} & = & \gamma^{task}\sum_{j=0}^{L}s_{j}^{task}\mathbf{h}_{k,j},
\end{eqnarray}
where $\gamma^{task}$ is task-specific scale factor, $s_{j}^{task}$
is a softmax-normalised weight and $\mathbf{h}_{k,j}$ is $j$-layer's
representation of token $k$. The weighted sum representation vectors
are then incorporated into task-specific architecture as same as traditional
token embeddings.

\paragraph{GPT models:}

The generative pre-training (GPT) \citep{radford2018improving} is
a language understanding training paradigm of a combination of two
training stages: unsupervised pre-training for training a language
model and supervised fine-tune on downstream tasks. The former stage
aims to produce universal contextualised word representation that
is transferable to a wide range of tasks in the latter stage with
little adaptation. Different from prior language models, the GPT models
(GPT-1 and GPT-2) use a multi-layer Transformer decoder \citep{vaswani2017attention}
for the language model to predict the next word given all the previous
words in a piece of text. The Transformer decoder, which mainly relies
on the self-attention mechanism, has been shown its great advantages
over recurrent neural networks such as LSTM in handling long-term
dependencies and the ease of paralleling process data. We will provide
more details of the Transformer later in Section \ref{subsec:chap3_Transformer}.
Being trained on large-scale textual datasets, specifically the BookCorpus
dataset \citep{zhu2015aligning} of more than 7,000 books across a
variety of genres for the GPT-1 and a corpus of 8 million web pages
for the GPT-2, the GPT models significantly outperform existing methods
on various tasks, including challenging tasks in zero-shot settings.

\paragraph{BERT:}

Bidirectional Encoder Representation from Transformers (BERT) \citep{devlin2018bert}
proposes to address some shortcomings of both the ELMo and GPT models.
The GPT models' main drawback is in their unidirectional design, where
it ignores the contextual information from the reverse direction (right-to-left).
The ELMo model, on the other hand, treats the backward and forward
pass in separation. The BERT offers a masked language model objective
in which it masks tokens in the given text on a random basis and asks
machines to predict the position of the masked input. By doing that,
BERT can learn the contextualised word representations of words by
combining the contextual information from both directions. Similar
to the GPT models, BERT is also powered by Transformer. However, it
makes use of a multi-layer Transformer encoder instead. BERT has demonstrated
its strong empirical success in which it achieves new state-of-the-art
performance on 11 NLP tasks.

\subsection{Feature Extraction\label{subsec:chap3_Feature-Extraction}}

The feature extraction module is to model the contextual information
of the question and the context separately. Much of neural MRC systems
utilise recurrent neural networks (RNNs) to extract sequential information
from the context and the question. Recent trends in NLP also reveal
the emerging of Transformer \citep{vaswani2017attention} as a popular
alternative. In this section, we will briefly describe these two standard
techniques.

\subsubsection{Recurrent Neural Networks\label{subsec:chap3_Recurrent-neural-networks}}

Long short-term memory (LSTM) \citep{hochreiter1997long} and Gated
Recurrent Units (GRU) \citep{cho2014learning} are two variants of
RNNs often used to model sequential dependencies between embedding
tokens in the given question and context. For better handling of long
sequences, it is a common choice that researchers use bidirectional
RNNs instead of a unidirectional one. In the case of neural MRC systems,
the extracted information for the question $q$ and the context $c$
is usually carried in the same way. In particular, we first tokenise
the question of length $S$ into a set of words and further embed
them into a vector space$\left\{ \mathbf{e}_{i}\right\} _{i=1}^{S}$,
$\mathbf{e}_{i}\in\mathbb{R}^{d}$ where $d$ is the embedding size.
We call word embeddings $\left\{ \mathbf{e}_{i}\right\} _{i=1}^{S}$
as context-free embeddings. We then compute a context-dependent representations
of words in the original question $\mathbf{L}=\left\{ \mathbf{w}_{i}^{q}\right\} _{i=1}^{S}$
where each question embedding $\mathbf{w}_{i}^{q}$ at time step $i$
is the concatenation of the hidden states of forward and backward
LSTM passes:\vspace{-1cm}

\begin{eqnarray}
\mathbf{w}_{i}^{q} & = & [\overrightarrow{\text{LSTM}}(\mathbf{e}_{i}^{q});\overleftarrow{\text{LSTM}}(\mathbf{e}_{i}^{q})],
\end{eqnarray}
where $\overrightarrow{\text{LSTM}}(\mathbf{e}_{i}^{q})$ and $\overleftarrow{\text{LSTM}}(\mathbf{e}_{i}^{q})$
denote the hidden states of the forward pass and the backward pass,
respectively. The operator $[\,;]$ denotes vector concatenation of
two vectors. In addition to the contextual hidden features, a joint
feature vector of the final states of the forward pass and backward
pass is used to represent the semantic representation of the whole
question\vspace{-1cm}

\begin{eqnarray}
\mathbf{q} & = & \left[\overleftarrow{\mathbf{w}_{1}^{q}};\overrightarrow{\mathbf{w}_{S}^{q}}\right].\label{eq:chap3_global_question_rep}
\end{eqnarray}
\vspace{-0.8cm}

We apply the same procedure to extract a set $\mathbf{C}$ of contextual
features $\mathbf{C}=\left\{ \mathbf{w}_{j}^{c}\right\} _{j=1}^{T}$
given the context $c$ of $T$ words.

\subsubsection{Transformer\label{subsec:chap3_Transformer}}

Transformer is proposed by \citet{vaswani2017attention}, originally
for machine translation tasks. Different from the RNN-based approaches
in modelling a sequence of text, Transformer relies entirely on self-attention
with parallel data processing capability. As a result, Transformer
is supposed to be more efficient than RNN-based approaches in terms
of computation.

\citet{vaswani2017attention} defines the attention function as a
mapping function between a query vector and a set of key-value pairs
to output a vector. Assuming that $\mathbf{Q},\mathbf{K},\mathbf{V}$
are matrices of stacked query vectors, stacked key vectors and stacked
value vectors, respectively, the attention weight matrix is given
by\vspace{-1cm}

\begin{eqnarray}
\text{Attention}(\mathbf{Q},\mathbf{K},\mathbf{V}) & = & \text{softmax}(\frac{\mathbf{Q}\mathbf{K}^{\top}}{\sqrt{d_{k}}})\mathbf{V},
\end{eqnarray}
where $d_{k}$ denotes the dimension of the queries and keys. A multi-head
attention mechanism that learns a joint representation from different
sets of $\mathbf{Q},\mathbf{K}$ and $\mathbf{V}$ is more effective
than using single head attention in practice. This is because the
multi-head attention mechanism allows models to learn the input components'
interactions from different subspaces rather than relying on a single
subspace.

A sequence encoder based on self-attention is a stack of multiple
multi-head attention followed by a fully connected feed-forward network.
QANet \citep{yu2018qanet} is an example of the MRC models that use
Transformer as a sequence encoder.

\subsection{Context-Question Interaction}

The context-question interaction module is to extract the correlation
between question and context to arrive at an answer. Much of the neural
MRC systems rely on attention mechanism to extract the correlation
between the context and the question. As mentioned in Sec. \ref{subsec:chap2_Reasoning-over-Set},
the attention mechanism is a technique for obtaining a fixed-size
representation of an arbitrary set of representations, dependent on
a query via selective summary. In this section, we discuss attention
mechanisms in the context of MRC systems. They are generally divided
into two categories: unidirectional attention and bidirectional attention.

\subsubsection{Unidirectional Attention}

Given a global vector representation of the question $\mathbf{q}$
as in Eq. \ref{eq:chap3_global_question_rep} and the extracted features
of context $\mathbf{C}$ as in Subsec. \ref{subsec:chap3_Recurrent-neural-networks},
the unidirectional attention mechanism is to determine parts in the
context $\mathbf{C}$ that are relevant to the question $\mathbf{q}$.
Formally, it computes a correlation score $s_{i}=f(\mathbf{q},\mathbf{w}_{j}^{c})$
between the question vector $\mathbf{q}$ and each contextual representation
$\mathbf{w}_{j}^{c}$ at time step $i$ of the context, followed by
a normalisation function, and eventually outputs an attention weight
$\alpha_{i}$ for each context word. The normalisation function is
usually the softmax function:\vspace{-1cm}

\begin{eqnarray}
\alpha_{i} & = & \frac{\text{exp}(\mathbf{W}s_{i})}{\sum_{j}\text{exp}(\mathbf{W}s_{j})},
\end{eqnarray}
where $\mathbf{W}\in\mathbb{R}^{1\times d}$. The attended context
vector is then calculated by\vspace{-1cm}

\begin{eqnarray}
\mathbf{i} & = & \sum_{i}\alpha_{i}\mathbf{w}_{i}^{c}.\label{eq:chap3_attended_vector}
\end{eqnarray}
The attended context feature is eventually combined with the question
representation $\mathbf{q}$ to predict an answer.

Different models have proposed different methods to compute the correlation
function $f\left(.\right)$. For example, \citet{hermann2015teaching}
computes alignment scores $s_{i}$ using a linear combination of the
question vector $\mathbf{q}$ and each vector $\mathbf{w}_{j}^{c}$
of the contextual representations:\vspace{-1cm}

\begin{eqnarray}
s_{i} & = & \text{tanh}(\mathbf{W}^{c}\mathbf{w}_{i}^{c}+\mathbf{W^{q}}\mathbf{q}),\label{eq:chap3_attention_tanh_func}
\end{eqnarray}
where $\mathbf{W^{c}}\in\text{\ensuremath{\mathbb{R}^{d\times d}}},\mathbf{W}^{q}\in\mathbb{R}^{d\times d}$
are learnable weights of neural network models. On the other hand,
\citet{chen2016thorough} makes use of a bilinear function using a
transformation matrix $\mathbf{W}^{s}\in\mathbb{R}^{d\times d}$ to
map the question vector into the vector space of the contextual representation:\vspace{-1cm}

\begin{eqnarray}
s_{i} & = & \mathbf{q}^{\top}\mathbf{W}^{s}\mathbf{w}_{i}^{c}.\label{eq:chap3_attention_bilinear_func}
\end{eqnarray}
Empirical results suggest that the bilinear function has the edge
over the additive form in Eq. \ref{eq:chap3_attention_tanh_func}.
However, both approaches are widely used in practice as the bilinear
function is more costly in computation than the other.

\subsubsection{Bidirectional Attention}

Although unidirectional attention mechanisms have significantly advanced
the neural MRC systems, it is beneficial to leverage selective summary
from both directions: query-to-context and context-to-query. \citep{xiong2016dynamic,seo2016bidirectional}
are representative studies in this line of work. In particular, \citep{seo2016bidirectional}
first computes a similarity score with function $f\left(.\right)$
between the question words $\mathbf{L}=\left\{ \mathbf{w}_{i}^{q}\right\} _{i=1}^{S}$
and the context words $\mathbf{C}=\left\{ \mathbf{w}_{j}^{c}\right\} _{j=1}^{T}$:\vspace{-1cm}

\begin{eqnarray}
s_{ij} & = & f\left(\mathbf{w}_{i}^{q},\mathbf{w}_{j}^{c}\right),
\end{eqnarray}
where $f\left(.\right)$ is a trainable function. The context-to-query
attention weights are then computed as the normalised output of the
softmax function across the row, and the query-to-context attention
weights are output across the column. Finally, the combination of
attended context vector and attended question vector is used for prediction.
The context-to-query based attended vector and query-to-context based
attended vector are both computed as similar as in Eq. \ref{eq:chap3_attended_vector}.

\subsection{Answer Decoder\label{subsec:chap3_Answer-Decoder}}

Depending on the tasks, different designs of an answer decoder are
required to deliver answer predictions from the context-question interactions.
There are three representative tasks in textual question answering:
open-ended answering, multi-choice answering and span extraction.

\subsubsection{Open-Ended Answering}

This task requires an MRC system to output a word or an entity name
in vocabulary as the answer. In this case, the answer decoder is usually
a composition of a few fully connect layers followed by a softmax
function to return the probabilities of words in the vocabulary chosen
for the answer.

\subsubsection{Multi-Choice Answering}

This task imitates the multi-choice test in which an MRC system is
asked to select one or multiple correct answers from a given set of
answer choices for each question. A common technique is to rank the
similarity between each answer choice and the question-context pair
and finally pick the one with the highest scores.

\subsubsection{Span Extraction}

The span extraction task requires machines to extract a subsequence
from the given context. \citet{wang2016machine} has designed an answer-pointer-based
decoder in which it predicts two indices of token in the context to
determine the boundary of the output subsequence. Thanks to its simplicity
and ease of implementation, the answer-pointer-based decoder has been
widely used in MRC models since invented.

\section{Visual and Language Reasoning\label{sec:chap3_Visual-and-Language-Reasoning}}

\begin{figure}
\begin{centering}
\medskip{}
\includegraphics[width=0.8\textwidth]{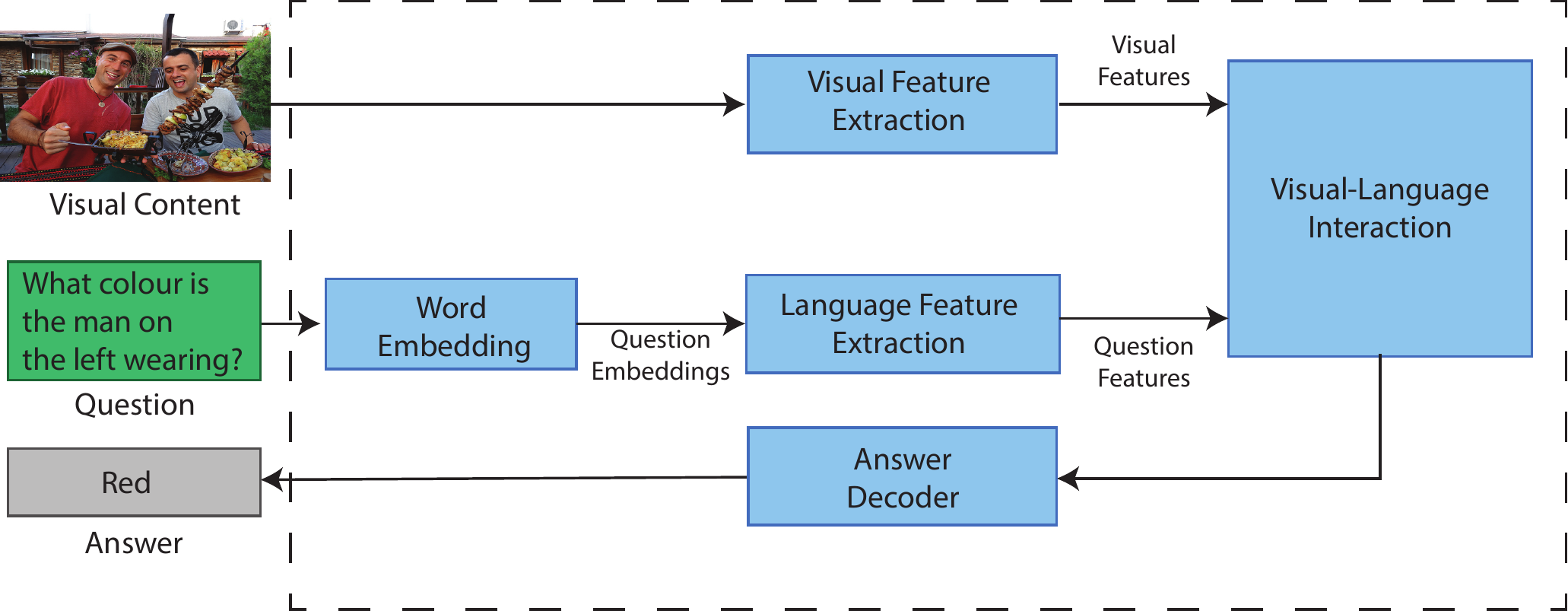}
\par\end{centering}
\caption{General architecture of a Visual Question Answering (VQA) system.
Image taken from GQA dataset \citep{hudson2019gqa}.\label{fig:chap3_General-VQA}}
\end{figure}

In the scope of this thesis, we focus on Visual Question Answering
(VQA) as a representative task of language and visual reasoning. VQA
is a task to assess whether a machine can understand and reason about
the visual content present in either a static image (Image QA) or
a video (Video QA) through a question answering process. Compared
to the MRC systems discussed in Section \ref{sec:chap3_Neural-Machine-Reading-Comp},
VQA can be seen as an extension of MRC where the context information
is present in the visual domain. In other words, VQA teaches machines
\emph{how to see}, while MCR teaches machines \emph{how to read}.\emph{
}Due to a multimodal problem by definition, solving VQA requires domain
knowledge from both modalities, linguistic and vision. Figure \ref{fig:chap3_General-VQA}
presents an example of VQA and the overall architecture of a VQA system.
As MRC tasks and VQA tasks are similar in terms of problem definition,
the high-level design of a VQA system is very similar to the design
of the MRC tasks in Section \ref{sec:chap3_Neural-Machine-Reading-Comp}.
It contains four key components: a visual feature extraction module,
a language feature extraction module, a visual-language interaction
module, and an answer decoder. Much of the existing VQA models use
a standard way to extract features from visual cues and linguistic
cues; however, they spend tremendous efforts to model the interaction
across domains. We will detail representative approaches in modelling
the visual-language interaction in VQA in Section \ref{sec:chap3_Neural-reasoning-for-ImageQA}
and Section \ref{sec:chap3_Neural-reasoning-for-VideoQA}.

\subsection{Visual Representation}

\subsubsection{Image Feature Representation for Image QA}

Image representation plays a critical role in computer vision systems.
Before the advent of deep neural networks with large scale datasets,
there is a significant amount of literature on image representation
methods such as SIFT \citep{lowe2004distinctive}, HOG \citep{dalal2005histograms}.
However, they are now replaced by high-level features extracted by
deep convolutional neural networks (CNNs). Early methods in Image
QA studies rely on the shallow VGG networks (VGG-16, VGG-19) \citep{simonyan2014very}
to extract visual features, while recent studies often use very deep
CNN architectures of hundreds of layers (e.g. ResNet models \citep{he2016deep}).
These CNN models are often pre-trained with large-scale image classification
datasets of thousands of classes, for example, ImageNet \citep{imagenet_cvpr09}.
Since the introduction of the bottom-up attention model \citep{anderson2018bottom},
object-centric representation of images with object proposals became
more popular as they support the reasoning process better. 

\subsubsection{Spatio-Temporal Video Representation for Video QA}

Video QA extends the problem of Image QA to a video which is a long
sequence of images. Hence, we need to represent visual information
across both space and time. Spatio-temporal video representation is
traditionally done by different variations of recurrent networks (RNNs),
of which many have been applied to the context of Video QA, such as
recurrent encoder-decoder \citep{zhu2017uncovering,zhao2019long},
bidirectional LSTM \citep{kim2017deepstory} and two-staged LSTM \citep{zeng2017leveraging}.
External memory can be added to these networks \citep{gao2018motion,zeng2017leveraging}
to increase the memorising ability of long-term dependencies in videos.
This technique is more beneficial for videos that are longer \citep{xu2016msr}
and with more complex structures such as movies \citep{tapaswi2016movieqa}
and TV programs \citep{lei2018tvqa} with extra accompanying channels
such as speech or subtitles. In these cases, memory networks \citep{kim2017deepstory,na2017read,wang2019holistic}
were used to store multimodal features \citep{wang2018movie} for
later retrieval. Memory augmented RNNs can also compress video into
heterogeneous sets \citep{fan2019heterogeneous} of dual appearance/motion
features. 

In RNNs-based video representation, appearance and motion are modelled
separately. Therefore, people also look for alternative approaches
to represent appearance and motion simultaneously. 3D and 2D/3D hybrid
convolutional operators \citep{tran2018closer,qiu2017learning} intrinsically
integrates spatio-temporal visual information, which greatly advances
the video understanding field. These are also used for Video QA in
recent studies \citep{jang2017tgif,li2019beyond}. Multi-scale temporal
structures can be modelled by either mixing short and long term convolutional
filters \citep{wu2019long} or combining pre-extracted frame features
non-local operators \citep{tang2018non,li2017temporal}. Within the
second approach, the TRN network \citep{zhou2018temporal} demonstrates
the role of temporal frame relations in human action recognition.
Relations of pre-detected objects were also considered in a separate
processing stream \citep{jin2019multi} and combined with other modalities
in late-fusion \citep{singh2019spatio}.

\subsection{Language Representation}

Similar to the language representation in the MRC systems explained
in Section \ref{sec:chap3_Neural-Machine-Reading-Comp}, much of VQA
works utilise RNNs to model the semantic meaning of questions. Question
words are first embedded into fixed-size vectors, often initialised
by pre-trained GloVe word embeddings \citep{pennington2014glove}.
Word embeddings are further fed into bidirectional RNNs, for example,
biLSTM, to output a set of contextual word embeddings and a global
representation of the whole sequence. Some recent works \citep{yang2020bert,garcia2020knowit}
also take advantage of the robust contextualised word embeddings learned
by pre-trained language models such as BERT or GPT for language representation
for VQA tasks.

\section{Neural-reasoning for Image QA\label{sec:chap3_Neural-reasoning-for-ImageQA}}

Assuming that we have vector representations of the visual information
and the question in an Image QA setting, early studies \citep{zhou2015simple,kafle2016answer}
adopt simple fusion operators, including vector concatenation, element-wise
addition, or element-wise multiplication, to synthesise a joint representation
of visual and language clues. Although these methods are simple and
straightforward, they have limitations in modelling the interaction
between the two input channels. This explains why they do not generalise
well in practice. A large number of studies address the problem of
cross-modality interaction in a more explicit way to improve performance.
In general, Image QA approaches can be classified into four categories:
attention-based methods, bilinear pooling methods, relation networks,
and neural-symbolic reasoning.

\subsection{Attention-based Methods\label{subsec:chap3_Attention-based-Methods}}

Similar to the MRC systems discussed in Sec. \ref{sec:chap3_Neural-Machine-Reading-Comp},
attention mechanisms are widely used in Image QA models. In the context
of Image QA, attention mechanism mostly means spatial attention to
obtain region-specific CNN features or selective terms in a given
question relevant to expected answers. In other words, Image QA models
have to identify which part of the image and which words in the question
are more important than the others. For example, considering a scene
of two men talking to each other as given in the example in Fig. \ref{fig:chap3_General-VQA},
to answer the question \emph{``what colour is the man on the left
wearing?''}, Image QA models need to pay attention to the \emph{man
on the left} instead of the \emph{man on the right} or any other irrelevant
image regions. Regarding the language part, the words \emph{``colour''},\emph{
``wearing''} and the phrase \emph{``man on the left''} contains
key information to find the correct answer. 

\citet{yang2016stacked} proposed a stacked attention network (SAN)
that repeatedly refines the representation of the query based on the
timely attention distribution over the image feature map. This method
achieved new state-of-the-art results on major Image QA datasets by
the time of publication, including COCO-QA \citep{ren2015exploring}
and VQA \citep{antol2015vqa}. Attention mechanisms also play a critical
role in many memory network based Image QA models. \citet{xu2016ask}
proposed a Spatial Memory Network which computes a spatial attention
distribution based on the correlations between image cells and words
in the question. Similarly, \citet{xiong2016dynamic} use a multi-hop
attention model to retrieve relevant information from visual input
facts to write into memory states. The attention weights are calculated
based on the interaction of both input modalities and the intermediate
memory states. \citet{lu2016hierarchical} introduced a co-attention
mechanism to generate attention on the image feature map and the question
simultaneously. In addition to the newly introduced attention mechanism,
they also used hierarchical encoding of the question at the word level,
the phrase level and the question level. Paying attention to different
semantic levels of the question makes the co-attention more meaningful,
hence, improves the overall performance of Image QA.

In the recent developments of Image QA, some models \citep{hudson2018compositional,hudson2019learning}
leverage attention mechanisms on the linguistic cues to generate a
chain of reasoning instructions to guide the reasoning process in
a multi-step manner. In particular, the design of MACNet \citep{hudson2018compositional}
mimics design principles of computer architectures with three interacting
components: a controller, a read unit and a write unit. These three
units seamlessly interact with each other in a recurrent design to
model the interactions between different question and input image
components.

\subsection{Bilinear Pooling Methods}

The primary goal of Image QA models is to learn a joint representation
of a given question its corresponding image. One of the straightforward
methods to obtain a joint representation of two modalities is via
a bilinear pooling operation \citep{fukui2016multimodal}. The bilinear
pooling operation is the outer product of two vectors. Let $\mathbf{x}=\left\{ x_{j}\right\} _{j=1}^{N},\mathbf{x}\in\mathbb{R}^{N}$
denotes a visual feature vector representing the image, $\mathbf{y}=\left\{ y_{k}\right\} _{k=1}^{M},\mathbf{y}\in\mathbb{R}^{M}$
denotes a question feature vector, a bilinear model uses a quadratic
expression across every pair of features from the two input modalities:\vspace{-1cm}

\begin{eqnarray}
z_{i} & = & \sum_{j=1}^{N}\sum_{k=1}^{M}w_{ijk}x_{j}y_{k}+b_{i}=\mathbf{x}^{\top}\mathbf{W}_{i}\mathbf{y}+b_{i},\label{eq:chap3_bilinear_pooling_op}
\end{eqnarray}
where $\mathbf{W}_{i}\in\mathbb{R}^{N\times M}$ is a weight matrix
and $b_{i}$ is a bias for the output $z_{i}$. In order to learn
a joint representation vector $\mathbf{z}=\left\{ z_{i}\right\} _{i=1}^{O}$,
where $O$ is the size of the output vector $\mathbf{z}$, from input
vector $\mathbf{x}$ and $\mathbf{y}$, we need to learn a series
of weight matrices $\mathbf{W}=\left\{ \mathbf{W}_{i}\right\} _{i=1}^{O}$.
Therefore, applying the bilinear pooling operation in Eq. \ref{eq:chap3_bilinear_pooling_op}
straightaway would be extremely computationally expensive in a high-dimensional
space. As a result, \citet{fukui2016multimodal} used the so-called
Count Sketch projection function \citep{charikar2002finding} to project
the input vectors $\text{x}$ and $\mathbf{y}$ into a lower-dimensional
space. The output represents the correlation between the question
and the input image and then serves as input for answer prediction.
Studies by \citet{fukui2016multimodal} also shown that incorporating
a spatial attention mechanism improved their bilinear pooling method
in performance. \citet{kim2016hadamard} introduced a multi-modal
low-rank bilinear pooling (MLB) stream to obtain an approximated bilinear
pooling by reducing the rank of the weight matrix $\mathbf{W}_{i}$,
making it more affordable for neural network training. \citet{kim2018bilinear}
further improved the low-rank bilinear pooling model with multi-head
attention, which yielded promising results on different real-world
Image QA datasets.

\subsection{Relation Networks for Visual Reasoning}

In Sec. \ref{sec:chap2_relational_reasoning}, we have mentioned learning
and reasoning with relation networks. Here, we discuss the application
of relation networks in Image QA setting specifically. \citet{santoro2017simple}
relies on a universal pairwise relation network to model the relationship
between every pair of image regions conditioning on the context information
given by the question:\vspace{-1cm}

\begin{eqnarray}
RN(\mathbf{O}) & = & f_{\bm{\phi}}\left(\sum_{i,j}g_{\bm{\theta}}\left(\mathbf{o}_{i},\mathbf{o}_{j},\mathbf{q}\right)\right),
\end{eqnarray}
where $\mathbf{o}_{i},\mathbf{o}_{j}$ denote visual spatial feature
vectors of $i$-th and $j$-th image regions in a set $\mathbf{O}$
of visual regions derived by standard convolution neural networks,
and $\mathbf{q}$ denotes the question vector representation. $f_{\bm{\phi}}\left(.\right),g_{\bm{\theta}}\left(.\right)$
are sub-networks with parameters $\bm{\phi}$ and $\bm{\theta}$,
respectively. The output of the conditional relation network is eventually
used for answer prediction. Although the design of these relation
networks for machine reasoning is relatively simple, its concept is
generic, hence, applicable to a variety of tasks. Specifically, \citet{santoro2017simple}
evaluated their proposed method on Image QA and Text QA in which they
have shown interesting findings. Relation networks are also successfully
applied to other computer vision tasks such as object detection \citep{hu2018relation},
action recognition \citep{zhou2018temporal}, and in other graph-based
methods for visual reasoning \citep{chen2018iterative,li2019relation}.

\subsection{Neural-symbolic Reasoning}

In visual reasoning, particularly in question answering settings,
questions often require interpretation through multiple steps to arrive
at a proper answer. Recall the example shown in Section \ref{subsec:chap3_Attention-based-Methods}\emph{
``what colour is the man on the left wearing?''} A human being would
first look for the \emph{``man on the left''}, then look at his
clothes \emph{(wearing)}, and finally identify the \emph{``colour''}
of his clothes. Therefore, to deal with these types of questions,
Image QA models need to understand the compositionality of the question
and accordingly localise the relevant information in the visual cues
to arrive at a valid answer. Other than the attention-based compositional
approach explained above, there is family of works called Neural Module
Networks (NMN) \citep{andreas2016neural}, which addresses compositional
visual reasoning based on structures of language constituents and
co-attention. Different from the end-to-end attention-based approach,
the NMN sits on the bridge between neural networks for representation
learning and symbolic program execution for reasoning.

\citet{andreas2016neural} defined a series of network modules where
each of them handles a sub-task. These modules can be assembled into
a meaningful layout to derive a proper answer given a structured representation
of the question. They leveraged an external language parser \citep{klein2003accurate}
for textual analysis, obtaining a function of question constituents
for each question. For example, a question like \emph{``what colour
is the vase?''} will be translated to \emph{describe{[}colour{]}(find{[}vase{]})}.
These functions are subsequently mapped to the predefined modules
to produce the module layouts. \citet{hu2017learning} improved the
NMN by introducing a layout predictor to learn the module layouts
automatically instead of manually mapping. Similarly, \citet{yi2018neural}
also used reinforcement learning to learn the layout predictor but
with less training data. Later, \citet{hu2018explainable} proposed
Stack-NMN to solve limitations of prior models in terms of relying
on policy gradient to train the layout predictor, making it the first
end-to-end differentiable NMN model. In one of the latest efforts,
\citet{vedantam2019probabilistic} approached NMN from a probabilistic
perspective, making the generated programs more interpretable while
improving the system's robustness with fewer training samples.

\section{Neural-reasoning for Video QA\label{sec:chap3_Neural-reasoning-for-VideoQA}}

Compared to Image QA, there are fewer studies on Video QA due to the
increasing complexity and challenges posed by the video representation.
Beyond information retrieval, answering questions on videos is, by
nature, a spatio-temporal reasoning problem. Not to mention that collecting
and annotating large-scale Video QA requires more effort than creating
Image QA datasets. Hence, there are a limited number of datasets to
benchmark Video QA models. As an extension of Image QA to handle temporal
dynamics, early works in Video QA tried to adapt techniques initially
developed for Image QA to Video QA. 

\citet{kim2017deepstory} and \citet{zeng2017leveraging} utilised
memory networks as a platform to retrieve the information in the video
features related to the question embedding. More recent Video QA methods
started interleaving simple reasoning mechanisms into the pattern
matching network operations. \citet{jang2017tgif} calculated the
attention weights on the video LSTM features queried by the question.
This attention-based reasoning mechanism aims to identify the spatio-temporal
regions relevant to the question, but it does not support deducing
new information based on the data provided. In an effort toward deeper
reasoning, \citet{gao2018motion} proposed to parse the two-stream
video features through a dynamic co-memory module which iteratively
refines the episodic memory. Lately, \citet{li2019beyond} used self-attention
mechanisms to internally contemplate about video and question first,
then put them through a co-attention block to match the information
contained in the two sources of data. For complex structured videos
with multimodal features such as in movies \citep{tapaswi2016movieqa}
and TV programs \citep{lei2018tvqa}, recent methods leveraged memories
\citep{fan2019heterogeneous,kim2017deepstory,na2017read} to store
multimodal features into episodic memory for later retrieval of related
information to the question. More sophisticated reasoning mechanisms
are also developed with hierarchical attention \citep{liang2018focal},
multi-head attention \citep{kim2018multimodal} or multi-step progressive
attention memory \citep{kim2019progressive} to jointly reason on
video/audio/text concurrent signals.

\section{Closing Remarks}

In this chapter, we have introduced key concepts in visual and language
reasoning. We also provided a comprehensive review of existing neural
reasoning methods with applications on different problems in machine
reasoning, including machine reading comprehension, image question
answering and video question answering. The following chapter will
detail the first contribution of this thesis on resembling the concept
of the dual process, which is originally from the human reasoning
systems, to visual reasoning with application on video question answering.

\newpage{}

\selectlanguage{american}%

\chapter{Dual System of Visual Reasoning\label{chap:DualProcess}}

\section{Introduction\label{sec:Chap4-Introduction}}

\selectlanguage{australian}%
A long standing goal in AI is to design a learning machine capable
of reasoning about the dynamic scenes it sees. A powerful demonstration
of such a capability is answering unseen natural questions about a
video. Recall that reasoning is the mental faculty to produce new
knowledge from the previously acquired knowledge base in response
to a query \citep{bottou2014machine}. Video QA systems must be able
to learn, acquire and manipulate visual knowledge distributed through
space-time conditioned on the compositional linguistic cues. Recent
successes in Image QA \citep{anderson2018bottom,hu2017learning,hudson2019learning,yi2018neural}
pave possible roads, but Video QA is largely under-explored \citep{song2018explore,li2019beyond}.
Compared to static images, video data poses new challenges, primarily
due to the inherent dynamic nature of visual content over time \citep{gao2018motion,wang2018movie}.
At the lowest level, we have correlated motion and appearance \citep{gao2018motion}.
At higher levels, we have objects that are persistent over time, actions
that are local in time, and relations that can span over an extended
length.

Searching for an answer from a video facilitates solving interwoven
sub-tasks in both the visual and lingual spaces, probably in an iterative
and compositional fashion. In the visual space, the sub-tasks at each
step involve extracting and attending to objects, actions, and relations
in time and space. In the textual space, the tasks involve extracting
and attending to concepts in the context of sentence semantics. A
plausible approach to Video QA is to prepare video content to accommodate
the retrieval of information specified in the question \citep{jang2017tgif,kim2017deepstory,zeng2017leveraging}.
However, this has not yet offered a more complex reasoning capability
that involves multi-step inference and handling of compositionality.
More recent works have attempted to add limited reasoning capability
into the system through memory and attention mechanisms that are tightly
entangled with visual representation \citep{gao2018motion,li2019beyond}.
These systems are thus non-modular and less comprehensible as a result.

Our approach to Video QA in this chapter is to disentangle the processes
of visual pattern recognition\emph{ }and\emph{ }compositional reasoning
\citep{yi2018neural}. This division of labour realises a \emph{dual
process} cognitive view that the two processes are qualitatively different:
visual cognition can be reactive and domain-specific, but reasoning
is usually deliberative and domain-general \citep{evans2008dual,kahneman2011thinking}.
In our system, pattern recognition precedes and makes its output accessible
to the reasoning process. More specifically, we derive a hierarchical
model over time at the visual understanding level, dubbed Clip-based
Relational Network (ClipRN), that can accommodate objects, actions,
and relations in space-time. This is followed by a generic differentiable
reasoning module, known as Memory-Attention-Composition network (MACNet)
\citep{hudson2018compositional}, which iteratively manipulates a
set of objects in a knowledge base given a set of cues in the query,
one step at a time. In our setting, the MACNet takes prepared visual
content as the knowledge base, and iteratively co-attends to the textual
concepts and the visual concepts/relations to extract clues for the
answer. The overall dual-process system is modular and fully differentiable,
making it easy to compose modules and train.

\begin{figure*}
\begin{centering}
\includegraphics[width=1\textwidth]{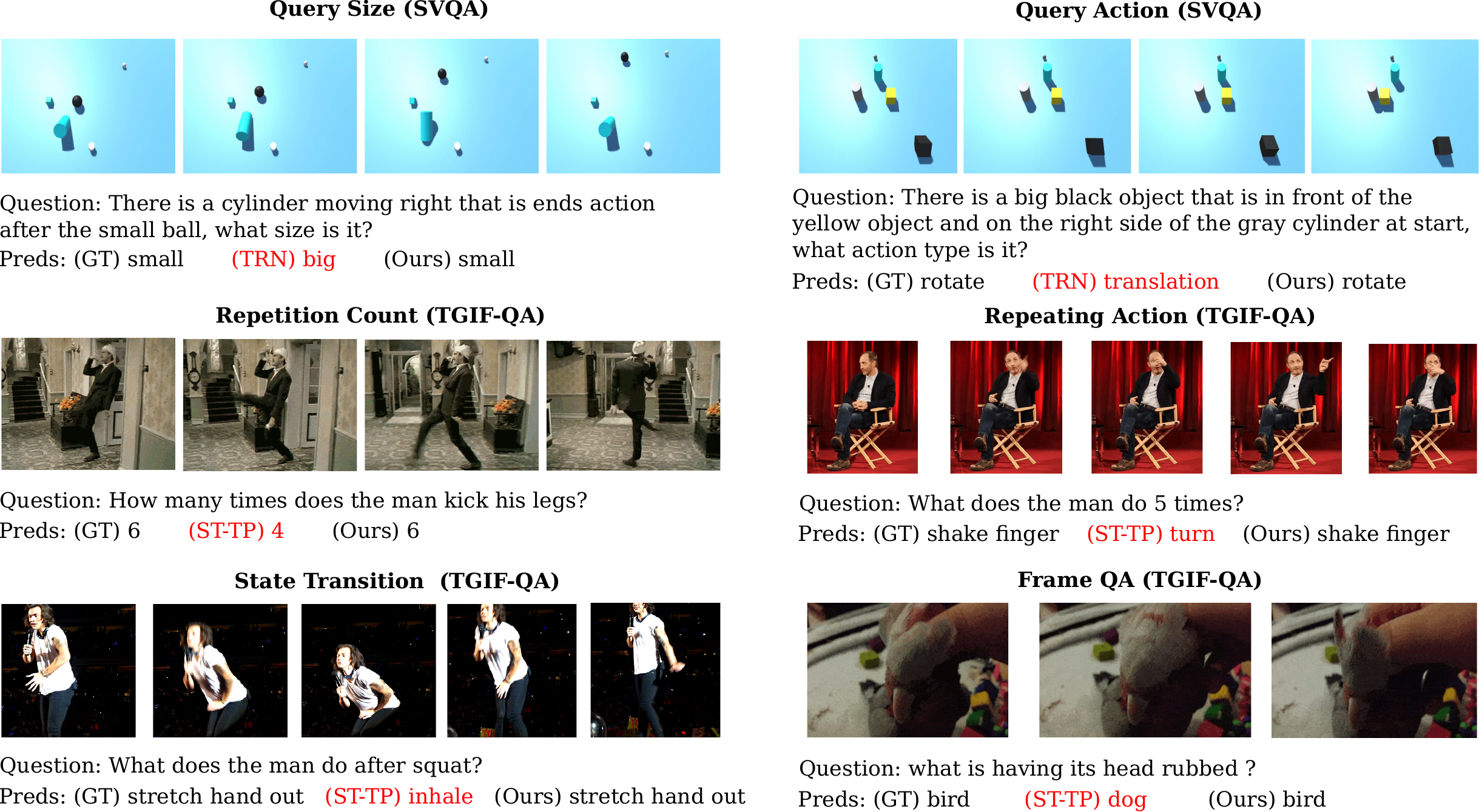}
\par\end{centering}
\caption{Examples of SVQA and TGIF-QA dataset. GT: ground truth; TRN: our baseline
utilising TRN \citep{zhou2018temporal}; ST-TP: method introduced
by \citet{jang2017tgif}. Best viewed in colour. \label{fig:chap4_qualcomp}}
\end{figure*}

We validate our dual process model on two large public datasets, TGIF-QA
\citep{jang2017tgif} and SVQA \citep{song2018explore}. The TGIF-QA
is a real dataset, and is relatively well-studied \citep{gao2018motion,jang2017tgif,li2019beyond}.
See Fig.~\ref{fig:chap4_qualcomp}, the last two rows for example
frames and question types. The SVQA is a new synthetic dataset designed
to mitigate the inherent linguistic biases in the real datasets and
promote multi-step reasoning \citep{song2018explore}. Several cases
of complex, multi-part questions are shown in Fig.~\ref{fig:chap4_qualcomp},
first row. The proposed model (ClipRN+MACNet) achieves new records
on both datasets, and the margins on the SVQA are qualitatively significant
from the best known results. Some example responses are displayed
in Fig.~\ref{fig:chap4_qualcomp}, demonstrating how our proposed
method works in different scenarios.

Our contributions in this chapter are: (1) Introducing a modular neural
architecture for learning to reason in Video QA. The system implements
dual process theory by disentangling reactive visual cognition and
language understanding from deliberative reasoning. (2) Proposing
a hierarchical model for preparing video representation taking into
account of query-driven frame selectivity within a clip and temporal
relations between clips. 

\selectlanguage{american}%

\section{Background \label{sec:Chap4-Background}}

\selectlanguage{australian}%
We provide an extensive background on vision and language reasoning,
in addition to a brief background on Video QA in Chapter \ref{chap:VisualLanguageReasoning}.
Here, we present more background on video representation and the dual
process systems that are closely related to our proposed method in
this chapter.

\paragraph*{Video representation in Video QA}

Available methods for Video QA typically relied on recurrent networks
or 3D convolutions to extract video features. Variations of LSTM were
used in \citep{kim2017deepstory} with a bidirectional LSTM and in
\citep{zeng2017leveraging} in the form of a two-staged LSTM. Likewise,
\citet{gao2018motion} and \citet{li2019beyond} used two levels of
GRU, of which one is in ``facts'' extraction, and the other one
is in each iteration of the memory based reasoning. In another direction,
convolutional networks were used to integrate visual information with
either 2D or 3D kernels \citep{jang2017tgif,li2019beyond}.

Different from these two traditional trends, in this chapter, we propose
ClipRN, a query-driven hierarchical relational feature extraction
strategy, which supports strong modelling for both near-term and far-term
spatio-temporal relations. The ClipRN supports multiple levels of
granularity in the temporal scale. This development is necessary to
address the non-deterministic queries in Video QA tasks.

\paragraph*{Dual process systems}

Reasoning systems that exhibit behaviours consistent with dual process
theories are typically neural-symbolic hybrids (e.g., see \citep{garcez2019neural}
for an overview). In \citep{yi2018neural}, visual pattern recognition
modules form elements of a symbolic program whose execution would
find answers for image question answering. Different from \citep{yi2018neural},
we rely on implicit reasoning capability in a fully differentiable
neural system \citep{bottou2014machine,hudson2018compositional}.\selectlanguage{american}%

\section{Method \label{sec:Chap4-Method}}

\selectlanguage{australian}%
\begin{figure*}
\centering{}\includegraphics[width=1\textwidth]{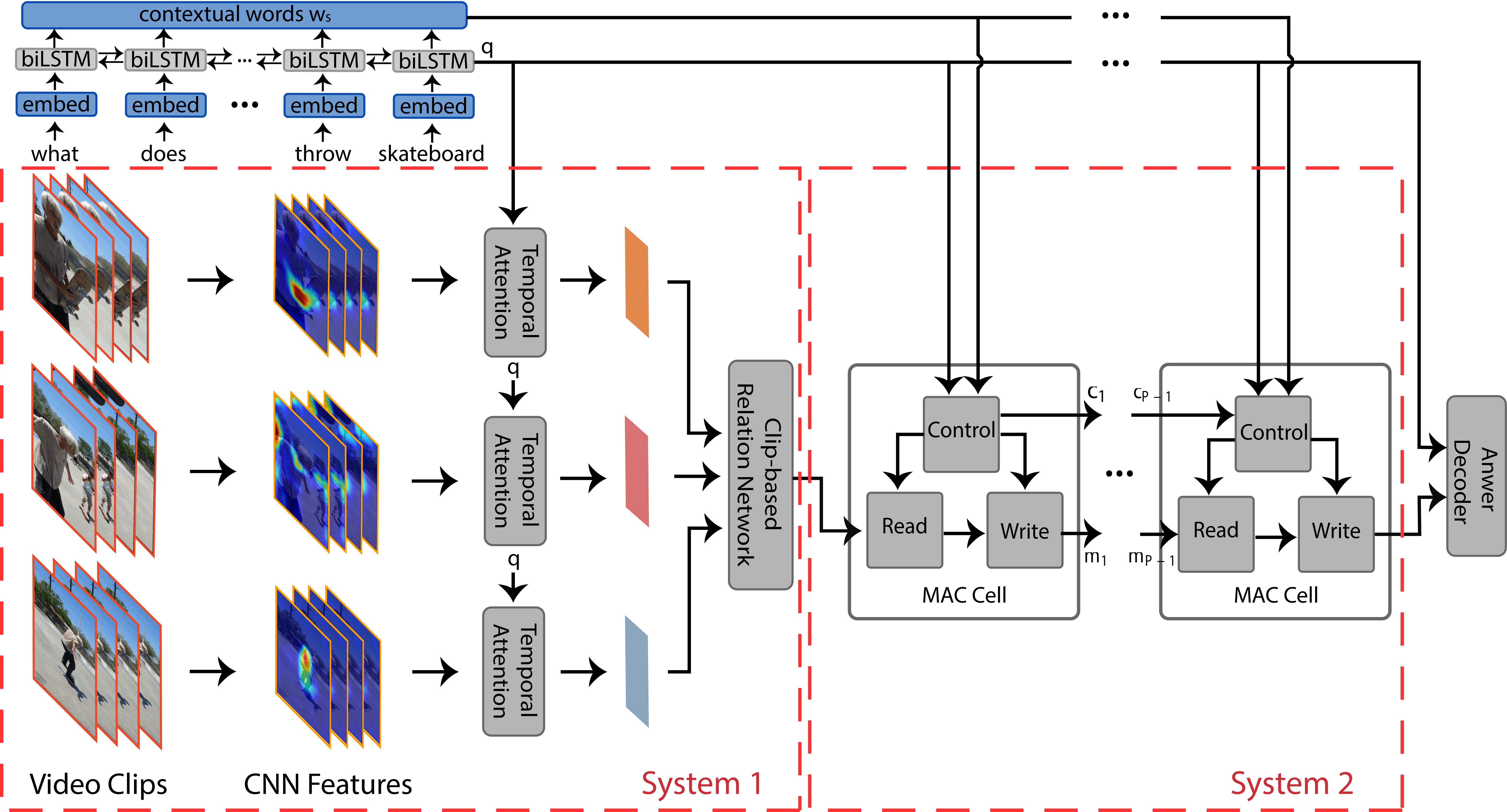}\caption{Overview of network architecture for Video QA. The model is viewed
as a dual process system of hierarchical video representation with
Clip-based Relation Network (ClipRN) and high-level multi-step reasoning
with MAC cells, in which textual cues guide the computation of both
processes. Inputs of ClipRN are the aggregated features of equal-size
clips obtained by a temporal attention mechanism. The high-level reasoning
module iteratively co-attends to the contextual words of a given question
and the visual concepts/relations prepared by the ClipRN unit to extract
relevant visual information to the answer. At the end of the network,
an answer decoder, taking as input the joint representation of the
question feature and the retrieved visual information, is used for
prediction. \label{fig:chap4_architecture}}
\end{figure*}

A Video QA system is a mapping function that returns an answer $\hat{y}$
in a pre-defined answer space $\mathcal{A}$ from the information
deduced from a given video $\mathcal{V}$ in response to a natural
question $\mathbf{q}$. Formally, the Video QA problem can be formulated
as follows:\vspace{-1cm}

\begin{eqnarray}
\hat{y} & = & \underset{a\in\mathcal{A}}{\text{argmax}}\mathcal{F}_{\bm{\theta}}\left(a;\mathbf{q},\mathcal{\mathcal{\boldsymbol{V}}}\right),\label{eq:chap4_prob_def}
\end{eqnarray}
where $\theta$ is the model parameters of scoring function $\mathcal{F}\left(.\right)$.
In this section, we present our main contribution in this chapter
to addressing the challenges posed in Video QA. In particular, we
propose a modular end-to-end neural architecture for the scoring function
$\mathcal{F}\left(.\right)$, as illustrated in Fig.~\ref{fig:chap4_architecture}. 

\subsection{Dual Process System View}

Our architecture is partly inspired by the \emph{dual process theory}
dictating that there are two loosely coupled cognitive processes serving
separate purposes in reasoning: the lower pattern recognition that
tends to be associative, and the higher-order reasoning faculty that
tends to be deliberative \citep{evans2008dual,kahneman2011thinking}.
Translated into our Video QA scenarios, we have the pattern recognition
process for extracting visual features, representing objects and relations,
and making the representation accessible to the higher reasoning process.
The interesting and challenging aspects come from two sources. First,
video spans over both space and time, hence calling for methods to
deal with object persistence, action span and repetition, and long-range
relations. Second, Video QA aims to respond to the textual query;
hence the two processes of a Video QA system, including the video
representation and reasoning, should be conditional and driven by
the textual cues.

For language coding, we make use of the standard practice as what
explained in Section \ref{sec:chap3_Visual-and-Language-Reasoning}.
In particular, we model the semantic meaning of the language cues
using biLSTM, taking as input question words initialised with GloVe
word embeddings \citep{pennington2014glove}. Given a length-$S$
question, we represent it with two sets of linguistic clues: contextual
words $\left\{ \mathbf{w}_{s}|\mathbf{w}_{s}\in\mathbb{R}^{d}\right\} _{s=1}^{S}$
which are the joint output states of the biLSTM at each time step
in both directions, and the question vector $\mathbf{q}=[\overleftarrow{\mathbf{w}_{1}};\overrightarrow{\mathbf{w}_{S}}],\mathbf{q}\in\mathbb{R}^{d}$
which is the joint representation of the final hidden states from
forward and backward LSTM passes.

We consider a video as a composition of video clips, in which each
clip can be viewed as an event/activity. While previous studies have
explored the significance of the hierarchical representation of video
\citep{zhu2018end}, we hypothesise that it is vital to model the
far-term relationships between clips. Inspired by \citep{santoro2017simple}
and recent work \citep{zhou2018temporal} on action recognition, known
as Temporal Relation Network (TRN), we propose a Clip-based Relation
Network (ClipRN) for video representation, where a clip feature is
a query-dependent summary of its video frames. Ideally, the ClipRN
is more effective in modelling a temporal sequence than the simplistic
TRN, which comes with only a sparse number of sampled frames. The
ClipRN represents the video as a tensor serving as a knowledge base
for a generic reasoning engine to later stages.

The reasoning process, due to its deliberative nature, involves multiple
steps in an iterative fashion. We utilise Memory-Attention-Composition
(MAC) cells \citep{hudson2018compositional} for the task due to its
generality and modularity. More specially, the MAC cells are called
repeatedly conditioned on the textual cues to manipulate information
from given video representations as a knowledge base. Finally, the
information prepared by MACNet, combined with the textual cues, is
presented to a decoder to produce an answer.

In short, our system consists of three components where the outputs
of one component are the inputs to another: (1) temporal relational
pattern recognition, (2) multi-step reasoning with MAC cells and (3)
answer decoders. We detail these components in what follows.

\subsection{Temporal Relational Pattern Recognition \label{subsec:chap4_ClipRN}}

\begin{figure}
\centering{}\includegraphics[width=0.8\columnwidth]{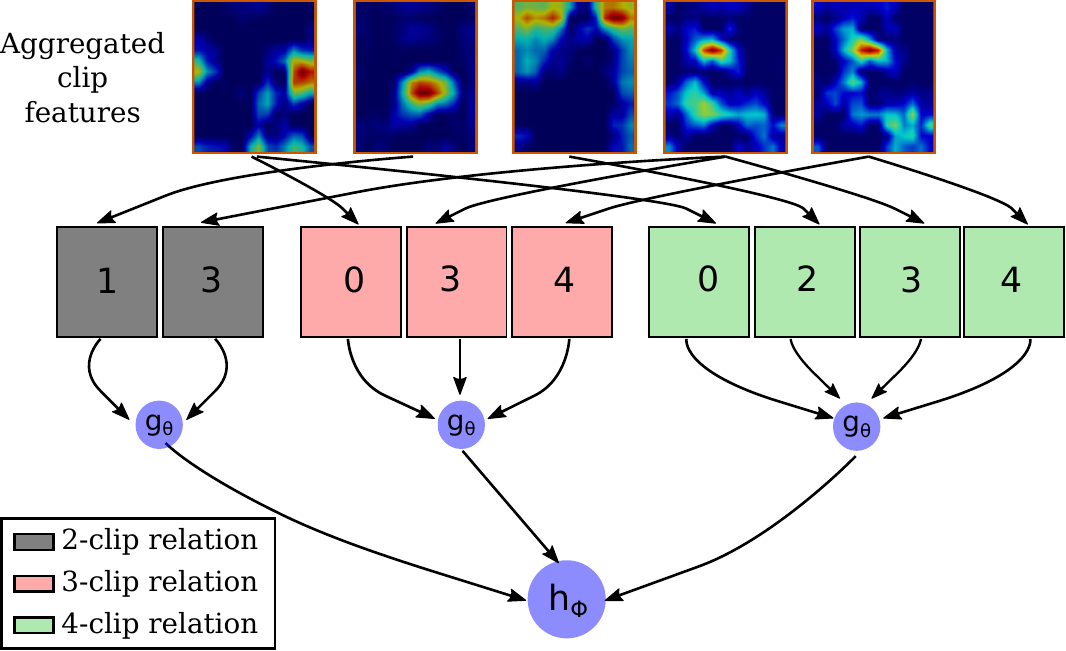}\caption{Illustration of Clip-based Relation Network (ClipRN). Aggregated features
of equal size clips are fed into $k$-clip relation modules. Inputs
to the relation modules are selected on a random basis whilst keeping
their temporal order unchanged. In this illustration, our ClipRN represents
a video as aggregated features of five video clips using 2-clip relation,
3-clip relation, and 4-clip relation modules. This results in an output
feature having the same dimensions as the input features.\label{fig:chapter_4_cliprn_illustration}}
\end{figure}
Given a video of continuous frames, we begin with dividing the video
into $N$ equal-length clips $\mathbf{C}=\left\{ \mathbf{C}_{i}\right\} _{i=1}^{N}$.
Each clip $\mathbf{C}_{i}$, which contains $T$ frames, is represented
by $\mathbf{C}_{i}=\{\mathbf{V}_{i,t}\mid\mathbf{V}_{i,t}\in\mathbb{R}^{1024\times W\times H}\}_{t=1}^{T}$,
where $\mathbf{V}_{i,t}$ is frame feature maps (tensor) of the $t$-th
frame in clip $\mathbf{C}_{i}$ extracted by ResNet-101 \citep{he2016deep};
$W,H$ are width and height of the extracted feature maps. Frame-level
features are subsequently projected to a $d$ dimensional space via
linear transformations, resulting in transformed representations for
each clip $\mathbf{C}_{i}$ as $\{\mathbf{V}_{i,t}^{\prime}\mid\mathbf{V}_{i,t}^{\prime}\in\mathbb{R}^{d\times W\times H}\}_{t=1}^{T}$.

As consecutive frames contain redundant or irrelevant information
to the question, it is crucial to attend to frames selectively. We
thus utilise a temporal attention mechanism conditioned on the question
vector $\mathbf{q}$ to compute a clip feature $\hat{\mathbf{C}}_{i}$
of the corresponding clip $\mathbf{C}_{i}$ as a weighted sum of its
video frames:\vspace{-1cm}

\begin{eqnarray}
\mathbf{v}_{i,t}^{\text{pool}} & = & \frac{1}{W\cdot H}\sum_{w=1}^{W}\sum_{h=1}^{H}\mathbf{V}_{i,t,w,h}^{\prime};\mathbf{v}_{i,t}^{\text{pool}}\in\mathbb{R}^{d},\\
\mathbf{s}_{i,t} & = & \mathbf{W}\left(\left(\mathbf{W}^{q}\mathbf{q}+b\right)\odot\left(\mathbf{W}^{v}\mathbf{v}_{i,t}^{\text{pool}}+\mathbf{b}\right)\right),\\
\hat{\mathbf{C}_{i}} & = & \sum_{t=1}^{T}\mathbf{V}_{i,t}^{\prime}\cdot\textrm{softmax}(\mathbf{s}_{i,t}),
\end{eqnarray}
where, $\mathbf{W},\mathbf{W}^{q},\mathbf{W}^{v}$and $\mathbf{b}$
are learnable weights, and $\odot$ is the element-wise multiplication.

To account for relations between clips, we borrow the strategy of
TRN described in \citep{zhou2018temporal} which adapts and generalises
the proposal by \citet{santoro2017simple} to the temporal domain.
Different from \citet{zhou2018temporal}, our relational network operates
at the clip level instead of frame level. More specifically, the $k$-order
relational representation of video is given as\vspace{-1cm}

\begin{eqnarray}
R^{(k)}\left(\mathbf{C}\right) & = & h_{\bm{\phi}}\left(\sum_{i_{1}<i_{2}...<i_{k}}g{}_{\bm{\theta}}\left(\hat{\mathbf{C}}_{i_{1}},\hat{\mathbf{C}}_{i_{2}},...,\hat{\mathbf{C}}_{i_{k}}\right)\right),\label{eq:chap4_k-level-relation}
\end{eqnarray}
for $k=2,3,..,K$, where $h_{\bm{\phi}}$ and $g_{\bm{\theta}}$ are
any aggregation function with parameters $\bm{\phi}$ and $\bm{\theta}$,
respectively. We term this resultant model as \emph{Clip-based Relation
Network} (ClipRN). Fig.~\ref{fig:chapter_4_cliprn_illustration}
illustrates our procedure for our ClipRN.

\paragraph{Remark}

The ClipRN subsumes the TRN as a special case when $T\rightarrow1$.
However, by computing the relations at the clip level, we can better
model the hierarchical structure of videos and avoid computational
complexity inherent in TRN. For example, we neither need to apply
sparse sampling of frames nor use the multi-resolution trick as in
TRN. Considering a long and complex video sequence, there is a high
chance of having pairs of distantly related frames with the TRN, hence
their relations are less important than those of near-term frames.
In the worst case scenario, those pairs can become noise to the feature
representation and mislead the reasoning process in later stages.
In contrast, our clips representation with ClipRN not only can preserve
such near-term relations but also guarantee the far-term relations
between short snippets of a video.

\subsection{Multi-step Centralised Reasoning}

Higher-order reasoning on the rich relational temporal patterns is
the key to reliably answering questions. Our approach is to disentangle
the slow, deliberative reasoning steps out of fast, automatic feature
extraction and temporal relation modelling. This ``slow-thinking''
reasoning is done with a dedicated module that repeatedly distils
and purifies the key relational information contained in the ClipRN
features.

In our experiments, we use the MACNet \citep{hudson2018compositional}
as an option for the reasoning module. At the core of MACNet are the
recurrent cells called \emph{control units}, collaborating with \emph{read
units} and \emph{write units} to iteratively make reasoning operations
on a knowledge base using a sequence of clues extracted from the query.
Compared to mixed-up feature extraction/reasoning mechanisms, the
control units give the MACNet distinctive features of a centralised
reasoning module that can make a well-informed decision on attention
and memory reads/writes. MACNet is also powered by the flexible retrieving/processing/reference
mechanism while processing the query and looking up in the knowledge
base. These characteristics are well suited to explore the rich, condensed
relational information in ClipRN features. The iterative reasoning
process of MACNet supports a level of error self-correcting ability
that also helps deal with the possible remaining redundancy and distraction.

In our setup, the knowledge base $\mathbf{B}\in\mathbb{R}^{d\times X\times Y}$
used in MACNet is gathered from the ClipRN features from all available
orders:
\begin{eqnarray}
\mathbf{B} & = & \sum_{k=2}^{K}R^{(k)}\left(\mathbf{C}\right),
\end{eqnarray}
where, $R^{(k)}\left(\mathbf{C}\right)$ are the $k$-order ClipRN
representations calculated as in Eq.~(\ref{eq:chap4_k-level-relation}).

For each reasoning step $r$, the relevant aspects of question to
this step are estimated from vector $\mathbf{q}\in\mathbb{R}^{d}$:\vspace{-1cm}

\begin{eqnarray}
\mathbf{q}_{r} & = & \mathbf{W}_{r}^{q}\mathbf{q}+\mathbf{b}_{r},
\end{eqnarray}
where, $\mathbf{W}_{r}^{q}$ and $\mathbf{b}_{r}$ are network weights.

Let $\left[\,;\right]$ denote the concatenation operator of two tensors.
Based on the pair of clues contextual words and step-aware question
vector ($\{\mathbf{w}_{s}\}_{s=1}^{S},\mathbf{q}_{r})$, recall that
$S$ is a given question's length, and the control state of the previous
reasoning step $\mathbf{c}_{r-1}\in\mathbb{R}^{d}$, the control unit
calculates a soft self-attention weight $\alpha_{r,s}^{\text{control}}$
over words in the question:
\begin{eqnarray}
\mathbf{c}_{r}^{\prime} & = & \mathbf{W}_{r}[\mathbf{W}_{r}^{c}\mathbf{c}_{r-1};\mathbf{q}_{r}],\\
\alpha_{r,s}^{\text{control}} & = & \textrm{softmax}(\mathbf{W}_{r}^{\alpha}(\mathbf{c}_{r}^{\prime}\odot\mathbf{w}_{s})+\mathbf{b}),
\end{eqnarray}
and infers the control state $\mathbf{c}_{r}\in\mathbb{R}^{d}$ at
current reasoning step $r$:
\begin{eqnarray}
\mathbf{c}_{r} & = & \sum_{s=1}^{S}\alpha_{r,s}^{\text{control}}\mathbf{w}_{s}.
\end{eqnarray}

The read unit uses this control signal and the prior memory $\mathbf{m}_{r-1}\in\mathbb{R}^{d}$
to calculate the read attention weights $\alpha_{r,x,y}^{\text{read}}$
for each location $x,y$ in the knowledge base $\mathbf{B}\in\mathbb{R}^{d\times X\times Y}$
and retrieves the related information:
\begin{eqnarray}
\mathbf{r}_{r} & = & \sum_{x,y}\alpha_{r,x,y}^{\text{read}}\mathbf{B}_{x,y},
\end{eqnarray}
where,\vspace{-1cm}

\begin{eqnarray}
\mathbf{i}_{r,x,y} & = & \left[\mathbf{m}_{r-1}\odot\mathbf{B}_{x,y};\mathbf{B}_{x,y}\right],\label{eq:chap4_mem_kb_interaction}\\
\mathbf{i}_{r,x,y}^{\prime} & = & \boldsymbol{W}_{r}\mathbf{i}_{r,x,y},\\
\alpha_{r,x,y}^{\text{read}} & = & \textrm{softmax}(\mathbf{W}_{r}^{\alpha}(\mathbf{c}_{r}\odot\mathbf{i}_{r,x,y}^{\prime})+\mathbf{b}).
\end{eqnarray}
The vector $\mathbf{i}_{r,x,y}$ in Eq. \ref{eq:chap4_mem_kb_interaction}
denotes the interaction between the previous memory state $\mathbf{m}_{r-1}$
and the knowledge base cell at $(x,y)$.

To finish each reasoning iteration, the write unit calculates the
intermediate reasoning result $\mathbf{m}_{r}$ by updating the previous
record $\mathbf{m}_{r-1}$ with the new information derived from the
retrieved knowledge $\mathbf{r}_{r}$. The memory state update is
done by a function $f\left(.\right)$: $\mathbf{m}_{r}=f\left(\mathbf{m}_{r-1},\mathbf{r}_{r}\right);\:\mathbf{m}_{r}\in\mathbb{R}^{d}$.
In our experiments, the function $f\left(.\right)$ is simply a linear
transformation on top of a vector concatenation operator.

At the end of the iterative process with $P$ steps ($P$ MAC cells),
the final memory state $\mathbf{m}_{P}$ emerges as the output of
the reasoning module that is eventually used by the answer decoders
for answer prediction.

\subsection{Answer Decoders\label{subsec:chap4_Answer-Decoders}}

Similar to prior works \citep{jang2017tgif,song2018explore}, we adopt
different answer decoders depending on the tasks. These include open-ended
QA and multi-choice QA.

For open-ended QA (e.g. those in Frame QA in TGIF-QA dataset and all
QA pairs in SVQA dataset \textendash{} see Section~\ref{subsec:chap4_Datasets}),
we treat them as multi-class classification problems of $\mid\mathcal{A}\mid$
labels defined in an answer space $\mathcal{A}$. We employ a classifier
that composes 2-fully connected layers, followed by the softmax function
to predict the probabilities of possible answers in the answer space
$\mathcal{A}$. The classifier takes as input the combination of the
memory state $\mathbf{m}_{P}$ and the question representation $\mathbf{q}$:\vspace{-1cm}

\begin{eqnarray}
\mathbf{p} & = & \textrm{softmax}(\mathbf{W}^{o2}\left(\mathbf{W}^{o1}\left[\mathbf{m}_{P};\mathbf{W}^{q}\mathbf{q}+\mathbf{b}\right]\right)+\mathbf{b}),
\end{eqnarray}
where, $\mathbf{p}\in\mathbb{R^{\mathcal{\mid\mathcal{A}\mid}}}$
. The cross-entropy loss is used as the loss function of the network
in this case.

Similarly, we use a linear regression function to predict real-value
numbers (repetition count) directly from the joint representation
of $\mathbf{m}_{P}$ and $\mathbf{q}$. We further pass the regression
output through a rounding function for prediction:\vspace{-1cm}

\begin{eqnarray}
s & = & \lfloor\mathbf{W}^{o2}\left(\mathbf{W}^{o1}\left[\mathbf{m}_{P};\mathbf{W}^{q}\mathbf{q}+\mathbf{b}\right]\right)+\mathbf{b}\rfloor,
\end{eqnarray}
where $\lfloor.\rfloor$ is the standard rounding function. Mean Squared
Error (MSE) is used as the loss function during the training process
in this case.

Regarding the multi-choice question type, we treat each answer candidate
of a short sentence in the same way as we process the question. In
particular, we reuse one MACNet for both the question and answer candidates
in which network parameters are shared. As a result, there are two
types of memory output, one derived by the question $\mathbf{m}_{q,P}$,
and the other one by the answer candidates $\mathbf{m}_{a,P}$. Inputs
to a classifier are from four sources, including $\mathbf{m}_{q,P}$,
$\mathbf{m}_{a,P}$ , question representation $\mathbf{q}$ and answer
candidates $\mathbf{a}$:\vspace{-1cm}

\begin{eqnarray}
\mathbf{y} & = & \left[\mathbf{m}_{q,P};\mathbf{m}_{a,P};\mathbf{W}^{q}q+b;\mathbf{W}^{a}\mathbf{a}+\mathbf{b}\right],\\
\mathbf{y}^{\prime} & = & \sigma(\mathbf{W}^{y}\mathbf{y}+\mathbf{b});\sigma=\text{ELU(.)}.
\end{eqnarray}
Finally, a linear regression is used to compute scores for respective
answer candidates:\vspace{-1cm}

\begin{eqnarray}
s & = & \mathbf{W}^{y^{\prime}}\mathbf{y}^{\prime},
\end{eqnarray}
where $\mathbf{W}^{y^{\prime}}\in\mathbb{R}^{1\times d}$. The model,
in this case, is trained with hinge loss of pairwise comparisons,
$\text{max}\left(0,1+s^{n}-s^{p}\right)$, between scores for incorrect
$s^{n}$ and correct answers $s^{p}$.\selectlanguage{american}%

\section{Experiments \label{sec:Chap4-Experiments}}

\selectlanguage{australian}%

\subsection{Datasets \label{subsec:chap4_Datasets}}

We evaluate our proposed method on two recent public datasets: Synthetic
Video Question Answering (SVQA) \citep{song2018explore} and TGIF-QA
\citep{jang2017tgif}.

\paragraph{SVQA}

This dataset is a recent benchmark for multi-step reasoning. Resembling
the CLEVR dataset \citep{johnson2017clevr} for traditional visual
question answering task, SVQA provides long questions with logical
structures along with spatial and temporal interactions between objects.
The SVQA is designed to mitigate several drawbacks of current Video
QA datasets, including language biases and the shortcoming of compositional
structures in questions. It contains 120K QA pairs generated from
12K videos covering various question types such as count, exist, object
attributes comparison and query.

\paragraph{TGIF-QA}

This is currently the largest dataset for Video QA, containing 165K
QA pairs collected from 72K animated GIFs. This dataset covers four
sub-tasks, mainly to address the unique properties of video, including
repetition count, repeating action, state transition and frame QA.
Of the four tasks, the first three require strong spatio-temporal
reasoning abilities. \textbf{\emph{Repetition Count}}: This is one
of the most challenging tasks in Video QA where machines are asked
to count the repetitions of an action. For example, one has to answer
a question like \emph{``how many times does the woman shake hips?''}
This is defined as an open-ended task with 11 possible answers in
total ranging from 0 to 10+. \textbf{\emph{Repeating Action}}: This
is a multiple choice task asking machines to choose one correct answer
out of five answer candidates per question. The task is to identify
a repeated action for a given number of times in the video (e.g. \emph{``what
does the dog do 4 times?''}). \textbf{\emph{State Transition}}: This
is also a multiple choice task asking machines to perceive the transition
between two states/events. There are certain states characterised
in the dataset, including facial expressions, actions, places and
object properties. Questions like \emph{``what does the woman do
before turning to the right side?''} and\emph{ ``what does the woman
do after looking to the left side?''} aim at identifying the previous
state and the next state, respectively. \textbf{\emph{Frame QA}}:
This task is akin to the traditional visual QA where the answer to
a question can be found in some video's frames. None of the temporal
relations is necessary to answer questions.

\subsection{Implementation Details \label{subsec:Chap4-Implementation-Details}}

Each video is segmented into five equal clips, each of which has eight
consecutive frames. The middle frame of each clip is determined based
on the length of the video. We take the \emph{conv4 }output features
of ResNet-101 \citep{he2016deep} pre-trained on ImageNet as the visual
features of each video frame. Each frame feature has dimensions of
$14\times14\times1024$. Words in a question and answer candidates
(if present) are embedded into vectors of 300 dimensions and initialised
by pre-trained GloVe embeddings \citep{pennington2014glove}. Unless
otherwise stated, we use $P=12$ MAC cells for multi-step reasoning
in our network, similar to what described in \citep{hudson2018compositional}.
All hidden state sizes are set to $512$ for both CRN and MAC cells.

Our network is trained using Adam, with a learning rate of $5\times10^{-5}$
for repetition count and $10^{-4}$ for other tasks, and with a batch
size of 16. The SVQA is split into three parts with proportions of
70-10-20\% for training, cross-validation, and testing set, accordingly.
As for the TGIF-QA dataset, we take 10\% of training videos in each
sub-task as the validation sets. Reported results are at the epochs
giving the best of performance on the corresponding validation sets.

\paragraph{Evaluation Metrics}

For the TGIF-QA, to be consistent with prior works \citep{jang2017tgif,gao2018motion,li2019beyond},
we use accuracy as the evaluation metric for all tasks except the
repetition count task, whose evaluation metric is Mean Square Error
(MSE). For the SVQA, we report accuracy for all sub-tasks.

\subsection{Results \label{subsec:Chap4-Results}}

\subsubsection{Selecting Model Parameters}

We conduct experiments to justify the choices of model parameters
for our method on subsets of the SVQA dataset. Results are shown in
Table~\ref{tab:chap4_ablSVQA}. As we train the network with only
a single GPU, we take up to 40 frames for each video. We tried experiments
with 10 segmented clips, 4 frames for each clip, but it invoked a
protracted training process compared to one with a smaller number
of clips. This is due to the increasing number of \emph{k}-order relations
in Eq.~\ref{eq:chap4_k-level-relation}. Hence, we chose to segment
each video into 5 clips to balance between the performance and training
time. To demonstrate the effect of the number of sampled frames on
the overall performance, we conduct a study with two settings, one
with 30 sampled frames and the other one with 40 frames, on subsets
of SVQA dataset consisting of 20,000 question/answer pairs for training,
and 3,000 pairs for cross-validation. All experiments on these subsets
are terminated after 15 epochs.

Experimental results show that more frames lead to slightly higher
overall performance. However, there is a trade-off of training time
caused by the longer processing time for reading external files. Last
but not least, we have noticed that applying mean pooling over spatial
dimension when calculating temporal attention in each clip produces
favourable performance over max pooling.

\begin{table}
\centering{}{\footnotesize{}}%
\begin{tabular}{c|c}
\hline 
\multirow{2}{*}{{\small{}Model}} & \multirow{2}{*}{{\small{}Val. Acc.}}\tabularnewline
 & \tabularnewline
\hline 
\hline 
{\small{}CRN 5 clips$\times$6 frames+mean pooling} & {\small{}55.9}\tabularnewline
\hline 
{\small{}CRN 5 clips$\times$8 frames+mean pooling} & {\small{}56.1}\tabularnewline
\hline 
{\small{}CRN 5 clips$\times$8 frames+max pooling} & {\small{}54.9}\tabularnewline
\hline 
\end{tabular}\medskip{}
{\small{}\caption{Parameter selection with experiments on SVQA subsets. \label{tab:chap4_ablSVQA}}
}
\end{table}

\begin{table}
\centering{}{\footnotesize{}}%
\begin{tabular}{l|c|cccc}
\hline 
\multirow{2}{*}{{\footnotesize{}Model}} & \multirow{2}{*}{{\footnotesize{}SVQA}{\small{}$\uparrow$}} & \multicolumn{4}{c}{{\footnotesize{}TGIF-QA ({*})}}\tabularnewline
\cline{3-6} \cline{4-6} \cline{5-6} \cline{6-6} 
 &  & {\footnotesize{}Action}{\small{}$\uparrow$} & {\footnotesize{}Trans.}{\small{}$\uparrow$} & {\footnotesize{}Frame}{\small{}$\uparrow$} & {\footnotesize{}Count}{\small{}$\downarrow$}\tabularnewline
\hline 
\hline 
{\footnotesize{}Linguistic only} & {\footnotesize{}42.6} & {\footnotesize{}51.5} & {\footnotesize{}52.8} & {\footnotesize{}46.0} & {\footnotesize{}4.77}\tabularnewline
{\footnotesize{}Ling.+S.Frame} & {\footnotesize{}44.6} & {\footnotesize{}51.3} & {\footnotesize{}53.4} & {\footnotesize{}50.4} & {\footnotesize{}4.63}\tabularnewline
{\footnotesize{}S.Frame+MACNet} & {\footnotesize{}58.1} & {\footnotesize{}67.8} & {\footnotesize{}76.1} & {\footnotesize{}57.1} & {\footnotesize{}4.41}\tabularnewline
{\footnotesize{}Avg.Pool+MACNet} & {\footnotesize{}67.4} & {\footnotesize{}70.1} & {\footnotesize{}77.7} & {\footnotesize{}58.0} & {\footnotesize{}4.31}\tabularnewline
{\footnotesize{}TRN+MACNet} & {\footnotesize{}70.8} & {\footnotesize{}69.0} & {\footnotesize{}78.4} & {\footnotesize{}58.7} & {\footnotesize{}4.33}\tabularnewline
{\footnotesize{}ClipRN+MLP} & {\footnotesize{}49.3} & {\footnotesize{}51.5} & {\footnotesize{}53.0} & {\footnotesize{}53.5} & {\footnotesize{}4.53}\tabularnewline
\hline 
\textbf{\footnotesize{}ClipRN+MACNet} & \textbf{\footnotesize{}75.8} & \textbf{\footnotesize{}71.3} & \textbf{\footnotesize{}78.7} & \textbf{\footnotesize{}59.2} & \textbf{\footnotesize{}4.23}\tabularnewline
\hline 
\end{tabular}{\small{}\medskip{}
}\caption{Ablation studies on SVQA dataset and TGIF-QA dataset, test split.
Accuracy is used as the evaluation metric in most cases, except the
count task in the TGIF where MSE is used. \label{tab:chap4-ablation1}}
\end{table}

\subsubsection{Ablation Studies}

\begin{table}
\centering{}{\small{}}%
\begin{tabular}{c|cccc}
\hline 
\multirow{2}{*}{{\footnotesize{}Reasoning iterations}} & \multicolumn{4}{c}{{\footnotesize{}TGIF-QA}}\tabularnewline
\cline{2-5} \cline{3-5} \cline{4-5} \cline{5-5} 
 & {\footnotesize{}Action}{\small{}$\uparrow$} & {\footnotesize{}Trans.}{\small{}$\uparrow$} & {\footnotesize{}Frame}{\small{}$\uparrow$} & {\footnotesize{}Count}{\small{}$\downarrow$}\tabularnewline
\hline 
\hline 
{\footnotesize{}4} & {\footnotesize{}69.9} & {\footnotesize{}77.6} & {\footnotesize{}58.5} & {\footnotesize{}4.30}\tabularnewline
{\footnotesize{}8} & {\footnotesize{}70.8} & \textbf{\footnotesize{}78.8} & {\footnotesize{}58.6} & {\footnotesize{}4.29}\tabularnewline
{\footnotesize{}12} & \textbf{\footnotesize{}71.3} & {\footnotesize{}78.7} & \textbf{\footnotesize{}59.2} & \textbf{\footnotesize{}4.23}\tabularnewline
\hline 
\end{tabular}{\small{}\medskip{}
}\caption{Ablation studies on test split with different numbers of reasoning
iterations. Accuracy is used as the evaluation metric in most cases,
except the count task in the TGIF where MSE is used.\label{tab:chap4-ablation2}}
\end{table}

To demonstrate the effectiveness of each component on the overall
performance of the proposed network, we first conduct a series of
ablation studies on both the SVQA and TGIF-QA datasets. The ablation
results are presented in Table~\ref{tab:chap4-ablation1}, \ref{tab:chap4-ablation2}
showing progressive improvements, which justify the added complexity.
We explain below the baselines.

\textbf{Linguistic only:} With this baseline, we aim to assess how
much linguistic information affects overall performance. From Table~\ref{tab:chap4-ablation1},
it is clear that the TGIF-QA dataset greatly suffers from language
biases, while the problem is mitigated with SVQA dataset to some extent.

\textbf{Ling.+S.Frame:} This is a basic model of VQA that combines
the encoded question vector with CNN features of a random frame taken
from a given video. As expected, this baseline offers modest improvements
over the model using only linguistic features.

\textbf{S.Frame+MACNet:} To demonstrate the significance of multi-step
reasoning in Video QA, we randomly select one video frame and then
use its CNN feature maps as the knowledge base of MACNet. As the SVQA
dataset contains questions with compositional sequences, it greatly
benefits from performing reasoning in a multi-step fashion.

\textbf{Avg.Pool+MACNet:} A baseline to assess the significance of
temporal information in the simplest form of average temporal pooling
comparing to ones using a single frame. We follow \citet{zhou2018temporal}
to sparely sample 8 frames which are the middle frames of equal size
segments from a given video. As can be seen, this model achieves significant
improvements compared to the previous baselines on both datasets.

\textbf{TRN+MACNet:} This baseline is a special case of our method
where we flatten the model's hierarchical design. In addition, the
temporal relation network operates on the frame level rather than
on the clip level, similar to what was proposed by \citet{zhou2018temporal}.
The model mitigates the limit of the feature engineering process for
video representations of a single frame and simply temporal average
pooling. Using a single frame loses crucial temporal information of
the video and mostly fails when strong temporal reasoning capability
plays a vital role, particularly in state transition and counting.
We use visual features processed in the Avg.Pool+MACNet experiment
to feed into a TRN module for fair comparisons. TRN improves by more
than 12\% of overall performance from the one using a single video
frame on the SVQA, while the improvement for state transition task
of the TGIF-QA is more than 2\%, around 1.5\% for both repeating action
and frame QA, and 0.08 MSE in case of repetition count. Although this
baseline produces significant increments on the SVQA comparing to
the experiment Avg.Pool+MACNet, the improvement on the TGIF-QA is
minimal.

\textbf{ClipRN+MLP: }In order to evaluate how the reasoning module
affects the overall performance, we conduct an experiment where we
use a simple MLP as the reasoning module with the proposed visual
representation ClipRN. 

\textbf{ClipRN+MAC:} This is our proposed method where the output
of ClipRN is used as a knowledge base for MACNet. We witness significant
improvements on all sub-tasks in the SVQA over the simplistic TRN,
whilst results on the TGIF-QA dataset are less noticeable. The results
reveal the strong spatio-temporal representation capacity for reasoning
of our ClipRN over the TRN, especially in the case of compositional
reasoning. The results also prove our earlier argument that sparsely
sampled frames from the video are insufficient to embrace fast-paced
actions/events such as repeating action and repetition count.

\begin{table*}
\centering{}%
\begin{tabular}{c|c|>{\centering}p{2cm}>{\centering}p{2cm}>{\centering}p{2cm}>{\centering}p{2cm}}
\hline 
\multicolumn{2}{c|}{{\small{}Model}} & {\small{}SA(S)} \citep{song2018explore} & {\small{}TA-GRU(T)} \citep{song2018explore} & {\small{}SA+TA-GRU} \citep{song2018explore} & \textbf{\small{}ClipRN+ MACNet}\tabularnewline
\hline 
\hline 
\multicolumn{2}{c|}{{\small{}Exist}} & {\small{}51.7} & {\small{}54.6} & {\small{}52.0} & \textbf{\small{}72.8}\tabularnewline
\hline 
\multicolumn{2}{c|}{{\small{}Count}} & {\small{}36.3} & {\small{}36.6} & {\small{}38.2} & \textbf{\small{}56.7}\tabularnewline
\hline 
\multirow{3}{*}{{\small{}Integer Comparison}} & {\small{}More} & {\small{}72.7} & {\small{}73.0} & {\small{}74.3} & \textbf{\small{}84.5}\tabularnewline
\cline{2-6} \cline{3-6} \cline{4-6} \cline{5-6} \cline{6-6} 
 & {\small{}Equal} & {\small{}54.8} & {\small{}57.3} & {\small{}57.7} & \textbf{\small{}71.7}\tabularnewline
\cline{2-6} \cline{3-6} \cline{4-6} \cline{5-6} \cline{6-6} 
 & {\small{}Less} & {\small{}58.6} & {\small{}57.7} & {\small{}61.6} & \textbf{\small{}75.9}\tabularnewline
\hline 
\multirow{5}{*}{{\small{}Attribute Comparison}} & {\small{}Color} & {\small{}52.2} & {\small{}53.8} & {\small{}56.0} & \textbf{\small{}70.5}\tabularnewline
\cline{2-6} \cline{3-6} \cline{4-6} \cline{5-6} \cline{6-6} 
 & {\small{}Size} & {\small{}53.6} & {\small{}53.4} & {\small{}55.9} & \textbf{\small{}76.2}\tabularnewline
\cline{2-6} \cline{3-6} \cline{4-6} \cline{5-6} \cline{6-6} 
 & {\small{}Type} & {\small{}52.7} & {\small{}54.8} & {\small{}53.4} & \textbf{\small{}90.7}\tabularnewline
\cline{2-6} \cline{3-6} \cline{4-6} \cline{5-6} \cline{6-6} 
 & {\small{}Dir} & {\small{}53.0} & {\small{}55.1} & {\small{}57.5} & \textbf{\small{}75.9}\tabularnewline
\cline{2-6} \cline{3-6} \cline{4-6} \cline{5-6} \cline{6-6} 
 & {\small{}Shape} & {\small{}52.3} & {\small{}52.4} & {\small{}53.0} & \textbf{\small{}57.2}\tabularnewline
\hline 
\multirow{5}{*}{{\small{}Query}} & {\small{}Color} & {\small{}29.0} & {\small{}22.0} & {\small{}23.4} & \textbf{\small{}76.1}\tabularnewline
\cline{2-6} \cline{3-6} \cline{4-6} \cline{5-6} \cline{6-6} 
 & {\small{}Size} & {\small{}54.0} & {\small{}54.8} & {\small{}63.3} & \textbf{\small{}92.8}\tabularnewline
\cline{2-6} \cline{3-6} \cline{4-6} \cline{5-6} \cline{6-6} 
 & {\small{}Type} & {\small{}55.7} & {\small{}55.5} & {\small{}62.9} & \textbf{\small{}91.0}\tabularnewline
\cline{2-6} \cline{3-6} \cline{4-6} \cline{5-6} \cline{6-6} 
 & {\small{}Dir} & {\small{}38.1} & {\small{}41.7} & {\small{}43.2} & \textbf{\small{}87.4}\tabularnewline
\cline{2-6} \cline{3-6} \cline{4-6} \cline{5-6} \cline{6-6} 
 & {\small{}Shape} & {\small{}46.3} & {\small{}42.9} & {\small{}41.7} & \textbf{\small{}85.4}\tabularnewline
\hline 
\multicolumn{2}{c|}{{\small{}Overall}} & {\small{}43.1} & {\small{}44.2} & {\small{}44.9} & \textbf{\small{}75.8}\tabularnewline
\hline 
\end{tabular}{\small{}\medskip{}
}\caption{Comparison against the state-of-the-art methods on SVQA. Reported
numbers are accuracy on the test split. \label{tab:chap4-svqa}}
\end{table*}

\subsubsection{Benchmarking against SOTAs}

We also compare our proposed model with other state-of-the-art methods
on both datasets, as shown in Table~\ref{tab:chap4-svqa} (SVQA)
and Table~\ref{tab:chap4-tgif} (TGIF-QA). As the TGIF-QA is older,
much effort has been spent on benchmarking it, and significant progress
has been made in recent years. The SVQA is new, and hence published
results are not very indicative of the latest advance in modelling.

For the SVQA, Table~\ref{tab:chap4-ablation1} and Table~\ref{tab:chap4-svqa}
reveal that the contributions of visual information to the overall
performance of the best known results are minimal. This means their
system is largely suffered from the linguistic bias of the dataset
for the decision making process. In contrast, our proposed methods
do not seem to suffer from this issue. We establish new qualitatively
different SOTAs on all sub-tasks and a massive jump from 44.9\% accuracy
to 75.8\% accuracy overall. 

\begin{table}
\centering{}%
\begin{tabular}{c|>{\centering}p{1.7cm}>{\centering}p{1.7cm}>{\centering}p{1.7cm}>{\centering}p{1.7cm}}
\hline 
{\small{}Model} & {\small{}Action (\%)$\uparrow$} & {\small{}Trans. (\%)$\uparrow$} & {\small{}Frame (\%)$\uparrow$} & {\small{}Count (MSE)$\downarrow$}\tabularnewline
\hline 
\hline 
{\small{}VIS+LSTM (aggr) \citep{ren2015exploring}} & {\small{}46.8} & {\small{}56.9} & {\small{}34.6} & {\small{}5.09}\tabularnewline
{\small{}VIS+LSTM (avg) \citep{ren2015exploring}} & {\small{}48.8} & {\small{}34.8} & {\small{}35.0} & {\small{}4.80}\tabularnewline
{\small{}VQA-MCB (aggr) \citep{fukui2016multimodal}} & {\small{}58.9} & {\small{}24.3} & {\small{}25.7} & {\small{}5.17}\tabularnewline
{\small{}VQA-MCB (avg) \citep{fukui2016multimodal}} & {\small{}29.1} & {\small{}33.0} & {\small{}15.5} & {\small{}5.54}\tabularnewline
{\small{}Yu et al. \citep{yu2017end}} & {\small{}56.1} & {\small{}64.0} & {\small{}39.6} & {\small{}5.13}\tabularnewline
{\small{}ST(R+C) \citep{jang2017tgif}} & {\small{}60.1} & {\small{}65.7} & {\small{}48.2} & {\small{}4.38}\tabularnewline
{\small{}ST-SP(R+C) \citep{jang2017tgif}} & {\small{}57.3} & {\small{}63.7} & {\small{}45.5} & {\small{}4.28}\tabularnewline
{\small{}ST-SP-TP(R+C) \citep{jang2017tgif}} & {\small{}57.0} & {\small{}59.6} & {\small{}47.8} & {\small{}4.56}\tabularnewline
{\small{}ST-TP(R+C) \citep{jang2017tgif}} & {\small{}60.8} & {\small{}67.1} & {\small{}49.3} & {\small{}4.40}\tabularnewline
{\small{}ST-TP(R+F) \citep{jang2017tgif}} & {\small{}62.9} & {\small{}69.4} & {\small{}49.5} & {\small{}4.32}\tabularnewline
{\small{}Co-memory(R+F) \citep{gao2018motion}} & {\small{}68.2} & {\small{}74.3} & {\small{}51.5} & {\small{}4.10}\tabularnewline
{\small{}PSAC(R) \citep{li2019beyond}} & {\small{}70.4} & {\small{}76.9} & {\small{}55.7} & {\small{}4.27}\tabularnewline
{\small{}HME(R+C) \citep{fan2019heterogeneous}} & \textbf{\small{}73.9} & {\small{}77.8} & {\small{}53.8} & \textbf{\small{}4.02}\tabularnewline
\hline 
\textbf{\small{}ClipRN+MACNet(R)} & {\small{}71.3} & \textbf{\small{}78.7} & \textbf{\small{}59.2} & {\small{}4.23}\tabularnewline
\hline 
\end{tabular}{\small{}\medskip{}
}\caption{Comparison with the SOTA methods on TGIF-QA. For count, the lower
the better. R: ResNet, C: C3D features, F: optical-flow features.
\label{tab:chap4-tgif}}
\end{table}

For the TGIF-QA dataset, \citet{jang2017tgif} extended winner models
of the VQA 2016 challenge to evaluate on Video QA task, namely VIS+LSTM
\citep{ren2015exploring} and VQA-MCB \citep{fukui2016multimodal}.
Early fusion and late fusion are applied to both two approaches. We
also list here some other methods provided by \citet{jang2017tgif},
including those proposed in \citep{fukui2016multimodal} and \citep{yu2017end}.
Interestingly, none of the previous works reported ablation studies
utilising only textual cues as the input to assess the linguistic
bias of the dataset and the fact that some of the reported methods
produced worse performance than this baseline. We suspect that the
improper integrating of visual information confused the systems and
resulted in such low performance. In Table~\ref{tab:chap4-tgif},
SP indicates spatial attention, ST presents temporal attention while
``R'', ``C'' and ``F'' indicate ResNet, C3D and optical-flow
features, respectively. Later, \citet{gao2018motion} greatly advanced
performance on this dataset with a co-memory mechanism on two video
feature streams. \citet{li2019beyond} recently achieved respected
performance on TGIF-QA with only ResNet features using a novel self-attention
mechanism. Our method, which also relies only on ResNet features,
achieves new state-of-the-art performance on the state transition
task and the frame QA task with a considerable margin compared to
prior works on the frame QA task. It appears that methods using both
appearance features (RestNet features) and motion features (C3D or
optical-flow features) perform poorly on the frame QA task, suggesting
the need for an adaptive feature selection mechanism. For action and
counting tasks, although we achieve competitive performance with prior
works \citep{gao2018motion,fan2019heterogeneous}, it is not directly
comparable since they utilised motion in addition to appearance features.
On the other hand, our method models the temporal relationships without
motion features; thus, the action boundaries are not clearly detected.
We hypothesise that counting task needs a specific network, as evident
in recent work \citep{levy2015live,trott2017interpretable}.

\subsubsection{Qualitative Results}

Fig.~\ref{fig:chap4_qualcomp} shows example frames and associated
question types in the TGIF-QA and SVQA datasets. The figure also presents
corresponding responses by our proposed method, and those by ST-TP
\citep{jang2017tgif} (on the TGIF-QA) and TRN+MACNet (a special case
of our proposed method, on the SVQA) for reference. The questions
clearly demonstrate challenges that video QA systems must face such
as visual ambiguity, subtlety, compositional language understanding,
and concepts grounding. The questions in the SVQA were designed for
multi-step reasoning, and the dual process system of ClipRN+MACNet
proves to be effective in these cases. We provide more analysis with
attention maps produced by our method on the SVQA dataset in \ref{sec:Appendix_A_dual_process}.\selectlanguage{american}%

\section{Closing Remarks\label{sec:Chap4-Discussion}}

\selectlanguage{australian}%
In this chapter, we have proposed a new differentiable architecture
for learning to reason in video question answering. The architecture
is founded on the premise that Video QA tasks necessitate a conditional
dual process of associative video cognition and deliberative multi-step
reasoning, given textual cues. The two processes are ordered in that
the former process prepares query-specific representation of video
to support the latter reasoning process. With that in mind, we designed
a hierarchical relational model for query-guided video representation
named Clip-based Relational Network (ClipRN) and integrated it with
a generic neural reasoning module (MACNet) to infer an answer. The
system is fully differentiable, hence amenable to end-to-end training.
Compared to existing state-of-the-arts in Video QA, the new system
is more modular and thus open to accommodate a wide range of low-level
visual processing and high-level reasoning capabilities. Tested on
SVQA (synthetic) and TGIF-QA (real) datasets, the proposed system
demonstrates new state-of-the-art performance in a majority of cases.
The gained margin is strongly evident in the case where the system
is defined for \textendash{} multi-step reasoning.

The proposed layered neural architecture is in line with proposals
in \citep{fodor1983modularity,harnad1990symbol}, where reactive perception
(System 1) precedes and is accessible to deliberative reasoning (System
2). Better perception capabilities will definitely make it easier
for visual reasoning. For example, action counting potentially benefits
more from activity motion rather than from the holistic feature maps
of visual scene as currently implemented. In addition, this chapter
used a simple implementation for hierarchy modelling of videos by
using a temporal attention to implicitly model the near-term relations
between video frames within a short clip, while the far-term temporal
relations between clips are done by the ClipRN. In the next chapter,
we will consider these aspects with a more homogeneous approach for
modelling near-term and far-term relations in videos.

We also observed that the generic reasoning scheme of MAC net is surprisingly
powerful for the domain of Video QA, especially for the problems that
demand multi-step inference (e.g., on the SVQA dataset). This suggests
that it is worthy to spend effort to advance reasoning functionalities
for both general cases and in spatio-temporal settings. We will address
this by bringing structured representations of data to assist the
reasoning process in Chapter \ref{chap:RelationalVisualReasoning}.
Finally, although we have presented a seamless feedforward integration
of System 1 and System 2, it is still open on how the two systems
interact. 

\selectlanguage{american}%

\selectlanguage{australian}%

\newpage{}

\selectlanguage{american}%

\chapter{Multimodal Reasoning\label{chap:MultimodalReasoning}}

\section{Introduction\label{sec:Chap5-Introduction}}

\selectlanguage{australian}%
In Chapter \ref{chap:DualProcess}, we have discussed that answering
natural questions about a video is a powerful demonstration of cognitive
capability. The task involves acquiring and manipulating spatio-temporal
visual, acoustic and linguistic representations from the video guided
by the compositional semantics of linguistic cues \citep{gao2018motion,lei2018tvqa,li2019beyond,song2018explore,tapaswi2016movieqa,wang2018movie}.
As questions are potentially unconstrained, Video QA requires deep
modelling capacity to encode and represent crucial multimodal video
properties such as linguistic content, object permanence, motion profiles,
prolonged actions, and varying-length temporal relations in a hierarchical
manner. For Video QA, the visual and textual representations should
ideally be question-specific and answer-ready. 

\begin{figure*}
\begin{minipage}[c][1\totalheight][b]{0.48\textwidth}%
\begin{center}
\includegraphics[width=1\textwidth]{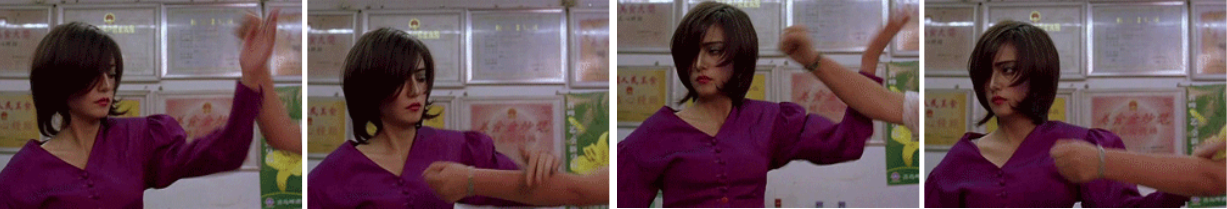}
\par\end{center}
{\small{}(1) }\textbf{\small{}Question}{\small{}: What does the girl
do 9 times?}{\small\par}

{\small{}\hspace{1.5em}}\textbf{\small{}Choice 1}{\small{}: walk}{\small\par}

{\small{}\hspace{1.5em}}\textbf{\small{}Choice 2}{\small{}: blocks
a person's punch}{\small\par}

{\small{}\hspace{1.5em}}\textbf{\small{}Choice 3}{\small{}: step}{\small\par}

{\small{}\hspace{1.5em}}\textbf{\small{}Choice 4}{\small{}: shuffle
feet}{\small\par}

{\small{}\hspace{1.5em}}\textbf{\small{}Choice 5}{\small{}: wag tail\medskip{}
}{\small\par}

{\small{}\hspace{1em}Baseline: }\textcolor{red}{\small{}walk}{\small\par}

{\small{}\hspace{1em}HCRN: }\textcolor{green}{\small{}blocks a person's
punch}{\small\par}

{\small{}\hspace{1em}Ground truth: }\textcolor{brown}{\small{}blocks
a person's punch}{\small\par}%
\end{minipage}\hfill{}%
\begin{minipage}[c][1\totalheight][b]{0.48\textwidth}%
\begin{center}
\includegraphics[width=1\textwidth]{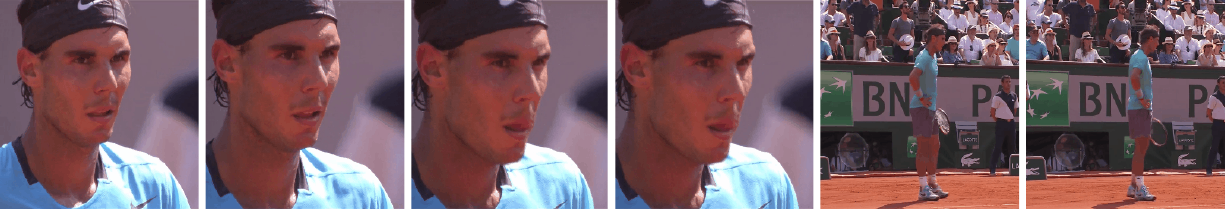}
\par\end{center}
{\small{}(2) }\textbf{\small{}Question}{\small{}: What does the man
do before turning body to left?}{\small\par}

{\small{}\hspace{1.5em}}\textbf{\small{}Choice 1}{\small{}: run across
a ring}{\small\par}

{\small{}\hspace{1.5em}}\textbf{\small{}Choice 2}{\small{}: pick
up the man's hand}{\small\par}

{\small{}\hspace{1.5em}}\textbf{\small{}Choice 3}{\small{}: flip
cover face with hand}{\small\par}

{\small{}\hspace{1.5em}}\textbf{\small{}Choice 4}{\small{}: raise
hand}{\small\par}

{\small{}\hspace{1.5em}}\textbf{\small{}Choice 5}{\small{}: breath\medskip{}
}{\small\par}

{\small{}\hspace{1em}Baseline: }\textcolor{red}{\small{}pick up the
man's hand}{\small\par}

{\small{}\hspace{1em}HCRN: }\textcolor{green}{\small{}breath}{\small\par}

{\small{}\hspace{1em}Ground truth: }\textcolor{brown}{\small{}breath}{\small\par}%
\end{minipage}

\caption{Examples of \emph{short-form Video QA} for which frame relations are
key toward correct answers. {\small{}(1) }\emph{\small{}Near-term
frame relations }{\small{}are required for counting of fast actions.
(2) }\emph{\small{}Far-term frame relations }{\small{}connect the
actions in long transition.} HCRN with the ability to model hierarchical
conditional relations handles successfully, while baseline struggles.\label{fig:chap5_Examples-of-short-form}}
\end{figure*}

\begin{figure*}
\noindent\begin{minipage}[t]{1\textwidth}%
\begin{center}
\includegraphics[width=0.98\textwidth]{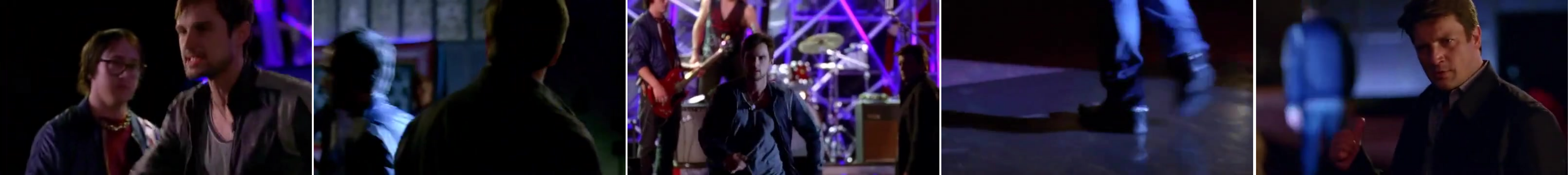}
\par\end{center}%
\end{minipage}

\medskip{}
\textbf{Subtitle}:\medskip{}

\noindent{\fboxrule 0.7pt\fboxsep 4pt\fbox{\begin{minipage}[c]{1\textwidth - 2\fboxsep - 2\fboxrule}%
\begin{center}
\begin{minipage}[t]{0.32\textwidth}%
\textcolor{teal}{\small{}00:00:0.395 -{}-> 00:00:1.896}{\small\par}

{\small{}(Keith:) I'm not gonna stand here and let you accuse me}{\small\par}%
\end{minipage}\hfill{}%
\begin{minipage}[t]{0.32\textwidth}%
\textcolor{teal}{\small{}00:00:1.897 -{}-> 00:00:4.210}{\small\par}

{\small{}(Keith:) of killing one of my best friends, all right?}{\small\par}%
\end{minipage}\hfill{}%
\begin{minipage}[t]{0.32\textwidth}%
\textcolor{teal}{\small{}00:00:8.851 -{}-> 00:00:10.394}{\small\par}

{\small{}(Castle:) You hear that sound?}{\small\par}%
\end{minipage}
\par\end{center}%
\end{minipage}}}
\begin{centering}
\medskip{}
\par\end{centering}
\noindent\begin{minipage}[t]{1\textwidth}%
\textbf{\small{}Question}{\small{}: What did Keith do when he was
on the stage?}{\small\par}

{\small{}}%
\begin{tabular}{>{\raggedright}p{0.49\textwidth}>{\raggedright}p{0.49\textwidth}}
{\small{}\enskip{}}\textbf{\small{}Choice 1}{\small{}: Keith drank
beer}{\small\par}

{\small{}\enskip{}}\textbf{\small{}Choice 2}{\small{}: Keith played
drum}{\small\par}

{\small{}\enskip{}}\textbf{\small{}Choice 3}{\small{}: Keith sang
to the microphone} & \textbf{\small{}Choice 4}{\small{}: Keith played guitar}{\small\par}

\textbf{\small{}Choice 5}{\small{}: Keith got off the stage and walked
out}\tabularnewline
\end{tabular}{\small\par}

{\small{}\medskip{}
}{\small\par}

{\small{}\hspace{0.7em}Baseline: }\textcolor{red}{\small{}Keith played
guitar}{\small\par}

{\small{}\hspace{0.7em}HCRN: }\textcolor{green}{\small{}Keith got
off the stage and walked out}{\small\par}

{\small{}\hspace{0.7em}Ground truth: }\textcolor{brown}{\small{}Keith
got off the stage and walked out}{\small\par}%
\end{minipage}

\caption{Example of \emph{long-form Video QA}. This is a typical question where
a model needs to collect sufficient relevant cues from both visual
content of a given video and textual subtitles to give the correct
answer. In this particular example, our baseline is likely to suffer
from the linguistic bias \emph{(``stage'' }and\emph{ ``played guitar''})
while our model successfully manages to arrive at the correct answer
by connecting the linguistic information from the first shot and visual
content in the second one.\label{fig:chap5_Examples-of-long-form}}
\end{figure*}

The typical approach toward modelling videos for QA is to build neural
architectures specially designed for a particular data format and
modality. In this perspective, the two common variants of Video QA
emerge: \emph{short-form Video QA,} where relevant information is
contained in the visual content of short video snippets of a single
event (see Fig.~\ref{fig:chap5_Examples-of-short-form} for examples),
and \emph{long-form Video QA} (also known as Video Story QA), where
clues to arrive at the answers are carried in the mixed visual-textual
data of longer movie or TV scenes (see Fig.~\ref{fig:chap5_Examples-of-long-form}
for example). Because of this specificity, such hand crafted architectures
tend to be non-optimal for changes in data modality \citep{lei2018tvqa},
varying video length \citep{na2017read} or question types (such as
frame QA \citep{li2019beyond} versus action count \citep{fan2019heterogeneous}).
This has resulted in proliferation of heterogeneous networks.

In Chapter \ref{chap:DualProcess}, we explored different aspects
that necessitate handling short-form Video QA. However, we have not
yet considered long-form Video QA. In this chapter, we wish to build
a family of effective models for both long-form and short-form Video
QA from reusable units that are more homogeneous and easier to construct,
maintain and comprehend. The content of this chapter is mainly based
on our preliminary work on short-form Video QA \citep{le2020hierarchical}.
We start by realising that Video QA involves two sub-tasks: (a) selecting
relevant content for each channel in the context of the linguistic
query, and (b) composing spatio-temporal concepts and relations hidden
in the data in response to the query. Much of sub-task (a) can be
abstracted into a conditional computational structure that computes
multi-way interaction between the query and the several objects. With
this ability, solving sub-task (b) can be approached by composing
the hierarchical structure of such abstraction from the ground up. 

Toward this goal, we propose a general-purpose \emph{reusable neural
unit} called Conditional Relation Network (CRN) that encapsulates
and transforms an array of objects into a new array of relations conditioned
on a contextual feature. The unit computes sparse high-order relations
between the input objects, then modulates the encoding through a specified
context (See Fig.~\ref{fig:chap5_Illustration-of-Multiscales}).
The flexibility of CRN and its encapsulating design allow it to be
replicated and layered to form deep hierarchical conditional relation
networks (HCRN) in a straightforward manner (See Fig.\ \ref{fig:chap5_visual_stream}
and Fig.\ \ref{fig:chap5_textual_stream}). In practice, we design
HCRN as a two-stream network of visual content and textual subtitles.
These two sub-networks share a similar design philosophy; however,
they are tailored for each input modality. While the visual stream
is built up by stacked CRN units at different granularities, the textual
stream is composed of a single CRN unit taking textual segments as
inputs. The stacked units in the visual stream thus provide contextualised
refinement of relational knowledge from visual objects. In a stage-wise
manner, it combines appearance features with clip activity flow and
linguistic context, and afterwards incorporates the context information
from the whole video motion and linguistic features. On the textual
side, the CRN unit functions in the same fashion but on textual objects.
The resultant HCRN is homogeneous, agreeing with the design philosophy
of networks built up from a building block such as InceptionNet \citep{szegedy2015going},
ResNet \citep{he2016deep} and FiLM \citep{perez2018film}.

The hierarchy of the HCRN for each input modality is shown as follows.
At the lowest level of the visual stream, the CRNs encode the relations
\emph{between} frame appearance in a clip and integrate the \emph{clip
motion as context}; this output is processed at the next stage by
CRNs that now integrate in the \emph{linguistic context}. In the following
stage, the CRNs capture the relation \emph{between} the clip encodings,
and integrate in \emph{video motion as context}. In the final stage
the CRN integrates the video encoding with the linguistic feature
as context (See Fig.~\ref{fig:chap5_visual_stream}). As for the
textual stream, due to its high-level abstraction and the diversity
in linguistic expressions compared to its visual counterpart, we only
use the CRN to encode relations between\emph{ }textual segments in
a given dialogue extracted from textual subtitles and leverage well-known
techniques in sequential modelling, such as LSTM \citep{hochreiter1997long}
or BERT \citep{devlin2018bert}, to understand sequential relations
at the word level. By allowing the CRNs to be stacked in a hierarchical
fashion, the model naturally supports modelling hierarchical structures
in video and relational reasoning. Likewise, by allowing appropriate
context to be introduced in stages, the model handles multimodal fusion
and multi-step reasoning. For long videos, deeper levels in hierarchy
design can be added, enabling the encoding of relations between distant
frames.

We demonstrate the capability of the HCRN in answering questions in
major Video QA datasets, including both short-form and long-form videos.
The hierarchical architecture with four layers of CRN units achieves
favourable performance against prior studies across all Video QA tasks.
Notably, it performs consistently well on questions involving either
appearance, motion, state transition, temporal relations, or action
repetition, demonstrating that the model can analyse and combine the
information in all of these channels. Furthermore, the HCRN scales
well on longer length videos simply with the addition of an extra
layer. Fig.~\ref{fig:chap5_Examples-of-short-form} and Fig.~\ref{fig:chap5_Examples-of-long-form}
demonstrate several representative cases that are difficult for a
simple baseline of flat visual-question interaction but can be handled
by our model. Our model and results demonstrate the impact of building
general-purpose neural reasoning units that support native multimodality
interaction in improving robustness and generalisation capacities
of Video QA models.

The rest of this chapter is organised as follows. Sec.~\ref{sec:Chap5-Background}
reviews related background, apart from the broader background in Chapter
\ref{chap:Background} and Chapter \ref{chap:VisualLanguageReasoning}.
Sec.~\ref{sec:Chap5-Method} details our main contributions in this
chapter \textendash{} the CRN, the HCRN for Video QA on both short-form
videos and long-form movie videos with textual subtitles. This section
also includes our complexity analysis of the CRN and HCRN as the video
length grows. The next section describes the results of the experimental
suite. Sec.~\ref{sec:Chap5-Remark} provides further discussion and
closing marks of the chapter.\selectlanguage{american}%

\section{Background \label{sec:Chap5-Background}}

\selectlanguage{australian}%
Apart from the background provided in Chapter \ref{chap:VisualLanguageReasoning}
and Chapter \ref{chap:DualProcess}, this section discusses relevant
works that are closely related to our proposed HCRN model. The HCRN
model advances the development of Video QA by addressing three key
challenges: (1) Efficiently representing videos as an amalgam of complementing
factors including appearance, motion and relations, (2) Effectively
allows the interaction of such visual features with the linguistic
query and (3) Allows integration of different input modalities in
Video QA with one unique building block.

\textbf{Long/short-form Video QA} has enjoyed rising attention in
recent years as mentioned in Sec.~\ref{sec:chap3_Neural-reasoning-for-VideoQA},
with the release of a number of large-scale short-form Video QA datasets,
such as TGIF-QA \citep{jang2017tgif,xu2016msr}, as well as long-form
MovieQA datasets with accompanying textual modalities, such as TVQA
\citep{lei2018tvqa}, MovieQA \citep{tapaswi2016movieqa} and PororoQA
\citep{kim2017deepstory}. All studies on Video QA treats short-form
and long-form Video QA as two separate problems in which proposed
methods \citep{fan2019heterogeneous,gao2018motion,jang2017tgif,kim2017deepstory,kim2019progressive,lei2018tvqa}
are deviated to handle either one of the two problems. Different from
those works, this chapter takes the challenge of designing a generic
method that covers both long-form and short-form Video QA with simple
exercises of block stacking and rearrangements to switch one unique
model between the two problems depending upon the availability of
input modalities for each of them.

\textbf{Spatio-temporal video representation and multimodal fusion}
- Earlier attempts for generic multimodal fusion for visual reasoning
include bilinear operators, either applied directly \citep{kim2018bilinear}
or through attention \citep{kim2018bilinear,yu2017multi}. While these
approaches treat the input tensors equally in a costly joint multiplicative
operation, HCRN separates conditioning factors from refined information,
hence it is more efficient and also more flexible on adapting operators
to conditioning types.

Temporal hierarchy has been explored for video analysis \citep{lienhart1999abstracting},
most recently with recurrent networks \citep{pan2016hierarchical,baraldi2017hierarchical}
and graph networks \citep{mao2018hierarchical}. However, we believe
we are the first to consider the hierarchical interaction of multi-modalities,
including linguistic cues for Video QA.

\textbf{Linguistic query\textendash visual feature interaction} \textbf{in
Video QA} has traditionally been formed as a visual information retrieval
task in a common representation space of independently transformed
question and referred video as explained in Sec.~\ref{sec:chap3_Neural-reasoning-for-VideoQA}.
The retrieval is more convenient with heterogeneous memory slots \citep{fan2019heterogeneous}.
On top of information retrieval, co-attention between the two modalities
provides a more interactive combination \citep{jang2017tgif}. Developments
along this direction include attribute-based attention \citep{ye2017video},
hierarchical attention \citep{liang2018focal,zhao2018multi,zhao2017video},
multi-head attention \citep{kim2018multimodal,li2019learnable}, multi-step
progressive attention memory \citep{kim2019progressive} or combining
self-attention with co-attention \citep{li2019beyond}. For higher
order reasoning, the question can interact iteratively with video
features via episodic memory or through a switching mechanism \citep{yang2019question}.
Multi-step reasoning for Video QA is also approached by \citep{xu2017video}
and \citep{song2018explore} with refined attention.

Unlike these techniques, our HCRN model supports conditioning video
features with linguistic clues as a context factor in every stage
of the multi-level refinement process. This allows the linguistic
cues to be involved earlier and deeper into video presentation construction
than any available methods.

\textbf{Neural building blocks} - Beyond the Video QA domain, CRN
unit shares the idealism of uniformity in neural architecture with
other general-purpose neural building blocks such as the block in
InceptionNet \citep{szegedy2015going}, Residual Block in ResNet \citep{he2016deep},
Recurrent Block in RNN, conditional linear layer in FiLM \citep{perez2018film},
and matrix-matrix-block in neural matrix net \citep{do2018learning}.
Our CRN departs significantly from these designs by assuming an array-to-array
block that supports conditional relational reasoning and can be reused
to build networks of other purposes in vision and language processing.
As a result, our HCRN is a perfect fit for bot short-form Video QA,
where questions are all about the visual content of a video snippet,
and long-form Video QA (Movie QA), where a model has to look at both
visual cue and textual cue (subtitles) to arrive at correct answers.
Due to the great challenges posed by the long-form Video QA and the
diversity in terms of model building, current approaches in Video
QA mainly spend efforts on handling the visual part while leaving
the textual part for common techniques such as LSTM \citep{lei2018tvqa}
or the latest advance in natural language processing BERT \citep{yang2020bert}.
To the best of our knowledge, HCRN is the first model that could solve
both short-form and long-form Video QA with models building up from
a generic neural block.\selectlanguage{american}%

\section{Method \label{sec:Chap5-Method}}

\selectlanguage{australian}%
\begin{figure}
\begin{centering}
\includegraphics[width=0.75\columnwidth]{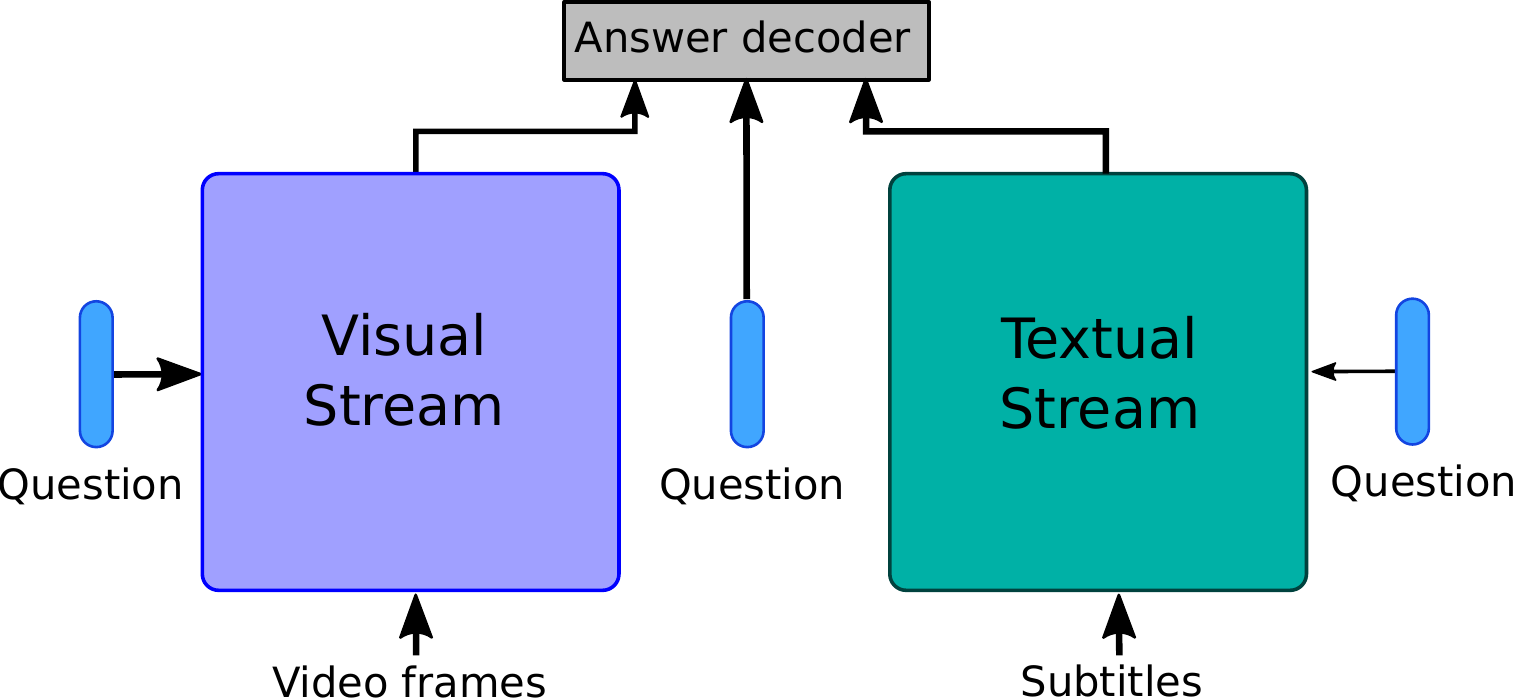}
\par\end{centering}
\caption{Overall multimodal Video QA architecture. The two streams handle visual
and textual modalities in parallel, followed by an answer decoder
for feature joining and prediction.\label{fig:chap5_Overall-VideoQA}}
\end{figure}

As we have formulated in Chapter \ref{chap:DualProcess}, the goal
of Video QA is to deduce an answer $\hat{y}$ from a video $\bm{\mathcal{V}}$
in response to a natural question $\mathbf{q}$. In this chapter,
we optionally consider an additional information channel such as subtitles
$\mathcal{S}$. This refers to the case where video $\mathcal{V}$
is a movie scene whose content is associated with conversational dialogues
between movie characters. The answer $\hat{y}$ can be found in an
answer space $\bm{\mathcal{A}}$ which is a pre-defined set of possible
answers for open-ended questions or a list of answer candidates in
the case of multi-choice questions. Hence, Video QA in the context
of this chapter can be formulated as follows:\vspace{-1cm}

\begin{eqnarray}
\hat{y} & = & \underset{a\in\mathcal{\bm{A}}}{\text{argmax}}\mathcal{F}_{\bm{\theta}}\left(a;\mathbf{q},\mathcal{\bm{V}},\mathcal{\bm{S}}\right),\label{eq:chap5_prob-def}
\end{eqnarray}
where $\bm{\theta}$ is the model parameters of scoring function $\mathcal{F}\left(.\right)$.

In this chapter, we address two common settings of Video QA: (a) short-form
Video QA where the visual content of a given single video shot singularly
suffice to answer the questions, similar to what we have discussed
in Chapter \ref{chap:DualProcess}, and (b) long-form Video QA where
the essential information disperses among visual content in the multi-shot
video sequences of a movie/TV program and the conversational content
in accompanying textual subtitles.

HCRN is designed in the endeavour for a homogeneous neural architecture
that can adapt to solve both problems. Its overall workflow is depicted
in Fig.~\ref{fig:chap5_Overall-VideoQA}. In long-form videos, when
both visual and textual streams are present, HCRN distils relevant
information from the visual stream and the textual stream. These two
streams are both are conditioned on the question. Eventually, it combines
them into an answer decoder for final prediction in late-fusion multimodal
integration. In short-form cases, where only video frames are available,
the visual stream is solely active, working with the single-input
answer decoder. One of the ambitions of the design is to build each
processing stream as a hierarchical network simply by stacking common
core processing units of the same family. Similar to all previous
deep-learning-based approaches in the literature \citep{jang2017tgif,li2019beyond,fan2019heterogeneous,gao2018motion,lei2018tvqa,le2neural},
our HCRN operates on top of the feature embeddings of multiple input
modalities. By doing this, our model can take advantage of the powerful
feature representations extracted by either common visual recognition
models pre-trained on large-scale datasets such as ResNet \citep{he2016deep},
ResNeXt \citep{xie2017aggregated,hara2018can} or pre-trained word
embeddings such as GloVe \citep{pennington2014glove} and BERT \citep{devlin2018bert}.

In the following subsections, we present the design of the core unit
in Sec.~\ref{subsec:chap5_Relation-Network}, the hierarchical designs
tailored to each modality in Sec.~\ref{subsec:chap5_HCRN}, the answer
decoder in Sec.~\ref{subsec:chap5_Answer-Decoders}. Theoretical
analysis of the computational complexity of the models follows in
Sec.~\ref{subsec:chap5_Complexity-Analysis}.

\subsection{Conditional Relation Network Unit\label{subsec:chap5_Relation-Network}}

\begin{algorithm}[t]
\small
\label{algo:CRN}
\caption{CRN Unit}
	\SetKwInOut{Input}{Input}
	\SetKwInOut{Output}{Output}
	\SetKwInOut{Metaparams}{Metaparams}
	\Input{Array $\bm{\mathcal{X}}=\{\mathbf{x}_i\}_{i=1}^n$, conditioning feature $\mathbf{c}$}
	\Output{Array $\mathbf{R}$}
	\Metaparams{$\{k_{\textrm{max}},t \mid k_{\textrm{max}}<n\}$}
	
	Initialize $\mathbf{R} \leftarrow \{\}$

	\For{$k \leftarrow 2$ \KwTo $k_{\SI{}{max}}$}
	{
		$\mathbf{Q}^{k}=$ randomly select $t$ subsets of size $k$ from $\bm{\mathcal{X}}$
		
		\For{\SI{}{\textbf{each}} subset $\mathbf{Q}_i^k$ $\in$ $\mathbf{Q}^{k}$}
		{	
			$\mathbf{g}_i = g^k(\mathbf{Q}_i^k)$
			
			$\mathbf{h}_i = h^k(\mathbf{g}_i,\mathbf{c})$
		}
        
		$\mathbf{r}^k=p^{k}(\{\mathbf{h}_i\})$
		
		add $\mathbf{r}^k$ to $\mathbf{R}$
		
	}
\end{algorithm}

\begin{figure}
\begin{centering}
\includegraphics[width=0.6\columnwidth]{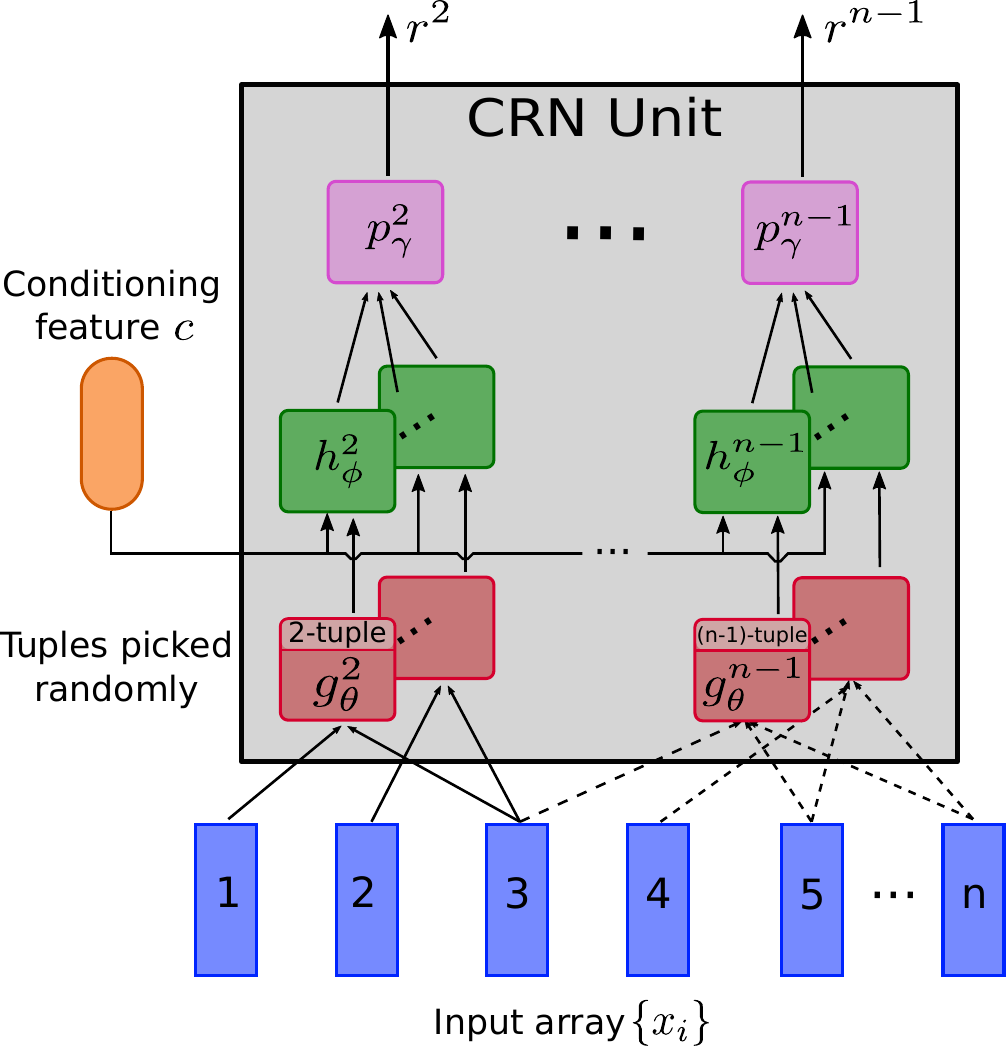}
\par\end{centering}
\caption{Conditional Relation Network. a) Input array {\small{}$\mathcal{\bm{X}}$}
of $n$ objects are first processed to model $k$-tuple relations
from $t$ sub-sampled size-$k$ subsets by sub-network $g^{k}\left(.\right)$.
The outputs are further conditioned with the context $\mathbf{c}$
via sub-network $h^{k}\left(.\,,.\right)$ and finally aggregated
by $p^{k}\left(.\right)$ to obtain a result vector $\mathbf{r}^{k}$
which represents $k$-tuple conditional relations. Tuple sizes can
range from $2$ to $(n-1)$, which outputs an $(n-2)$-dimensional
output array.\label{fig:chap5_Illustration-of-Multiscales}}
\medskip{}
\end{figure}

\begin{table}
\begin{centering}
\begin{tabular}{l|l}
\hline 
{\small{}Notation} & {\small{}Role}\tabularnewline
\hline 
\hline 
{\small{}$\mathcal{\bm{X}}$} & {\small{}Input array of $n$ objects (e.g. frames, clips)}\tabularnewline
\hline 
{\small{}$\mathbf{c}$} & {\small{}Conditioning feature (e.g. query, motion feat.)}\tabularnewline
\hline 
{\small{}$k_{\text{max}}$} & {\small{}Maximum subset (also tuple) size considered}\tabularnewline
\hline 
{\small{}$k$} & {\small{}Each subset size from $2$ to $k_{max}$}\tabularnewline
\hline 
{\small{}$t$} & {\small{}Number of size-$k$ subsets of $\mathcal{\bm{X}}$ randomly
sampled}\tabularnewline
\hline 
{\small{}$\mathbf{Q}^{k}$} & {\small{}Set of $t$ size-$k$ subsets sampled from $\mathcal{\bm{X}}$}\tabularnewline
\hline 
{\small{}$g^{k}\left(.\right)$} & {\small{}Sub-network processing each size-$k$ subset}\tabularnewline
\hline 
{\small{}$h^{k}\left(.\,,.\right)$} & {\small{}Conditioning sub-network}\tabularnewline
\hline 
{\small{}$p^{k}\left(.\right)$} & {\small{}Aggregating sub-network}\tabularnewline
\hline 
{\small{}$\mathbf{R}$} & {\small{}Result array of CRN unit on $\mathcal{\bm{X}}$ given $\mathbf{c}$}\tabularnewline
\hline 
{\small{}$\mathbf{r}^{k}$} & {\small{}Member result vector of $k$-tuple relations}\tabularnewline
\hline 
\end{tabular}\medskip{}
\par\end{centering}
\centering{}\caption{Notations of CRN unit operations.\label{tab:chap5_notations}}
\end{table}

We introduce a general computation unit, termed Conditional Relation
Network (CRN), which takes as input an array of $n$ objects $\mathcal{\bm{X}}=\left\{ \mathbf{x}_{i}\right\} _{i=1}^{n}$
and a conditioning feature $\mathbf{c}$ serving as global context.
Objects are assumed to live either in the same vector space $\mathbb{R}^{d}$
or tensor space, for example, $\mathbb{R}^{d\times W\times H}$ in
the case of images (or video frames), where $W$ and $H$ are the
width and height of a video frame, respectively. CRN generates an
output array of objects of the same dimensions containing high-order
object relations of input features given the global context. The global
context is problem-specific, serving as a modulator to the formation
of the relations. When in use for Video QA, CRN's input array is composed
of features at either frame level, short-clip levels, or textual features.
Examples of global context include the question and motion profiles
at a given hierarchical level. 

Given input set of object $\mathcal{\bm{X}},$ CRN first considers
the set of subsets $\mathcal{\bm{Q}}=\left\{ \mathbf{Q}^{k}\right\} _{k=2}^{k_{\textrm{max}}}$
where each set $\mathbf{Q}^{k}$ contains $t$ size-$k$ subsets randomly
sampled from $\mathcal{\bm{X}}$, where $t$ is the sampling frequency,
$t<C(n,k)$. On each collection $\mathbf{Q}^{k}$, CRN then uses member
relational sub-networks to infer the joint representation of $k$-tuple
object relations. In videos, due to temporal coherence, the objects
$\left\{ \mathbf{x}_{i}\right\} _{i=1}^{n}$ share a great amount
of mutual information, therefore, it is reasonable to use a subsampled
set of $t$ combinations instead of considering all possible combinations
for $\mathbf{Q}^{k}$. This subsampling trick is inspired by \citet{zhou2018temporal}.
However, we sample $t$ size-$k$ subsets directly from the original
$\mathcal{\mathcal{\bm{X}}}$ rather than randomly take subsets in
a pool of all possible size-$k$ subsets as in their method. By doing
this, we can reduce the computational complexity of the CRN as it
is much cheaper to sample elements out of a set than building all
combinations of subsets which cost $\mathcal{O}(2^{n})$ time. We
provide more analysis on the complexity of the CRN unit later in Sec.
\ref{subsec:chap5_Complexity-Analysis}.

Regarding relational modelling, each member subset of $\mathbf{Q}^{k}$
then goes through a set function $g^{k}(.)$ for relational modelling.
The relations across objects are then further refined with a conditioning
function $h^{k}(.,\mathbf{c})$ in the light of conditioning feature
$\mathbf{c}$. Finally, the $k$-tuple relations are summarised by
the aggregating function $p^{k}(.)$. The similar operations are done
across the range of subset size from 2 to $k_{\text{max}}$. Regarding
the choice of $k_{\text{max}}$, we use $k_{\text{max}}=n-1$ in later
experiments, resulting in the output array of size $n-2$ if $n>2$
and an array of size $1$ if $n=2$.

The detailed operation of the CRN unit is presented formally as pseudo-code
in Alg.~1 and visually in Fig.~\ref{fig:chap5_Illustration-of-Multiscales}.
Table~\ref{tab:chap5_notations} summarises the notations used across
these presentations.

\subsubsection{Networks Implementation}

\paragraph{Set aggregation:}

The set functions $g^{k}(.)$ and $p^{k}(.)$ can be implemented as
any aggregation sub-networks that join a random set into a single
representation. As a choice in implementation, the function $g^{k}(.)$
is either average pooling or a simple concatenation operator while
$p^{k}(.)$ is average pooling.

\paragraph{Conditioning function:}

The design of the conditioning sub-network that implements $h^{k}(.,\mathbf{c})$
depends on the relationship between the input set $\bm{\mathcal{X}}$
and the conditioning feature $\mathbf{c}$ as well as the properties
of the channels themselves. Here we present four forms of neural operation
implementing this function.
\begin{itemize}
\item \textbf{\emph{Additive form}}\textbf{:}
\end{itemize}
A simple form of $h^{k}(.,\mathbf{c})$ is feature concatenation followed
by a MLP that models the non-linear relationships between multiple
input modalities:\vspace{-1cm}

\begin{eqnarray}
h^{k}\left(.,\mathbf{c}\right) & = & \text{ELU}\left(\mathbf{W}^{h}\left[.;\mathbf{c}\right]\right),\label{eq:chap5_conditioning_concat}
\end{eqnarray}
where $\mathbf{W}^{h_{1}}\in\mathbb{R}^{d\times d}$ is a weight matrix;
$[\thinspace;]$ denotes the tensor concatenation operation and ELU
is the non-linear activation function introduced by \citet{clevert2015fast}.
Eq.~\ref{eq:chap5_conditioning_concat} is sufficient when $\mathbf{x}$
and $\mathbf{c}$ are additively complementary. 
\begin{itemize}
\item \textbf{\emph{Multiplicative form}}\textbf{:}
\end{itemize}
To support more complex the relationship between the input $\mathbf{x}$
and the conditioning feature $\mathbf{c}$, a more sophisticated joining
operation is warranted. For example, when $\mathbf{c}$ implies a
selection criterion to modulate the relationship between elements
in $\mathbf{x}$, the multiplicative relation between them can be
represented in conditioning function by:\vspace{-1cm}

\begin{eqnarray}
h^{k}\left(\mathbf{x},\mathbf{c}\right) & = & \text{ELU}\left(\mathbf{W}^{h}\left[\mathbf{x};\mathbf{x}\odot\mathbf{c};\mathbf{c}\right]\right),\label{eq:chap5_conditioning_mul}
\end{eqnarray}
where $\odot$ denotes Hadamard product.
\begin{itemize}
\item \textbf{\emph{Sequential form}}\textbf{:}
\end{itemize}
As aforementioned, how to properly choose the sub-network $h^{k}(.,\mathbf{c})$
is also driven by the properties of the input set $\mathcal{\bm{X}}$
itself. In the context of Video QA, elements in $\mathcal{\bm{X}}$
(visual features or dialogue-based textual features) would contain
strong temporal relationships, we additionally integrate sequential
modelling capability, using biLSTM in this thesis, along with the
conditioning sub-network as presented in Eq.~\ref{eq:chap5_conditioning_concat}
and Eq.~\ref{eq:chap5_conditioning_mul}. Formally, $h^{k}(.,\mathbf{c})$
is defined as:\vspace{-1cm}

\begin{eqnarray}
\mathbf{s} & = & \left[\mathbf{x};\mathbf{x}\odot\mathbf{c};\mathbf{c}\right],\label{eq:chap5_conditioning}\\
\mathbf{s}^{\prime} & = & \text{BiLSTM}(\mathbf{s}),\label{eq:chap5_conditioning_bilstm}\\
h^{k}(.,\mathbf{c}) & = & \text{maxpool}(\mathbf{s}^{\prime}).\label{eq:chap5_max_temporally}
\end{eqnarray}

\begin{itemize}
\item \textbf{\emph{Dual conditioning form}}\textbf{:}
\end{itemize}
In the later use of CRN in Video QA settings, where it can happen
that two concurrent signals $\mathbf{c}_{1},\mathbf{c}_{2}$ are used
as conditioning features, we simply extend Eq.~\ref{eq:chap5_conditioning_concat}
and Eq.~\ref{eq:chap5_conditioning_mul} and Eq.~\ref{eq:chap5_conditioning}
as:
\begin{align}
h^{k}\left(\mathbf{x},\mathbf{c}\right)= & \text{\ ELU}\left(\mathbf{W}^{h}\left[\mathbf{x};\mathbf{c}_{1};\mathbf{c}_{2}\right]\right),\\
h^{k}\left(\mathbf{x},\mathbf{c}\right)= & \text{\ ELU}\left(\mathbf{W}^{h}\left[\mathbf{x};\mathbf{x}\odot\mathbf{c}_{1};\mathbf{x}\odot\mathbf{c}_{2};\mathbf{c}_{1};\mathbf{c}_{2}\right]\right),\\
\mathbf{s}= & \left[\mathbf{x};\mathbf{x}\odot\mathbf{c}_{1};\mathbf{x}\odot\mathbf{c}_{2};\mathbf{c}_{1};\mathbf{c}_{2}\right],
\end{align}
respectively.

\subsection{Hierarchical Conditional Relation Network for Multimodal Video QA
\label{subsec:chap5_HCRN}}

We use CRN blocks to build a deep network architecture that supports
multiple Video QA settings. In particular, two variations are specifically
designed to work on short-form and long-form Video QA. For each of
these settings, the network design adapts to exploit the inherent
characteristics of a video sequence, namely temporal relations, motion,
linguistic conversation and the hierarchy of video structure, and
support reasoning guided by linguistic questions. We term the proposed
network architecture Hierarchical Conditional Relation Networks (HCRN).
The design of the HCRN by stacking reusable core units is partly inspired
by modern CNN network architectures, of which InceptionNet \citep{szegedy2015going}
and ResNet \citep{he2016deep} are the most well-known examples. In
the general form of Video QA, HCRN is a multi-stream end-to-end differentiable
neural network in which one stream handles visual content and the
other one handles textual dialogues in subtitles. The network is modular,
and each network stream plays a plug-and-play role adaptively to the
presence of input modalities.

\subsubsection{Preprocessing\label{subsec:chap5_Preprocessing}}

With the HCRN as defined in Eq. \ref{eq:chap5_prob-def}, HCRN takes
input and question represented as sets of visual or textual objects
and computes the answer. In this section, we describe the preprocessing
of raw videos into appropriate input sets for HCRN.

\paragraph{Visual representation:}

We begin by dividing the video $\mathcal{\bm{V}}$ of $L$ frames
into $N$ equal length clips $\mathbf{C}=\{\mathbf{C}_{1},...,\mathbf{C}_{N}\}$.
For short-form videos, each clip $\mathbf{C}_{i}$ of length $T=\left\lfloor L/N\right\rfloor $
is represented by two sources of information: frame-wise appearance
feature vectors $\mathbf{V}_{i}=\left\{ \mathbf{v}_{i,j}\mid\mathbf{v}_{i,j}\in\mathbb{R^{\text{2048}}}\right\} _{j=1}^{T}$
, and a motion feature vector at clip level $\mathbf{f}_{i}\in\mathbb{R^{\text{2048}}}$.
Appearance features are vital for video understanding as the visual
saliency of objects/entities in the video is usually of interest to
human questions. In short clips, moving objects and events are primary
video artefacts that capture our attention. Hence, it is common to
see the motion features coupled with the appearance features to represent
videos in the video understanding literature. On the contrary, in
long-form videos such as those in movies and TV programs, the concerns
can be less about specific motions but more into movie plot or film
grammar. As a result, we use the frame-wise appearance as the only
feature for the long-form Video QA. In our experiments, $\mathbf{v}_{i,j}$
are the \emph{pool5} output of ResNet \citep{he2016deep} features
and $\mathbf{f}_{i}$ are derived by ResNeXt-101 \citep{xie2017aggregated,hara2018can}.

Subsequently, linear feature transformations are applied to project
$\{\mathbf{v}_{ij}\}$ and $\mathbf{f}_{i}$ into a standard $d$-dimensions
feature space to obtain $\left\{ \hat{\mathbf{v}}_{ij}\mid\hat{\mathbf{v}}_{ij}\in\mathbb{R}^{d}\right\} $
and $\hat{\mathbf{f}}_{i}\in\mathbb{R}^{d}$, respectively.

\paragraph{Linguistic representation:}

Linguistic objects are built from the question, answer choices and
long-form videos' subtitles. We explore two options of representation
learning for them using biLSTM and BERT.
\begin{itemize}
\item \textbf{\emph{Sequential embedding with biLSTM:}}
\end{itemize}
All words in linguistic cues, including those in the question, answer
choices and subtitles, are first embedded into vectors of 300 dimensions
by pre-trained GloVe word embeddings \citep{pennington2014glove}. 

For the question and answer choices, we further pass these context-independent
embedding vectors through a biLSTM. Output hidden states of the forward
and backward LSTM passes are finally concatenated to form the overall
query representation $\mathbf{q}\in\mathbb{R}^{d}$ for the questions
and \textbf{$\mathbf{a}\in\mathbb{R}^{d}$} for the answer choices
if available (multi-choice question-answer pairs). 

For the accompanying subtitles provided in long-form Video QA, instead
of treating them as one big passage as in prior works \citep{lei2018tvqa,kim2019progressive},
we dissect the subtitle passage $\mathcal{\bm{S}}$ into a fixed number
of $M$ overlapping \emph{segments} $\mathcal{\bm{U}}=\{\mathcal{\bm{U}}_{1},..,\mathcal{\bm{U}}_{M}\}$.
The number of words in sibling segments is identical $T=\textrm{length}(\mathcal{\bm{S}})/M$
but varies from one video to another depending on the overall length
of the given subtitle passage $\bm{\mathcal{S}}$. 

We process each segment with pre-trained word embeddings followed
by a biLSTM as similar to the way we process the question. The $T$
hidden states of the biLSTM are then used as textual objects:
\begin{eqnarray}
\mathbf{\bm{U}}_{i} & = & \textrm{biLSTM}(\mathcal{\bm{U}}_{i}),
\end{eqnarray}
where $\left\{ \mathbf{U}_{i}\right\} _{i=1}^{M}$ are the final representations
ready to be used by the textual stream which will be described in
Subsec.~\ref{subsec:chap5_Textual-stream}.
\begin{itemize}
\item \textbf{\emph{Contextual embedding using BERT:}}
\end{itemize}
As an alternative option for linguistic representation, we utilise
pre-trained BERT network \citep{devlin2018bert} to extract contextual
word embeddings. Instead of encoding words independently, as in GloVe,
BERT embeds each word in the context of its surrounding words using
a self-attention mechanism.

For short-form Video QA, we tokenise the given question and answer
choices in multi-choice questions and subsequently feed the tokens
into the pre-trained BERT network. The averaged embedding of words
in a sentence is used as a unified representation for that sentence.
This applies to generate both the question representation $\mathbf{q}$
and answer choices $\{\mathbf{a}_{i}\}_{i=1,...,A}$.

For long-form Video QA, with each answer choice $i$, we form a long
string $\mathbf{L}_{i}$ by stacking it with subtitles $\bm{\mathcal{S}}$
and the question sentence. We then tokenise and embed each string
$\mathbf{L}_{i}$ with BERT into contextual hidden matrix $\mathbf{H}$
of size $d\times m$ where $m$ is the maximum number of tokens in
the input and $d$ is hidden dimensions of BERT. This tensor $\mathbf{H}$
is then split up into corresponding embedding vectors of the subtitles
$\mathbf{H}^{s}$, of the question $\mathbf{H}^{q}$ and of the answer
choice $\mathbf{H}^{a_{i}}$:
\begin{eqnarray}
\left(\mathbf{H}^{s},\mathbf{H}^{q},\mathbf{H}^{a_{i}}\right) & = & \textrm{BERT}(\mathbf{L}_{i}).
\end{eqnarray}

Eventually, we suppress the contextual tokens question representation
and answer choices into their respective single representation by
mean pooling:
\begin{eqnarray}
\mathbf{q} & = & \textrm{mean}(\mathbf{H}^{q});\:\mathbf{a}_{i}=\textrm{mean}(\mathbf{H}^{a_{i}}),
\end{eqnarray}
while keeping those of subtitles $\mathbf{H}^{s}$ as a set of textual
objects. $\left\{ \mathbf{U}_{i}\right\} _{i=1}^{M}$ are obtained
by sliding overlapping windows of the same length over $\mathbf{H}^{s}$.

\subsubsection{Visual Stream\label{subsec:chap5_Visual-stream}}

\begin{figure*}
\begin{centering}
\includegraphics[width=0.8\textwidth]{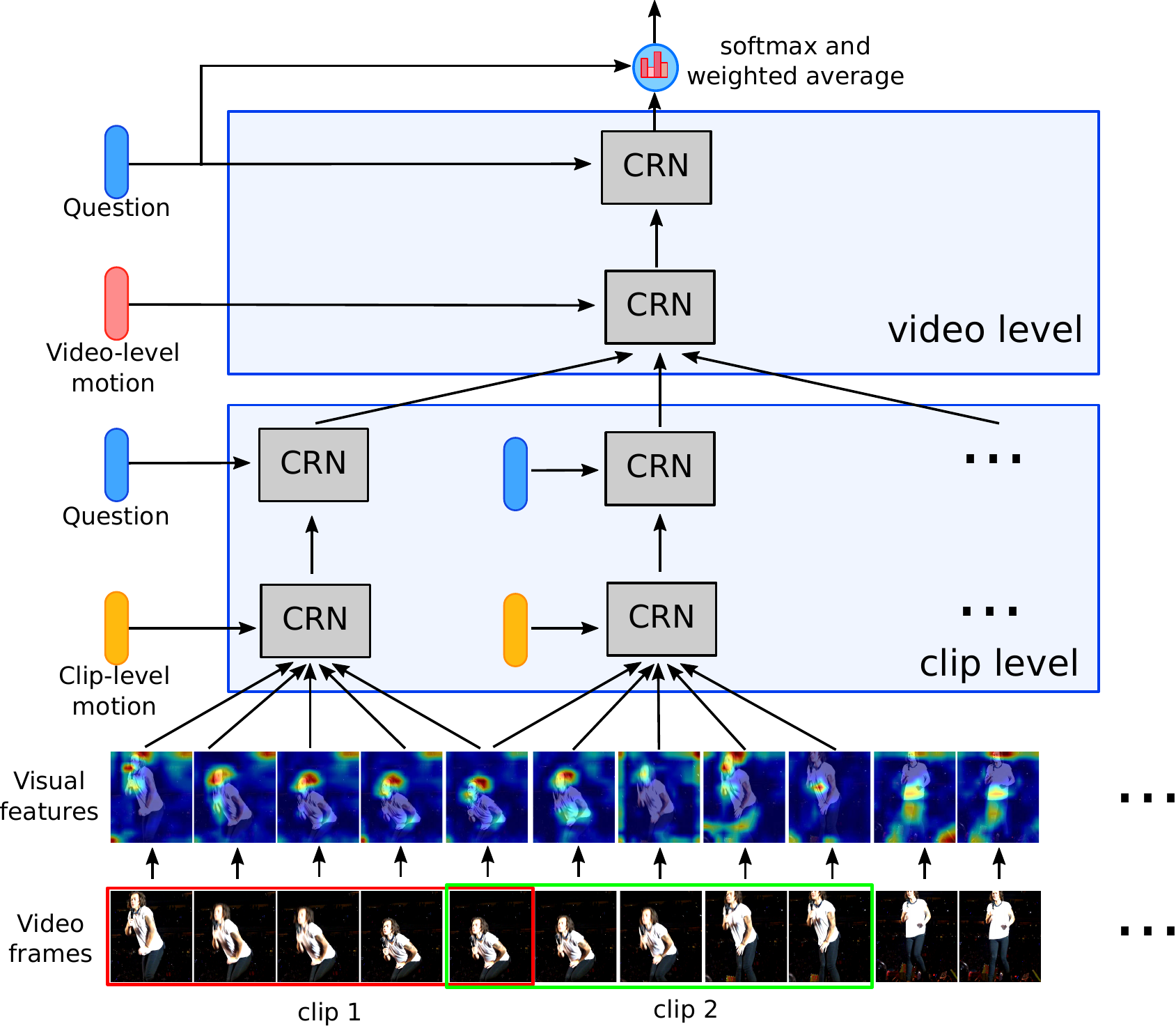}
\par\end{centering}
\caption{Visual stream. The CRNs are stacked in a hierarchical fashion. They
embed the video input at different granularities including frame,
short clip and entire video levels. At each level of granularity,
the video feature embedding is conditioned on the respective level-wise
motion feature and universal linguistic cue. \label{fig:chap5_visual_stream}}
\end{figure*}

An effective model for Video QA needs to distil the visual content
in the context of the question and filter out the usually large portion
of the data that is not relevant to the question. Drawing inspiration
from the hierarchy of video structure, we boil down the problem of
Video QA into a process of video representation in which a given video
is encoded progressively at different granularities, including short
clip (a sequence of video frames) and entire video levels (a sequence
of clips). It is crucial that \emph{the whole process conditions on
linguistic cues}.

With that in mind, we design a two-level structure to represent a
video, one at the clip level and the other one at the video level,
as illustrated in Fig\@.~\ref{fig:chap5_visual_stream}. We use
two stacked CRN units at each hierarchy level, one conditioned on
motion features followed by one conditioned on linguistic cues. Intuitively,
the motion feature serves as a dynamic context shaping the temporal
relations found among frames (at the clip level) or clips (at the
video level). It also provides a saliency indicator of which relations
are worth the attention \citep{mahapatra2008motion}.

As the shaping effect is applied to all relations in a complementary
way, selective (multiplicative) relation between the relations and
the conditioning feature is not needed, and thus a simple concatenation
operator followed by an MLP suffices. On the other hand, the linguistic
cues are by nature selective; that is, not all relations are equally
relevant to the question. Thus we utilise the multiplicative form
for feature fusion as in Eq.~\ref{eq:chap5_conditioning_mul} for
the CRN units which condition on question representation. 

With this particular design of network architecture, the input array
at clip level consists of frame-wise appearance feature vectors $\{\hat{\mathbf{v}}_{ij}\}$,
while that at a video level is the output at the clip level. The
motion conditioning feature at clip level CRNs is corresponding clip
motion feature vector $\hat{\mathbf{f}}_{i}$. They are further passed
to a LSTM, whose final state is used as video-level motion features.
Note that this particular implementation is not the only option. We
believe we are the first to progressively incorporate multiple modalities
of input in such a hierarchical manner in contrast to the typical
approach of treating appearance features and motion features as a
two-stream network.

\paragraph{Deeper hierarchy:}

To handle a video of longer size, up to thousands of frames which
is equivalent to dozens of short-term clips, there are two options
to reduce the computational cost of CRN in handling large sets of
subsets $\left\{ \mathbf{Q}^{k}\mid k=2,...,k_{\text{max}}\right\} $
given an input array $\mathcal{\bm{X}}$: (i) limit the maximum subset
size $k_{\text{max}}$, or (ii) extend the visual stream networks
to deeper hierarchy. For the former option, this choice of sparse
sampling may have the potential to lose critical relation information
of specific subsets. The latter, on the other hand, is able to densely
sample subsets for relation modelling. Specifically, we can group
$N$ short-term clips into $N_{1}\times N_{2}$ hyper-clips, of which
$N_{1}$ is the number of the hyper-clips and $N_{2}$ is the number
of short-term clips in one hyper-clip. By doing this, the visual stream
now becomes a 3-level of hierarchical network architecture. See Sec.~\ref{subsec:chap5_Complexity-Analysis}
for the effect of going deeper on running time and Subsec.~\ref{subsec:chap5_Deepening-model-hierarchy}
on accuracy.

\paragraph{Computing the output:}

At the end of the visual stream, we compute the average visual feature
which is driven by the question representation $\mathbf{q}$. Assuming
that the outputs of the last CRN unit at the video level are an array
$\mathbf{O}=\left\{ \mathbf{O}_{i}\mid\mathbf{O}_{i}\in\mathbb{R}^{d\times H}\right\} _{i=1}^{N-4}$,
we first stack them together, resulting in an output tensor $\mathbf{o}\in\mathbb{R}^{d\times(N-4)\times H}$.
We further vectorise this output tensor to obtain the final output
$\mathbf{O}^{\prime}=\left\{ \mathbf{o}_{h}^{\prime}\right\} _{h=1}^{H^{\prime}},\:\mathbf{O}^{\prime}\in\mathbb{R}^{d\times H^{\prime}},H^{\prime}=(N-4)\times H$.
The weighted average information is given by:\vspace{-1cm}

\begin{eqnarray}
\mathbf{i}_{h} & = & \left[\mathbf{W}^{o^{\prime}}\mathbf{o}_{h}^{\prime};\mathbf{W}^{o^{\prime}}\mathbf{o}_{h}^{\prime}\odot\mathbf{W}^{q}\mathbf{q}\right],\:\text{for}\:h=1,..,H^{\prime},\label{eq:chap5_attention_vision}\\
\mathbf{i}_{h}^{\prime} & = & \text{ELU}\left(\mathbf{W}^{i}\mathbf{i}_{h}+\mathbf{b}\right),\\
\gamma_{h} & = & \text{softmax}\left(\mathbf{W}^{i^{\prime}}\mathbf{i}_{h}^{\prime}+b_{h}\right),\\
\tilde{\mathbf{o}} & = & \sum_{h=1}^{H^{\prime}}\gamma_{h}\mathbf{o}_{h}^{\prime};\,\tilde{\mathbf{o}}\in\mathbb{R}^{d},
\end{eqnarray}
where, $\mathbf{W}^{o^{\prime}}\in\mathbb{R}^{d\times d},\mathbf{W}^{q}\in\mathbb{R}^{d\times d},\mathbf{W}^{i}\in\text{\ensuremath{\mathbb{R}^{d\times d}}}$
and \textbf{$\mathbf{W}^{i^{\prime}}\in\mathbb{R}^{1\times d}$} are
weight matrices. $\left[.\ ,.\right]$ denotes concatenation operation,
and $\odot$ is the Hadamard product. In the case of multi-choice
question answering where answer choices $\{\mathbf{a}_{i}\}$ are
available, Eq.\ \ref{eq:chap5_attention_vision} becomes $\mathbf{i}_{i,h}=\left[\mathbf{W}^{o^{\prime}}\mathbf{o}_{h}^{\prime};\mathbf{W}^{o^{\prime}}\mathbf{o}_{h}^{\prime}\odot\mathbf{W}^{q}\mathbf{q};\mathbf{W}^{o^{\prime}}\mathbf{o}_{h}^{\prime}\odot\mathbf{W}^{a}\mathbf{a}_{i}\right]$.

\subsubsection{Textual Stream\label{subsec:chap5_Textual-stream}}

\begin{figure*}
\begin{centering}
\includegraphics[width=0.8\textwidth]{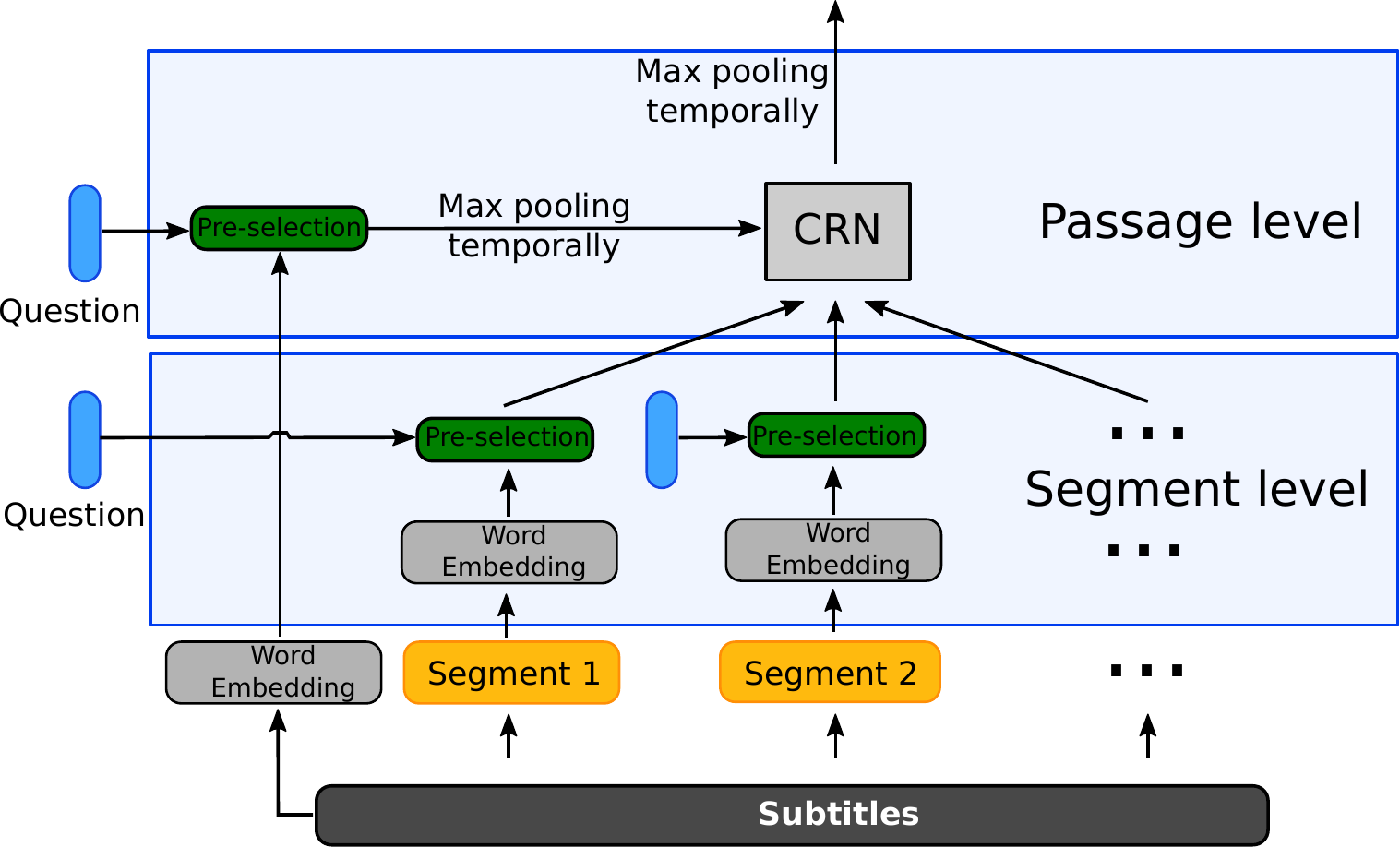}
\par\end{centering}
\caption{Textual stream. Both segment-level and passage-level textual objects
are modulated with the question by a \emph{pre-selection }module.
They then serve as input and conditioning features for a CRN which
models long-term relationships between segments.\label{fig:chap5_textual_stream}}
\end{figure*}

HCRN architecture is readily applicable to the accompanying textual
subtitles in a similar bottom-up fashion as in the visual stream.
The HCRN textual stream consists of two-level hierarchical structures
that process textual objects of each segment and join segments into
passage level (See Fig.~\ref{fig:chap5_textual_stream}).

The input of the stream is preprocessed subtitles, question, and answer
choices (Subsec.~\ref{subsec:chap5_Preprocessing}). The subtitles
$\bm{\mathcal{S}}$ is represented as its overall representation $\mathbf{H}^{s}$
and also a sequence of equal-length segments, each of which has been
encoded into a set of textual objects $\mathbf{U}_{i}=\left\{ \mathbf{u}_{i}^{t}\right\} _{t=1,..,T}\in\mathbb{R}^{d\times T}$.
Meanwhile, the question is encoded into a single vector $\mathbf{q}\in\mathbb{R}^{d}$.

\paragraph{Question-relevant pre-selection:}

Unlike video frames, subtitles $\bm{\mathcal{S}}$ contains irregularly
timed conversations between movie characters. Furthermore, while relevant
visual features are abundant throughout the video, only a small portion
of the subtitles is relevant to the query and reflective of the answers
$\mathbf{a}$. To assure such relevance, antecedent to CRN unit, we
modulate the representation of passage $\mathbf{H}^{s}$ and those
of $M$ segments $\left\{ \mathbf{U}_{i}\right\} _{i=1}^{M}$ with
both use the question and the answer choice. It is done with the \emph{pre-selection}
operator described below.

At the segment level, the modulated representation $\hat{\mathbf{U}}_{i}\in\mathbb{R}^{d\times T}$
of each segment $i$ of $T$ objects are produced by 
\begin{eqnarray}
\hat{\mathbf{u}}_{ij} & = & \mathbf{W}^{u}\left[\mathbf{u}_{ij};\mathbf{u}_{ij}\odot\mathbf{q}\right],\:\text{for}\:j=1,..,T\:\text{and}\:\mathbf{u}_{ij}\in\mathbb{R}^{d}.\label{eq:chap5_preselect_segment_level}
\end{eqnarray}
Similarly, at the video level, the subtitle modulated representation
$\hat{\mathbf{H}^{s}}\in\mathbb{R}^{S\times d}$ is built as\vspace{-1cm}

\begin{eqnarray}
\hat{\mathbf{h}}_{k}^{s} & = & \mathbf{W}^{h}\left[\mathbf{h}_{k}^{s};\mathbf{h}_{k}^{s}\odot\mathbf{q}\right],\:\text{for}\:k=1,..,S\:\text{and}\:\mathbf{h}_{k}^{s}\in\mathbb{R}^{d}.\label{eq:chap5_preselection_subtitle}
\end{eqnarray}

\paragraph{Textual CRN unit:}

A single CRN unit of the stream operates at the passage level, which
models the relationships between segments. The modulated output features
$\left\{ \hat{\mathbf{U}}_{i}\right\} _{i=1}^{M}$ are passed as input
objects to a CRN unit. The conditioning feature of the CRN is the
max-pooled vector of the modulated representation of the whole subtitle
passage $\hat{\mathbf{H}}^{s}$. At the end of the textual stream,
a temporal max-pooling operator is applied over the outputs of the
CRN to obtain a single vector. 

\subsubsection{Adaptation \& Implementation}

\paragraph{Short-form Video QA:}

Recall that short-form Video QA in this thesis refers to QA about
single-shot videos of a few seconds without accompanying textual data.
For these cases, we employ the standard visual stream as described
in Subsec.~\ref{subsec:chap5_Visual-stream} to distil video/question
joint representation. This representation is ready to be used by the
answer decoder (See Sec.~\ref{subsec:chap5_Answer-Decoders}) for
generating the final output.

\paragraph{Long-form Video QA:}

Different from the short-form Video QA, long-form Video QA involves
reasoning on both visual information from video frames and textual
content from subtitles. Compared to short snippets where local low-level
motion saliency plays a critical role, discovering high-level semantic
concepts associated with movie characters is more important \citep{sang2010character}.
Such semantics interleave in the data of both modalities. The further
difference comes from the fact that long-form videos are of greater
duration hence require appropriate processing.

Although long-form videos share similar traits with short-forms in
having a hierarchical structure, they are distinctive in terms of
semantic compositionality and length. We employ visual and textual
streams as described in Subsec.~\ref{subsec:chap5_Visual-stream}
and Subsec.~\ref{subsec:chap5_Textual-stream} with some adaptation
for better suitability with the data format and structure and use
the joint representation of the two streams for answer prediction. 

Ideally, long-form Video QA requires modelling interactions between
a question, visual content and textual content in subtitles. Although
both the visual stream and textual stream described above involve
early integration of the question representation into a visual representation
and textual representation of subtitles, we do not opt for early interaction
between visual content and textual content in subtitles in this thesis.
Pairwise joint semantic representation between visual and language
has proven to be useful in Video QA and closely related tasks \citep{yu2018joint},
however, it assumes the existence of pairs of multimodal sequence
data in which they are highly semantically compatible. Those pairs
are either a video sequence and a textual description of the video
or a video sequence and positive/negative answer choices to a question
about the visual content in the video. This is not always the case
for the visual content and textual content in subtitles in long-form
videos such as those taken from movies. Although the visual content
and textual content may complement each other to some extent, in many
cases, they may be greatly distant from each other. Let's take the
following scenario as an example: two characters are standing and
chatting with each other in a movie scene. While the visual content
may provide information about where the conversation takes place,
it hardly contains any information about the topic of their conversation.
As a result, combining the visual information and textual information
at an early stage, in this case, has the potential to cause information
distortion and make it difficult for information retrieval. In addition,
treating visual stream and textual stream in separation makes it easier
to justify the benefits of using CRN units in modelling the relational
information in each modality, hence, easier to assess the generic
use of the CRN unit.

Note that in the visual stream in the HCRN architecture for long-form
Video QA, we use CRN units to handle a subset of sub-sampled frames
in each clip at the clip level. At the video level, a CRN gathers
long-range dependencies between this clip-level information sent up
from lower level CRN outputs.

\begin{figure*}
\begin{centering}
\includegraphics[width=0.8\textwidth]{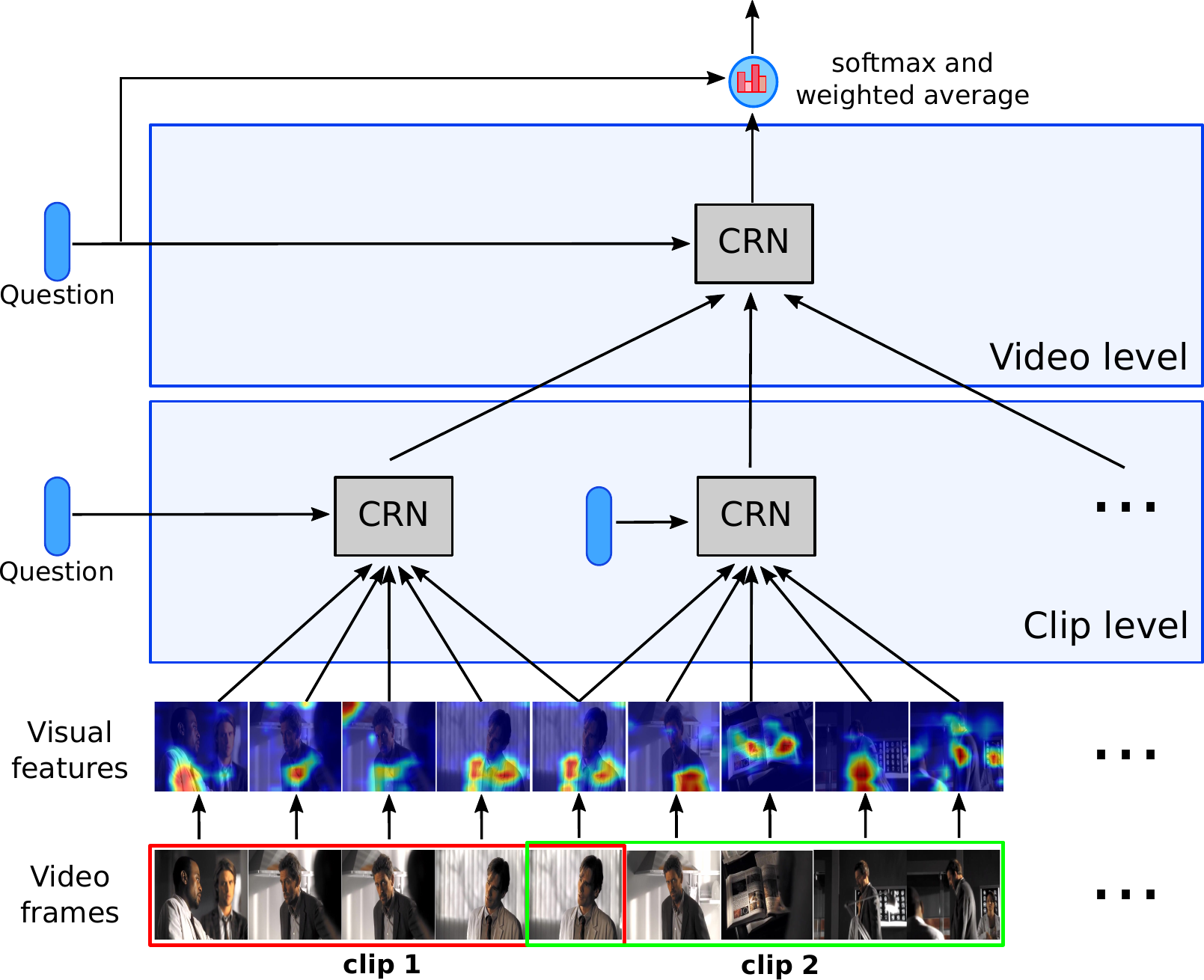}
\par\end{centering}
\caption{The adapted architecture of visual stream (Fig.\ref{fig:chap5_visual_stream})
for long-form Video QA. At each level, only question-conditioned CRNs
are employed. The motion-conditioned CRNs are unnecessary as the low-level
motion features are less relevant in long-form media.\label{fig:chap5_visual_stream-long}}
\end{figure*}

All CRN units at both levels take the question representation as conditioning
features (See Fig. \ref{fig:chap5_visual_stream-long}). Compared
to the standard architecture introduced in Subsec.~\ref{subsec:chap5_Visual-stream}
and Fig. \ref{fig:chap5_visual_stream}, we drop all CRN units that
condition on motion features. This adaptation is to accommodate the
fact that low-level motion is less relevant than overall semantic
flow in both clip- and video-level. 

\subsection{Answer Decoders\label{subsec:chap5_Answer-Decoders}}

Similar to Sec.~\ref{subsec:chap4_Answer-Decoders} in the previous
chapter, we follow previous works \citep{fan2019heterogeneous,jang2017tgif,song2018explore}
to adopt different answer decoders depending on the task, as follows:
\begin{itemize}
\item QA pairs with open-ended answers are treated as multi-class classification
problems. For these, we employ a classifier which takes as input
the combination of the retrieved information from visual stream $\tilde{\mathbf{o}}^{v}$,
the retrieved information from textual stream $\tilde{\mathbf{o}}^{t}$
and the question representation $\mathbf{q}$, and computes answer
probabilities $\mathbf{p}\in\mathbb{R}^{|\mathcal{\bm{A}}|}$:
\begin{eqnarray}
\mathbf{y} & = & \text{ELU}\left(\mathbf{W}^{o}\left[\tilde{\mathbf{o}}^{v};\tilde{\mathbf{o}}^{t};\mathbf{W}^{q}\mathbf{q}+\mathbf{b}\right]+\mathbf{b}\right),\label{eq:chap5_joint-embedding}\\
\mathbf{y}^{\prime} & = & \text{ELU}\left(\mathbf{W}^{y}\mathbf{y}+\mathbf{b}\right),\label{eq:chap5_proj_joint_rep}\\
\mathbf{p} & = & \textrm{softmax}\left(\mathbf{W}^{y^{\prime}}\mathbf{y}^{\prime}+\mathbf{b}\right).
\end{eqnarray}
\item In the case of multi-choice questions where answer choices are available,
we iteratively treat each answer choice $\{\mathbf{a}_{i}\}_{i=1,...,A}$
as a query in exactly the way we handle the question. Eq.~\ref{eq:chap5_preselect_segment_level}
and Eq.~\ref{eq:chap5_preselection_subtitle}, therefore, take each
of the answer choices' representation $\{\mathbf{a}_{i}\}$ as a conditioning
feature along with the question representation. Regarding Eq.~\ref{eq:chap5_joint-embedding},
it is replaced by:
\begin{eqnarray}
\mathbf{y} & = & \text{ELU}\left(\mathbf{W}^{o}\left[\tilde{\mathbf{o}}^{v_{\text{qa}}};\tilde{\mathbf{o}}^{t_{\text{qa}}};\mathbf{W}^{q}\mathbf{q}+\mathbf{b};\mathbf{W}^{a}\mathbf{a}+\mathbf{b}\right]+\mathbf{b}\right),\label{eq:chap5_joint-embedding-mc}
\end{eqnarray}
where $\tilde{\mathbf{o}}^{v_{\text{qa}}}$ is output of visual stream
respect to queries as question and answer choices, whereas $\tilde{\mathbf{o}}^{t_{\text{qa}}}$
is output of the textual stream counterpart.
\item For short-form Video QA, we simply drop the retrieved information
from textual stream $\tilde{\mathbf{o}}_{t}$ in Eq.~\ref{eq:chap5_joint-embedding}
and Eq.~\ref{eq:chap5_joint-embedding-mc}. We also use the popular
hinge loss as what presents in \citep{jang2017tgif} for pairwise
comparisons, $\text{max}\left(0,1+s^{n}-s^{p}\right)$, between scores
for incorrect $s^{n}$ and correct answers $s^{p}$ to train the network:
\begin{eqnarray}
s & = & \mathbf{W}^{y^{\prime}}\mathbf{y}^{\prime}+b,
\end{eqnarray}
where $\mathbf{W}^{y^{\prime}}\in\mathbb{R}^{1\times d}$ is a weight
matrix. Regarding long-form Video QA, we use the cross entropy as
the training loss for fair comparisons with prior works.
\item For repetition count task, we use a linear regression function taking
$\mathbf{y^{\prime}}$ in Eq.~\ref{eq:chap5_proj_joint_rep} as input,
followed by a rounding function for integer count results. The loss
for this task is Mean Squared Error (MSE).
\end{itemize}

\subsection{Complexity Analysis \label{subsec:chap5_Complexity-Analysis}}

We now provide a theoretical analysis of running time for CRN units
and HCRN. We will show that adding layers saves computation time,
just providing a justification for deep hierarchy.

\subsubsection{CRN Units}

For clarity, let us recall the notations introduced in our CRN units:
$k_{\text{max}}$ is maximum subset (also tuple) size considered from
a given input array of $n$ objects, subject to $k_{\text{max}}<n$;
$t$ is number of size-$k$ subsets, ($k=2,3,...,k_{\text{max}}$),
randomly sampled from the input set; $g^{k}(.),h^{k}(.,.)$ and $p^{k}(.)$
are sub-networks for relation modelling, conditioning and aggregating,
respectively. In our implementation, $g^{k}(.)$ and $p^{k}(.)$ are
chosen to be set functions and $h^{k}(.,.)$ is a nonlinear transformation
that fuses modalities.

Each input object in CRN is arranged into a matrix of size $K\times F$,
where $K$ is the number of object elements and $F$ is the embedding
size for each element. All the operations involving input objects
are element-wise, that is, they are linear in time w.r.t. $K$. Assume
that the set function of order $k$ in the CRN's operation in Alg.
1 has linear time complexity in $k$. This holds true for most aggregation
functions such as mean, max, sum or product. With the relation orders
ranging from $k=2,3,...,k_{\text{max}}$ and sampling frequency $t$,
inference cost in time for a CRN is:
\begin{eqnarray}
\text{cost}_{CRN}\left(t,k_{\text{max}},K,F\right) & = & \text{cost}(g)+\text{cost}(h),\label{eq:chap5_CRN-complexity}
\end{eqnarray}
where,
\begin{eqnarray*}
\text{cost}(g) & = & \mathcal{O}\left(\frac{t}{2}k_{\text{max}}(k_{\text{max}}-1)KF\right),\\
\text{cost}(h) & = & \mathcal{O}\left((4t+2)(k_{max}-1)KF^{2}\right).
\end{eqnarray*}
Here the running time of $\text{cost}(g)$ is quadratic in length
because each $g(.)$ that takes $k$ objects as input will cost $k$
time, for $k=2,3,...,k_{\text{max}}$. The running time of $\text{cost}(g)$
is quadratic in $F$ due to the feature transformation operation that
costs $F^{2}$ time. When $k_{\text{max}}\ll F$, $\text{cost}(h)$
will dominate the time complexity. However, since the function $h(.)$
involves matrix operations only, it is usually fast.

The unit produces an output array of length $k_{max}-1$, where each
output object is of the same size as the input objects.

\subsubsection{HCRN Models}

We adhere to the complexity analysis of visual stream only which increases
linearly in complexity with a video length. The overall complexity
of HCRN depends on the design choice for each CRN unit and the specific
arrangement of CRN units. For clarity, let $t=2$ and $k_{\max}=n-1$,
which are found to work well in experiments. Let $L$ be the video
length, organised into $N$ clips of length $T$ each, i.e., $L=NT$.

\paragraph{2-level HCRN:}

Consider, for example, the 2-level architecture HCRN, representing
clips and video. Each level is a stack of two CRN layers, one for
motion conditioning followed by the other for linguistic conditioning.
The clip-level CRNs cost $N\times\text{cost}_{CRN}\left(2,T-1,1,F\right)$
time for motion conditioning and $N\times\text{cost}_{CRN}\left(2,T-3,1,F\right)$
time for question conditioning, where $\text{cost}_{CRN}$ is the
cost estimator in Eq.~(\ref{eq:chap5_CRN-complexity}). This adds
to roughly $\mathcal{O}\left(2TLF\right)+\mathcal{O}\left(10LF^{2}\right)$
time.

Now the output array of size $(T-4)\times F$ for the question-conditioned
clip-level CRN becomes one in $N$ input objects the video-level CRNs.
The CRNs at the video level, therefore, take a cost of $\text{cost}_{CRN}\left(2,N-1,T-4,F\right)$
time for the motion-conditioned one and $\text{cost}_{CRN}\left(2,N-3,T-4,F\right)$
time for the question-conditioned one, respectively, totalling $\mathcal{O}\left(2NLF\right)+\mathcal{O}\left(10LF^{2}\right)$
in order. Here we have made use of the identity $L=NT$. The total
cost is therefore in the order of $\mathcal{O}\left(2(T+N)LF\right)+\mathcal{O}\left(20LF^{2}\right)$.

\paragraph{3-level HCRN:}

Let us now analyse a 3-level architecture HCRN that generalises the
2-level HCRN. The $N$ clips are organised into $P$ sub-videos, each
has $Q$ clips, i.e., $N=PQ$. Since the CRNs at clip level remain
the same, the first level costs $2TLF$ time to compute as before.
Moving to the next level, each sub-video CRN takes as input an array
of length $Q$, whose elements have size $(T-4)\times F$. Applying
the same logic as before, the set of sub-video-level CRNs cost roughly
$P\times\text{cost}_{CRN}\left(2,Q-1,T-4,F\right)$ time or approximately
$\mathcal{O}\left(2\frac{N}{P}LF\right)+\mathcal{O}\left(10LF^{2}\right)$.
Here we have used the identities $N=PQ$ and $L=NT$.

A stack of two sub-video CRNs now produces an output array of size
$(Q-4)(T-4)\times F$, serving as an input object in an array of length
$P$ for the video-level CRNs. Thus the video-level CRNs cost roughly
\begin{eqnarray*}
\text{cost}_{CRN}\left(2,P-1,(Q-4)(T-4),F\right)
\end{eqnarray*}
time or approximately $\mathcal{O}\left(2PLF\right)+\mathcal{O}\left(10LF^{2}\right)$
. Here we again have used the identities $L=NT$ and $N=PQ$. Thus
the total cost is in the order of $\mathcal{O}\left(2(T+\frac{N}{P}+P)LF\right)+\mathcal{O}\left(30LF^{2}\right)$.

\subsubsection*{Deeper models might save time for long videos:}

Recall that the 2-level HCRN has time cost of 
\begin{eqnarray*}
\mathcal{O}\left(2(T+N)LF\right)+\mathcal{O}\left(20LF^{2}\right),
\end{eqnarray*}
and the 3-level HCRN the cost of 
\begin{eqnarray*}
\mathcal{O}\left(2(T+\frac{N}{P}+P)LF\right)+\mathcal{O}\left(30LF^{2}\right).
\end{eqnarray*}
Here the cost that is linear in $F$ is due to the $g$ functions,
and the quadratic cost in $F$ is due to the $h$ functions.

When going from 2-level to 3-level architectures, the cost for the
\emph{$g$ }functions\emph{ drops} by $\mathcal{O}\left(2(N-\frac{N}{P}-P)LF\right)$,
and the cost for the \emph{$h$} functions\emph{ increases} by $\mathcal{O}\left(10LF^{2}\right)$.
Now assuming $N\gg\max\left\{ P,\frac{N}{P}\right\} $, for example
$P\approx\sqrt{N}$ and the number of clips $N>20$. Then the linear
drop can be approximated further as $\mathcal{O}\left(2NLF\right)$.
As $N=\frac{L}{T}$, this can be written as $\mathcal{O}\left(2\frac{L^{2}}{T}F\right)$.
In practice the clip size $T$ is often fixed, thus the drop in the
$g$ functions scales quadratically with video length $L$, whereas
the increase in the $h$ functions scales linearly with $L$. This
suggests that \emph{going deeper in hierarchy could actually save
the running time for long videos}.

See Subsec.~\ref{subsec:chap5_Deepening-model-hierarchy} for empirical
validation of the saving.\selectlanguage{american}%

\section{Experiments \label{sec:Chap5-Experiments}}

\selectlanguage{australian}%

\subsection{Datasets}

We evaluate the effectiveness of the proposed CRN unit and the HCRN
architecture on a series of short-form and long-form Video QA datasets.
In particular, we use three different datasets as benchmarks for the
short-form Video QA, namely TGIF-QA \citep{jang2017tgif}, MSVD-QA
\citep{xu2017video} and MSRVTT-QA \citep{xu2016msr}. All those three
datasets are collected from real-world videos. We also evaluate HCRN
on long-form Video QA using one of the largest datasets publicly available,
TVQA \citep{lei2018tvqa}. Details of each benchmark are as below.

\paragraph{TGIF-QA:}

This is currently the most prominent dataset for Video QA, containing
165K QA pairs and 72K animated GIFs. The dataset covers four tasks
addressing the unique properties of video data. Of which, the first
three require strong spatio-temporal reasoning abilities\emph{: }\textbf{\emph{Repetition
Count}} - to retrieve the number of occurrences of an action, \textbf{\emph{Repeating
Action}} - multi-choice task to identify the action that is repeated
for a given number of times, \textbf{\emph{State Transition}} - multi-choice
tasks regarding temporal order of events. The last task - \textbf{\emph{Frame
QA}} - is akin to image QA where the answer to a given question can
be found from one particular video frame. Please see Sec. \ref{subsec:chap4_Datasets}
for more details of the TGIF-QA dataset.

\paragraph{MSVD-QA:}

This is a small dataset of 50,505 question answer pairs annotated
from 1,970 short clips. Questions are of five types, including what,
who, how, when and where, of which 61\% of questions are used for
training whilst 13\% and 26\% are used as the validation set and test
set, respectively.

\paragraph{MSRVTT-QA:}

The dataset contains 10K videos and 243K question answer pairs. Similar
to MSVD-QA, questions are of five types. Splits for train, validation
and test are with the proportions are 65\%, 5\%, and 30\%, respectively.
Compared to the other two datasets, videos in MSRVTT-QA contain more
complex scenes. They are also much longer, ranging from 10 to 30 seconds
long, equivalent to 300 to 900 frames per video.

\paragraph{TVQA:}

This is one of the largest long-form Video QA datasets annotated from
6 different TV shows:\emph{ The Big Bang Theory}, \emph{How I Met
Your Mother}, \emph{Friends}, \emph{Grey's Anatomy}, \emph{House},
\emph{Castle}. There are total 152.5K question-answer pairs associated
with 5 answer choices each from 21,8K long clips of 60/90 secs which
comes down to 122K, 15,25K and 15,25K for train, validation and test
set, respectively. The dataset also provides start and end timestamps
for each question to limit the video portion where one can find corresponding
answers.

Regarding the evaluation metric, we mainly use accuracy in all experiments,
excluding those for repetition count on TGIF-QA dataset where Mean
Square Error (MSE) is used.

\subsection{Implementation Details}

\subsubsection{Feature Extraction}

For short-form Video QA datasets, each video is preprocessed into
$N$ short clips of fixed lengths, 16 frames each. In detail, we first
locate $N$ equally spaced anchor frames. Each clip is then defined
as a sequence of 16 consecutive video frames taking a pre-computed
anchor as the middle frame. For the first and the last clip where
frame indices may exceed the boundaries, we repeat the first frame
or the last frame of the video multiple times until it fills up the
clip\textquoteright s size.

Given segmented clips, we extract motion features for each clip using
a pre-trained model of the ResNeXt-101\footnote{https://github.com/kenshohara/video-classification-3d-cnn-pytorch}
\citep{xie2017aggregated,hara2018can}. Regarding the appearance feature
used in the experiments, we take the \emph{pool5} output of ResNet
\citep{he2016deep} features as a feature representation of each frame.
This means we completely ignore the 2D structure of spatial information
of video frames which is likely to be beneficial for answering questions
particularly interested in object's appearance, such as those in the
Frame QA task in the TGIF-QA dataset. We are aware of this but deliberately
opt for light-weighted extracted features, and drive the main focus
of this chapter on the significance of temporal relation, motion,
and the hierarchy of video data by nature. Note that most of the videos
in the datasets in the short-form Video QA category, except those
in the MSRVTT-QA dataset, are short. Hence, we intentionally divide
each video into 8 clips (8$\times$16 frames) to produce partially
overlapping frames between clips to avoid temporal discontinuity.
Longer videos in MSRVTT-QA are additionally segmented into 24 clips
of 16 frames each, primarily aiming at evaluating the model's ability
to handle very long sequences.

For the TVQA long-form dataset, we did a similar strategy as for short-video
where we divide each video into$N$ clips. However, as TVQA videos
are longer and only recorded at 3fps, we adapt by choosing $N=6$,
each clip contains 8 frames based on empirical experiences. The \emph{pool5}
output of ResNet features is also used as the feature representation
of each frame.

For the subtitles, the maximum subtitle's length is set at $256$.
We simply cut off those who are longer than that and do zero paddings
for those who are shorter. Subtitles are further divided into 6 segments
overlapping at half a segment size.

\subsubsection{Network Training}

HCRN and its variations are implemented in Python 3.6 with Pytorch
1.2.0. Common settings include $d=512$, $t=2$ for both visual and
textual streams. For all experiments, we train the model using Adam
optimiser with a batch size of 32 , initially at a learning rate of
$10^{-4}$ and decay by half after every 5 epochs for counting task
in the TGIF-QA dataset and after every 10 epochs for the others. All
experiments are terminated after 25 epochs and reported results are
at the epoch giving the best validation accuracy. Depending on the
amount of training data and hierarchy depth, it may take around 4-30
hours of training on one single NVIDIA Tesla V100 GPU. Pytorch implementation
of the model is publicly available. \footnote{https://github.com/thaolmk54/hcrn-videoqa}.

As for experiments with the large-scale language representation model
BERT, we use the latest pre-trained model provided by Hugging Face\footnote{https://github.com/huggingface/transformers}.
We fine-tune the BERT model during training with a learning rate of
$2\times10^{-5}$ in all experiments.

\subsection{Quantitative Results\label{subsec:chap5_Quantitative-results}}

We compare our proposed model with state-of-the-art methods (SOTAs)
on the aforementioned datasets. By default, we use pre-trained GloVe
embedding \citep{pennington2014glove} for word embedding following
by a BiLSTM for sequential modelling. Experiments using contextual
embeddings by a pre-trained BERT network \citep{devlin2018bert} is
explicitly specified with\emph{ ``(BERT)''}. Detailed experiments
for short-form Video QA and long-form Video QA are as the following.

\subsubsection{Short-form Video QA}

For TGIF-QA, we compare with most recent SOTAs, including \citep{fan2019heterogeneous,gao2018motion,jang2017tgif,li2019beyond},
over four tasks. These works, except for \citep{li2019beyond}, make
use of motion features extracted from either optical flow or 3D CNNs
and its variants.

\begin{figure}
\begin{centering}
\includegraphics[width=0.6\textwidth]{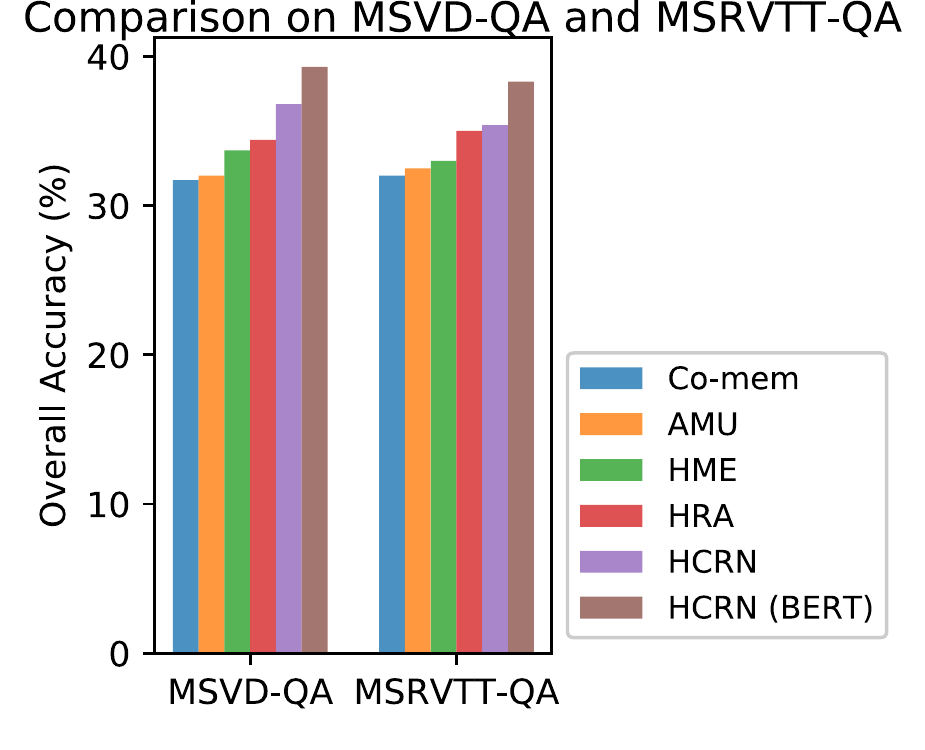}
\par\end{centering}
\caption{Performance comparison on MSVD-QA and MSRVTT-QA dataset with state-of-the-art
methods: Co-mem \citep{gao2018motion}, HME \citep{fan2019heterogeneous},
HRA \citep{chowdhury2018hierarchical}, and AMU \citep{xu2017video}.
\label{fig:chap5_msvd-msrvtt}}
\medskip{}
\end{figure}

The results are summarised in Table~\ref{tab:chap5_tgif} for TGIF-QA,
and in Fig.~\ref{fig:chap5_msvd-msrvtt} for MSVD-QA and MSRVTT-QA.
Results of the previous works are taken from either the original papers
or what reported by \citet{fan2019heterogeneous}. It is clear that
our model consistently outperforms or achieves competitive performance
with the SOTA models on all tasks across all datasets. The improvements
are particularly noticeable when strong temporal reasoning is required,
i.e., for the questions involving actions and transitions in TGIF-QA.
These results confirm the significance of the modelling of both near-term
and far-term temporal relations toward finding correct answers. In
addition, results on the TGIF-QA dataset over 10 runs of different
random seeds confirm the robustness of the CRN units against the randomness
in input subsets selection. More analysis on how relations affect
the model's performance is provided in later ablation studies.

Regarding results with the recent advance in language representation
model BERT, it does not have much effect on the results across tasks
requiring strong temporal reasoning in the TGIF-QA. This can be explained
by the fact that questions in this dataset are relatively short and
all questions are created from a limited number of patterns, hence,
contextual embeddings do not account much benefit. Whereas the Frame
QA task relies on much richer vocabulary, thanks to its free-form
questions, hence, contextual embeddings extracted by BERT greatly
boost the model's performance on this task. We also empirically find
out that fine-tune only the last two layers of the BERT network gives
more favourable performance comparing to fine-tuning all layers.

The MSVD-QA and MSRVTT-QA datasets represent highly challenging benchmarks
for machine compared to the TGIF-QA, thanks to their open-ended nature.
Our model HCRN outperforms existing methods on both datasets, achieving
36.8\% and 35.4\% accuracy which are 2.4 points and 0.4 points improvement
on MSVD-QA and MSRVTT-QA, respectively. This suggests that the model
can handle both small and large datasets better than existing methods.
We also fine-tune the contextual embeddings by BERT on these two
datasets. Results show that the contextual embeddings bring great
benefits on both MSVD-QA and MSRVTT-QA, achieving 39.3\% and 38.3\%
accuracy, respectively. These figures are consistent with the result
on the Frame QA task of the TGIF-QA dataset. Please note that we also
fine-tune only last two layers of the BERT network in these experiments
simply due to favourable empirical results.

Finally, we provide a justification for the competitive performance
of our HCRN against existing rivals by comparing model features in
Table~\ref{tab:chap5_Model-design-choices}. Whilst it is not straightforward
to compare head-to-head on internal model designs, it is evident that
effective video modelling necessitates handling of motion, temporal
relation and hierarchy at the same time. We will back this hypothesis
by further detailed studies in Sec.~\ref{subsec:chap5_Ablation-Studies}
(for motion, temporal relations, shallow hierarchy) and Subsec.~\ref{subsec:chap5_Deepening-model-hierarchy}
(deep hierarchy).

\subsubsection{Long-form Video QA}

For TVQA, we compare our method to several baselines and recent state-of-the-art
methods (See Table~\ref{tab:chap5_tvqa}). The dataset is relatively
new, and due to the challenges with long-form Video QA, there are
only several attempts in benchmarking it. Comparisons of the proposed
method with the baselines and other methods are made on the validation
set. Results of compared methods are taken as reported in their original
publications. Experiments using timestamp information are indicated
by \emph{w/ ts} and those with full-length subtitles are with \emph{w/o
ts. }If not indicated explicitly, ``V.'', ``S.'', ``Q.'' short
for visual features, subtitle features and question features, respectively.
The evaluated settings are as follows:
\begin{itemize}
\item \emph{(B) Q. }: This is the simplest baseline where we only use question
features for predicting the correct answer. Results in this setting
will reveal how much our model relies on the linguistic bias to arrive
at correct answers.
\item \emph{(B) S. + Q. (w/o pre-selection)}: In this baseline, we use question
and subtitle features for prediction. Both question and subtitle features
are simply obtained by concatenating the final output hidden states
of forward and backward LSTM passes and further pass to a classifier
as in Sec.~\ref{subsec:chap5_Answer-Decoders}.
\item \emph{(B) S. + Q. (w/ pre-selection)}: This baseline is to show the
significance of pre-selection in the textual stream as in Eq.~\ref{eq:chap5_preselection_subtitle}.
Question features and selective output between the question and the
subtitles as explained in Eq.~\ref{eq:chap5_preselection_subtitle}
are fed into a classifier for prediction.
\item \emph{(B) V. + Q.}: In this baseline, we simply apply average pooling
to squeeze the visual features of entire videos into a vector and
further combine it with question features for prediction.
\item \emph{(B) S. + V. + Q. (w/ pre-selection)}: We combine two baselines
\emph{``(B) S. + Q. (w/ pre-selection)''} and \emph{``(B) V. +
Q.''} right above. Given that, subtitle features are extracted with
pre-selection as in Eq.~\ref{eq:chap5_preselection_subtitle} and
video features are smashed over space-time before going through a
classifier.
\item \emph{(HCRN) S. + Q.}: This is to evaluate the effect of the textual
stream alone in our proposed network architecture as described in
Fig.~\ref{fig:chap5_textual_stream}.
\item \emph{(HCRN) V. + Q.}: This is to evaluate the effect of the visual
stream counterpart alone in Fig.~\ref{fig:chap5_visual_stream-long}.
\item \emph{(HCRN) S. + V. + Q.}: Our full proposed architecture with the
presence of both two modalities.
\end{itemize}
As shown in Table~\ref{tab:chap5_tvqa}, using BERT for contextual
word embeddings significantly improves performance comparing to GloVe
embeddings and BiLSTM counterpart. The results are consistent with
what reported by \citet{yang2020bert}. Although our model only achieves
competitive performance with \citet{yang2020bert} in the setting
of using timestamps (71.3 vs. 72.1), we significantly outperform them
by 3.0 absolute points on the more challenging setting when we do
not have access to timestamp indicators of where answers located.
This clearly shows that our bottom-up approach is promising to solve
long-form Video QA in its general setting. 

Even though BERT is designed to do contextual embedding which pre-computes
some word relations, HCRN still shows additional benefit in modelling
relations between segments in the passage. However, this benefit is
not as clearly demonstrated as in the case of using GloVe embedding
coupled with BiLSTM (Table~\ref{tab:chap5_tvqa} - Exp. 9 vs. Exp.
7 and Exp. 13 vs. Exp. 14). To deeper understanding the behaviour
of HCRN, we will concentrate our further analysis using comparison
with GloVe + LSTM.

It clearly shows that using output hidden states of BiLSTM to represent
subtitle features directly for classification has a very limited effect
(See Table~\ref{tab:chap5_tvqa} - Exp. 3 vs Exp. 4). The results
could be explained as LSTM fails to handle such long subtitles of
hundreds of words. In the meantime, pre-selection plays a critical
role to find relevant information in the subtitles to a question which
boosts performance from 42.3\% to 57.9\% when using full subtitles
and from 41.8\% to 62.1\% when using timestamps (See Exp. 5). Our
HCRN further improves approximately 1.0 points (without timestamp
annotation) and 1.5 points (with timestamp annotation) comparing to
the baseline of GloVe embeddings and BiLSTM with pre-selection for
the textual stream only setting. As for the effects on visual stream,
our HCRN gains around 1.0 points over the baseline of averaging visual
features. This leads to an improvement of north of 1.0 points when
leveraging both visual and textual stream together (See Exp. 12 vs.
Exp. 15) on both settings with timestamp annotation and without timestamp
annotation. Even though our results are behind the method developed
by \citet{lei2018tvqa}, we wish to point out that their context matching
model with the bi-directional attention flow (BiDAF) \citep{seo2016bidirectional}
significantly contributes to the performance. Whereas, our model only
makes use of vanilla BiLSTM to sequence modelling. Therefore, a direct
comparison between the two methods is unfair. We instead mainly compare
our method with a simple variant of \citet{lei2018tvqa} by dropping
off the BiDAF to show the contribution of our CRN unit as well as
the HCRN network. This is equivalent to the baseline \emph{(B) S.
+ V. + Q. (w/ pre-selection) }in this thesis.

\begin{table*}
\begin{centering}
\begin{tabular}{l|cccc}
\hline 
{\small{}Model} & {\small{}Action} & {\small{}Trans.} & {\small{}Frame} & {\small{}Count}\tabularnewline
\hline 
\hline 
{\small{}ST-TP \citep{jang2017tgif}} & {\small{}62.9} & {\small{}69.4} & {\small{}49.5} & {\small{}4.32}\tabularnewline
{\small{}Co-mem \citep{gao2018motion}} & {\small{}68.2} & {\small{}74.3} & {\small{}51.5} & {\small{}4.10}\tabularnewline
{\small{}PSAC \citep{li2019beyond}} & {\small{}70.4} & {\small{}76.9} & {\small{}55.7} & {\small{}4.27}\tabularnewline
{\small{}HME \citep{fan2019heterogeneous}} & {\small{}73.9} & {\small{}77.8} & {\small{}53.8} & {\small{}4.02}\tabularnewline
\hline 
{\small{}HCRN{*}} & \textbf{\small{}75.2}{\small{}$\pm$0.4} & \textbf{\small{}81.3}{\small{}$\pm$0.2} & {\small{}55.9$\pm$0.3} & \textbf{\small{}3.93}{\small{}$\pm$0.03}\tabularnewline
{\small{}HCRN (embeddings with BERT)} & {\small{}69.8} & {\small{}79.8} & \textbf{\small{}57.9} & {\small{}3.96}\tabularnewline
\hline 
\end{tabular}\medskip{}
\par\end{centering}
\caption{Comparison with the state-of-the-art methods on TGIF-QA dataset. For
count, the lower the better (MSE) and the higher the better for the
others (accuracy). {*}Means with standard deviations over 10 runs.\label{tab:chap5_tgif}}
\end{table*}

\begin{table}
\begin{centering}
{\scriptsize{}}%
\begin{tabular}{l|cccc}
\hline 
{\small{}Model} & {\small{}Appearance} & {\small{}Motion} & {\small{}Hierarchy} & {\small{}Relation}\tabularnewline
\hline 
\hline 
{\small{}ST-TP \citep{jang2017tgif}} & {\small{}$\checkmark$} & {\small{}$\checkmark$} &  & \tabularnewline
{\small{}Co-mem \citep{gao2018motion}} & {\small{}$\checkmark$} & {\small{}$\checkmark$} &  & \tabularnewline
{\small{}PSAC \citep{li2019beyond}} & {\small{}$\checkmark$} &  &  & \tabularnewline
{\small{}HME \citep{fan2019heterogeneous}} & {\small{}$\checkmark$} & {\small{}$\checkmark$} &  & \tabularnewline
{\small{}HCRN} & {\small{}$\checkmark$} & {\small{}$\checkmark$} & {\small{}$\checkmark$} & {\small{}$\checkmark$}\tabularnewline
\hline 
\end{tabular}\medskip{}
\par\end{centering}
\caption{Model design choices and input modalities in comparison. See Table~\ref{tab:chap5_tgif}
for corresponding performance on TGIF-QA dataset. \label{tab:chap5_Model-design-choices}}
\medskip{}
\end{table}

\begin{table*}
\begin{centering}
\begin{tabular}{l|l|cc}
\hline 
\multirow{2}{*}{{\small{}Exp.}} & \multirow{2}{*}{{\small{}Model}} & \multicolumn{2}{c}{{\small{}Val. Acc.}}\tabularnewline
\cline{3-4} \cline{4-4} 
 &  & {\small{}w/o ts} & {\small{}w/ ts}\tabularnewline
\hline 
\hline 
 & \textbf{\small{}State-of-the-art methods} &  & \tabularnewline
{\small{}1} & {\small{}$\quad$TVQA S.+Q.+V. \citep{lei2018tvqa,yang2020bert}} & {\small{}64.7} & {\small{}67.7}\tabularnewline
{\small{}2} & {\small{}$\quad$VideoQABERT \citep{yang2020bert}} & {\small{}63.1} & \textbf{\small{}72.1}\tabularnewline
\hline 
 & \textbf{\small{}Textual stream only} &  & \tabularnewline
{\small{}3} & {\small{}$\quad$(B) Q.} & {\small{}41.6} & {\small{}41.6}\tabularnewline
{\small{}4} & {\small{}$\quad$(B) S. + Q. (w/o pre-selection)} & {\small{}42.3} & {\small{}41.8}\tabularnewline
{\small{}5} & {\small{}$\quad$(B) S. + Q. (w/ pre-selection)} & {\small{}57.9} & {\small{}62.1}\tabularnewline
{\small{}6} & {\small{}$\quad$(B) Q. (BERT)} & {\small{}44.2} & {\small{}44.2}\tabularnewline
{\small{}7} & {\small{}$\quad$(B) S. + Q. (BERT)} & {\small{}65.1} & {\small{}71.1}\tabularnewline
{\small{}8} & {\small{}$\quad$(HCRN) S. + Q.} & {\small{}59.0} & {\small{}63.6}\tabularnewline
{\small{}9} & {\small{}$\quad$(HCRN) S. + Q. (BERT)} & {\small{}66.0} & {\small{}71.1}\tabularnewline
\hline 
 & \textbf{\small{}Visual stream only} &  & \tabularnewline
{\small{}10} & {\small{}$\quad$(B) V. + Q.} & {\small{}42.2} & {\small{}42.2}\tabularnewline
{\small{}11} & {\small{}$\quad$(HCRN) V. + Q.} & {\small{}43.1} & {\small{}43.1}\tabularnewline
\hline 
 & \textbf{\small{}Two streams} &  & \tabularnewline
{\small{}12} & {\small{}$\quad$(B) S. + V. + Q. (w/ pre-selection)} & {\small{}58.5} & {\small{}63.2}\tabularnewline
{\small{}13} & {\small{}$\quad$(B) S. + V. + Q. (BERT)} & {\small{}65.7} & {\small{}71.4}\tabularnewline
{\small{}14} & {\small{}$\quad$(HCRN) S. + V. + Q. (BERT)} & \textbf{\small{}66.1} & {\small{}71.3}\tabularnewline
{\small{}15} & {\small{}$\quad$(HCRN) S. + V. + Q.} & {\small{}59.1} & {\small{}64.7}\tabularnewline
\hline 
\end{tabular}\medskip{}
\par\end{centering}
\caption{Comparison with baselines and state-of-the-art methods on TVQA dataset.
\emph{w/o ts}: without using timestamp annotation to limit the search
space of where to find answers; \emph{w/ ts}: making use of the timestamp
annotation.\label{tab:chap5_tvqa}}
\end{table*}

\subsection{Ablation Studies \label{subsec:chap5_Ablation-Studies}}

\begin{table*}
\centering{}%
\begin{tabular}{l|cccc}
\hline 
{\small{}Model} & {\small{}Action} & {\small{}Transition} & {\small{}FrameQA} & {\small{}Count}\tabularnewline
\hline 
\hline 
\textbf{\small{}Relations $(k_{max},t)$} &  &  &  & \tabularnewline
{\small{}$\quad$$k_{max}=1,t=1$} & {\small{}72.8} & {\small{}80.7} & {\small{}56.3} & {\small{}3.87}\tabularnewline
{\small{}$\quad$$k_{max}=1,t=3$} & {\small{}73.4} & {\small{}81.1} & {\small{}56.0} & {\small{}3.94}\tabularnewline
{\small{}$\quad$$k_{max}=1,t=5$} & {\small{}74.0} & {\small{}80.3} & {\small{}56.4} & {\small{}3.86}\tabularnewline
{\small{}$\quad$$k_{max}=1,t=7$} & {\small{}74.8} & {\small{}80.0} & {\small{}56.5} & {\small{}3.85}\tabularnewline
{\small{}$\quad$$k_{max}=1,t=9$} & {\small{}73.7} & {\small{}81.1} & {\small{}56.4} & {\small{}3.82}\tabularnewline
{\small{}$\quad$$k_{max}=2,t=2$} & {\small{}72.8} & {\small{}80.8} & {\small{}56.3} & {\small{}3.80}\tabularnewline
{\small{}$\quad$$k_{max}=2,t=9$} & {\small{}73.1} & {\small{}81.4} & {\small{}56.0} & {\small{}3.85}\tabularnewline
{\small{}$\quad$$k_{max}=4,t=2$} & {\small{}74.2} & {\small{}81.5} & {\small{}56.8} & {\small{}3.88}\tabularnewline
{\small{}$\quad$$k_{max}=4,t=9$} & {\small{}73.4} & {\small{}81.8} & {\small{}56.5} & {\small{}3.83}\tabularnewline
{\small{}$\quad$$k_{max}=\left\lfloor n/2\right\rfloor ,t=2$} & {\small{}74.7} & {\small{}81.1} & {\small{}55.7} & {\small{}3.85}\tabularnewline
{\small{}$\quad$$k_{max}=\left\lfloor n/2\right\rfloor ,t=9$} & {\small{}74.7} & {\small{}81.0} & {\small{}55.4} & {\small{}3.94}\tabularnewline
{\small{}$\quad$$k_{max}=n-1,t=1$} & {\small{}74.2} & {\small{}80.8} & {\small{}55.3} & {\small{}3.98}\tabularnewline
{\small{}$\quad$$k_{max}=n-1,t=3$} & {\small{}75.2} & {\small{}81.1} & {\small{}56.2} & {\small{}4.00}\tabularnewline
{\small{}$\quad$$k_{max}=n-1,t=5$} & {\small{}75.0} & {\small{}81.3} & {\small{}55.6} & {\small{}3.95}\tabularnewline
{\small{}$\quad$$k_{max}=n-1,t=7$} & {\small{}75.3} & {\small{}81.9} & {\small{}55.8} & {\small{}4.00}\tabularnewline
{\small{}$\quad$$k_{max}=n-1,t=9$} & {\small{}74.9} & {\small{}81.1} & {\small{}55.6} & {\small{}3.94}\tabularnewline
{\small{}$\quad$Fix $k=k_{max},k_{max}=n-1,t=2$} & {\small{}72.9} & {\small{}80.2} & {\small{}56.5} & {\small{}3.90}\tabularnewline
\hline 
\textbf{\small{}Hierarchy} &  &  &  & \tabularnewline
{\small{}$\quad$$1$-level, video CRN only} & {\small{}72.7} & {\small{}81.4} & {\small{}57.2} & {\small{}3.88}\tabularnewline
{\small{}$\quad$$1.5$-level, clips$\rightarrow$pool} & {\small{}73.1} & {\small{}81.2} & {\small{}57.2} & {\small{}3.88}\tabularnewline
\hline 
\textbf{\small{}Motion conditioning} &  &  &  & \tabularnewline
{\small{}$\quad$w/o motion} & {\small{}69.8} & {\small{}78.4} & {\small{}57.9} & {\small{}4.38}\tabularnewline
{\small{}$\quad$w/o short-term motion} & {\small{}74.1} & {\small{}80.9} & {\small{}56.4} & {\small{}3.87}\tabularnewline
{\small{}$\quad$w/o long-term motion} & {\small{}75.8} & {\small{}80.8} & {\small{}56.8} & {\small{}3.97}\tabularnewline
\hline 
\textbf{\small{}Linguistic conditioning} &  &  &  & \tabularnewline
{\small{}$\quad$w/o linguistic condition} & {\small{}68.7} & {\small{}80.5} & {\small{}56.6} & {\small{}3.92}\tabularnewline
{\small{}$\quad$w/o question at the clip level} & {\small{}75.0} & {\small{}81.0} & {\small{}56.0} & {\small{}3.85}\tabularnewline
{\small{}$\quad$w/o question at the video level} & {\small{}74.2} & {\small{}81.0} & {\small{}55.3} & {\small{}3.90}\tabularnewline
\hline 
\textbf{\small{}Multiplicative relation} &  &  &  & \tabularnewline
{\small{}$\quad$w/o MUL. in all CRNs} & {\small{}74.0} & {\small{}81.7} & {\small{}55.8} & {\small{}3.86}\tabularnewline
{\small{}$\quad$w/ MUL. for both question and motion} & {\small{}75.4} & {\small{}80.2} & {\small{}55.1} & {\small{}3.98}\tabularnewline
\hline 
\textbf{\small{}Subset sampling} &  &  &  & \tabularnewline
{\small{}$\quad$Sampled from pre-computed superset} & {\small{}75.2} & {\small{}81.3} & {\small{}55.9} & {\small{}3.90}\tabularnewline
{\small{}$\quad$Directly sampled} & {\small{}75.3} & {\small{}81.8} & {\small{}55.3} & {\small{}3.89}\tabularnewline
\hline 
\end{tabular}\medskip{}
\caption{Ablation studies on TGIF-QA dataset. For count, the lower the better
(MSE) and the higher the better for the others (accuracy). When not
explicitly specified, we use $k_{max}=n-1,t=2$ for relation order
and sampling resolution. \label{tab:chap5_Ablation-tgif}}
\medskip{}
\end{table*}

\begin{table*}
\begin{centering}
\begin{tabular}{l|l|cc}
\hline 
\multirow{2}{*}{{\small{}Exp.}} & \multirow{2}{*}{{\small{}Model}} & \multicolumn{2}{c}{{\small{}Val. Acc.}}\tabularnewline
\cline{3-4} \cline{4-4} 
 &  & {\small{}w/o ts} & {\small{}w/ ts}\tabularnewline
\hline 
\hline 
 & \textbf{\small{}Textual stream only} &  & \tabularnewline
{\small{}1} & {\small{}S. + Q. (w/ MUL. + w/ LSTM)} & {\small{}59.0} & {\small{}63.6}\tabularnewline
{\small{}2} & {\small{}S. + Q. (w/ MUL. + w/o LSTM)} & {\small{}57.1} & {\small{}63.6}\tabularnewline
{\small{}3} & {\small{}S. + Q. (w/o MUL. + w/ LSTM)} & {\small{}59.1} & {\small{}63.9}\tabularnewline
{\small{}4} & {\small{}S. + Q. (w/o MUL. + w/o LSTM)} & {\small{}57.4} & {\small{}63.5}\tabularnewline
\hline 
 & \textbf{\small{}Visual stream only} &  & \tabularnewline
{\small{}5} & {\small{}V. + Q. (w/ MUL. + w/ LSTM)} & {\small{}42.7} & {\small{}42.7}\tabularnewline
{\small{}6} & {\small{}V. + Q. (w/ MUL. + w/o LSTM)} & {\small{}43.1} & {\small{}43.1}\tabularnewline
{\small{}7} & {\small{}V. + Q. (w/o MUL. + w/ LSTM)} & {\small{}42.2} & {\small{}42.2}\tabularnewline
{\small{}8} & {\small{}V. + Q. (w/o MUL. + w/o LSTM)} & {\small{}42.1} & {\small{}42.1}\tabularnewline
\hline 
 & \textbf{\small{}Two streams} &  & \tabularnewline
{\small{}9} & {\small{}S. + V. + Q. (S. w/ MUL. + LSTM, V. w/ MUL.)} & {\small{}59.1} & {\small{}64.7}\tabularnewline
\hline 
\end{tabular}\medskip{}
\par\end{centering}
\caption{Ablation studies on TVQA dataset.\label{tab:chap5_tvqa-ablation}}
\medskip{}
\end{table*}

To provide more insights about the roles of CRN's components, we conduct
extensive ablation studies on the TGIF-QA and TVQA dataset with a
wide range of configurations.\emph{ }The results are reported in Table~\ref{tab:chap5_Ablation-tgif}
for the short-form Video QA and in Table\ \ref{tab:chap5_tvqa-ablation}
for the long-form Video QA. 

\subsubsection{Short-form Video QA}

Overall, we find that our sampling method does not hurt the HCRN performance
that much while it is much more effective than the one used by \citet{zhou2018temporal}
in terms of complexity. We also find that ablating any of the design
components or CRN units would degrade the performance for temporal
reasoning tasks (actions, transition and action counting). The effects
are detailed as follows.

\paragraph{Effect of relation order $k_{max}$ and resolution $t$:}

Without relations ($k_{max}=1,t=1$) the performance suffers, specifically
on actions and events reasoning whereas counting tends to be better.
This is expected since those questions often require putting actions
and events in relation with a larger context (e.g., what happens before
something else) while motion flow is critical for counting but for
far-term relations. In this case, most of the tasks benefit from increasing
sampling resolution $t$ $(t>1)$ because of better chance to find
a relevant frame as well as the benefits of the far-term temporal
relation learned by the aggregating sub-network $p^{k}(.)$ of the
CRN unit. However, when taking relations into account ($k_{max}>1$),
we find that HCRN is robust against sampling resolution $t$ but
depends critically on the maximum relation order $k_{max}$. The relative
independence w.r.t. $t$ can be due to visual redundancy between frames,
so that resampling may capture almost the same information. On the
other hand, when considering only low-order object relations, the
performance is significantly dropped in action and transition tasks
while it is slightly better for counting and frame QA. These results
confirm that high-order relations are required for temporal reasoning.
As the frame QA task requires only reasoning on a single frame, incorporating
temporal information might confuse the model. Similarly, when the
model only makes use of the high-order relations (\emph{Fix} $k=k_{max},k_{max}=n-1,t=2$),
the performance suffers, suggesting combining both low-order object
relations and high-order object relations is a lot more efficient.

\paragraph{Effect of hierarchy:}

We design two simpler models with only one CRN layer: 
\begin{itemize}
\item $1$\emph{-level, $1$ CRN video on key frames only}: Using only one
CRN at the video-level whose input array consists of key frames of
the clips. Note that video-level motion features are still maintained. 
\item $1.5$-\emph{level, clip CRNs $\rightarrow$} \emph{pooling}: Only
the clip-level CRNs are used, and their outputs are mean-pooled to
represent a given video. The pooling operation represents a simplistic
relational operation across clips. The results confirm that a hierarchy
is needed for high performance on temporal reasoning tasks.
\end{itemize}

\paragraph{Effect of motion conditioning:}

We evaluate the following settings: 
\begin{itemize}
\item \emph{w/o short-term motions}: Remove all CRN units that condition
on the short-term motion features (clip level) in the HCRN.
\item \emph{w/o long-term motions}: Remove the CRN unit that conditions
on the long-term motion features (video level) in the HCRN.\emph{ }
\item \emph{w/o motions}: Remove motion feature from being used by HCRN.
We find that motion, in agreeing with prior arts, is critical to detect
actions, hence computing action count. Long-term motion is particularly
significant for the counting task, as this task requires maintaining
a global temporal context during the entire process. For other tasks,
short-term motion is usually sufficient. E.g. in the action task,
wherein one action is repeatedly performed during the entire video,
long-term context contributes little. Not surprisingly, motion does
not play a positive role in answering questions on single frames as
only appearance information needed.
\end{itemize}

\paragraph{Effect of linguistic conditioning and multiplicative relation:}

Linguistic cues represent a crucial context for selecting relevant
visual artefacts. For that we test the following ablations:
\begin{itemize}
\item \emph{w/o question at the clip level}: Remove the CRN unit that conditions
on question representation at clip level. 
\item \emph{w/o question at the video level}: Remove the CRN unit that conditions
on question representation at video level.
\item \emph{w/o linguistic condition: }Exclude all CRN units conditioning
on linguistic cue while the linguistic cue is still in the answer
decoder. Likewise, the multiplicative relation form offers a selection
mechanism. Thus we study its effect as follows:
\item \emph{w/o MUL.} \emph{in all CRNs}: Exclude the use of multiplicative
relations in all CRN units.
\item \emph{w/ MUL. relation for question and motion}: Leverage multiplicative
relations in all CRN units. 
\end{itemize}
We find that the conditioning question provides an important context
for encoding video. Conditioning features (motion and language), through
the multiplicative relation as in Eq.~\ref{eq:chap5_conditioning_mul},
offers further performance gain in all tasks rather than Frame QA,
possibly by selectively passing question-relevant information up the
inference chain.

\paragraph{Effect of subset sampling}

We conduct experiments with the full model of Fig.~\ref{fig:chap5_visual_stream}
(a) with $k_{max}=n-1,t=2$. The experiment \emph{``Sampled from
pre-computed superset''} refers to our CRN units of using the same
sampling trick as what is in \citep{zhou2018temporal}, where the
set $\mathbf{Q}^{k}$ is sampled from a pre-computed collection of
all possible size-$k$ subsets. \emph{``Directly sampled''}, in
contrast, refers to our sampling method as described in Alg. 1. The
empirical results show that directly sampling size-$k$ subsets from
an input set does not degrade much of the performance of the HCRN
for short-form Video QA, suggesting it a better choice when dealing
with large input sets in size to reduce the complexity of the CRN
units.

\subsubsection{Long-form Video QA}

We focus on studying the effect of different options for the conditioning
sub-network $h^{k}(.,.)$ in Sec.~\ref{subsec:chap5_Relation-Network}
which reflects the flexibility of our model when dealing with different
forms of input modalities. The options are:
\begin{itemize}
\item \emph{S. + Q. (w/ MUL. + w/ LSTM)}: Only textual stream is used.
The conditioning sub-network $h^{k}(.,.)$ is with multiplicative
relation between tuples of segments and conditioning feature, and
coupled with BiLSTM as formulated in Eqs.~(\ref{eq:chap5_conditioning},
\ref{eq:chap5_conditioning_bilstm} and \ref{eq:chap5_max_temporally}).
\item \emph{S. + Q. (w/ MUL. + w/o LSTM)}: Remove the BiLSTM network in
the experiment \emph{S. + Q. (w/ MUL. + w/ LSTM)} to evaluate the
effect of sequential modelling in textual stream.
\item \emph{S. + Q. (w/o MUL. + w/ LSTM)}: The multiplicative relation in
the experiment \emph{S. + Q. (w/ MUL. + w/ LSTM)} is now replaced
with a simple concatenation of conditioning feature and tuples of
segments.
\item \emph{S. + Q. (w/o MUL. + w/o LSTM)}: Remove the BiLSTM network in
the right above experiment\emph{ S. + Q. (w/o MUL. + w/ LSTM)} to
evaluate the effect of when both selective relation and sequential
modelling are missing.
\item \emph{V. + Q. (w/ MUL. + w/ LSTM)}: Only visual stream is under consideration.
We use the multiplicative form as in Eq.~\ref{eq:chap5_conditioning_mul}
for the conditioning sub-network instead of a simple concatenation
operation in Eq.~\ref{eq:chap5_conditioning_concat}. We additionally
use a BiLSTM network for sequential modelling as same as the way we
have done with the textual stream. This is because both visual content
and textual content are temporal sequences by nature.
\item \emph{V. + Q. (w/ MUL. + w/o LSTM)}: Similar to the above experiment
\emph{V. + Q. (w/ MUL. + w/ LSTM)} but without the use of the BiLSTM
network.
\item \emph{V. + Q. (w/o MUL. + w/ LSTM)}: The conditioning sub-network
is simply a tensor concatenation operation. This is to compare against
the one using multiplicative form \emph{V. + Q. (w/ MUL. + w/ LSTM)}.
\item \emph{V. + Q. (w/o MUL. + w/o LSTM)}: Similar to \emph{V. + Q. (w/o
MUL. + w/ LSTM)} but without the use of the BiLSTM network for sequential
modelling.
\item \emph{S. + V. + Q. (S. w/ MUL. + LSTM, V. w/ MUL.)}: Both two streams
are present for prediction. We combine the best option of each network
stream, \emph{S. + Q. (w/ MUL. + w/ LSTM)} for the textual stream
and \emph{V. + Q. (w/ MUL. + w/o LSTM) }for the visual stream for
comparison. 
\end{itemize}
It is empirically shown that the simple concatenation as in Eq.~\ref{eq:chap5_conditioning_concat}
is insufficient to combine conditioning features and output of relation
network in this dataset. In the meantime, the multiplicative relation
between the conditioning features and those relations is a better
fit as it greatly improves the model's performance, especially in
the visual stream. This shows the consistency in empirical results
between long-form Video QA and shot-form Video QA where the relation
between visual content and the query is more about multiplicative
(selection) rather than simple additive. 

On the other side, sequential modelling with BiLSTM is more favourable
for textual stream than visual stream even though both streams are
sequential by nature. This well aligns with our analysis in Sec.~\ref{subsec:chap5_Relation-Network}.

\begin{table}
\begin{centering}
\begin{tabular}{l|cc}
\hline 
\multirow{2}{*}{{\small{}Depth of hierarchy}} & \multicolumn{2}{c}{{\small{}Overall Acc.}}\tabularnewline
\cline{2-3} \cline{3-3} 
 & {\small{}val.} & {\small{}test}\tabularnewline
\hline 
\hline 
{\small{}$2$-level, $24$ clips $\rightarrow$ $1$ vid} & {\small{}35.4} & {\small{}35.5}\tabularnewline
{\small{}$3$-level, $24$ clips $\rightarrow$ $4$ sub-vids $\rightarrow$
$1$ vid} & {\small{}35.1} & {\small{}35.4}\tabularnewline
\hline 
\end{tabular}\medskip{}
\par\end{centering}
\caption{Results for going deeper hierarchy on MSRVTT-QA dataset. Run time
is reduced by factor of $4$ for going from 2-level to 3-level hierarchy.
\label{tab:chap5_scale_analysis-msrvtt}}
\medskip{}
\end{table}

\subsubsection{Deepening Model Hierarchy Saves Time \label{subsec:chap5_Deepening-model-hierarchy}}

We test the scalability of the HCRN on long videos in the MSRVTT-QA
dataset, which are organised into 24 clips (3 times longer than the
other two datasets). We consider two settings:
\begin{itemize}
\item \emph{$2$-level hierarchy,} \emph{$24$ clips}$\rightarrow$$1$
\emph{vid}: The model is as illustrated in Fig.~\ref{fig:chap5_visual_stream},
where 24 clip-level CRNs are followed by a video-level CRN.
\item \emph{$3$-level hierarchy}, $24$ \emph{clips}$\rightarrow$$4$\emph{
sub-vid}s$\rightarrow$$1$ \emph{vid}: Starting from the 24 clips
as in the $2$-level hierarchy, we group 24 clips into 4 sub-videos,
each is a group of 6 consecutive clips, resulting in a $3$-level
hierarchy. These two models are designed to have a similar number
of parameters, approx. 44M.
\end{itemize}
The results are reported in Table~\ref{tab:chap5_scale_analysis-msrvtt}.
Unlike existing methods which usually struggle with handling long
videos, our method is scalable for them by offering deeper hierarchy,
as analysed theoretically in Sec.~\ref{subsec:chap5_Complexity-Analysis}.
The theory suggests that using a deeper hierarchy can reduce the training
time and inference time for HCRN when the video is long. This is validated
in our experiments, where we achieve \emph{4 times reduction in training
and inference time} by going from 2-level HCRN to 3-level counterpart
whilst maintaining competitive performance.\selectlanguage{american}%

\section{Discussion\label{sec:Chap5_discussion}}

\selectlanguage{australian}%
HCRN presents a new class of neural architectures for multimodal Video
QA, pursuing the ease of model construction through reusable uniform
building blocks. Different from temporal attention based approaches,
which put effort into selecting objects, HCRN concentrates on modelling
relations and hierarchy in different input modalities in Video QA.
In the scope of this chapter, we study how to deal with visual content
and subtitles where applicable. The visual and text streams share
a similar structure in terms of near-term, far-term relations and
information hierarchy. The difference in methodology and design choices
between ours and the existing works leads to distinctive benefits
in different scenarios as empirically proven. 

Within the scope of this chapter, we did not consider the audio
channel and early fusion between modalities, leaving these open for
future work. Since the CRN is generic, we envision a design for the
HCRN for the audio stream similar to the visual stream. The modalities
can be combined at any hierarchical level, or any step within the
same level. For example, as the audio, subtitle and visual content
are partly synchronised, we can use a sub-HCRN to represent the three
streams per segment. Alternatively, we can fuse the modalities into
the same CRN as long as the feature representations are projected
onto the same tensor space. Future works will also include object-oriented
representation of videos as these are native to our CRN unit, thanks
to its generality. As the CRN is a relational model, both cross-object
relations and cross-time relations can be modelled. These are likely
to improve the interpretability of the model and get closer to how
human reasons across multiple modalities. CRN units can be further
augmented with attention mechanisms to cover better object selection
ability, so that related tasks such as frame QA in the TGIF-QA dataset
can be further improved.

\selectlanguage{american}%

\section{Closing Remarks\label{sec:Chap5-Remark}}

\selectlanguage{australian}%
We introduced a general-purpose neural unit called Conditional Relation
Network (CRN) and a method to construct hierarchical networks for
multimodal Video QA using the CRN as a building block. The CRN is
a relational transformer that encapsulates and maps an array of tensorial
objects into a new array of relations, all conditioned on a contextual
feature. In the process, high-order relations among input objects
are encoded and modulated by a conditioning feature. This design allows
flexible construction of sophisticated structures such as stacking
and hierarchy and supports iterative reasoning, making it suitable
for QA over multimodal inputs and structured domains like video. The
HCRN was evaluated on multiple Video QA datasets, covering both short-form
Video QA whose questions are mainly about activities/events happening
in a short snippet (TGIF-QA, MSVD-QA, MSRVTT-QA) and long-form Video
QA whose questions are related to both long movie scenes and associated
movie subtitles (TVQA dataset). HCRN demonstrates its competitive
reasoning capability in a wide range of different settings against
state-of-the-art methods. The examination of CRN in Video QA highlights
the importance of building a generic neural reasoning unit that supports
native multimodal interaction in improving robustness of visual reasoning.

Viewing our HCRN under the dual process neural architecture introduced
in Chapter \ref{chap:DualProcess}, it plays the role of System 1
with advanced visual perception by the capability to encode a wider
range of information and relations needed for the reasoning process
than exisiting methods including our own Clip-based Relational Network
in the previous chapter. However, HCRN has not yet considered the
explicit relationships between \emph{semantic objects }(e.g. person,
car, cat etc.). The current implementation of HCRN assumed frame features
at a time step as abstract objects. The following chapter will address
this issue by using the categorical and relational structures of semantic
objects to advance reasoning functionalities. \selectlanguage{american}%

\selectlanguage{australian}%

\newpage{}

\selectlanguage{american}%

\chapter{Relational Visual Reasoning\label{chap:RelationalVisualReasoning}}

\section{Introduction\label{sec:Chap6-Introduction}}

\selectlanguage{australian}%
Human visual reasoning involves analysing linguistic aspects of the
query and continuously inter-linking them with visual objects through
a series of information aggregation steps \citep{lake2017building}.
Artificial reasoning engines mimic this ability by using structured
representations (e.g. scene graphs) \citep{shi2019explainable} to
discover categorical and relational information about visual objects. 

In the previous chapters, we studied the importance of reasoning for
an intelligent agent by asking it to answer questions about the temporal
dynamics in videos. The findings showed that temporal relations, compositionality
and the hierarchical design of the models are vital for machine agents
to perform visual reasoning. However, we neglect the explicit associations
and interactions of cross-modality components in our models. In this
chapter, we focus on exploring the significance of structured representations
of inputs in visual reasoning. For simplicity, we use Image Question
Answering (Image QA) as the testbed for evaluating the benefits of
these structures to neural networks in learning and reasoning. 

We address two key abstractions in this chapter: How can we extend
structured representations of input modalities seamlessly across both
visual-lingual borders? Furthermore, how can we extend these structures
to be dynamic and responsive to the reasoning process rather than
static representations as in prior studies? We explore the dynamic
relational structures of visual scenes that are proactively discovered
within reasoning context and their adaptive connections to the components
of a linguistic query to answer visual questions effectively. 

\begin{figure}
\begin{centering}
\includegraphics[width=0.65\columnwidth]{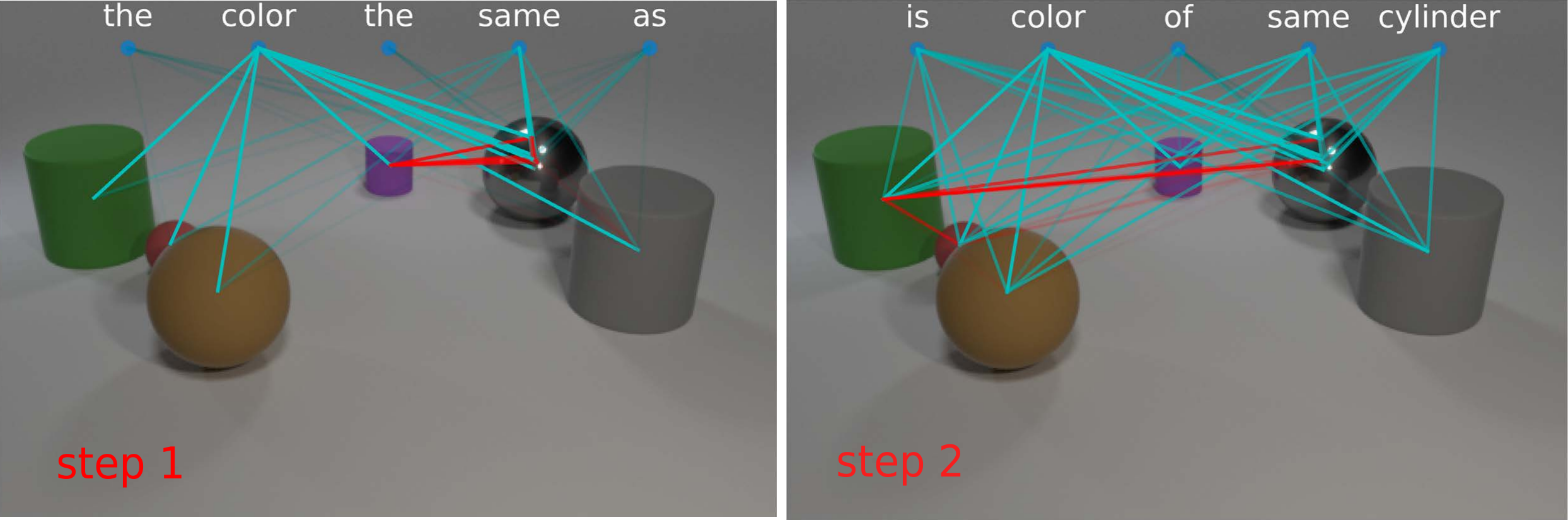}
\par\end{centering}
\caption{We aim to dynamically construct visual graphs (red edges) and linguistic-visual
bindings (cyan edges (most prominent words shown)) adaptively to reasoning
steps for each image-question pair. \label{fig:chap6-illustration}}
\end{figure}

Recent history observes the success of compositional reasoning which
iteratively pays attention to a subset of clues in the query and simultaneously
looks up a corresponding subset of facts from a static unstructured
knowledge source to construct a representation related to the answer
\citep{hudson2018compositional}. Concurrently, findings in visual
relational modelling show that the information in visual scenes is
significantly distributed at the interconnections between semantic
factors of visual objects and linguistic objects from both the image
and query \citep{baradel2018object}. These observations suggest that
relational structures can improve compositional reasoning \citep{xu2019can}.
However, direct application of attention mechanisms on a static structuralised
knowledge source \citep{velivckovic2018graph} would miss the full
advantage of compositionality. Moreover, object relations are naturally
rich and multifaceted \citep{kim2018visual}, therefore an \emph{a
priori} defined set of semantic predicates such as visual scene graphs
\citep{hudson2019learning} and language grounding \citep{huang2019multi}
are either incomplete \citep{xu2017scene}, or too complicated and
irrelevant to use without further pruning. 

We approach this dilemma by dynamically constructing relevant object
connections on-demand according to the evolving reasoning states.
There are two types of connections: links that relate visual objects
and links that bind visual objects in the image to linguistic counterparts
in the query (See Fig. \ref{fig:chap6-illustration}). Conceptually,
this dynamic structure constitutes a relational working memory that
temporarily links and refines concepts both within and across modalities.
These relations are compact and readily support structural inference.

The output model in this chapter, called Language-binding Object Graph
Networks (LOGNet) for Image QA, includes an iterative operation of
a LOG unit that uses a contextualised co-attention to identify pairs
of visual objects that are temporally related. Another co-attention
head is concurrently used to provide a cross-domain binding between
visual concepts and linguistic clues. A progressive chain of dynamic
graphs is inferred by our model (see Fig.~\ref{fig:chap6-illustration}).
These dynamic structures enable representation refinement with residual
graph convolution iterations. The refined information will be added
to an internal working memory progressing toward predicting the answer.
The modules are interconnected through co-attention signals making
the model end-to-end differentiable.

We evaluate our model on major Image QA datasets. Both qualitative
and quantitative results indicate that LOGNet has advantages over
state-of-the-art methods in answering long and complex questions.
Our results show superior performance even when trained on just 10\%
of data. These questions require complex high-order reasoning which
necessitates our model's ability to dynamically couple entities to
build a predicate, and then chain these predicates in the correct
order. The structured representation provides guidance to the reasoning
process, improving the learning fitness, particularly with limited
training data.

\selectlanguage{american}%

\section{Background\label{sec:Chap6-Background}}

\selectlanguage{australian}%
\begin{figure*}
\begin{centering}
\includegraphics[width=0.95\textwidth]{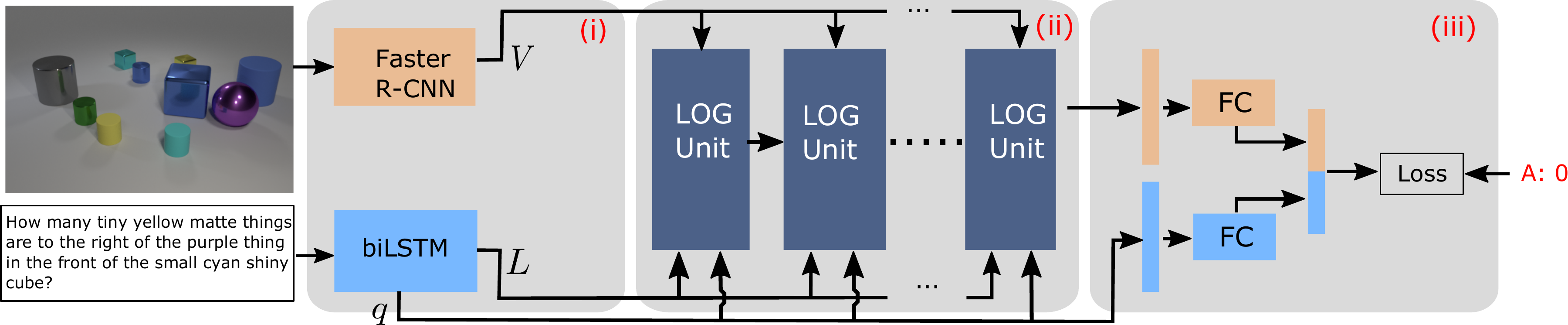}
\par\end{centering}
\caption{Overall architecture of LOGNet. (i) Linguistic and visual representations
(ii) Information refinement with LOG modules, and (iii) Multimodal
fusion and answer prediction.\label{fig:chap6-Overall-Architecture}}
\end{figure*}

In Sec.~\ref{sec:chap3_Neural-reasoning-for-ImageQA}, we have mentioned
general approaches in Image QA. This section provides more background
on compositional and relational reasoning, which are our main focuses.
Recent compositional reasoning research aims at either structured
symbolic program execution using custom-built modules \citep{hu2017learning}
or working through recurrent implicit reasoning steps on an unstructured
representation \citep{perez2018film}. Relational structures have
been demonstrated to be crucial for reasoning \citep{xu2019can}.
End-to-end relational modelling considers pair-wise predicates of
CNN features \citep{santoro2017simple}. With reliable object detection,
visual reasoning can use semantic objects as cleaner representations
\citep{anderson2018bottom,desta2018object}. When semantic or geometrical
predicate labels are available, either as provided \citep{hudson2019gqa}
or by learning \citep{xu2017scene} to form semantic scene graphs,
such structures can be leveraged for visual reasoning \citep{shi2019explainable,li2019relation}. 

In contrast to these methods, our relational graphs presented in this
chapter are not limited by the predefined predicates but liberally
form them according to the reasoning context. Our model is also different
from previous question-conditioned graph construction \citep{norcliffe2018learning}
in the dynamic nature of the multiform graphs where only relations
that are relevant emerge. Dynamic graph modelling has been considered
by recurrent modelling \citep{palm2018recurrent}, and although their
states transform, the graph structures stay fixed. A related idea
uses language conditioned message passing to extract context-aware
features \citep{hu2019language}. In contrast, LOGNet does not treat
linguistic cues as a single conditioning vector, it instead allows
them to live as a set of active objects that interact with visual
objects through binding and individually contribute to the joint representation.
The language binding also differentiates LOGNet from MUREL \citep{cadene2019murel}
where the contributions of linguistic cues to visual objects are the
same though an expensive bilinear operator.\selectlanguage{american}%

\section{Language-binding Object Graph Network\label{sec:Chap6-Language-binding-Object-Graph}}

\selectlanguage{australian}%
We formulate the Image QA task similar to the formulation of the Video
QA tasks in the previous chapters. The goal of the Image QA task is
to deduce an answer $\hat{y}$ in an answer space $\mathcal{\bm{A}}$
from an image $\mathcal{\bm{I}}$ in response to a natural question
$\mathbf{q}$. Mathematically, the Image QA task is presented as:
\begin{eqnarray}
\hat{y} & = & \arg\max_{a\in\mathbb{\bm{\mathcal{A}}}}\mathcal{F}_{\bm{\theta}}\left(a;\mathbf{q},\mathcal{\bm{I}}\right),
\end{eqnarray}
where, $\bm{\theta}$ is the model parameters of the scoring function
$\mathcal{F}\left(.\right)$. 

This chapter envisions Image QA as a process of relational reasoning
over a scene of multiple visual objects conditioned on a set of linguistic
cueing objects. Crucially, a pair of co-appearing visual objects may
induce multiple relations, whose nature may be unknown \emph{a priori},
and hence must be inferred dynamically in adaptive interaction with
the linguistic cues.

We present a new neural model $\mathcal{F}\left(.\right)$ called
LOGNet (See Fig.~\ref{fig:chap6-Overall-Architecture}) to realise
this vision. At the high level, for each image and query pair, LOGNet
first normalises them into two individual sets of linguistic and visual
objects. Then, it performs iterative multi-step reasoning by iteratively
summoning Language-binding Object Graph (LOG) units to achieve a compact
multi-modal representation in a recurrent manner. This representation
is finally combined with the query representation to reach the answers.
We detail these steps in Sec.~\ref{subsec:chap6_Language-binding-Object-Graph}.

\begin{figure*}
\begin{centering}
\includegraphics[width=1\textwidth]{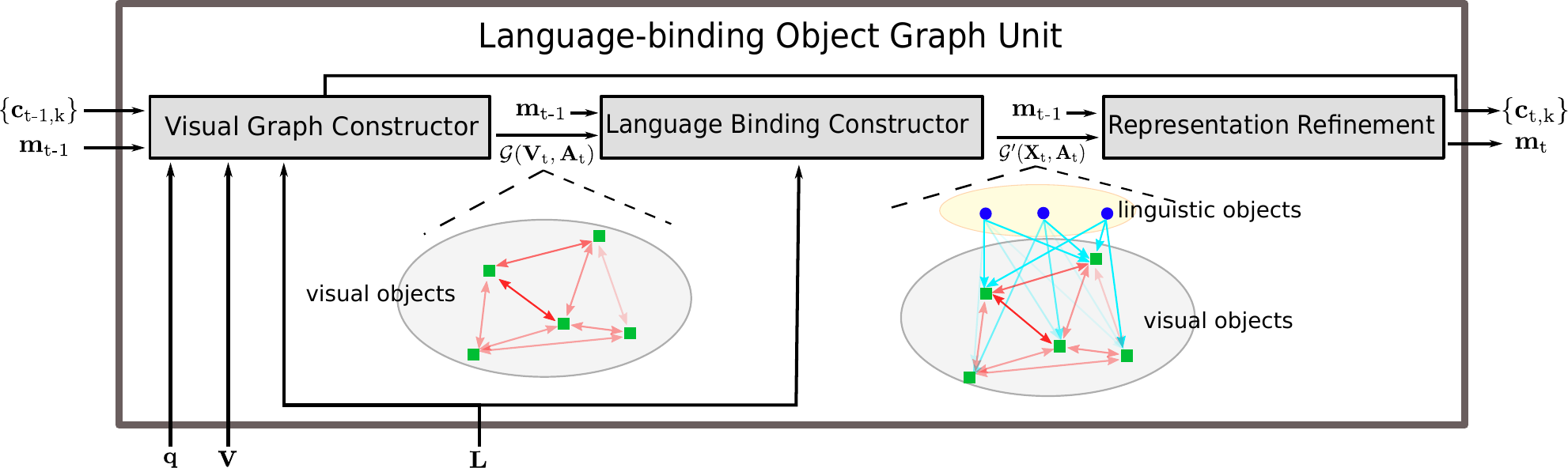}
\par\end{centering}
\caption{Language-binding Object Graph (LOG) unit. $\mathbf{L}$: linguistic
objects, $\mathbf{V}$: visual objects, red edges: visual graph, cyan
edges: language-visual binding. The following elements are dynamic
at pass $t$: $\mathbf{q}_{t}$ \textendash{} query semantic ; $\{\mathbf{c}_{t,k}\}$
\textendash{} language-based controlling signals; $\mathbf{m}_{t}$
- working memory state. \label{fig:Chap6-LOGUnit}}
\end{figure*}

\subsection{Linguistic and Visual Objects}

Similar to what presented in Sec.~\ref{sec:chap3_Visual-and-Language-Reasoning}
of Chapter \ref{chap:VisualLanguageReasoning} in terms of language
embedding, we embed words in a given length-$S$ query into 300-D
vectors, which are subsequently passed through a biLSTM. The hidden
states of LSTM representing the context-dependent word embeddings
$\mathbf{w}_{s}$ which are collected into a chain of contextual embeddings
$\mathbf{L}=\left\{ \mathbf{w}_{s}\right\} _{s=1}^{S}$ $\in\mathbb{R}^{d\times S}$,
where $d$ is the vector length. These contextual embeddings are used
as linguistic objects in reasoning. We also retain the overall query
semantic as $\mathbf{q}=\left[\overleftarrow{\mathbf{w}_{1}};\overrightarrow{\mathbf{w}_{S}}\right]\in\mathbb{R}^{d}$
joining the final states of forward and backward LSTM passes. Unless
otherwise specified, we use $[.\thinspace;.]$ to denote the concatenation
operator of two vectors.

The input image $\mathcal{\bm{I}}$ is first processed into a set
of appearance/spatial features $\mathbf{O}=\left\{ (\mathbf{a}_{i},\mathbf{p}_{i})\right\} _{i=1}^{N}$
of $N$ regions extracted by an off-the-shelf object detection such
as Faster R-CNN \citep{ren2015faster}. The appearance component $\mathbf{a}_{i}\in\mathbb{R^{\text{2048}}}$
are ROI pooling features and the spatial $\mathbf{p}_{i}$ are normalised
coordinates of the region box \citep{yu2017joint}. In particular,
the spatial feature $\mathbf{p}_{i}$ is calculated by:\vspace{-1cm}

\begin{eqnarray}
\mathbf{p}_{i} & = & \left[\frac{x_{i}^{tl}}{W},\frac{y_{i}^{tl}}{H},\frac{x_{i}^{br}}{W},\frac{y_{i}^{br}}{H},\frac{w_{i}}{W},\frac{h_{i}}{H},\frac{w_{i}*h_{i}}{W*H}\right],
\end{eqnarray}
where $(x_{i}^{tl},y_{i}^{tl}),\,(x_{i}^{br},y_{i}^{br})$ are coordinates
of the top left and bottom right point of the $i^{th}$ bounding box,
respectively; $w_{i},h_{i}$ are the width and height of the bounding
boxes; and $W,H$ are the width and height of the image $\bm{\mathcal{I}}$,
respectively.

The appearance feature and the spatial feature are further combined
and projected by trainable linear embeddings to produce a set of visual
objects $\mathbf{V}=\{\mathbf{v}_{i}\}_{i=1}^{N}\in\mathbb{R}^{d\times N}$.
We transform all vectors into a $d$-dimensional vector space in all
network components for ease of implementation. The pair $(\mathbf{L},\mathbf{V})$
are now readily used as input for a chain of LOG reasoning operations.

\subsection{Language-binding Object Graph Unit\label{subsec:chap6_Language-binding-Object-Graph}}

Each Language-binding Object Graph (LOG) is essentially a recurrent
unit whose state is kept in a compact working memory $\mathbf{m}_{t}\in\mathbb{R}^{d}$
and a controlling signal $\mathbf{c}_{t}\in\mathbb{R}^{d}$. Input
of each LOG operation includes the visual and linguistic objects $(\mathbf{V},\mathbf{L})$,
and the overall query semantic $\mathbf{q}$.

Each LOG consists of three submodules: (i) a \emph{visual graph constructor
}to build a context-aware adjacency matrix of visual graph $\mathcal{G}_{t}$,
(ii) a \emph{language binding constructor }to compute the adaptive
linkage between linguistic and visual objects and form a multi-modal
graph $\mathcal{G}_{t}^{\prime}$, and (iii)\emph{ representation
refinement }module\emph{ }to update object representation using the
graphs. (See Fig. \ref{fig:Chap6-LOGUnit}).

\subsubsection{Visual Graph Constructor}

At each LOG operation, we construct an undirected graph $\mathcal{G}_{t}=\left(\mathit{\mathit{\mathbf{V}}}_{t},\mathbf{A}_{t}\right)$
from $N$ visual objects $\mathbf{V}=\{\mathbf{v}_{i}\}_{i=1}^{N}$
by finding adaptive features $\mathbf{V}_{t}$ and constructing the
weighted adjacency matrix $\mathbf{A}_{t}$. Different from the widely
used static semantic graphs \citep{xu2017scene}, our graph $\mathcal{G}_{t}$
is dynamically constructed at each reasoning step $t^{th}$ and is
modulated by the recurrent controlling signal $\mathbf{c}_{t}$ and
overall linguistic cue $\mathbf{q}$. This reflects the dynamic relations
of objects triggered by both the question and reasoning context. In
fact, this design is consistent with how human reasons. For example,
to answer different questions about a visual scene, we connect different
pairs of objects although their geometrical and appearance similarities
were unchanged. Moreover, our mind traverses through multiple types
of object relationships in different steps of reasoning, especially
when a query contains multiple or nested relations. Let $\mathbf{W}_{t}\in\mathbb{R}^{d\times d}$
denote sub-networks' weights at step $t^{th}$, we first augment the
nodes' features $\mathbf{V}_{t}=\left\{ \mathbf{v}_{t,i}\right\} _{i=1}^{N}$
as
\begin{eqnarray}
\mathbf{v}_{t,i} & = & \mathbf{W}_{t}^{v}\left[\mathbf{v}_{i};\mathbf{m}_{t-1}\varodot\mathbf{v}_{i}\right]+\mathbf{b}^{v},\:\text{for}\:i=1,..,N.\label{eq:chap6_step_wise_vis_feats}
\end{eqnarray}

The controlling signals $\{\mathbf{c}_{t,k}\}$ is derived from its
previous state and a step-specific query semantic $\mathbf{q}_{t}$
through a set of $K$ attention heads $\{\mathbf{\bm{\alpha}}_{t,k}\}_{k=1}^{K}$
on the linguistic objects $\mathbf{L}=\left\{ \mathbf{w}_{s}\right\} _{s=1}^{S}$:
\begin{eqnarray}
\mathbf{c}_{1} & = & \mathbf{q}_{1},\qquad\mathbf{q}_{t}=\mathbf{W}_{t}^{q}\mathbf{q}+\mathbf{b}_{t}^{q},\\
\mathbf{q}_{t}^{\prime} & = & [\mathbf{q}_{t};\sum_{k=1}^{K}(\gamma_{t,k}*\mathbf{c}_{t-1,k})],\qquad\sum_{k=1}^{K}\gamma_{t,k}=1,\\
\alpha_{s,t,k} & = & \text{softmax}_{s}\left(\mathbf{W}_{t,k}^{\alpha}(\mathbf{w}_{s}\varodot\mathbf{q}_{t}^{\prime})\right),\\
\mathbf{c}_{t,k} & = & \sum_{s=1}^{S}\alpha_{s,t,k}*\mathbf{w}_{s},\,\,\,\mathbf{c}_{t}=\{\mathbf{c}_{t,k}\},
\end{eqnarray}
where, $\gamma_{t,k}$ is the weights of the past controlling signals
being added to the current query semantic $\mathbf{q}_{t}^{\prime}$.

While single attention can be used to guide the multi-step reasoning
process \citep{hudson2018compositional}, we notice that it tends
to focus on one object attribute at a time neglecting the inter-aspect
relations because of the softmax operation. In Image QA, multiple
object attributes are usually necessary - e.g. to answer \emph{``what
is the colour of the small shiny object having the same shape with
the cyan sphere?''}, the object aspects \emph{``colour''} and \emph{``shape''}
both need to be attended to. Our development of using multi-head attention
enables such a goal. The controlling signals are then used to build
the context modulated node description matrix of $r$ rows, $\tilde{\mathbf{V}}_{t}\in\mathbb{R}^{r\times N}$
:\vspace{-1cm}

\begin{eqnarray}
\tilde{\mathbf{V}}_{t} & = & \textrm{norm}\left(\mathbf{W}_{t}^{\tilde{v}}\sum_{k=1}^{K}(\mathbf{V}\varodot\mathbf{c}_{t,k})\right),
\end{eqnarray}
where, $norm$ is a normalisation function for numerical stabilisation
which is the softmax function in our implementation.

Finally, we estimate the symmetric adjacency matrix $\mathbf{A}_{t}$$\in\mathbb{R}^{N\times N}$
by relating node features in $\tilde{\mathbf{V}_{t}}$. The adjacency
matrix $\mathbf{A}_{t}$ is a rank $r$ symmetric matrix representing
the first-order proximity in appearance and spatial features of the
nodes:
\begin{eqnarray}
\mathbf{A}_{t} & = & \tilde{\mathbf{V}}_{t}^{\top}\tilde{\mathbf{V}}_{t}.
\end{eqnarray}

The motivation behind the estimation of $\mathbf{A}_{t}$ is similar
to recent works \citep{santoro2017simple,cadene2019murel} on modelling
\emph{implicit} relations of visual objects, in which they do not
reflect any semantic or spatial relations but indicate the probabilities
of object-pair co-occurrences given a query.

\subsubsection{Language Binding Constructor}

The visual graph explored by the visual graph constructor is powerful
in representing dynamic object relation albeit still lacking the two-way
complementary object-level relation between visual and textual data.
In one direction, visual features provide grounding to ambiguous linguistic
words so that objects of the same category can be differentiated \citep{nagaraja2016modeling}.
Imagine the question ``what is the colour of the cat eating the cake''
in a scene with many cats visible, then appearance and spatial features
will clarify the selection of the cat of interest. In the opposite
direction, linguistic cues provide more precise information than visual
features of segmented regions. In the previous example, the ``eat''
relation between ``cat'' and ``cake'' is clear from the query
words and is useful to connect these two visual objects in the image.
These predicative advantages are even more important in the case of
higher order relationships.

Drawing inspiration from that observation, we build a multi-modal
graph $\mathcal{G}_{t}^{\prime}=(\mathbf{X}_{t},\mathbf{\mathbf{A}}_{t})$
from the constructed graph $\mathcal{G}_{t}=(\mathbf{V}_{t},\mathbf{A}_{t})$.
Each node $\mathbf{x}_{t,i}\in\mathbf{X}_{t}$ of $\mathcal{G}_{t}^{\prime}$
is a binding of the corresponding visual node $\mathbf{v}_{t,i}$
of $\mathcal{G}_{t}$ with its linguistic supplement given by the
context-aware function $f_{t}(.)$:\vspace{-1cm}

\begin{eqnarray}
\mathbf{x}_{t,i} & = & [\mathbf{v}_{t,i};f_{t}(\mathbf{w}_{1},...,\mathbf{w}_{S}\mid\mathbf{v}_{t,i})].
\end{eqnarray}

Designing $f_{t}(.)$ is key to make this representation meaningful.
In particular, we design this function as the weighted composition
of contextual words $\{\mathbf{w}_{s}\}_{s=1}^{S}$:
\begin{eqnarray}
f_{t}(\mathbf{w}_{1},...,\mathbf{w}_{S}|\mathbf{v}_{t,i}) & = & \sum_{s=1}^{S}\beta_{t,i,s}*\mathbf{w}_{s}.
\end{eqnarray}
Here combination weights $\beta_{t,i,s}$ represent the cross-modality
partnership between a visual object $\mathbf{v}_{t,i}$ and a linguistic
word $\mathbf{w}_{s}$, essentially forming the contextualised pair-wise
bipartite relations between the $\mathbf{V}$ and $\mathbf{L}$. 

To calculate $\beta_{t,i,s}$, we first preprocess them by modulating
$\mathbf{V}$ with the previous memory state $\hat{\mathbf{V}}_{t}=\mathbf{W}_{t}^{\hat{v}}\left[\mathbf{V};\mathbf{m}_{t-1}\varodot\mathbf{V}\right]+\mathbf{b}^{\hat{v}}$
and softly classifying each word $s$ into multiple lexical types
using a scalar weight $z_{s}$ similar to \citep{yang2019dynamic},
$z_{s}=\sigma(\mathbf{W}_{s}^{z1}(\mathbf{W}_{s}^{z0}\mathbf{w}_{s}+b_{s}^{z0})+b_{s}^{z1})$,
where $\mathbf{W}^{z0}\in\mathbb{R}^{d\times d}$ and $\mathbf{W}^{z1}\in\mathbb{R}^{1\times d}$
are the weight matrices. Subsequently, the normalised cross-modality
relation weights are calculated as:\vspace{-1cm}

\begin{eqnarray}
\beta_{t,i,s}= & z_{s} & *\textrm{softmax}_{s}(\mathbf{W}_{t,s}^{\beta}(\text{tanh}(\mathbf{W}_{t,s}^{\hat{v}}\hat{\mathbf{v}}_{t,i}+\mathbf{W}_{t,s}^{w}\mathbf{w}_{s}))).
\end{eqnarray}

By doing this, we allow per-object communication between the two modalities,
differentiating our method from prior works where the linguistic cues
is squeezed into a single vector for conditioning or combined with
visual signal in a late fusion fashion.

\subsubsection{Representation Refinement}

At the last step of LOG operation, we rely on the newly built multi-modal
graph $\mathcal{G}_{t}^{\prime}=(\mathbf{X}_{t},\mathbf{A}_{t})$
as the structure to refine the representation of objects by employing
a graph convolutional network (GCN) \citep{kipf2016semi} of $H$
hidden layers. Generally, vanilla GCNs have a difficulty of stacking
deep layers due to the common vanishing gradient and numerical instability.
Therefore, a GCN is usually shallow of 2 or 3 layers. We solve this
problem by borrowing the residual skip-connection trick from ResNet
\citep{he2016deep} to create more direct gradient flow. Concretely,
the refined node representation is given by:\vspace{-1cm}

\begin{eqnarray}
\mathbf{R}_{1} & = & \mathbf{X}_{t},\\
F_{h}\left(\mathbf{R}_{h-1}\right) & = & \mathbf{W}_{h-1}^{2}\rho\left(\mathbf{W}_{h-1}^{1}\mathbf{R}_{h-1}\mathbf{\mathbf{A}}_{t}+\mathbf{b}_{h-1}\right),\\
\mathbf{R}_{h} & = & \rho\left(\mathbf{R}_{h-1}+F_{h}\left(\mathbf{R}_{h-1}\right)\right),
\end{eqnarray}
where, $h=1,2..,H$, and $\rho$ is an activation function which is
the ELU operation in our later experiments. The parameters $\left(\mathbf{W}_{h-1}^{1},\mathbf{W}_{h-1}^{2}\right)$
can be optionally tied across $H$ layers.

As we obtain the refined representation $\mathbf{R}_{t,H}=\{\mathbf{r}_{t,i,H}\}_{i=1}^{N}$
after the $H$ refinement layers, we compute the overall final representation
by smashing the graph into one single vector:\vspace{-1cm}

\begin{eqnarray}
\tilde{\mathbf{x}_{t}} & = & \sum_{i=0}^{N}\delta_{t,i}*\mathbf{r}_{t,i,H},
\end{eqnarray}
where, $\delta_{t,i}=\text{softmax}_{i}(\mathbf{W}_{t}^{\delta}\mathbf{r}_{t,i,H})$.
Finally, we update LOG's working memory state:\vspace{-1cm}

\begin{eqnarray}
\mathbf{m}_{t} & = & \mathbf{W}_{t}^{m}\left[\mathbf{m}_{t-1};\tilde{\mathbf{x}}_{t}\right]+\mathbf{b}^{m}.
\end{eqnarray}

\subsection{Answer Prediction}

After $T$ passes of LOG iterations, LOGNet combines the final memory
state $\mathbf{m}_{T}$ with the sequential expression $\mathbf{q}$
of the question by concatenation followed by a linear layer to get
the final representation $\mathbf{y}=\mathbf{W}\left[\mathbf{m}_{T};\mathbf{q}\right]+\mathbf{b},\ \mathbf{y}\in\mathbb{R}^{d}$.

For answer prediction, we adopt a 2-layer multi-layer perceptron (MLP)
and a batch normalisation layer in between as a classifier. The network
is trained using cross-entropy loss (multi-class classification) or
binary cross-entropy loss (multi-label classification) according to
types of questions.\selectlanguage{american}%

\section{Experiments\label{sec:Chap6-Experiments}}

\selectlanguage{australian}%

\subsection{Datasets\label{subsec:chap6_Datasets}}

We evaluate our model on multiple Image QA datasets including: 

\textbf{CLEVR} \citep{johnson2017clevr}: presents several reasoning
tasks such as transitive relations and attribute comparison. Each
question is generated using a functional program of 13 pre-defined
basic functions. As we aim to challenge the reasoning abilities, each
question in this dataset is long and compositional; therefore, one
can only compute the correct answer after a complex reasoning chain.
We intentionally design experiments to evaluate the generalisation
capability of our model on various subsets of CLEVR, where most existing
works fail, sampled by the number of images and their corresponding
questions.

\textbf{CLEVR-Human }\citep{johnson2017inferring}:\textbf{ }composes
natural language question-answer pairs on images from CLEVR. Due to
diverse linguistic variations, this dataset requires stronger visual
reasoning ability than CLEVR.

\textbf{GQA} \citep{hudson2019gqa}:\textbf{ }is currently the largest
VQA dataset. The dataset contains over 22M question-answer pairs and
over 113K images covering various reasoning skills and requiring multi-step
inference, hence significantly reducing biases as in previous VQA
datasets. Each question is generated based on an associated scene
graph and pre-defined structural patterns. GQA has served as a standard
benchmark for most advanced compositional visual reasoning models
\citep{hudson2019gqa,hu2019language,hudson2019learning,le2020dynamic,shevchenko2020visual}.
We use the balanced splits of the dataset in our experiments. Because
LOGNet does not need prior predicates, we ignore these static graphs
using only the image and textual query as input.

\textbf{VQA v2 }\citep{goyal2017making}: As a large portion of questions
is short and can be answered by looking for facts in images, we design
experiments with splits of only long questions (>7 words). In particular,
the train split accounts for 21.6\% of the original training questions
and the validation split accounts for 21.5\% of the original validation
questions. The splits, hence, assesses the ability to model the relations
between objects, e.g.: \emph{``What is the white substance on the
left side of the plate and on top of the cake?''} For the splits
of VQA v2, we report performance with accuracies calculated by standard
VQA accuracy metric: $\text{min}(\frac{\text{\#humans that provided that answer}}{3},1)$
\citep{antol2015vqa}.

\begin{figure}[t]
\begin{centering}
\includegraphics[width=0.7\textwidth]{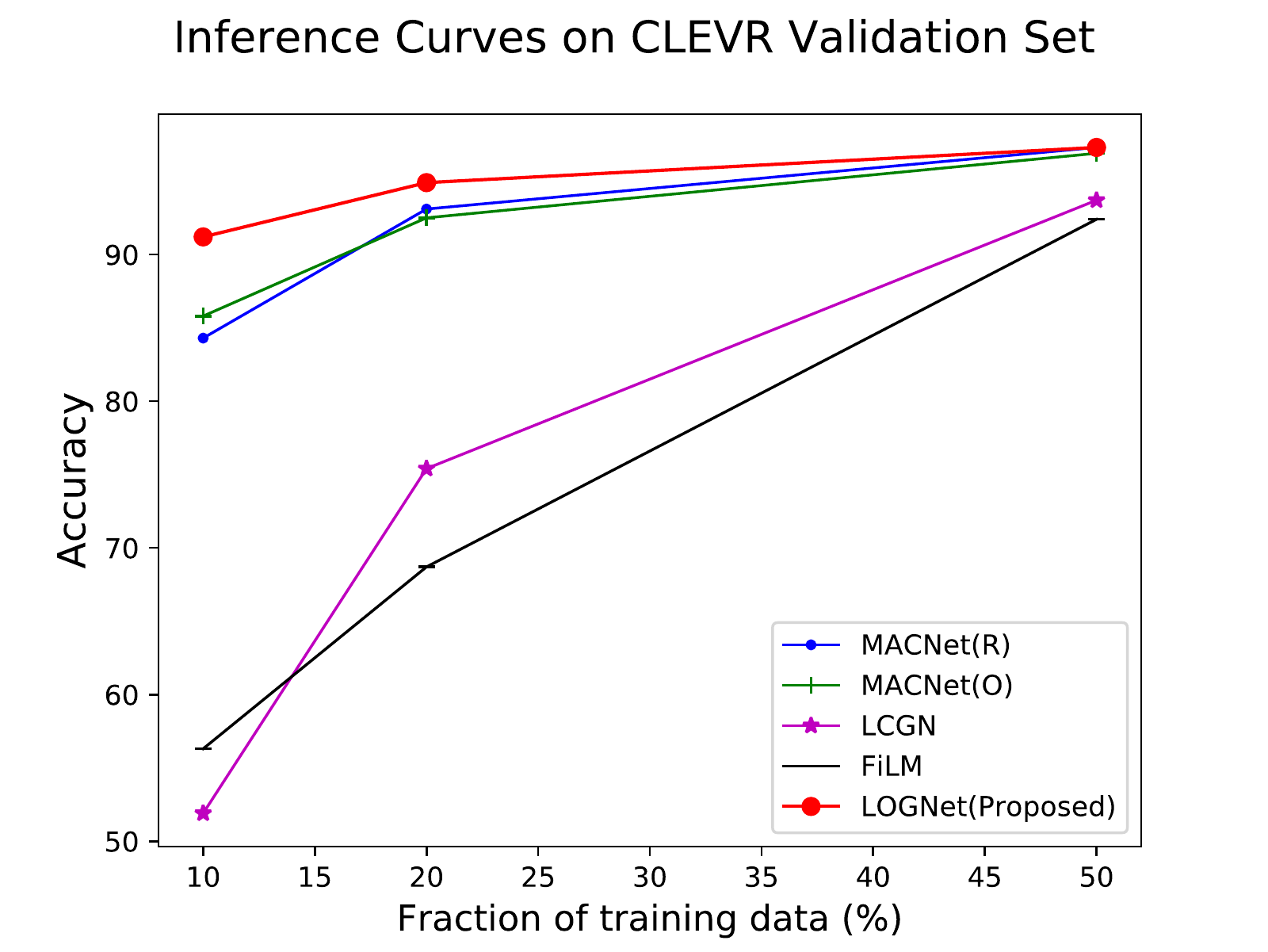}
\par\end{centering}
\caption{Quantitative performance on CLEVR subsets.\label{fig:Chap6-performance-CLEVR-subsets}}
\end{figure}

\subsection{Performance against SOTAs}

\begin{figure*}
\begin{centering}
\noindent\begin{minipage}[t]{1\textwidth}%
\begin{center}
\includegraphics[width=1\textwidth]{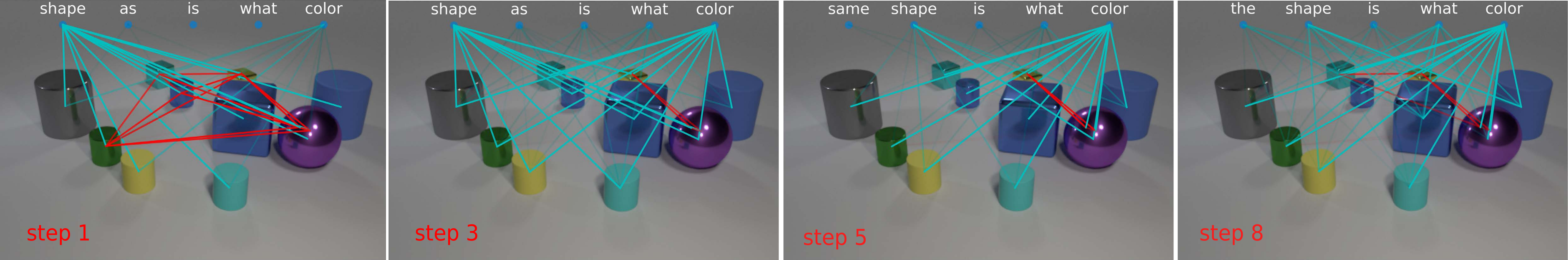}
\par\end{center}
\begin{center}
\begin{minipage}[t]{0.95\columnwidth}%
\textbf{\small{}Question}{\small{}: The other small shiny thing that
is the same shape as the tiny yellow shiny object is what colour?}{\small\par}

\textbf{\small{}Prediction}{\small{}: }\textcolor{green}{\small{}cyan}\textbf{\small{}\qquad{}Answer}{\small{}:
cyan}{\small\par}%
\end{minipage}
\par\end{center}%
\end{minipage}\medskip{}
\par\end{centering}
\begin{centering}
\noindent\begin{minipage}[t]{1\textwidth}%
\begin{center}
\includegraphics[width=1\textwidth]{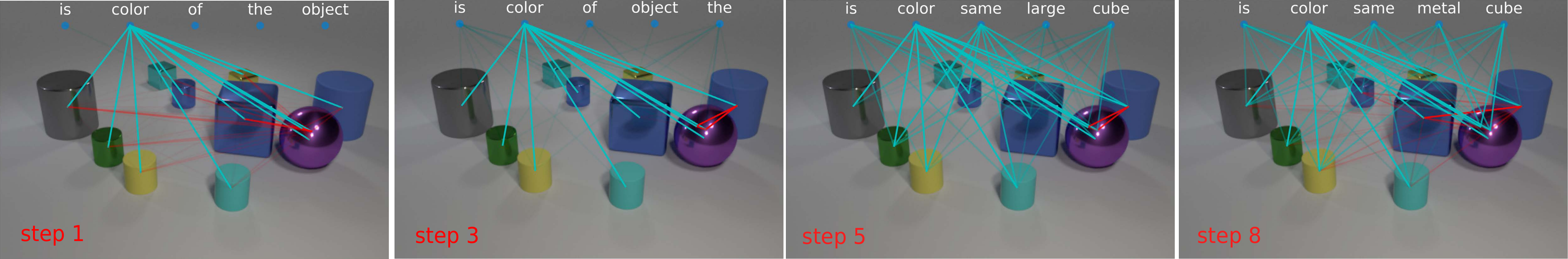}
\par\end{center}
\begin{center}
\begin{minipage}[t]{0.95\columnwidth}%
\textbf{\small{}Question}{\small{}: Is the colour of the big matte
object the same as the large metal cube?}{\small\par}

\textbf{\small{}Prediction}{\small{}: }\textcolor{green}{\small{}yes}\textbf{\small{}\qquad{}Answer}{\small{}:
yes}{\small\par}%
\end{minipage}
\par\end{center}%
\end{minipage}\medskip{}
\par\end{centering}
\begin{centering}
\ %
\noindent\begin{minipage}[t]{1\textwidth}%
\begin{center}
\includegraphics[width=1\textwidth]{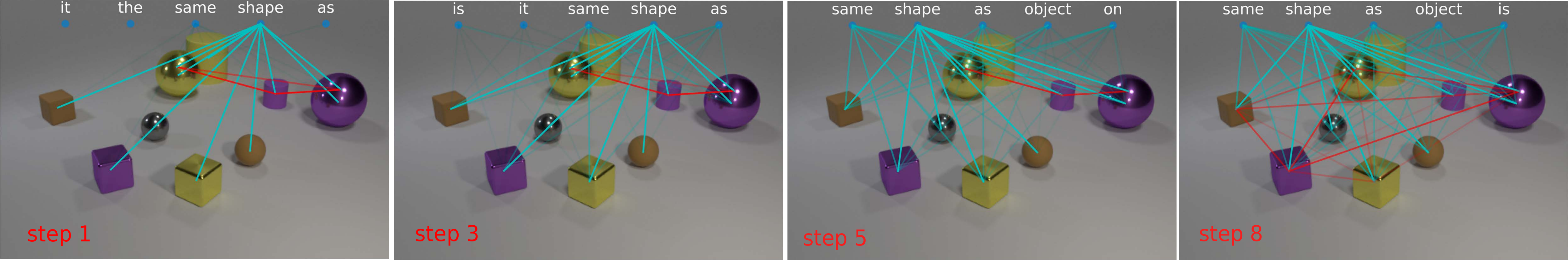}
\par\end{center}
\begin{center}
\begin{minipage}[t]{0.95\columnwidth}%
\textbf{\small{}Question}{\small{}: There is a tiny purple rubber
thing; does it have the same shape as the brown object that is on
the left side of the rubber sphere?}{\small\par}

\textbf{\small{}Prediction}{\small{}: }\textcolor{green}{\small{}no}\textbf{\small{}\qquad{}Answer}{\small{}:
no}{\small\par}%
\end{minipage}
\par\end{center}%
\end{minipage}\medskip{}
\par\end{centering}
\begin{centering}
\noindent\begin{minipage}[t]{1\textwidth}%
\begin{center}
\includegraphics[width=1\textwidth]{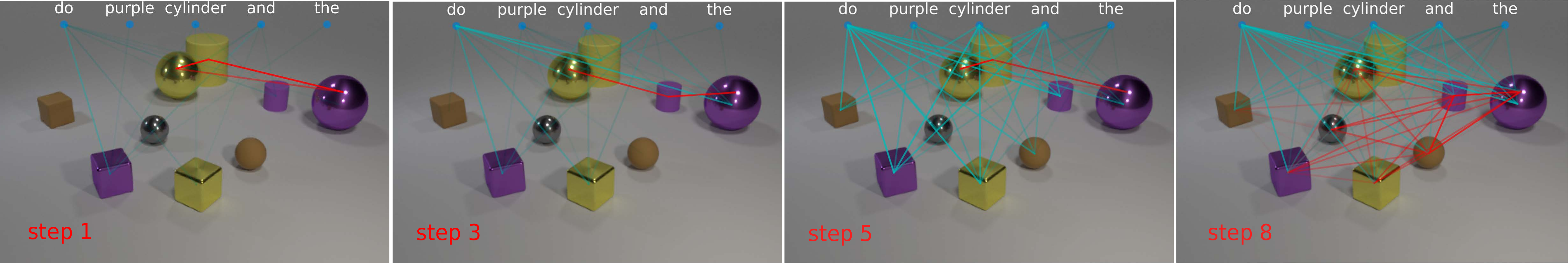}
\par\end{center}
\begin{center}
\begin{minipage}[t]{0.95\columnwidth}%
\textbf{\small{}Question}{\small{}: Do the purple cylinder and the
yellow rubber thing have the same size?}{\small\par}

\textbf{\small{}Prediction}{\small{}: }\textcolor{green}{\small{}no}\textbf{\small{}\qquad{}Answer}{\small{}:
no}{\small\par}%
\end{minipage}
\par\end{center}%
\end{minipage}
\par\end{centering}
\caption{Chains of visual object relation (in red) with language binding (in
cyan) constructed for two image-question pairs. Visual relations are
found adaptively to the specific questions and reasoning stages. Language
binding was sharp on key cross-modality relations at several early
steps, then flats out as memory converges. Only five words included
for visualisation purposes. Best viewed in colour.\label{fig:Chap6-qualitative}}
\end{figure*}

\begin{table}
\begin{centering}
\begin{tabular}{l|c}
\hline 
\multirow{1}{*}{{\small{}Method}} & \multicolumn{1}{c}{{\small{}Val. Acc. (\%)}}\tabularnewline
\hline 
\hline 
{\small{}FiLM }\citep{perez2018film} & {\small{}56.6}\tabularnewline
{\small{}MACNet(R) \citep{hudson2018compositional}} & {\small{}57.4}\tabularnewline
{\small{}LCGN }\citep{hu2019language} & {\small{}46.3}\tabularnewline
{\small{}BAN \citep{shrestha2019answer}} & {\small{}60.2}\tabularnewline
{\small{}RAMEN \citep{shrestha2019answer}} & {\small{}57.9}\tabularnewline
\textbf{\small{}LOGNet} & \textbf{\small{}62.5}\tabularnewline
\hline 
\end{tabular}
\par\end{centering}
\caption{Performance on CLEVR-Human. \label{tab:Chap6-clevr-human}}
\end{table}

Our LOGNet model is generally implemented with feature dimension $d=512$,
reasoning depth $T=8$, GCN depth $H=8$ and attention-width $K=2$.
The number of regions is $N=14$ for CLEVR and CLEVR-Human, and 100
for GQA and 36 for VQA v2 to match with other related methods. We
also match the word embeddings with others by using random vectors
of a uniform distribution for CLEVR/CLEVR-Human and pre-trained GloVe
vectors for the other datasets. We start training our model at learning
rate $10^{-4}$ and decaying by half after every 10 epochs in case
of VQA v2. All experiments are terminated after $25$ training epochs,
except those using less number of passes terminated after $30$ epochs.
Pytorch implementation of our model is available online\footnote{https://github.com/thaolmk54/LOGNet-VQA}.

We compare with state-of-the-art methods reporting performance as
in their papers or obtained with their public code. For the better
judgement of whether the improvement is from the model designs or
from the use of better visual embeddings, we reimplement MACNet \citep{hudson2018compositional}
with their feature choice of ResNet - MACNet(R), and additionally
try it out on our ROI pooling features - MACNet(O).

\begin{table}
\begin{centering}
\begin{tabular}{ll|cc}
\hline 
\multirow{2}{*}{{\small{}Training size}} & \multirow{2}{*}{{\small{}Method}} & \multicolumn{2}{c}{{\small{}Accuracy (\%)}}\tabularnewline
\cline{3-4} \cline{4-4} 
 &  & {\small{}val} & {\small{}test}\tabularnewline
\hline 
\hline 
\multirow{5}{*}{\textbf{\small{}Full}} & {\small{}CNN+LSTM} & {\small{}49.2} & {\small{}46.6}\tabularnewline
 & {\small{}Bottom-Up} & {\small{}52.2} & {\small{}49.7}\tabularnewline
 & {\small{}MACNet(O)} & {\small{}57.5} & {\small{}54.1}\tabularnewline
 & {\small{}LCGN} & {\small{}63.9} & {\small{}56.1}\tabularnewline
 & {\small{}LOGNet} & {\small{}63.2} & {\small{}55.2}\tabularnewline
\hline 
\multirow{2}{*}{\textbf{\small{}50\%}} & {\small{}LCGN} & {\small{}60.6} & {\small{}-}\tabularnewline
 & {\small{}LOGNet} & {\small{}61.0} & {\small{}-}\tabularnewline
\hline 
\multirow{2}{*}{\textbf{\small{}20\%}} & {\small{}LCGN} & {\small{}53.2} & {\small{}-}\tabularnewline
 & {\small{}LOGNet} & {\small{}53.8} & {\small{}-}\tabularnewline
\hline 
\end{tabular}
\par\end{centering}
\caption{Performance on GQA and subsets. \label{tab:chap6_GQA}}
\end{table}

\subsubsection{CLEVR and CLEVR-Human Dataset}

Fig.~\ref{fig:Chap6-performance-CLEVR-subsets} demonstrates the
large improvement of LOGNet over SOTAs including MACNet \citep{hudson2018compositional},
FiLM{\small{} }\citep{perez2018film} and LGCN{\small{} }\citep{hu2019language}
particularly with limited training data. With enough data, all models
converge in performance. With smaller training data, other methods
struggle to generalise, while LOGNet maintains stable performance.
With 10\% of training data, FiLM quickly drops to 51.9\%, and only
48.9\% in case of LGCN, which barely surpasses the linguistic bias
performance of 42.1\% reported by \citet{johnson2017clevr}. Behind
LOGNet (91.2\%), MACNet is the runner up in generalisation with around
85.8\%.

Our model shows significant improvement over other works, including
FiLM, MACNet, LCGN, BAN \citep{shrestha2019answer}, RAMEN \citep{shrestha2019answer},
on CLEVR-Human dataset (See Table \ref{tab:Chap6-clevr-human}) where
language vocabulary is richer than the original CLEVR. We only report
results without fine-tune on CLEVR for better judgement of the generalisation
ability of the methods. This suggests that LOGNet can better handle
the linguistic variations by its advantage in modelling cross-modality
interactions.

\begin{table}[t]
\begin{centering}
{\small{}}%
\begin{tabular}{l|c}
\hline 
\multirow{2}{*}{{\small{}Method}} & \multirow{2}{*}{{\small{}Val. Acc. (\%)}}\tabularnewline
 & \tabularnewline
\hline 
\hline 
{\small{}XNM} & {\small{}43.4}\tabularnewline
{\small{}MACNet(R)} & {\small{}40.7}\tabularnewline
{\small{}MACNet(O)} & {\small{}45.5}\tabularnewline
\textbf{\small{}LOGNet} & \textbf{\small{}46.8}\tabularnewline
\hline 
\end{tabular}{\small\par}
\par\end{centering}
\caption{Experiments on VQA v2 subset of long questions. \label{tab:chap6_vqa}}
\end{table}

\subsubsection{GQA}

LOGNet outperforms previous works including simple fusion approaches
CNN+LSTM and Bottom-Up \citep{anderson2018bottom}{\small{}, }and
the recent advanced multi-step inference MACNet. Although LOGNet achieves
competitive performance as compared with LCGN on the full training
set, it shows its advantage in generalisation and robustness against
overfitting in limited data experiments (20\% and 50\% splits) - see
Table~\ref{tab:chap6_GQA}.

\subsubsection{VQA v2 - Subset of Long Questions}

LOGNet is finally applied to the most complex questions of VQA v2.
Empirical results show that our model achieves favourable performance
over compared methods, MACNet and XNM {\large{}\citep{shi2019explainable},
}on this subset even though this dataset is not designed for compositional
questions. It is worth to note that visual embedding greatly influences
the performance of a model. In particular, MACNet(R) can only achieve
40.7\% on validation set while the same model with ROI pooling features
(MACNet(O)) can perform approximately 5\% better. Due to the rich
language vocabulary of human annotated datasets, the improvements
are less noticeable as compared with those on synthetic datasets such
as CLEVR.

\begin{table}[t]
\begin{centering}
{\footnotesize{}}%
\begin{tabular}{cl|c}
\hline 
{\small{}No.} & {\small{}Model} & {\small{}Val. Acc. (\%)}\tabularnewline
\hline 
\hline 
{\small{}1} & {\small{}Default config. (8 LOG units, 8 GCNs)} & {\small{}91.2}\tabularnewline
{\small{}2} & {\small{}w/o bounding box features} & {\small{}86.5}\tabularnewline
{\small{}3} & {\small{}Graph constructor w/o previous memory} & {\small{}86.5}\tabularnewline
{\small{}4} & {\small{}Graph constructor w/o language} & {\small{}56.2}\tabularnewline
{\small{}5} & {\small{}Single-head attn. controlling signal} & {\small{}86.3}\tabularnewline
{\small{}6} & {\small{}Rep. refinement w/ 1 GCN layers} & {\small{}75.9}\tabularnewline
{\small{}7} & {\small{}Rep. refinement w/ 4 GCN layers} & {\small{}89.4}\tabularnewline
{\small{}8} & {\small{}Rep. refinement w/ 12 GCN layers} & {\small{}91.1}\tabularnewline
{\small{}9} & {\small{}Rep. refinement w/ 16 GCN layers} & {\small{}89.5}\tabularnewline
{\small{}10} & {\small{}Language binding w/o previous memory} & {\small{}90.8}\tabularnewline
{\small{}11} & {\small{}w/o language binding} & {\small{}89.9}\tabularnewline
{\small{}12} & {\small{}1 LOG unit} & {\small{}69.0}\tabularnewline
{\small{}13} & {\small{}4 LOG units} & {\small{}76.3}\tabularnewline
{\small{}14} & {\small{}12 LOG units} & {\small{}91.6}\tabularnewline
{\small{}15} & {\small{}16 LOG units} & {\small{}91.1}\tabularnewline
\hline 
\end{tabular}{\footnotesize\par}
\par\end{centering}
\caption{Ablation studies - CLEVR dataset: 10\% subset.\label{tab:Chap6-Ablation-studies}}
\end{table}

\subsection{Ablation Studies}

We conduct ablation studies with our model on CLEVR subset of 10\%
training data (See Table~\ref{tab:Chap6-Ablation-studies}). We observe
consistent improvements responding to the increase in the number of
reasoning steps as well as going deeper with the representation refinement
process. We have tried up to $p=16$ LOG units and $H=16$ GCN layers
for representation refinement in each time step, establishing a very
deep reasoning process over hundreds of layers. The results strongly
prove the ability to leverage recurrent cells (row 12-14) and the
significance of the deep refinement layers (row 6-9). It is also
clear that linguistic cue plays a crucial role in all the components
of LOGNet and language binding contributes noticeably to performance
(row 1 and 11).

\subsection{Behaviour Analysis}

To understand the behaviour of the dynamic graphs during LOG iterations,
we visualise them for complex questions from CLEVR (see Fig.~\ref{fig:Chap6-qualitative}).
As seen, the linguistic objects most selected for binding are from
objects of interest or their attributes which give a hint to the model
of what aspect of the visual cue to look at. Question types (e.g.
yes-no/wh-question, object counting) and other function words (e.g.
``the'', ``is'', ``on'') are also paid much attention to. Note
that as linguistic objects are outputs of LSTM passes, those of function
words, such as articles and conjunctions connect nearby content words
and holds their aggregated information through the LSTM operations.

Progressing through the reasoning steps, LOGNet accumulates multiple
aspects of joint domain information in a compositional manner. In
earlier steps when most crucial reasonings happen, it is apparent
in Fig.~\ref{fig:Chap6-qualitative} that language binding concentrates
on sharp linguistic-visual relations such as from attribute and predicate
words (e.g. ``colour'', ``shape'', ``same'') to their related
objects. They constitute the most principal components of the working
memory. Later in the reasoning process, when the memory gets close
to the convergence, the binding weights flat out as not much critical
information is being added anymore. This agrees with the ablation
study result in the last four rows of Table \ref{tab:Chap6-Ablation-studies}
where the performance raises sharply in the early steps and gradually
converges.\selectlanguage{american}%

\section{Closing Remarks\label{sec:Chap6-Closing-Remarks}}

\selectlanguage{australian}%
In this chapter, we have presented a new neural recurrent model, Language-binding
Object Graph Network (LOGNet), for compositional and relational reasoning
over a knowledge base with implicit intra- and inter- modality connections.
Distinct from existing neural reasoning methods, our method computes
dynamic dependencies \emph{on-demand} as reasoning proceeds. Our focus
is on Image QA tasks, where raw visual and linguistic features are
given but their relations are unknown. The model facilitates a multi-step
reasoning process, in which implicit relations between objects are
constructed on-the-fly conditioned on the linguistic cue found for
each reasoning step. The experimental results demonstrated the superior
performance of LOGNet on multiple datasets. It also showed a strong
capability to generalise on unseen data when trained on just 10\%
of the training data of the CLEVR dataset. 

Although the impressive capability to work with limited training data
on the CLEVR dataset, LOGNet has limitations in identifying the affiliations
between linguistic components and visual objects in real-world data.
This is because modelling the relationships between visual objects
and the cross-domain associations in natural scenes is much more complex
than those in controlled settings as in CLEVR. It is necessary to
utilise external knowledge of these associations to guide the learning
process to address this problem. Motivated by this observation, our
next chapter will explain how to obtain such external knowledge and
use it as an extra source of information to improve the reasoning
capacity of visual reasoning models.

\selectlanguage{american}%

\selectlanguage{australian}%

\newpage{}

\chapter{Towards Robust Generalisation in Visual Reasoning\label{chap:RobustnessVR}}

\section{Introduction\label{sec:Chap7-Introduction}}

\begin{figure}
\begin{centering}
\includegraphics[width=0.6\columnwidth]{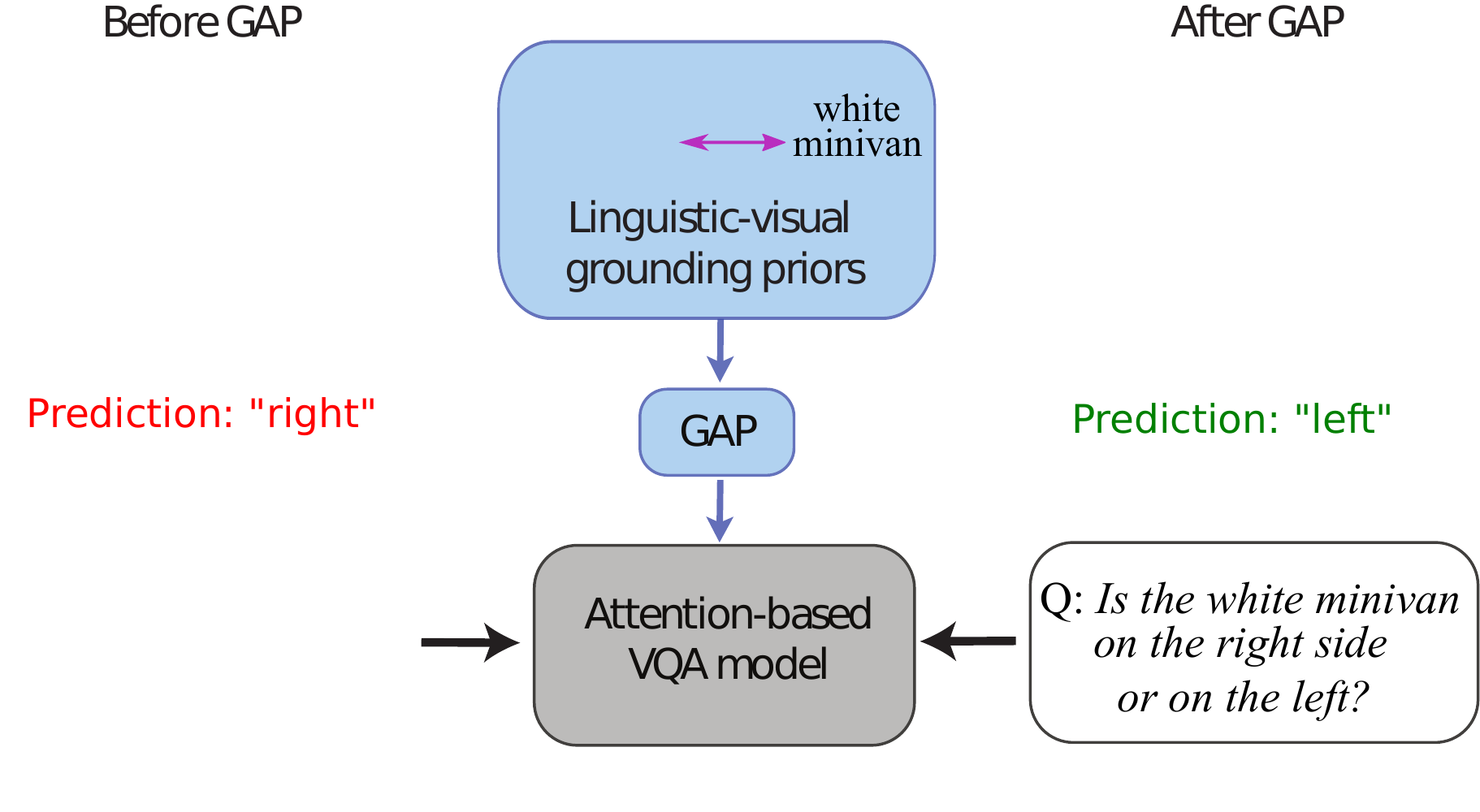}
\par\end{centering}
\caption{We introduce Grounding-based Attention Prior (GAP) mechanism (blue
boxes), which considers the linguistic-visual groundings between query
and image and regulates the attentions inside Image QA models (grey
box). This regularisation boosts the performance of Image QA models,
fortifies them against linguistic variations and increases their interpretability.
Illustration using a real case in the GQA dataset.}
\end{figure}
In the previous chapters, we showed that visual reasoning models extract
facts from visual data and present them into higher-level knowledge
in response to a query. In Chapter \ref{chap:RelationalVisualReasoning}
in particular, our proposed method LOGNet and other state-of-the-art
visual reasoning models on static scenes estimate the cross-domain
associations between the query and visual entities in the form of
attention weights. Such associations direct the knowledge distillation
process that results in a joint representation that can be decoded
into an answer.

The meaningfulness and appropriateness of these attention scores are
pivotal to the performance of Image QA systems \citep{lu2016hierarchical}.
However, the automatic adjustment of these scores by using only the
gradient from the labelled answers is inadequate, as shown by usually
meaningless and unintuitive attentions \citep{das2017human}. The
problem also extends to multi-step reasoning models such as those
in \citep{hudson2018compositional} and our proposed LOGNet \citep{le2020dynamic},
wherein unregulated attention errors may accumulate along the reasoning
iterations. Regularising attention mechanisms using an external source
of supervision is a promising solution for the problem. Early works
use human attention as the labels for supervising machine visual attention
\citep{qiao2018exploring,selvaraju2019taking}. This simple and direct
attention perceived by humans is not guaranteed to be optimal for
machine reasoning. Furthermore, because annotating attentions is
a complex labelling task, this process is inherently costly, inconsistent
and unreliable \citep{selvaraju2019taking}. Finally, these methods
only work for visual attention while ignoring linguistic attention
channel. \emph{Regulating dual-channel attention-based reasoning with
an objective, generic external knowledge without further extra labelling
remains a desire but missing capability.}

To this end, we aim at mining generic knowledge from available external
sources and leveraging it in guiding attention-based reasoning models
in this chapter. We explore the fact that the linguistic-visual associations
exist as common external knowledge in the form of image-caption grounding
data. In the scope of this chapter, we limit our study with static
visual scenes in images rather than considering dynamic visual scenes
as in Chapter \ref{chap:DualProcess} and Chapter \ref{chap:MultimodalReasoning}.
This is because obtaining grounding knowledge between language expressions
and dynamic events and actions in videos is greatly challenging at
the moment. With that in mind, we propose to extract these pair-wise
association facts and use them as guidelines to regulate the attention
mechanisms of Image QA models. The key challenge is that the Image
QA queries are expressed in free form natural language and the popular
recurrent feature extraction fails to extract the semantic concepts
that can match the appropriate image regions.

To address this challenge, we construct a grammatical parse tree of
the query and extract the nested phrasal expressions. These expressions
are then grounded to the visual objects and regions in the image,
providing the visual-language bindings information that can serve
as the attention supervision signals for Image QA models. The structured
information from the parse tree can help improve the query representation,
mitigating the risk of relying on surface statistics of word collocations
in common sequential models. This reduces the sensitivity against
variations in linguistic expressions such as rephrasing.

Through the extensive experiment, we prove that this methodology is
effective in boosting the performance of attention-based Image QA
models across representative methods and datasets. These improvements
surpass directly labelled supervision that requires extra annotation.
Furthermore, the ability to extract and match up linguistic concepts
underlying the changing syntactic variations improves the robustness
over question rephrasing, solving a key challenge of the field. Our
analysis further shows that the interpretable training process successfully
captures the cross-modality association inductive bias of the attention
model and extends to new and manipulated test data.

Our key contributions in this chapter are:

1. A new framework to distil external grounding knowledge into sets
of weakly supervision signals to regulate visual reasoning.

2. Innovative methods to syntactically represent the linguistic query
through a constituent parse tree enabling meaningful cross-modality
associations.

3. The first generic dual-modality regulation mechanism that can fortify
any attention-based Image QA model in performance and robustness to
linguistic variations.

\section{Background \label{sec:Chap7-Related-Work}}

In addition to related background on vision-and-language reasoning
presented in Chapter \ref{chap:VisualLanguageReasoning}, we discuss
relevant works that closely related to techniques proposed in this
chapter as follows.

\textbf{Attention regularisation }using direct supervision is well
studied in many problems such as machine translation \citep{liu2016neural}
and image captioning \citep{liu2017attention,ma2020learning,zhou2020more}.
In Image QA, attentions can be self-regulated through internal constraints
\citep{ramakrishnan2018overcoming,liu2021answer}. The early external
regularisation methods rely on human annotations either as textual
explanations \citep{wu2019self} or visual attention \citep{qiao2018exploring,selvaraju2019taking}.
Unlike these approaches, we take an indirect supervision approach
using external grounding data. We are also the first to regulate both
visual and linguistic attention channels.

\textbf{Robustness to question variations }is approached by simply
relying on a more powerful embedding platform \citep{jolly2020can}
and counterfactual data augmentation \citep{chen2020counterfactual}.
Internal regularisation methods look into attention scores and check
for anomalies \citep{lee2020regularizing}. More systematic approaches
include constraining the model with an external constraint on model
consistency \citep{shah2019cycle,whitehead2020learning,ray2019sunny}
and correlation estimation \citep{zhu2020overcoming}. We approach
this problem differently by addressing the root cause of the problem
which is the instability of the unstructured linguistic representation.

\textbf{Visual-Linguistic association} includes the tasks of text-image
matching \citep{lee2018stacked}, grounding referring expressions
\citep{yu2018mattnet} and cross-domain joint representation \citep{lu2019vilbert,su2020vl}.
The association has been showed to support tasks such as captioning
\citep{zhou2020more,karpathy2015deep}. The key challenges of the
topic include finding a linguistic representation \citep{cirik2018using}
that can match to visual objects \citep{kazemzadeh2014referitgame}.
We approach this problem particularly for Image QA and propose that
using a grammatical structure to replace word-based representation
is crucial for association.

The content presented in this chapter is also related to \textbf{Knowledge
distillation} paradigm \citep{hinton2015distilling}, including cross-task
\citep{albanie2018emotion} and cross modality \citep{gupta2016cross,liu2018multi,wang2020improving}
tasks. Particularly, we distil visual-linguistic grounding knowledge
for training attention of Image QA model.

\section{Preliminaries\label{sec:Chap7-Preliminary}}

To make the chapter self-contained and easy to follow, we repeat
the formulation of Image QA task presented in Chapter \ref{chap:RelationalVisualReasoning}
here:\vspace{-1cm}

\begin{eqnarray}
\hat{y} & = & \underset{a\in\mathcal{\bm{A}}}{\text{argmax}}\mathcal{F}_{\bm{\theta}}\left(a;\mathbf{q},\mathcal{\bm{I}}\right),\label{eq:chap7_scoring}
\end{eqnarray}
where $\hat{y}$ is a predicted answer, $\mathcal{\bm{I}}$ is an
input image, $\mathbf{q}$ is a natural question (also query), and
$\bm{\theta}$ is the model parameters of the scoring function $\mathcal{F}\left(.\right)$.
The answer $\hat{y}$ exists in a pre-defined answer space $\bm{\mathcal{A}}$.

\begin{figure*}
\begin{centering}
\includegraphics[width=1\textwidth]{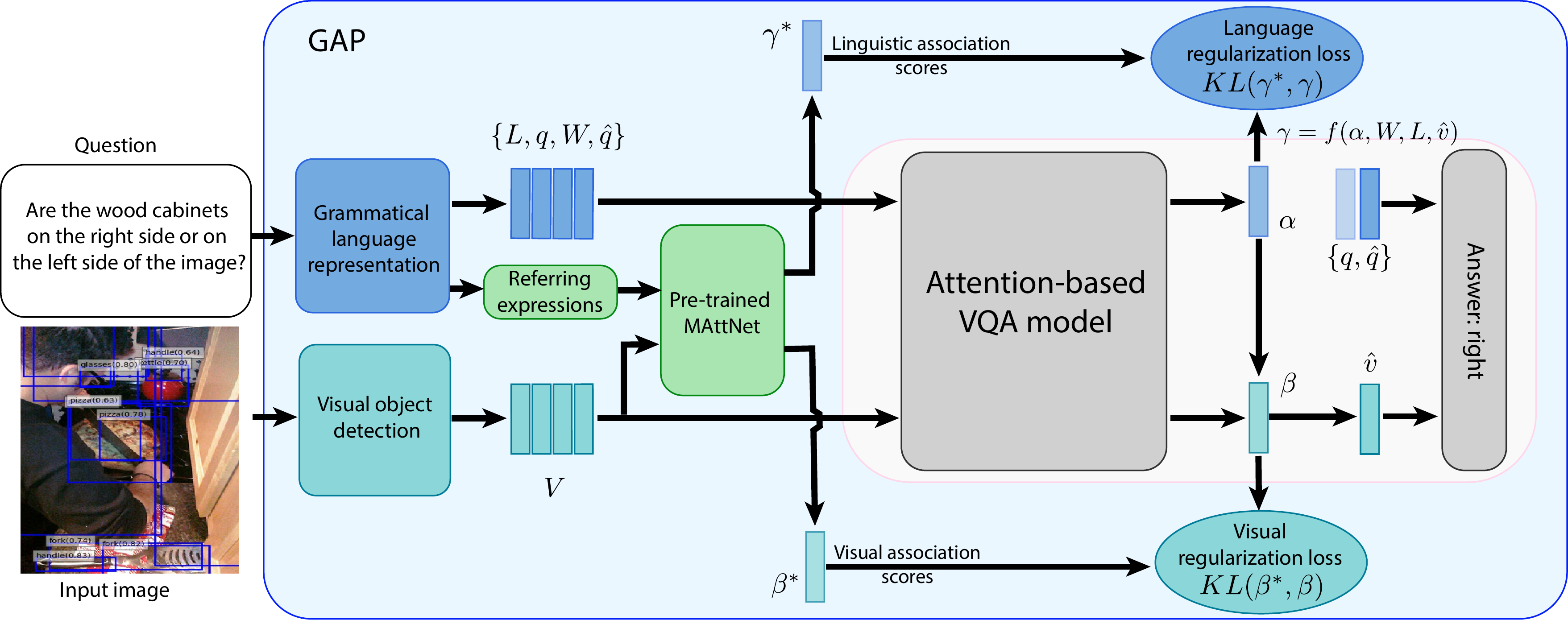}
\par\end{centering}
\caption{Overall architecture of an attention-based Image QA model using Grounding-based
Attention Prior (GAP) to regulate the computation of attention weights.
Gray components are standard for attention-based Image QA models while
components within the large blue rectangle are our contributions in
this chapter. \label{fig:chap7_Overall}}
\end{figure*}

The query $\mathbf{q}$ is first decomposed into a set of $T$ linguistic
entities $\mathbf{L}=\left\{ \mathbf{l}_{i}\right\} _{i=1}^{T}$.
These entities and the query $\mathbf{q}$ are then embedded into
a feature vector space $\mathbf{q}\in\mathbb{R}^{d},\,\mathbf{l}_{i}\in\mathbb{R}^{d}$.
In the case of sequential embedding popularly used for visual question
answering including both Image QA and Video QA as in the previous
chapters, entities are simply query words. They are usually encoded
with GloVe for word-level embedding \citep{pennington2014glove} followed
by RNNs such as biLSTM for sentence-level embedding. Likewise the
image $\bm{\mathcal{I}}$ is often segmented into a set of $N$ visual
regions with features $\mathbf{V}=\left\{ \mathbf{v}_{j}\mid\mathbf{v}_{j}\in\mathbb{R}^{d'}\right\} _{j=1}^{N}$
by an object detector such as Faster R-CNN \citep{ren2015faster}. 

A large family of Image QA systems \citep{lu2016hierarchical,anderson2018bottom,hudson2018compositional,le2020dynamic}
rely on attention mechanisms to distribute conditional computations
on linguistic entities $\mathbf{L}$ and visual counterparts $\mathbf{V}$.
In these models, query entities $\left\{ \mathbf{l}_{j}\right\} _{j=1}^{T}$
and visual objects $\left\{ \mathbf{v}_{j}\right\} _{j=1}^{N}$ are
weighted in consideration with each other. The weights are calculated
as attention scores for linguistic entities $\bm{\alpha}=\left\{ \alpha_{i}\right\} _{i=1}^{T}\in\mathbb{R}^{T}$
and visual entities $\bm{\beta}=\left\{ \beta_{j}\right\} _{j=1}^{N}\in\mathbb{R}^{N}$.

In these models, $\bm{\alpha}$ and $\bm{\beta}$ are calculated based
on the instantiation of input data. They can be implemented in different
ways such as direct single-shot attention \citep{anderson2018bottom},
co-attention \citep{lu2016hierarchical} or multi-step attention \citep{hudson2018compositional}.
We denote $P_{\bm{\theta}}(.)$ as a sub-network that is used to calculate
$\bm{\alpha}$, while $Q_{\bm{\theta}}(.)$ is used to calculate $\bm{\beta}$.
There are two generic attention mechanisms: \emph{Parallel} (where
$\bm{\alpha}$ and $\bm{\beta}$ are calculated concurrently) and
\emph{Alternating} (where one is calculated after and based on the
other, for example, $\bm{\beta}$ calculated after $\bm{\alpha}$
as in \citep{hudson2018compositional} and our LOGNet model in Chapter
\ref{chap:RelationalVisualReasoning} (also \citep{le2020dynamic})
and vice versa as in \citep{lu2016hierarchical}). In our experiment,
we concentrate on the \emph{Alternating mechanism} where the visual
attention is found based on the linguistic attention counterpart because
it is more popular among the successful methods. Here, linguistic
attention scores $\bm{\alpha}$ are found using the sub-network $P_{\bm{\theta}}(.)$:\vspace{-1cm}

\begin{eqnarray}
\alpha_{i} & = & P_{\bm{\theta}}(\mathbf{l}_{i}\mid\mathbf{L},\mathbf{q}),\label{eq:chap7_word_attn}
\end{eqnarray}
followed by the attended linguistic feature of the entire query $\mathbf{c}=\sum_{i=1}^{T}\alpha_{i}*\mathbf{l}_{i};\mathbf{c}\in\mathbb{R}^{d}$.
This attended linguistic feature is then used to calculate the visual
attention scores $\bm{\beta}$ through the sub-network $Q_{\bm{\theta}}(.)$:
\begin{eqnarray}
\beta_{j} & = & Q_{\bm{\theta}}(\mathbf{v}_{j}\mid\mathbf{V},\mathbf{c},\mathbf{q}).\label{eq:chap7_visual_attn}
\end{eqnarray}

In the case of multi-step reasoning models such as those in \citep{hudson2018compositional,le2020dynamic,hudson2019learning},
a pair of linguistic attention $\bm{\alpha}_{k}=\left\{ \alpha_{i,k}\right\} _{i=1}^{T}$
and visual attention $\bm{\beta}_{k}=\left\{ \beta_{j,k}\right\} _{j=1}^{N}$
at each reasoning step $k$ is found separately, forming two matrices
of attention scores. In the case of single-shot attention, the visual
attention $\bm{\beta}$ is calculated directly from the inputs $(\mathbf{V},\mathbf{q})$.

These attention scores drive the whole reasoning process to produce
a joint representation of visual and language which finally gets decoded
into a predicted answer $\hat{a}$:\vspace{-1cm}

\begin{eqnarray}
\hat{a} & = & \text{\ensuremath{\underset{a\in\mathcal{\bm{A}}}{\text{argmax}}}}P(a\mid\mathbf{q},\hat{\mathbf{v}}).
\end{eqnarray}
Here, $P(a\mid\mathbf{q},\hat{\mathbf{v}})=\text{softmax}_{a}(\mathcal{F}_{\bm{\theta}}(a;\mathbf{q},\hat{\mathbf{v}}))$
is computed by the answer decoder taking as input the attended visual
feature $\hat{\mathbf{v}}=\sum_{j=1}^{N}\beta_{j}*\mathbf{v}_{j}$
and the global representation $\mathbf{q}$ of the query.

The model is trained by minimising the question answering loss: \vspace{-1cm}

\begin{eqnarray}
\mathcal{L}_{\text{vqa}} & = & -\frac{1}{D}\sum_{\ell=1}^{D}\text{log}P\left(\hat{a}^{(\ell)}=y^{(\ell)}\mid\mathbf{q}^{(\ell)},\hat{\mathbf{v}}^{(\ell)}\right),\label{eq:chap7_Lvqa}
\end{eqnarray}
where $D$ is the number of training samples, $y$ is the groundtruth
label.

In this process, the attention mechanism is the key component controlling
how effectively the reasoning engine proceeds. In training, these
mechanisms are credit assigned by the gradient flowed back from the
labelled answers which are sparse and indirect. The following section
describes our method to directly supervise the attention calculation
using the grounding priors learned from external data.

\section{Grounding-based Attention Priors \label{sec:chap7-Method}}

This section presents our approach in this chapter, Grounding-based
Attention Priors (GAP), to extract the concept-level association between
query and image and use this external knowledge to regulate the cross-modality
attentions inside Image QA systems. The approach consists of two main
stages. First, we mine the external visual-linguistic grounding data
and extract the knowledge related directly to the specific QA pair
(Sec.~\ref{subsec:chap7_Question-VG-as-Prior-Structure}, green boxes
in Fig. \ref{fig:chap7_Overall}). Second, we use such knowledge as
priors to regulate the attention mechanism in Image QA (Sec.~\ref{subsec:chap7_Regularizing-VQA},
Elipses in Fig. \ref{fig:chap7_Overall}).

\subsection{Finding Query and Image Associations\label{subsec:chap7_Question-VG-as-Prior-Structure}}

Linguistic-visual Grounding aims to find the association between linguistic
and vision entities in a shared context. Modern methods match image
regions to linguistic phrases that share the common semantic meaning,
making them \emph{linguistic referring expressions (REs)} \citep{mao2016generation}.
In the context of Image QA problem, such expressions are neither individual
query words nor the whole question. Instead, these phrases are embedded
in the complex syntactic structure of the question, e.g \emph{``the
white car''} is an RE of the question \emph{``who is driving the
white car?''} For effective and meaningful matching, we propose to
explore this structure to extract the referring expressions and link
such linguistic structures to the visual structure of the image-query
pair. The next subsections detail this process.

\subsubsection{Grammatical Representation of Query\label{subsec:chap7_grammatical_structure}}

\begin{figure}
\begin{centering}
\includegraphics[width=0.65\columnwidth]{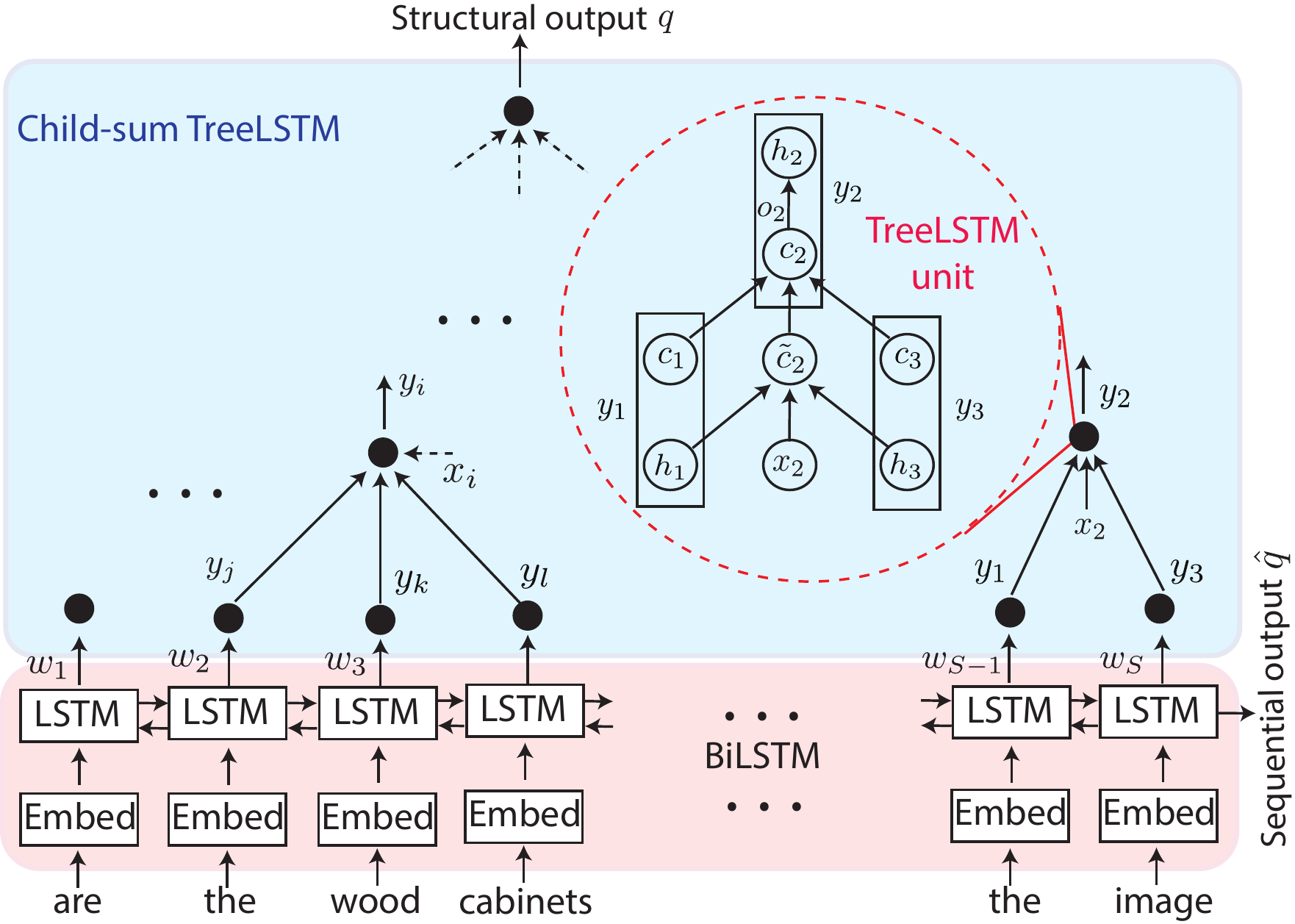}
\par\end{centering}
\caption{Grammatical representation of the query. The query is represented
as a constituency parse tree where each node corresponds to a phrase
of the query. The representations of the nodes are done through TreeLSTM
units. Leaf-node features are the usual sequential biLSTM output of
contextual embeddings.\label{fig:chap7_hybrid_lstm}}
\end{figure}

Attention-based Image QA works on a set of linguistic entities $\mathbf{L}=\left\{ \mathbf{l}_{i}\right\} _{i=1}^{T}$
(see Sec.~\ref{sec:Chap7-Preliminary}). In most existing Image QA
models \citep{anderson2018bottom,hudson2018compositional,hudson2019gqa,cadene2019murel,le2020dynamic},
this set is generated by encoding the input query by sequential modelling
such as RNNs. This results in the linguistic entities representing
contextual words $\mathbf{L}\equiv\mathbf{W}=\left\{ \mathbf{w}_{i}\right\} _{i=1}^{S}$
where $S$ is the query length in words. However, as visual regions
are matched with phrases instead of words, we instead propose to build
the representation $\mathbf{L}$ based on the phrases nested in the
syntactic structures of the query. This new representation approach
will be beneficial for both core reasoning and the grounding process.

A linguistic query is a full sentence that contains multiple entities
and their interactions. We find the linguistic entities in the grammatical
structure of the sentence \citep{cirik2018using} using a constituency
parse tree \citep{chomsky1956three}. This syntactic tree $\bm{\mathcal{T}}$
presents the hierarchical grammar structure of the sentence. This
structure starts at the leaves representing words ordered as they
appear in the query. Traversing up the tree, the lower nodes are joined
up into upper nodes representing higher-level phrases formally known
as constituents toward the root representing the whole sentence. We
gather the nodes of the tree to form a set of $T$ linguistic entities
$\mathbf{L}=\left\{ \mathbf{l}_{i}\right\} _{i=1}^{T}$. Each entity
in this set corresponds to a grammatical phrase of the question. The
feature $\mathbf{l}_{i}$ at each node is calculated based on the
structure of the tree $\mathcal{\bm{T}}$ using the process described
next.

We start by assigning the usual sequential LSTM output of contextual
word embeddings $\mathbf{W}=\left\{ \mathbf{w}_{i}\right\} _{i=1}^{S}$
to be the input of the leaves of the parse tree: $\left\{ \mathbf{x}_{i}\coloneqq\mathbf{w}_{i}\right\} _{i\in\textrm{Leaves of}\,\mathscr{\bm{\mathcal{T}}}}$
. By doing this, we take advantage of both forms of the query (sequence
of words and syntactic structure) in one representation. In traversing
up the tree, we utilise TreeLSTM \citep{tai2015improved} to propagate
information from children to their parent\footnote{In particular, we use the child-sum TreeLSTM variant that supports
varying number of children.}: \vspace{-1cm}

\begin{eqnarray*}
\mathbf{h}_{i},\mathbf{c}_{i} & = & \textrm{TreeLSTM}\left(\left\{ \mathbf{h}_{k},\mathbf{c}_{k},\mathbf{x}_{k}\right\} _{k\in\textrm{children}(i)}\right)
\end{eqnarray*}
Here, each node takes memory cells $\mathbf{c}_{k}$ and hidden states
$\mathbf{h}_{k}$ of its child nodes to compute its own memory $\mathbf{c}_{i}$
and hidden $\mathbf{h_{i}}$. Leaf nodes have these vectors initialised
to zeros. Meanwhile, internal nodes do not use input (See Fig. \ref{fig:chap7_hybrid_lstm}
for illustration).

Similar to a standard LSTM unit, each TreeLSTM unit also contains
different types of gates to control the information flow. In particular,
there are four gates including an input gate $\mathbf{i}_{i}$, an
output gate $\mathbf{o}_{i}$, a memory cell $\mathbf{c}_{i}$ and
hidden state $\mathbf{h}_{i}$ controlling the transition at a TreeLSTM
unit in which they play similar roles as in the standard LSTM unit.
Mathematically, if $k$ is a child node in the children set $\mathbf{C}(i)$
of node $i$, $\mathbf{f}_{ik}$ is the forget gate controlling the
information being sent from node $k$ to node $i$, the transition
equations for each of the gates at node $i$ are given by
\begin{eqnarray}
\mathbf{f}_{ik} & = & \sigma\left(\mathbf{W}^{(f)}\mathbf{x}_{i}+U^{(f)}\mathbf{h}_{k}+\mathbf{b}^{(f)}\right),\\
\tilde{\mathbf{c}}_{i} & = & \text{tanh}\left(\mathbf{W}^{(c)}\mathbf{x}_{i}+\sum_{k\in\mathbf{C}(i)}\mathbf{U}^{(c)}\mathbf{h}_{k}+\mathbf{b}^{(c)}\right),\\
\mathbf{i}_{i} & = & \sigma\left(\mathbf{W}^{(i)}\mathbf{x}_{i}+\sum_{k\in\mathbf{C}(i)}\mathbf{U}^{(i)}\mathbf{h}_{k}^{\text{}}+\mathbf{b}^{(i)}\right),\\
\mathbf{c}_{i} & = & \mathbf{i}_{i}\varodot\tilde{\mathbf{c}}_{i}+\sum_{k\in\mathbf{C}(i)}\mathbf{f}_{ik}\varodot\mathbf{c}_{k},\\
\mathbf{o}_{i} & = & \sigma\left(\mathbf{W}^{(o)}\mathbf{x}_{i}+\sum_{k\in\mathbf{C}(i)}\mathbf{U}^{(o)}\mathbf{h}_{k}^{\text{}}+\mathbf{b}^{(o)}\right),\\
\mathbf{h}_{i}^{\text{}} & = & \mathbf{o}_{i}\varodot\text{tanh}\left(\mathbf{c}_{i}\right),
\end{eqnarray}
where $\mathbf{x}_{i}$ is the input vector at node $i$; $\mathbf{W}$and
$\mathbf{U}$ are network parameters. 

At the end of the TreeLSTM inference, we assign the hidden states
to be the features of the linguistic entities: $\mathbf{l}_{i}\coloneqq\mathbf{h}_{i}$
and the global representation of the query to be the hidden of the
root: $\mathbf{q}\coloneqq\mathbf{h}_{\textrm{root}}$.

This method of linguistic representation allows attention-based Image
QA models to attend to linguistic expressions and visual objects where
they refer to entities of the compatible semantic level. As a result,
it maximises the efficiency of the linguistic-visual attention mechanism
and improves the performance (See experimental results in table \ref{tab:chap7_Ablation-studies}).
In addition, having phrase-level representation of the query opens
up the ability to connect Image QA attentions to external grounding
knowledge allowing direct regularisation. This vision is realised
in the next section.

\subsubsection{Grounding Query to Visual Regions\label{subsec:chap7_Grounding-query-to_vis_regions}}

The grounding is defined by the pairwise association between a set
of referring expressions $\mathbf{E}=\left\{ \mathbf{e}_{r}\right\} _{r=1}^{R}$
extracted from the query and a set of visual regions $\mathbf{V}=\left\{ \mathbf{v}_{j}\right\} _{j=1}^{N}$
from the image. 

\begin{figure}
\begin{centering}
\includegraphics[width=0.65\columnwidth]{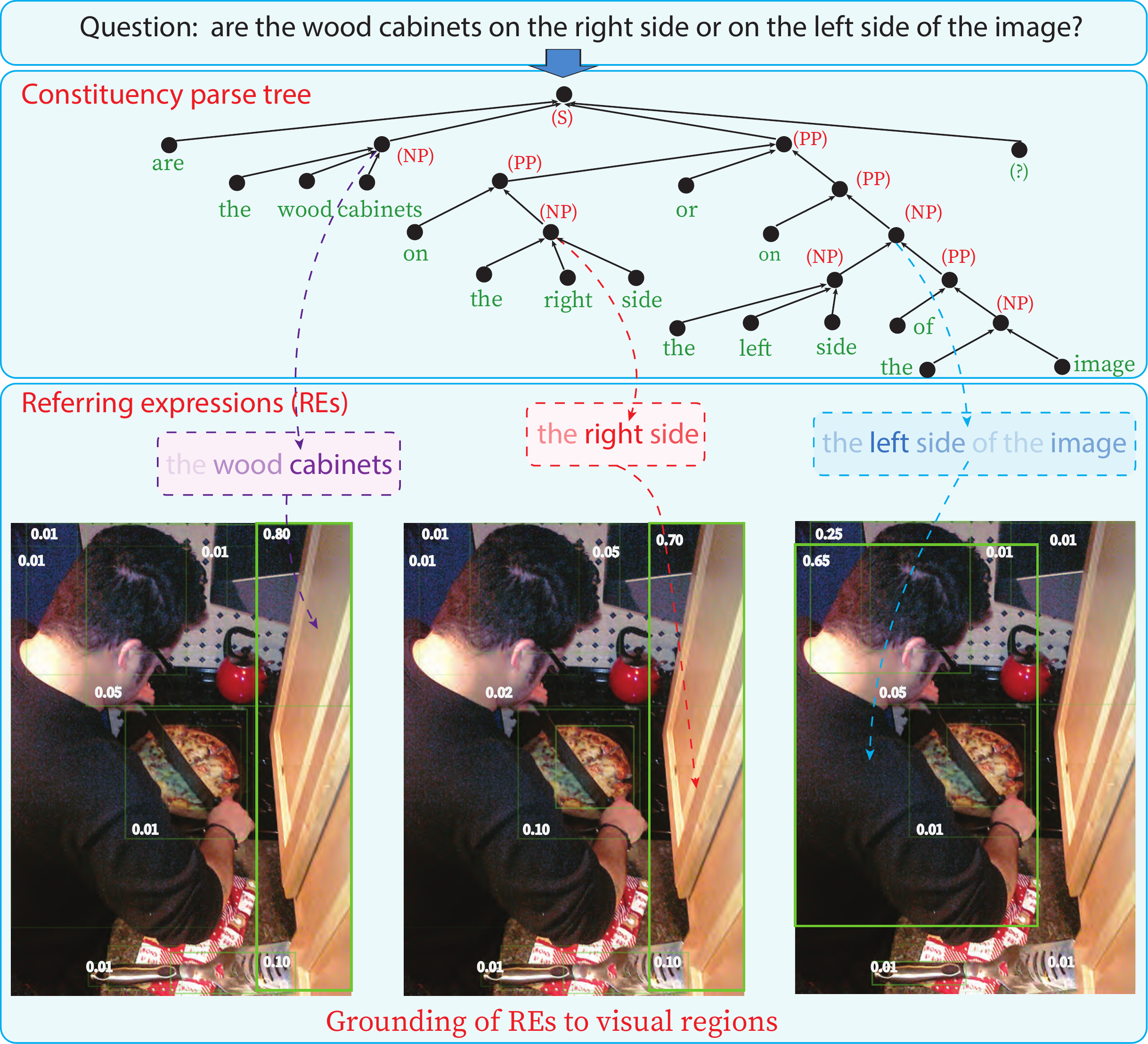}
\par\end{centering}
\caption{The query is parsed into a syntactic tree where the referring expressions
(REs) are exposed. Those REs are grounded to image regions to provide
priors for Image QA models.\label{fig:chap7_Groundeing_illustration}}
\end{figure}

In extracting $\mathbf{E}$, we revisit the grammatical structure
$\bm{\mathcal{T}}$ which provides the ability not only to break down
a query into a set of phrases $\mathbf{L}$ but also to associate
them with specific grammatical roles such as \emph{noun-phrases (NP),
verb-phrases (VP)} \citep{bies1995bracketing}. E.g.\emph{ ``the
white car''} is marked as a NP while \emph{``driving the white car''}
is a VP. As visual objects and regions are naturally associated with
noun-phrases, we select the set $\mathbf{E}\subset\mathbf{L}$ of
all the noun phrases and wh-noun phrases\footnote{noun phrases prefixed by a pronoun, e.g. ``which side of the photo'',
``whose bag''} as the referring expressions. Fig. \ref{fig:chap7_Groundeing_illustration}
demonstrates the selection of three REs from the question \emph{``are
the wood cabinets on the right side or on the left side of the image?''}
We use the Berkeley Neural Parser \citep{kitaev2018constituency}
for the constituent tree building operations.

The REs are then grounded to visual objects $\mathbf{V}=\{\mathbf{v}_{j}\}_{j=1}^{N}$
by a pre-trained grounding method such as those are in \citep{deng2018visual,yu2018mattnet,wang2019neighbourhood}
which learn to attend to critical words in the expressions and identify
which visual regions are relevant to them. In our implementation,
we use the popular Modular Attention Network (MAttNet) \citep{yu2018mattnet}
trained on RefCOCO dataset \citep{kazemzadeh2014referitgame}. 

For each RE $\mathbf{e}_{r}$ containing its member words $\left\{ \mathbf{w}_{ir}\right\} _{i=1}^{M}$,
where $M$ is the number of words in $\mathbf{e}_{r}$, we get two
sets of \emph{association scores} $\left\{ \gamma_{ir}^{*}\right\} _{i=1}^{M}$
for linguistic side and $\left\{ \beta_{jr}^{*}\right\} _{j=1}^{N}$
for the visual side. Consequentially, the linguistic association scores
$\gamma_{ir}^{*}$ weight the importance of word $\mathbf{w}_{ir}$
in the grounding to $\mathbf{V}$, and visual scores $\beta_{jr}^{*}$
reflects the association between region $\mathbf{v}_{j}$ and expression
$\mathbf{e}_{r}$:\vspace{-1cm}

\begin{eqnarray}
\gamma_{ir}^{*} & = & P(\mathbf{w}_{ir}\mid\mathbf{e}_{r}),\\
\beta_{jr}^{*} & = & P(\mathbf{v}_{j}\mid\hat{\mathbf{c}}_{r},\mathbf{V}),
\end{eqnarray}
where $\hat{\mathbf{c}}_{r}=\sum_{i=1}^{M}\gamma_{ir}^{*}*\mathbf{w}_{ir}$
is the attended linguistic feature.

These scores form the prior association between visual regions and
REs. As the REs are nested phrases, one word can appear across REs.
Hence, we calculate word-level association scores by probabilistic
opinion pooling \citep{dietrich2014probabilistic}, such as by simple
voting among all REs:
\begin{eqnarray}
\gamma_{i}^{*} & = & \frac{1}{R}\sum_{r=1}^{R}\gamma_{ir}^{*}\:\textrm{for}\:i=1,...,S.\label{eq:chap7_textual_grounding_scores-1}
\end{eqnarray}

Similarly, we gather visual association scores for each visual object
$\mathbf{v}_{j}$ by 
\begin{eqnarray}
\beta_{j}^{*} & = & \frac{1}{R}\sum_{r=1}^{R}\beta_{jr}^{*}\:\textrm{for}\:j=1,...,N.\label{eq:chap7_visual_grounding_scores}
\end{eqnarray}

These association scores $\bm{\gamma}^{*}$ and $\bm{\beta}^{*}$
are the key output of the grounding process. They act as the \emph{attention
priors} to regulate Image QA models' attention weights in the process
detailed in the next section.

\subsection{Regularising Image QA with Grounding-based Attention Priors\label{subsec:chap7_Regularizing-VQA}}

Recall from Sec.~\ref{sec:Chap7-Preliminary} that a Image QA models
generally take a set of linguistic entities $\mathbf{L}=\left\{ \mathbf{l}_{i}\right\} _{i=1}^{T}$
and a set of visual regions $\mathbf{V}=\{\mathbf{v}_{j}\}_{j=1}^{N}$
to deduce the answer through an attention mechanism. Although the
attention mechanisms have many variations, they can all be generalised
into a pair of linguistic attention over query entities $\bm{\alpha}\in\mathbb{R}^{T}$
and visual attention over visual regions $\bm{\beta}\in\mathbb{R}^{N}$.
In case of multi-step reasoning, we consider $\bm{\alpha}$ and $\bm{\beta}$
as the average pooling of step-wise weights $\alpha_{i,k}$ and $\beta_{j,k}$
of all step $k$.

The regularisation uses the association scores priors $\bm{\gamma}^{*}$
and $\bm{\beta}^{*}$ (Eq. \ref{eq:chap7_textual_grounding_scores-1},\ref{eq:chap7_visual_grounding_scores})
as supervision signals to regulate the computation of attention $\bm{\alpha}$
and $\bm{\beta}$. This effectively forces the reasoning process to
agree with the external common knowledge extracted from the grounding.

While $\bm{\gamma}^{*}$ are defined on words, $\bm{\alpha}$ are
generally calculated on linguistic entities such as phrases. To connect
these two, we use a learnable mapping function that distributes the
linguistic attention from entity-level down to word-level: \vspace{-1cm}

\begin{eqnarray}
\bm{\gamma} & = & f(\bm{\alpha},\mathbf{L},\mathbf{W},\hat{\mathbf{v}}),
\end{eqnarray}
where $\hat{\mathbf{v}}=\sum_{j=1}^{N}\beta_{j}*\mathbf{v}_{j}$. 

Standing out from all other attention supervision works, we apply
the regularisation on both attention channels. On linguistic attention
regularisation, we minimise a loss $\mathcal{L}_{\text{ling\_reg}}$
in the form of the Kullback-Leibler divergence \citep{kullback1951information}
between the attention weights and the prior association scores:\vspace{-1cm}

\begin{eqnarray}
\mathcal{L}_{\text{ling\_reg}} & = & KL(\bm{\gamma}^{*}\parallel\bm{\gamma})=\sum_{i}\gamma_{i}^{*}\text{log}\left(\frac{\gamma_{i}^{*}}{\gamma_{i}}\right).\label{eq:chap7_ling_reg_loss}
\end{eqnarray}

On visual channel, we similarly force the visual attention to resemble
the prior visual association scores: \vspace{-1cm}

\begin{eqnarray}
\mathcal{L}_{\text{vis\_reg}} & = & KL(\bm{\beta}^{*}\parallel\bm{\beta})=\sum_{j}\beta_{j}^{*}\text{log}\left(\frac{\beta_{j}^{*}}{\beta_{j}}\right).\label{eq:chap7_vis_reg_loss}
\end{eqnarray}

\subsection{Training}

As coming from the external source, the grounding-based attention
priors are imperfect and may contain noises. We design to effectively
use this valuable but noisy guidance signal by only use the regulation
in training but not in inference time. The training process aims to
use this general knowledge to softly guide the learning process of
attention mechanisms while avoiding being directly affected by noises.

In this process, we train the attention-based Image QA models end-to-end
with the weighted combined loss between the answering cross-entropy
loss (Eq.\ref{eq:chap7_Lvqa}) and the regulation losses (Eq.\ref{eq:chap7_ling_reg_loss}
and \ref{eq:chap7_vis_reg_loss}):\vspace{-1cm}

\begin{eqnarray}
\mathcal{L} & = & \mathcal{L}_{\text{vqa}}+\lambda_{l}\mathcal{L}_{\text{ling\_reg}}+\lambda_{v}\mathcal{L}_{\text{vis\_reg}},\label{eq:chap7_sum_loss}
\end{eqnarray}
where $\lambda_{l},\lambda_{v}>0$ are regularisation weights.

\section{Experiments \label{sec:Chap7-Experiments}}

\subsection{Datasets and Image QA Backbones }

\subsubsection{Backbone Networks}

We evaluate our approach on three representatives of the major model
families of Image QA, namely Bottom-Up Top-Down Attention (UpDn) \citep{anderson2018bottom}
for single-shot dual-attention, MACNet \citep{hudson2018compositional}
for multi-step compositional attention, and LOGNet \citep{le2020dynamic}
for relational object-centric reasoning.

\paragraph{Bottom-Up Top-Down Attention (UpDn)}

UpDn is the first to introduce the use of bottom-up attention mechanisms
to Image QA by utilising image region features extracted by Faster
R-CNN \citep{ren2015faster} pre-trained on Visual Genome dataset
\citep{krishna2017visual}. A top-down attention network driven by
the question representation is then used on top of the image region
features to find the attended image region for answering the given
question. The UpDn model won the VQA Challenge in 2017 and became
a standard baseline Image QA model since then.

\paragraph{MACNet}

MACNet is a multi-step co-attention based model to perform sequential
reasoning where they use Image QA as the testbed. Given a set of contextual
word embeddings and a set of visual region features, at each time
step, a MAC cell learns the interactions between the two sets with
the consideration of their history interaction at previous time steps
via a memory state. In particular, a controller first computes a vector
of linguistic attention scores. The linguistic attention vector is
then coupled with the memory state of the previous reasoning step
to compute the visual attention scores. At the end of a reasoning
step, the attended visual feature is finally used to update the memory
state of the reasoning process. The process is repeated over multiple
steps, resembling the way humans reason about the world. In our experiments,
we use a Pytorch equivalent implementation\footnote{https://github.com/tohinz/pytorch-mac-network}
of MACNet instead of using the original Tensorflow-based implementation.

\paragraph{LOGNet}

LOGNet is our relational visual reasoning model that we introduced
in Chapter \ref{chap:RelationalVisualReasoning}. It also falls in
the multi-step compositional attention reasoning family. However,
different from other approaches, it considers both intra- and inter-modality
interactions. For the intra-modality interactions, it uses a dynamic
graph to model the pairwise relations between visual objects while
the inter-modality interactions are done via a linguistic binding
process followed by representation refinement. The graph structure
representation of input modalities and the linguistic binding not
only benefits in terms of performance but also the interpretability
of the model, making the reasoning operation under the hood more transparent. 

To reduce the training time and complexity of LOGNet, we conduct experiments
with only 6 reasoning steps and 4 GCN layers at each step instead
of 8-12 reasoning steps and 8 GCN layers at each step as in the in
Chapter \ref{chap:RelationalVisualReasoning}.

\subsubsection{Datasets}

Experiments are done on three large scale datasets: VQA v2 \citep{goyal2017making},
VQA-Rephrasings \citep{shah2019cycle} and GQA \citep{hudson2019gqa}.
Among these, the GQA and VQA v2 are popular benchmarks for Image QA
models and VQA-Rephrasings is a large-scale benchmark measuring the
robustness of Image QA models against linguistic variations. The VQA-Rephrasings
includes two parts:\emph{ original (ORI)} split being a subset of
the VQA v2 val set and \emph{rephrased (REP) }split containing questions
rephrased by human annotators. 

\paragraph{VQA v2 }

is a large scale VQA dataset entirely based on human annotation and
is the most popular benchmark for VQA models. It contains 1.1M questions
with more than 11M answers annotated from over 200K MSCOCO images
\citep{lin2014microsoft}, of which 443,757 questions, 214,354 questions
and 447,793 questions in train, val and test split, respectively. 

We choose correct answers in the training set that appear more than
8 times, similar to existing works \citep{teney2018tips,anderson2018bottom}.
We report performance with accuracies calculated by standard VQA accuracy
metric: $\text{min}(\frac{\text{\#humans that provided that answer}}{3},1)$
\citep{antol2015vqa}.

\paragraph{VQA-Rephrasings dataset}

Despite the popularity of the VQA v2 dataset, it has limitations in
terms of supporting the measurement of the robustness of VQA models
against linguistic variations. The VQA-Rephrashings dataset is particularly
designed to fill the gap by providing three human-annotated rephrasings
for each of 40,054 questions across 40,504 images in the standard
VQA v2 validation split. While state-of-the-art VQA models show impressive
performance on the standard VQA v2 dataset, they struggle to understand
the same question expressed in slightly different ways.

\paragraph*{GQA}

is currently the largest VQA dataset. Please refer back to Sec.~\ref{subsec:chap6_Datasets}
for more details about this dataset.

\begin{figure*}
\begin{centering}
\includegraphics[width=1\textwidth]{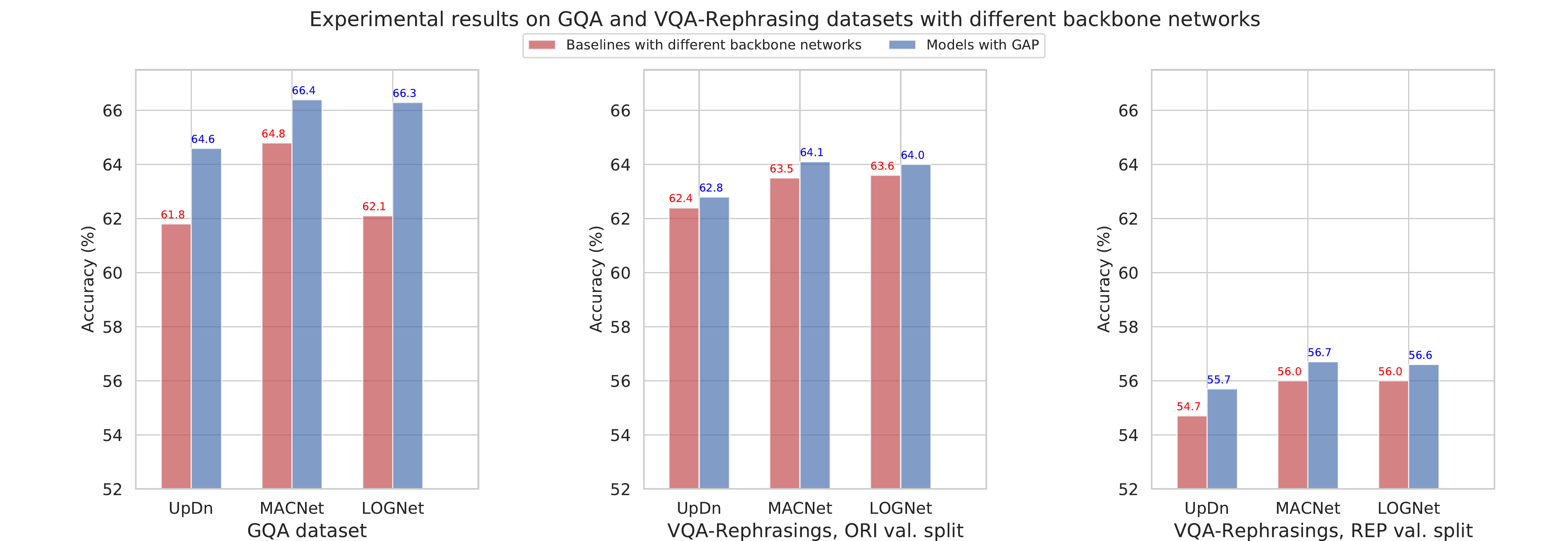}
\par\end{centering}
\caption{Experimental results on validation set of GQA dataset and both ORI
and REP split of VQA-Rephrasings dataset. GAP is effective in enhancing
all of the three representative backbone Image QA models (UpDn, MACNet,
and LOGNet).\label{fig:chap7_comparing_against_baselines_gqa_vqav2}}
\end{figure*}

\subsection{Implementation Details}

\subsubsection{Initial Language and Visual Embedding}

\paragraph{Textual embedding }

Given a length-$S$ question query, we first tokenise it into a sequence
of words and further embed each word into the vector space of 300
dimensions. We initialise word embeddings by the popular pre-trained
vectors representations in GloVe \citep{pennington2014glove}.

To model the sequential structure of language, we use bidirectional
LSTMs (biLSTMs) taking as input the word embedded vectors. The biLSTMs
result in hidden states $\overrightarrow{\mathbf{h}_{i}^{\text{seq}}}$
and $\overleftarrow{\mathbf{h}_{i}^{\text{seq}}}$ at time step $i$
for the forward pass and backward pass, respectively. We further combine
every pair $\overrightarrow{\mathbf{h}_{i}^{\text{seq}}}$ and $\overleftarrow{\mathbf{h}_{i}^{\text{seq}}}$
into a single vector $\mathbf{w}_{i}=\left[\overrightarrow{\mathbf{h}_{i}^{\text{seq}}};\overleftarrow{\mathbf{h}_{i}^{\text{seq}}}\right]$,
where $\left[;\right]$ indicates the vector concatenation operator.
The contextual words are then obtained by gathering these combined
vectors $\mathbf{W}=\left\{ \mathbf{w}_{i}\mid\mathbf{w}_{i}\in\mathbb{R}^{d}\right\} _{i=1}^{S}$.
The global representation $\mathbf{q}$ of the query sequential structure
is obtained by combining the the ends of the LSTM passes $\mathbf{q}=\left[\overrightarrow{\mathbf{h}_{1}^{\text{seq}}};\overleftarrow{\mathbf{h}_{S}^{\text{seq}}}\right]$.
In the case when we do not use the grammatical representation of the
query, $\mathbf{W}$ and $\mathbf{q}$ are used as the linguistic
entities and global representation of the query, respectively.

\paragraph{Visual embedding }

Visual regions are extracted by the popular object detection Faster
R-CNN \citep{ren2015faster} pre-trained on Visual Genome \citep{krishna2017visual}.
We use public code\footnote{https://github.com/MILVLG/bottom-up-attention.pytorch}
making use of the Facebook Detectron2 v2.0.1 framework\footnote{https://github.com/facebookresearch/detectron2}
for this purpose. For each image, we extract a set of $N$ RoI pooling
features with bounding boxes $\left\{ \left(\mathbf{a}_{j},\mathbf{b}_{j}\right)\right\} _{j=1}^{N}$,
where $\mathbf{a}_{j}\in\mathbb{R}^{2048},\mathbf{b}_{j}\in\mathbb{R}^{4}$
are appearance features of object regions and bounding box's coordinators,
respectively. We follow \citep{yu2017joint} to encode the bounding
box's coordinators into a spatial vector of 7 dimensions. We further
combine the appearance features with the encoded spatial features
by using a sub-network of two linear transformations to obtain a set
of visual objects $\mathbf{V}=\left\{ \mathbf{v}_{j}\mid\mathbf{v}_{j}\in\mathbb{R}^{d^{\prime}}\right\} _{j=1}^{N}$,
where $d^{\prime}$ is the vector length of the joint features between
the appearance features and the spatial features. For ease of implementation,
we choose the feature size of the linguistic side $d$ and the visual
size $d^{\prime}$ to be the same.

\subsubsection{Pre-training MAttNet}

We make use of the public source code\footnote{https://github.com/lichengunc/MAttNet}
of MAttNet to extract linguistic-visual grounding priors. However,
we extract visual features using the same Detectron2 framework that
we use for visual feature extraction for VQA instead of using the
original object detection implemented in the MAttNet's code. We train
the MAttNet ourselves on RefCOCO dataset \citep{kazemzadeh2014referitgame}
with the newly extracted visual features. 

We then use the pre-trained MAttNet model by feeding a linguistic
expression extracted as in Subsec. \ref{subsec:chap7_Grounding-query-to_vis_regions}
and its corresponding visual features to extract linguistic-visual
grounding priors.

\subsubsection{Hyperparameter Selection}

We choose the number of visual objects $N=100$ per image following
previous works \citep{hu2019language,hudson2019gqa}. The feature
dimensions $d=512$ and $d^{\prime}=512$ for all models in all experiments.
For MACNet and LOGNet, we choose the number of reasoning steps to
be $6$ in all experiments. As mentioned before, while results in
Chapter \ref{chap:RelationalVisualReasoning} suggested the best performance
of LOGNet with $8$ GCN layers in each reasoning step, we conduct
experiments with only 4 GCN layers each step in this chapter to reduce
the training and inference time.

Through experiments, we realise that GAP is not sensitive to the loss
weights $\lambda_{v}$ and $\lambda_{l}$ in Eq. \ref{eq:chap7_sum_loss}.
We simply have all losses share the same weights in all experiments. 

\subsection{Experimental Results}

\subsubsection{Enhancing Image QA Performance }

We compare GAP against the available VQA attention regularization
methods using UpDn model \citep{anderson2018bottom} on VQA v2 dataset.
Two of these methods use internal regularisation: \emph{adversarial
regularisation} (AdvReg){\small{} }\citep{ramakrishnan2018overcoming},
\emph{attention alignment} (Attn. Align) \citep{selvaraju2019taking},
the other two use human attention as external supervision: \emph{self-critical
reasoning} (SCR) \citep{wu2019self} and \emph{HINT} \citep{selvaraju2019taking}.
\emph{UpDn+word-based GAP} indicates our method without the grammatical
representation (word-level embeddings instead) of the query while
UpDn+GAP refers to the full model. 

\begin{table}
\begin{centering}
\begin{tabular}{l|cccc}
\hline 
\multirow{2}{*}{{\small{}Method}} & \multicolumn{4}{c}{{\small{}VQA v2 standard val$\uparrow$}}\tabularnewline
\cline{2-5} \cline{3-5} \cline{4-5} \cline{5-5} 
 & {\small{}All} & {\small{}Yes/ No} & {\small{}Num} & {\small{}Other}\tabularnewline
\hline 
\hline 
{\small{}UpDn} & {\small{}62.8} & {\small{}80.9} & {\small{}42.8} & {\small{}54.4}\tabularnewline
{\small{}UpDn+Attn. Align \citep{selvaraju2019taking}} & {\small{}63.2} & {\small{}81.0} & {\small{}42.6} & {\small{}55.2}\tabularnewline
{\small{}UpDn+AdvReg \citep{ramakrishnan2018overcoming}} & {\small{}62.7} & {\small{}79.8} & {\small{}42.3} & {\small{}55.2}\tabularnewline
{\small{}UpDn+SCR (w. ext.) \citep{wu2019self}} & {\small{}62.2} & {\small{}78.8} & {\small{}41.6} & {\small{}54.5}\tabularnewline
{\small{}UpDn+SCR (w/o ext.) \citep{wu2019self}} & {\small{}62.3} & {\small{}77.4} & {\small{}40.9} & {\small{}56.5}\tabularnewline
{\small{}UpDn+DLR \citep{jing2020overcoming}} & {\small{}58.0} & {\small{}76.8} & {\small{}39.3} & {\small{}48.5}\tabularnewline
{\small{}UpDn+HINT \citep{selvaraju2019taking}} & {\small{}63.4} & \textbf{\small{}81.2} & {\small{}43.0} & {\small{}55.5}\tabularnewline
\hline 
{\small{}UpDn+word-based GAP} & {\small{}63.5} & {\small{}79.6} & \textbf{\small{}43.2} & {\small{}56.5}\tabularnewline
{\small{}UpDn+GAP} & \textbf{\small{}63.6} & {\small{}79.7} & {\small{}43.1} & \textbf{\small{}56.9}\tabularnewline
\hline 
\end{tabular}\medskip{}
\par\end{centering}
\caption{Performance comparison between GAP and other attention regularisation
methods using UpDn on VQA v2. Results of other methods are taken from
corresponding papers.\label{tab:chap7_Comparing-att-reg-methods}}
\end{table}

\begin{table}
\begin{centering}
\begin{tabular}{l|cccc|c>{\centering}p{2cm}}
\hline 
\multirow{2}{*}{{\small{}Method}} & \multicolumn{4}{>{\centering}p{2.5cm}|}{{\small{}CS(k)$\uparrow$}} & \multicolumn{2}{c}{{\small{}Accuracy}}\tabularnewline
\cline{2-7} \cline{3-7} \cline{4-7} \cline{5-7} \cline{6-7} \cline{7-7} 
 & {\small{}k=1} & {\small{}k=2} & {\small{}k=3} & {\small{}k=4} & {\small{}REP$\uparrow$} & {\small{}Degrade from ORI$\downarrow$}\tabularnewline
\hline 
\hline 
{\small{}UpDn \citep{shah2019cycle}} & {\small{}60.5} & {\small{}47.0} & {\small{}40.5} & {\small{}34.5} & {\small{}51.2} & {\small{}10.3}\tabularnewline
{\small{}UpDn+CC \citep{shah2019cycle}} & {\small{}61.7} & {\small{}50.8} & {\small{}44.7} & {\small{}42.5} & {\small{}52.6} & {\small{}9.8}\tabularnewline
{\small{}UpDn+BERT (CLS)} & {\small{}50.3} & {\small{}43.3} & {\small{}39.5} & {\small{}36.9} & {\small{}42.7} & {\small{}6.1}\tabularnewline
{\small{}UpDn+BERT (Avg)} & {\small{}60.4} & {\small{}52.7} & {\small{}48.3} & {\small{}45.4} & {\small{}52.1} & \textbf{\small{}5.1}\tabularnewline
\hline 
{\small{}UpDn+word-based GAP} & {\small{}63.8} & {\small{}55.3} & {\small{}50.1} & {\small{}46.6} & {\small{}55.1} & {\small{}7.6}\tabularnewline
{\small{}UpDn+GAP} & \textbf{\small{}64.5} & \textbf{\small{}56.2} & \textbf{\small{}51.3} & \textbf{\small{}47.9} & \textbf{\small{}55.7} & {\small{}7.1}\tabularnewline
\hline 
\end{tabular}\medskip{}
\par\end{centering}
\caption{Image QA models's robustness to linguistic variations. CS(k): consensus
performance denoting the proportion of\emph{ at least k rephrasings
}answered\emph{ }correctly or zero otherwise. BERT(CLS): using output
of the CLS-token as the query representation; BERT(Avg): using mean
of all output as the query representation.\label{tab:chap7_exps_vqa_rephrasings}}
\end{table}

Results in Table \ref{tab:chap7_Comparing-att-reg-methods} shows
that our approach \emph{(UpDn+GAP)} improves around 0.8 percentage
points on top of the UpDn baseline and outperforms all other regularisation
methods. The favourable performance is consistent across most question
types. 

Compared directly to \emph{UpDn+SCR }and \emph{UpDn+HINT}, the results
suggest that our methodology of regulating visual reasoning model
using linguistic-visual grounding attention priors is more effective
than using human annotation. This is on top of the fact that using
available data is cheaper and more feasible in practice than collecting
new annotations.

\subsubsection{Robustness to Question Rephrasing}

Besides the overall performance, we evaluate the performance of GAP
in robustness against linguistic variations by experimenting on the
VQA-Rephrasings dataset. We compare GAP with two regularisation techniques
popular for this challenge, namely \emph{cycle consistency} \citep{shah2019cycle}
and\emph{ BERT pre-training} \citep{devlin2018bert}.

\begin{table}
\begin{centering}
\begin{tabular}{c|l|c|>{\centering}p{1.7cm}>{\centering}p{1.7cm}}
\hline 
\multirow{2}{*}{{\footnotesize{}No.}} & \multirow{2}{*}{{\small{}Models}} & \multirow{2}{*}{{\small{}GQA val. $\uparrow$}} & \multicolumn{2}{c}{{\small{}VQA- Rephrasings}}\tabularnewline
\cline{4-5} \cline{5-5} 
 &  &  & {\small{}ORI$\uparrow$} & {\small{}REP$\uparrow$}\tabularnewline
\hline 
\hline 
 & \textbf{\small{}UpDn backbone} &  &  & \tabularnewline
{\footnotesize{}1} & {\small{}$\quad$UpDn baseline} & {\small{}61.8} & {\small{}62.4} & {\small{}54.7}\tabularnewline
{\footnotesize{}2} & {\small{}$\quad$UpDn+word-based GAP} & {\small{}62.9} & {\small{}62.7} & {\small{}55.1}\tabularnewline
{\footnotesize{}3} & {\small{}$\quad$UpDn+TreeLSTM} & {\small{}62.3} & {\small{}62.4} & {\small{}55.1}\tabularnewline
{\footnotesize{}4} & {\small{}$\quad$UpDn+vis.GAP} & \textbf{\small{}64.9} & \textbf{\small{}63.1} & {\small{}55.4}\tabularnewline
{\footnotesize{}5} & {\small{}$\quad$UpDn+ling.GAP} & {\small{}62.4} & {\small{}62.3} & {\small{}54.6}\tabularnewline
{\footnotesize{}6} & \textbf{\small{}$\quad$}{\small{}UpDn+GAP} & {\small{}64.6} & {\small{}62.8} & \textbf{\small{}55.7}\tabularnewline
\hline 
 & \textbf{\small{}MACNet backbone} &  &  & \tabularnewline
{\footnotesize{}1} & {\small{}$\quad$MACNet baseline} & {\small{}64.8} & {\small{}63.5} & {\small{}56.0}\tabularnewline
{\footnotesize{}2} & {\small{}$\quad$MACNet+word-based GAP} & {\small{}66.1} & {\small{}63.8} & {\small{}56.2}\tabularnewline
{\footnotesize{}3} & {\small{}$\quad$MACNet+TreeLSTM} & {\small{}65.3} & {\small{}63.5} & {\small{}56.0}\tabularnewline
{\footnotesize{}4} & {\small{}$\quad$MACNet+vis.GAP} & {\small{}66.4} & {\small{}63.9} & {\small{}56.5}\tabularnewline
{\footnotesize{}5} & {\small{}$\quad$MACNet+ling.GAP} & {\small{}65.7} & {\small{}63.7} & {\small{}55.8}\tabularnewline
{\footnotesize{}6} & {\small{}$\quad$MACNet+GAP} & \textbf{\small{}66.4} & \textbf{\small{}64.1} & \textbf{\small{}56.7}\tabularnewline
\hline 
 & \textbf{\small{}LOGNet backbone} &  &  & \tabularnewline
{\footnotesize{}1} & {\small{}$\quad$LOGNet baseline} & {\small{}62.1} & {\small{}63.6} & {\small{}56.0}\tabularnewline
{\footnotesize{}2} & {\small{}$\quad$LOGNet+word-based GAP} & {\small{}65.8} & {\small{}63.8} & {\small{}56.1}\tabularnewline
{\footnotesize{}3} & {\small{}$\quad$LOGNet+TreeLSTM} & {\small{}62.3} & {\small{}63.1} & {\small{}56.2}\tabularnewline
{\footnotesize{}4} & {\small{}$\quad$LOGNet+vis.GAP} & {\small{}66.1} & {\small{}63.6} & {\small{}56.6}\tabularnewline
{\footnotesize{}5} & {\small{}$\quad$LOGNet+ ling.GAP} & {\small{}65.4} & {\small{}62.2} & {\small{}55.7}\tabularnewline
{\footnotesize{}6} & {\small{}$\quad$LOGNet+GAP} & \textbf{\small{}66.3} & \textbf{\small{}64.0} & \textbf{\small{}56.6}\tabularnewline
\hline 
\end{tabular}\medskip{}
\par\end{centering}
\caption{Ablation studies on GQA and VQA-Rephrasing dataset.  vis.GAP and
ling.GAP refer to models having only visual attention and only linguistic
attention regulated, respectively.\label{tab:chap7_Ablation-studies}}
\end{table}

Results in table \ref{tab:chap7_exps_vqa_rephrasings} show that comparing
to \emph{cycle consistency (CC)} \citep{shah2019cycle}, GAP has better
performance in large margins in both consensus performance and accuracy.
In particular, GAP significantly reduces language bias by gaining
over 3.0 percentage points on the VQA-Rephrashings REP split and reduces
2.7 points in performance degradation. 

Comparing to two variants of the BERT pre-training (either using CLS-token
output or mean of outputs as the query representation), GAP also shows
clear advantages over BERT which has much lower performance on both
ORI and REP splits. 

Looking deeper into the consensus performance of our method with and
without TreeLSTM, it shows that exploiting the grammatical structure
helps produce more consistent answers across rephrased questions when
$k$ increases. 

\begin{figure*}
\begin{centering}
\noindent\begin{minipage}[t]{1\textwidth}%
\begin{center}
\begin{minipage}[t]{0.46\textwidth}%
\begin{center}
\includegraphics[width=0.9\columnwidth]{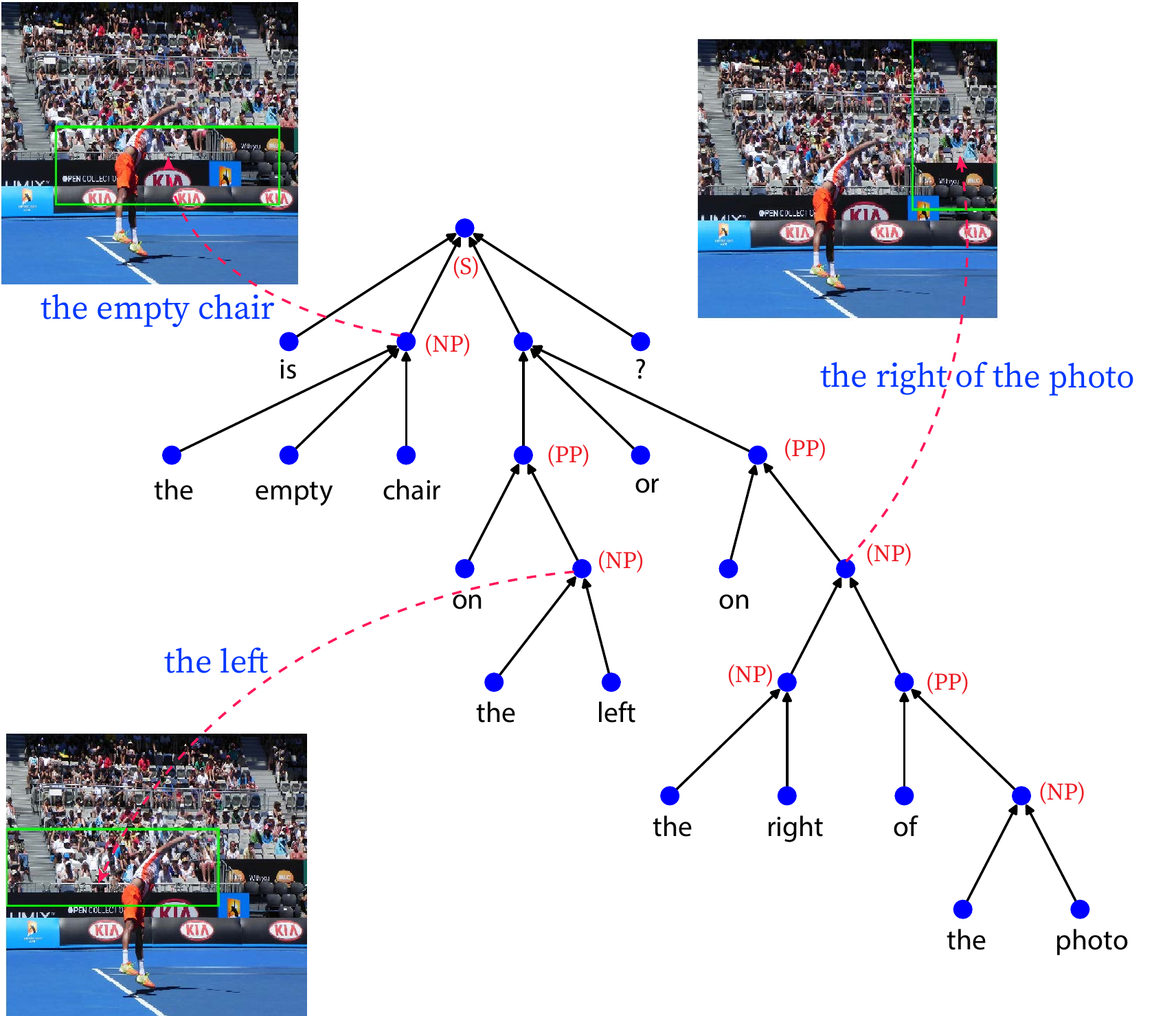}\vspace{-1em}
\par\end{center}
\begin{center}
(a)
\par\end{center}%
\end{minipage}\quad{}{\color{blue}\vrule}\quad{}%
\begin{minipage}[t]{0.46\textwidth}%
\begin{flushright}
\includegraphics[width=1\columnwidth]{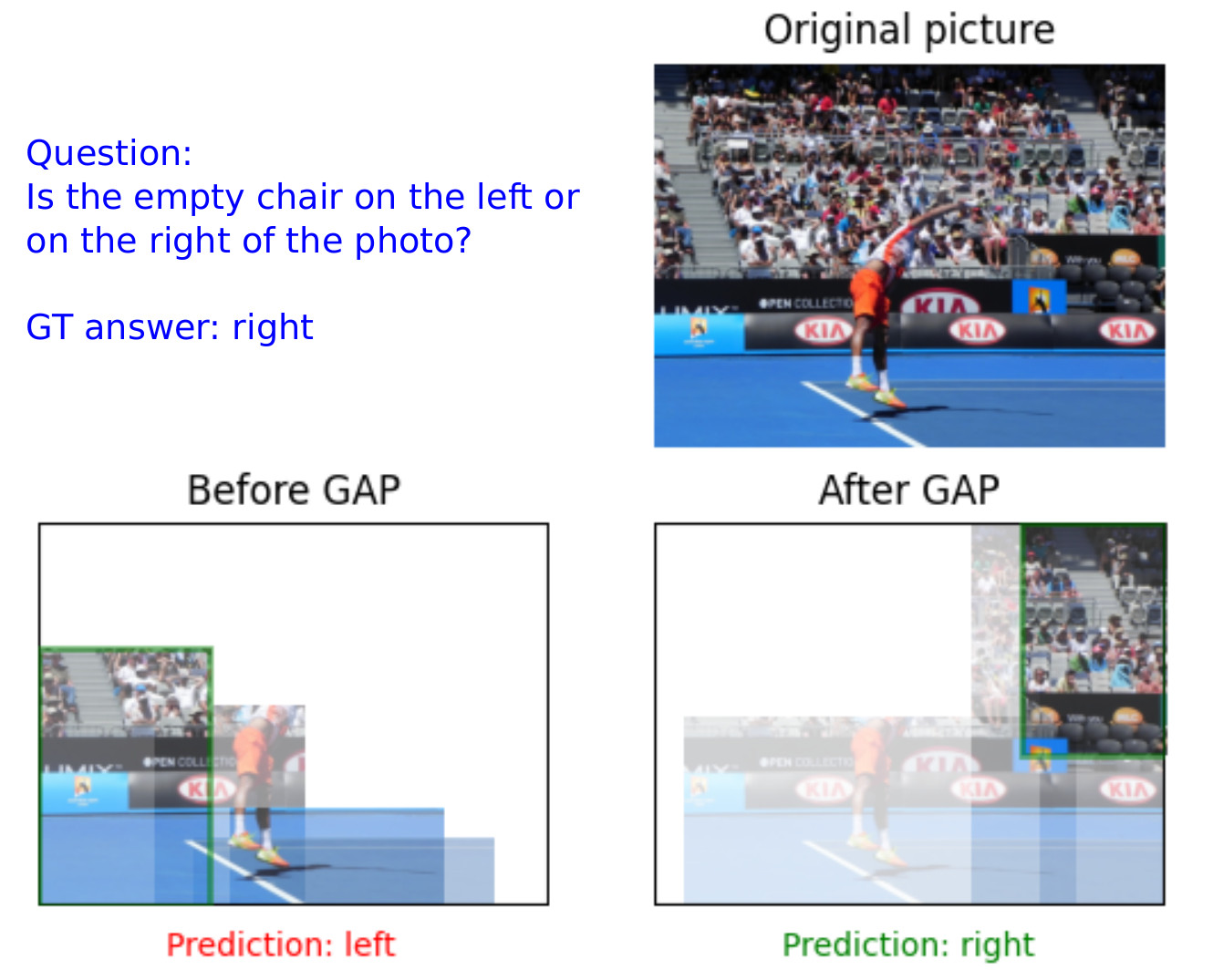}\vspace{-1em}
\par\end{flushright}
\begin{center}
(b)
\par\end{center}%
\end{minipage}
\par\end{center}%
\end{minipage}\medskip{}
\par\end{centering}
\caption{Qualitative open-box analysis of GAP's operation and effects. (a)
Constituency parse tree built from the question with three extracted
REs grounded to visual regions (green rectangles). (b) Visual attentions
and prediction of UpDn model before (left) vs. after applying GAP
(right). GAP shifts the model's highest visual attention (green rectangle)
to more appropriate regions while the original puts attention on irrelevant
parts.\label{fig:chap7_Qualitative-results}}
\end{figure*}

\subsubsection{Universality across Image QA Models}

Our regularisation method is theoretically applicable to any attention-based
VQA models. We evaluate the effectiveness of GAP on a wider range
of backbone models and datasets. Figure \ref{fig:chap7_comparing_against_baselines_gqa_vqav2}
summarises the improvements from applying GAP on UpDn, MACNet and
LOGNet on GQA and VQA-Rephrasings datasets. 

It can be clearly seen that GAP consistently improves all baselines
over all datasets. In particular, our method achieves the greatest
improvement on GQA and the REP split of VQA-Rephrasings. These two
sets contain more diverse variations of questions on which it is clearest
to see the effects of the grounding attention priors and the structured
representation of linguistic introduced in GAP.

Among the baselines, GAP brings more effect with LOGNet than the others.
We speculate that the cross-modality grounding priors in GAP are well
aligned with the linguistic binding mechanism in LOGNet. Therefore,
GAP helps bring out LOGNet's advantage in better joint representation
towards correct answers. 

\subsection{Model Analysis}

\subsubsection{Ablation Studies}

To provide more insights into our method, we conduct an extensive
set of ablation studies on the GQA and VQA-Rephrasings datasets (see
Table \ref{tab:chap7_Ablation-studies}). Throughout this experiment,
we witness the roles of every component toward the optimal performance
of the full model. 

The grammatical structure of questions (\emph{Baselines+TreeLSTM}
in Table \ref{tab:chap7_Ablation-studies}) provides a more robust
linguistic representation across all baselines used across datasets.
The role of this component is particularly significant in robustness
to question rephrased question (last column) with large improvement
(row 2 vs row 6). This can be explained by the fact that although
rephrased questions very different in sequential content, they share
a consistent grammatical structure which GAP exploits.

Visual attention regularisation (vis.GAP) seems to have more significant
effects than the linguistic counterpart (ling.GAP). This is reasonable
because visual attention itself is calculated based on linguistic
attention hence singularly regularising it has effect on both channels.

The last column of Table \ref{tab:chap7_Ablation-studies} also witnesses
the new state-of-the-art performance on the VQA-Rephrasings REP split
by the enhanced \emph{MACNet+GAP.}

\subsubsection{Qualitative Results}

We analyse the internal operation of GAP by visualising the tree structure
and grounding results. Fig. \ref{fig:chap7_Qualitative-results}.a
shows the parse tree structure built from a question and the linguistic-visual
grounding on visual regions. This structure is well built and the
groundings are more interpretable.

We also visualise the effect of GAP on VQA models by drawing their
visual attentions before and after applying GAP. Fig. \ref{fig:chap7_Qualitative-results}.b
shows visual attention of the UpDn model before and after applying
GAP. The visualisation clearly demonstrates that GAP helps shift the
attention towards appropriate visual regions while the attention produced
by the original UpDn baseline is less intuitive.

More examples of these analyses are provided in the supplementary
materials.

\section{Closing Remarks\label{sec:Chap7-Conclusion}}

We have presented a generic methodology to regulate Image QA models'
attention with Grounding-based Attention Prior (GAP) obtained by matching
concepts lying in visual and linguistic modalities. Through extensive
experiments across a variety of Image QA datasets, we demonstrated
the effectiveness of our approach in boosting attention-based Image
QA models' performance and improving their robustness against linguistic
variations. We also showed qualitative analysis to prove the benefits
of leveraging grounding-based attention priors in improving the interpretability
and trustworthiness of attention-based Image QA models.

GAP clearly offers a less expensive framework to extract external
knowledge from available sources to regulate the computational process
in Image QA models compared with the standard approaches of relying
on human attention annotation, making it more feasible and scalable
in real-world applications. In a broader view, representing the associations
between components across input modalities in the form of common knowledge
is essential towards systematic generalisation when dealing with a
multimodal problem such as visual and language reasoning.

\newpage{}

\selectlanguage{british}%

\chapter{Conclusions\label{chap:Conclusion}}

\section{Summary}

The capacity to perceive, understand and reason jointly over visual
scenes and natural language is essential for an intelligent computer
agent. This thesis presented neural network architectures and frameworks
to solve vision-and-language reasoning tasks, particularly Image Question
Answering (Image QA) and Video Question Answering (Video QA). We exploited
different pivotal aspects of vision and language interactions to provide
proper inductive biases for the reasoning process. Our contributions
in this thesis are four-fold. First, we introduced a highly effective
neural framework that simulated the dual process in the human cognitive
system for Video QA (Chapter \ref{chap:DualProcess}). We later argued
that high-order temporal relations between entities and events across
temporal dynamics and video hierarchy modelling are critical for video
understanding and reasoning (Chapter \ref{chap:MultimodalReasoning}).
We also exploited the significance of structured representations of
input modalities and modelling the explicit inter- and intra-modality
relations of input modality's components on the fly as reasoning proceeds
(Chapter \ref{chap:RelationalVisualReasoning}). Finally, we addressed
a fundamental drawback of most existing attention-based visual reasoning
models regarding their unregulated attention mechanisms by using visual-linguistic
grounded priors obtained from visual grounding (Chapter \ref{chap:RobustnessVR}).

In Chapter \ref{chap:DualProcess}, we argued that a complex reasoning
task over space-time such as Video QA necessitates a two-phase reasoning
system: one phase is an associative video cognition (System 1), and
the other one is deliberative multi-step reasoning (System 2). Both
phases are conditioned on the linguistic query. In particular, we
designed a module called Clip-based Relational Network (ClipRN) for
video representation and further integrated it with a generic neural
reasoning module to predict an answer to a question. Tested and analysed
on both synthetic and real Video QA datasets, the proposed framework
demonstrated its strong empirical performance. The results provided
strong evidence for the necessity of the two-phase reasoning system
for visual reasoning.

We explored the significance of relation network in video representation
in response to a linguistic query, leaving the entire reasoning process
for a generic reasoning module to later stages. This leads to ambiguity
of how the temporal relations help the reasoning process; hence, there
is a need for better perception capabilities of System 1 to facilitate
the reasoning process. These missing aspects were the focus of Chapter
\ref{chap:MultimodalReasoning}. We introduced a general-purpose neural
unit called Conditional Relational Network (CRN) as a building block
for learning to reason in a multimodal setting. CRN is a domain-independent
relational transformer that maps a set of tensorial objects to a new
set of relation-encoded objects. The flexibility and genericity of
the CRN unit allowed us to construct complex network architectures
by simply stacking the units along the hierarchical structure of videos.
We easily adapted the resultant network architecture, termed Hierarchical
Conditional Relational Networks (HCRN), which is built from a unique
building block for solving different settings of Video QA: short-form
Video QA where an agent is solely asked to respond to questions related
to the visual content in a given short video, and long-form Video
QA where the agent has to deal with longer and more complex movie
scenes, and information needed for question answering is located in
both the visual content and associated subtitles of movie scenes.
We extensively evaluated our proposed method on major short-term Video
QA datasets (TGIF-QA, MSVD-QA and MSRVTT-QA) and the current largest
long-term Video QA dataset (TVQA). Results showed the powerful reasoning
capabilities of HCRN against state-of-the-art methods.

Much of the questions in visual question answering are fine-grained.
Hence, it requires a strong inference structure to reason about the
relationships between objects within a visual scene as well as their
relationships with the linguistic components. However, most of the
existing works in visual reasoning represent visual scenes with holistic
local appearance features. In Chapter \ref{chap:RelationalVisualReasoning},
we built a recurrent neural model for compositional and relational
visual reasoning over a contextualised structured representation of
visual objects in consideration of linguistic binding. The intra-
and inter-modality dependencies are found on-demand as reasoning proceeds.
When tested on Image QA task, our proposed method demonstrated superior
performance on a wide range of datasets against state-of-the-art methods
even when trained on just 10\% of training data.

The structured representation of a visual scene studied in Chapter
\ref{chap:RelationalVisualReasoning} demonstrated its advantages
in allowing machines to exploit the inductive biases given by the
dynamic associations between vision and language for reasoning. However,
these vision-and-language associations are unregulated and often meaningless.
In Chapter \ref{chap:RobustnessVR}, we utilised external linguistic-vision
grounded data to improve the cross-modal binding by regularising the
internal attention mechanism within Image QA models. In order to ground
the linguistic query to visual parts, we leveraged its grammatical
structure to obtain linguistic referring expressions in which each
referring expression specifies a relevant region on the visual scene.
We then introduced a generic dual-modality regulation mechanism that
fortified attention-based visual reasoning models in performance and
robustness to linguistic variations. The effectiveness of the proposed
framework is justified through rigorous experiments with various existing
attention-based visual reasoning models across major Image QA datasets. 

\section{Future Directions}

Future work includes possible extensions of the proposed models in
the thesis. In Chapter \ref{chap:DualProcess}, although we have shown
the effectiveness of the dual process design in visual reasoning with
a seamless feed-forward integration of System 1 and System 2, how
the two systems interact is still open. It is ideal to have a mechanism
to softly decide whether performing inference at System 1 suffices
for a particular task without going through System 2. This will significantly
reduce the computational cost of the system as a whole. 

In Chapter \ref{chap:MultimodalReasoning}, we have successfully demonstrated
the significance of modelling near-term and far-term temporal relationships
within a video. However, the system still relies on simple local appearance
features of video frames. As pointed out in Chapter \ref{chap:RelationalVisualReasoning},
structured representations of input modalities greatly benefit visual
reasoning systems in learning and generalisation. We have partly addressed
this in our recent studies with an object-centric approach \citep{dang2021hierarchical}.
However, it remains a challenge how to perform temporal localisation
of actions and events over temporal dynamics and their complex spatio-temporal
interactions in the way humans reason about real-world surroundings. 

In Chapter \ref{chap:RelationalVisualReasoning}, all parameters regarding
the length of the recurrent computation and the depth of refinement
layers are chosen based on empirical evidence. Future work includes
the development of an adaptive algorithm to flexibly adjust these
parameters depending on the complexity of the given query. Modelling
the complexity of the query prior to reasoning also offers a promising
approach for the interaction between System 1 and System 2 in the
dual process system discussed above. If the query only requires the
perception of visual facts or simple relationships between objects
to arrive at an answer, visual reasoning models can perform inference
immediately at System 1 without going through a deliberative reasoning
process. By contrast, if the query if complex and requires a multi-step
reasoning process, System 1 will trigger the involvement of System
2. 

Finally, in Chapter \ref{chap:RobustnessVR}, the linguistic-visual
grounding priors are simply used as a regularisation term during training,
however are ignored during inference. This may not be ideal for out-of-distribution
problems. As we pursue future investigations, we are interested in
how to leverage the linguistic-visual grounding priors as extra input
for inference.

The questions raised in this thesis, and the solutions provided, open
up new directions for future work in improving the generalisation
and extensibility of vision and language reasoning models towards
systematic generalisation. Findings in Chapters (\ref{chap:MultimodalReasoning},
\ref{chap:RelationalVisualReasoning} and \ref{chap:RobustnessVR})
suggest that compositionality, multi-step reasoning and linguistic-to-vision
grounding are key for strong generalisation of visual reasoning systems.
Exploiting the compositionality of data, in particular, offers a potential
solution to understand novel concepts during inference, breaking out
of the assumption of classical supervised learning that training data
and test data share similar distributions. For example, even when
there is no sample of ``a red banana'' present in training, a machine
agent is still able to understand it during test time by combining
their understanding of ``red'' and ``banana'' as individual concepts.
Our results in Chapter \ref{chap:MultimodalReasoning} also suggest
more investigation is needed to represent long and untrimmed videos.
Although hierarchical modelling is useful, processing a video as a
whole at all times is costly and not ideal. We believe a top-down
approach to ``zoom in'' only in parts that are relevant to the query
is a promising direction. Finally, proper model structures and architectures
as in Chapter \ref{chap:RelationalVisualReasoning} could help minimise
the level of supervision and enhance the learning efficiency on simple
data, such as those in the CLEVR dataset, however the performance
significantly drops when tested on natural data. This poses a quest
for future work to learn to reason in an unsupervised manner.\selectlanguage{australian}%

\newpage{}

\appendix

\chapter{Supplementary}

\section{Dual Process in Visual Reasoning \label{sec:Appendix_A_dual_process}}

\subsection{Error Analysis and Extended Examples}

\begin{figure*}
\begin{centering}
\includegraphics[width=0.95\textwidth]{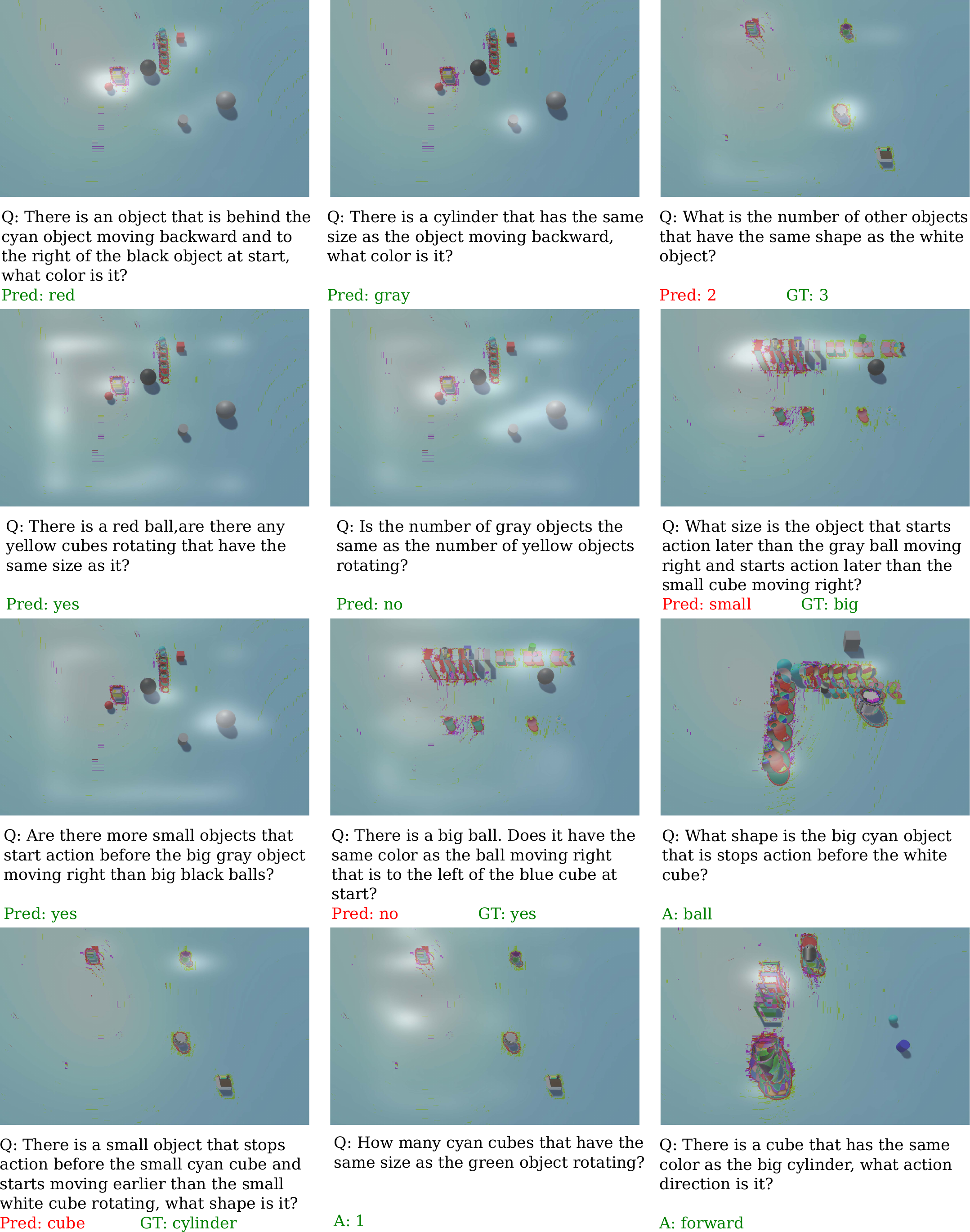}
\par\end{centering}
\caption{Examples of attention maps produced by our proposed model on the SVQA
dataset. Each video is represented as a still image with object motions.
Pred: Prediction; GT: groundtruth. Correct predictions are marked
with green colour, while wrong predictions are marked with red colour.
Examples covered a wide range of question types in the dataset, including
existing, counting and attribute queries.\label{fig:appendix_attentionmaps}}
\end{figure*}

For a better understanding of the model, we analyse failure modes
on the SVQA and TGIF-QA datasets. Fig.~\ref{fig:appendix_attentionmaps}
provides several examples of feature maps matching with the visualisation
of the video in the SVQA dataset, which is represented as a still
image with object motions. We notice that incorrect attention results
in most of the failures. In addition, for counting task, early iterations
of reasoning seem to produce more reasonable attention maps. This
suggests that it may require an external working memory to store intermediate
counting results rather than make predictions entirely based on the
memory state at the last reasoning step.

For the TGIF-QA dataset, there are many cases where questions require
the understanding of visual information that is not presented in ImageNet
dataset. Hence, the pre-trained ResNet model likely fails to capture
these features. These include questions about the movement of the
tiny parts of human body such as mouth, lips, arms, etc. For example:
\emph{``what does the boy in jacket do before bite lip?''}, \emph{``what
does the boy in jacket do after close mouth?''}, \emph{``what does
the man do after lick lips?''} However, due to the severe linguistic
bias of the TGIF-QA, the model can still give correct answers in some
cases since the correct answers share the closest semantic meaning
to the given question. This also suggests the use of either a richer
visual features extractor or additional features from a particular
human action dataset. Findings in chapter \ref{chap:MultimodalReasoning}
have partly solved this problem.

\section{Towards Robust Generalisation in Visual Reasoning}

\begin{figure*}
\begin{centering}
\noindent\begin{minipage}[t]{1\textwidth}%
\begin{center}
\noindent\begin{minipage}[t]{1\textwidth}%
\begin{center}
\textbf{\small{}Question:}{\small{} is the white minivan on the right
side or the left?}\textbf{\small{} }{\small{}\quad{}-\quad{}}\textbf{\small{}
GT answer:}{\small{} left}
\par\end{center}%
\end{minipage}
\par\end{center}
\begin{center}
\begin{minipage}[t]{0.45\textwidth}%
\begin{center}
\includegraphics[width=0.9\columnwidth]{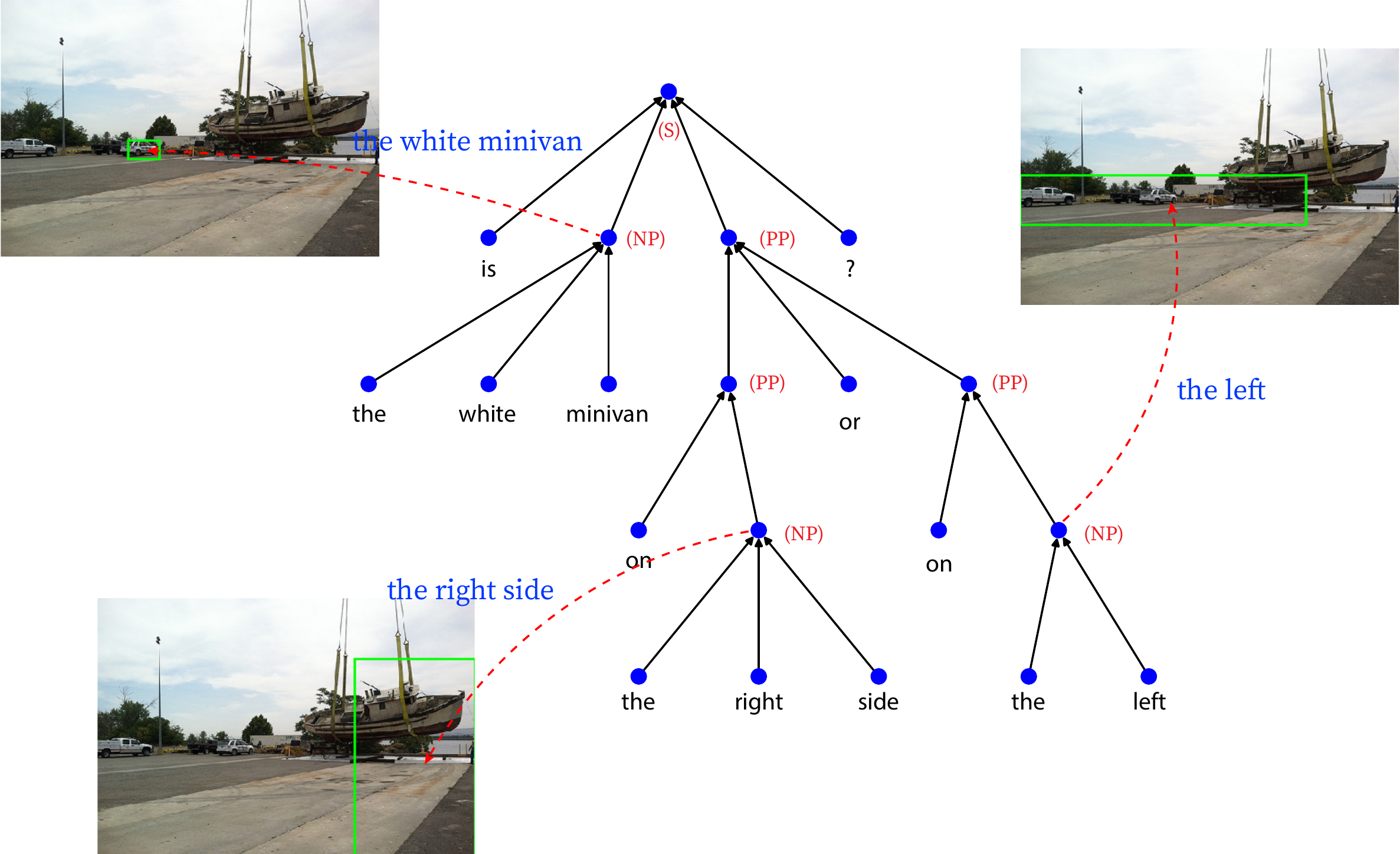}
\par\end{center}
\begin{center}
{\small{}(Constituency parse tree)}{\small\par}
\par\end{center}%
\end{minipage}\quad{}{\color{blue}\vrule}\quad{}%
\begin{minipage}[t]{0.4\textwidth}%
\begin{center}
\includegraphics[width=0.98\columnwidth]{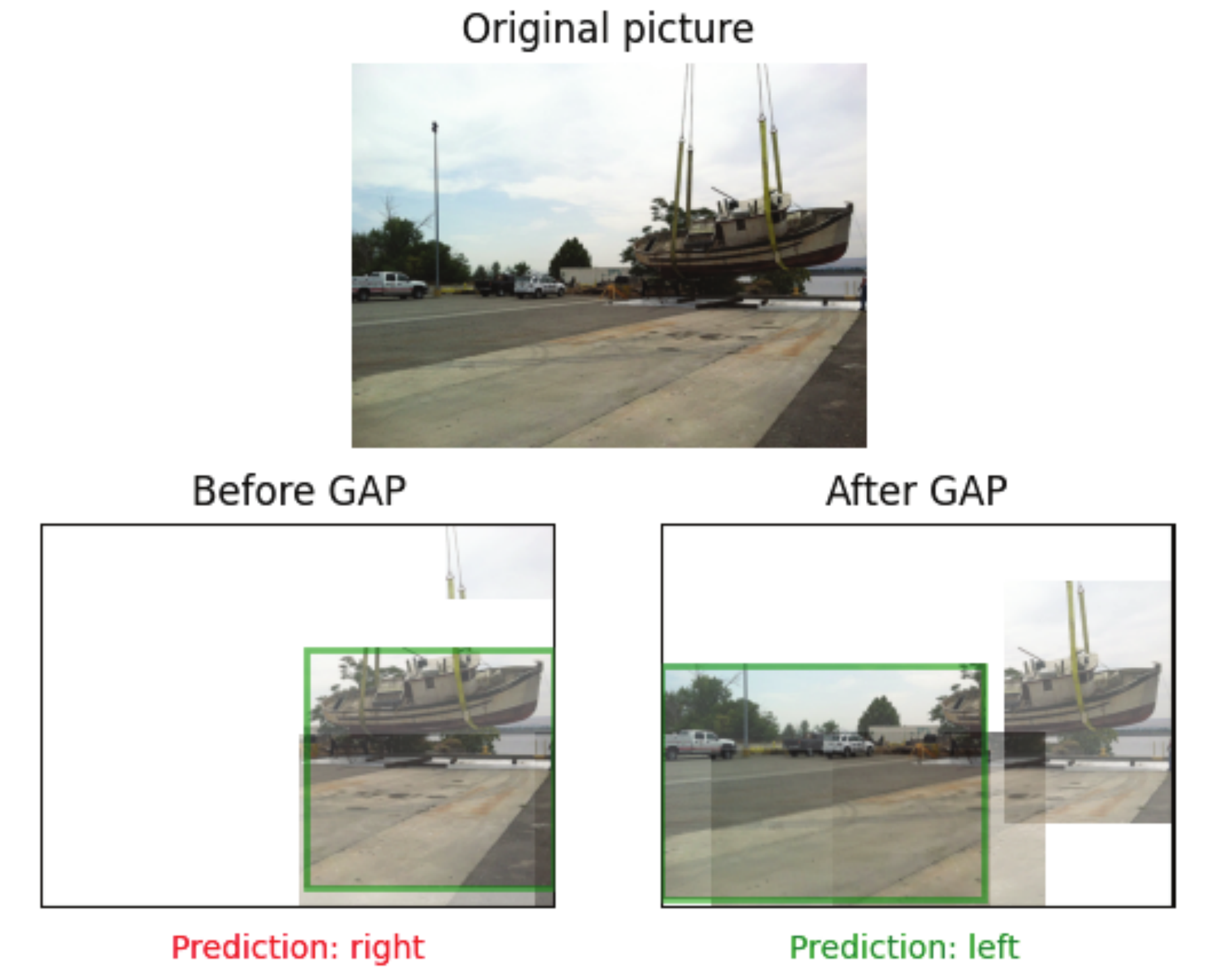}
\par\end{center}
\begin{center}
{\small{}(Visual attentions and predictions)}{\small\par}
\par\end{center}%
\end{minipage}
\par\end{center}%
\end{minipage}\vspace{1cm}
\par\end{centering}
\begin{centering}
\noindent\begin{minipage}[t]{1\textwidth}%
\begin{center}
\noindent\begin{minipage}[t]{1\textwidth}%
\begin{center}
\textbf{\small{}Question:}{\small{} which side is the car on?}\textbf{\small{}
}{\small{}\quad{}-\quad{}}\textbf{\small{} GT answer:}{\small{}
left}
\par\end{center}%
\end{minipage}
\par\end{center}
\begin{center}
\begin{minipage}[t]{0.45\textwidth}%
\begin{center}
\includegraphics[width=0.9\columnwidth]{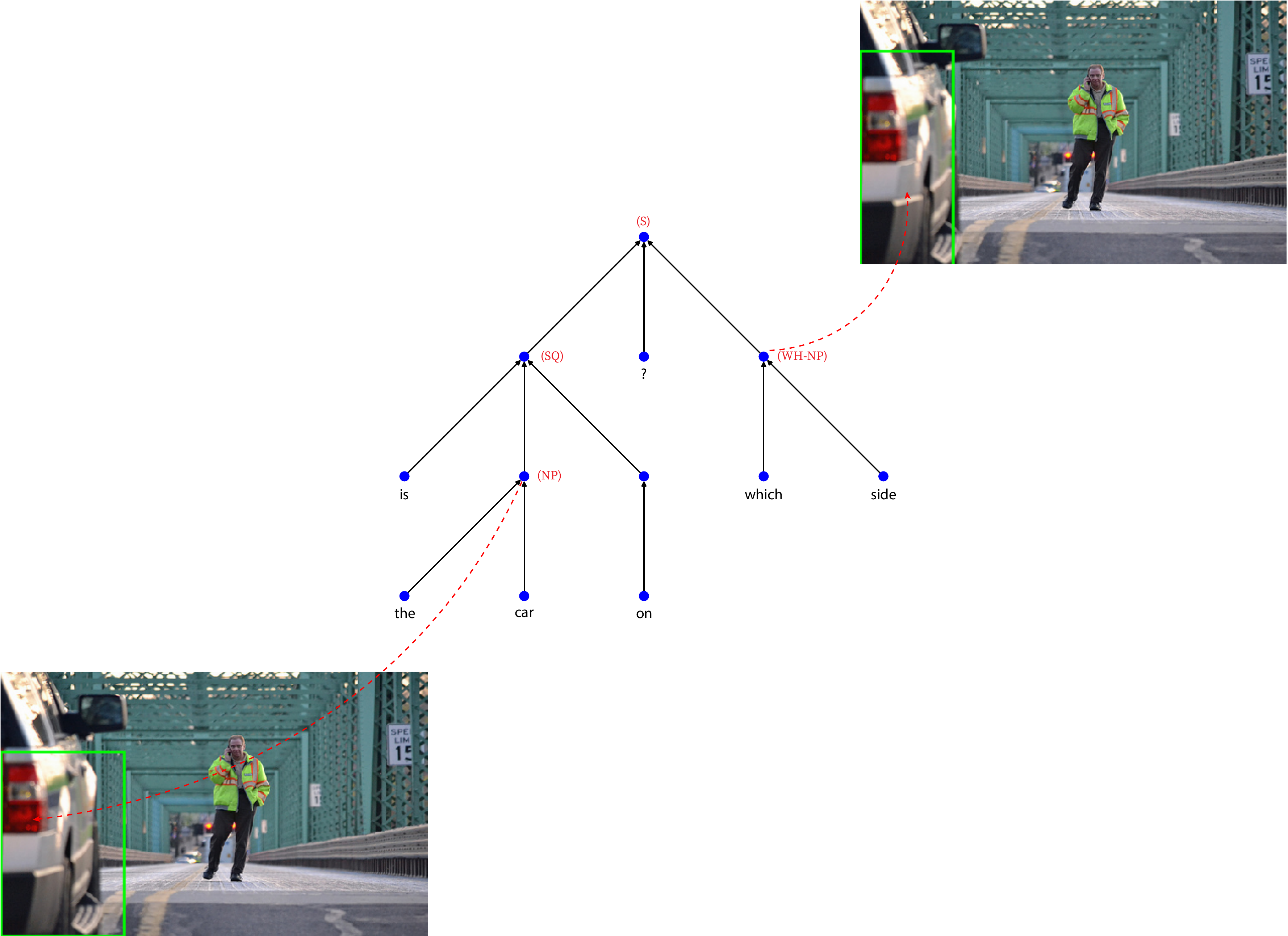}
\par\end{center}
\begin{center}
{\small{}(Constituency parse tree)}{\small\par}
\par\end{center}%
\end{minipage}\quad{}{\color{blue}\vrule}\quad{}%
\begin{minipage}[t]{0.4\textwidth}%
\begin{center}
\includegraphics[width=1\columnwidth]{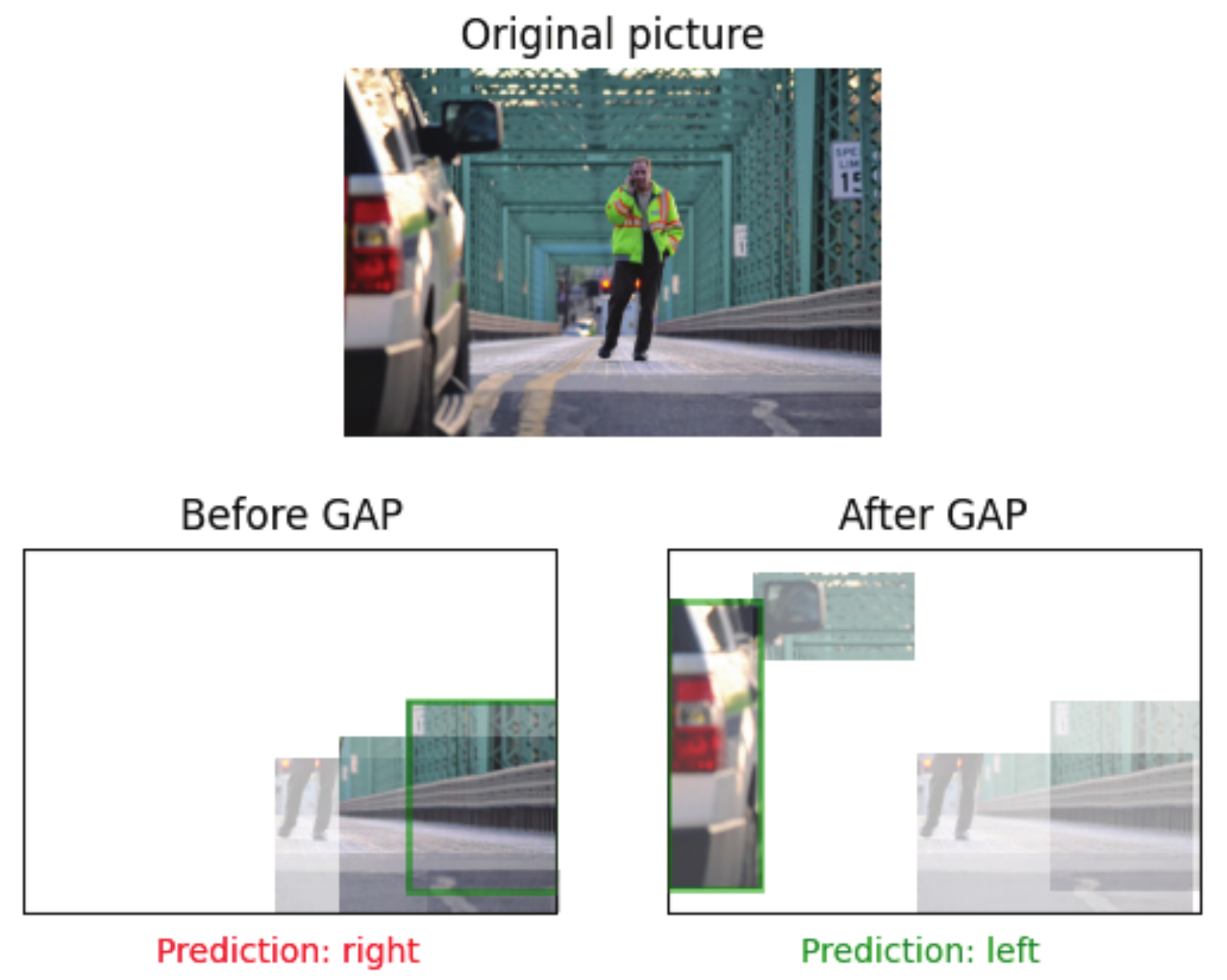}
\par\end{center}
\begin{center}
{\small{}(Visual attentions and predictions)}{\small\par}
\par\end{center}%
\end{minipage}
\par\end{center}%
\end{minipage}\medskip{}
\par\end{centering}
\caption{Qualitative open-box analysis of GAP\textquoteright s operation and
effects on UpDn backbone. (Left) Constituency parse tree built from
the question with three extracted REs grounded to visual regions (green
rectangles). (Right) Visual attentions and prediction of UpDn model
before (left) vs. after applying GAP (right). GAP shifts the model\textquoteright s
highest visual attention (green rectangle) to more appropriate regions
while the original puts attention on irrelevant parts.\label{fig:appendix_Qualitative-results-butd}}
\end{figure*}

\begin{figure*}
\begin{centering}
\noindent\begin{minipage}[t]{1\textwidth}%
\begin{center}
\noindent\begin{minipage}[t]{1\textwidth}%
\begin{center}
\textbf{\small{}Question:}{\small{} is the fence on the left side
of the photo?}\textbf{\small{} }{\small{}\quad{}-\quad{}}\textbf{\small{}
GT answer:}{\small{} no}
\par\end{center}%
\end{minipage}
\par\end{center}
\begin{center}
\begin{minipage}[t]{0.45\textwidth}%
\begin{center}
\includegraphics[width=0.9\columnwidth]{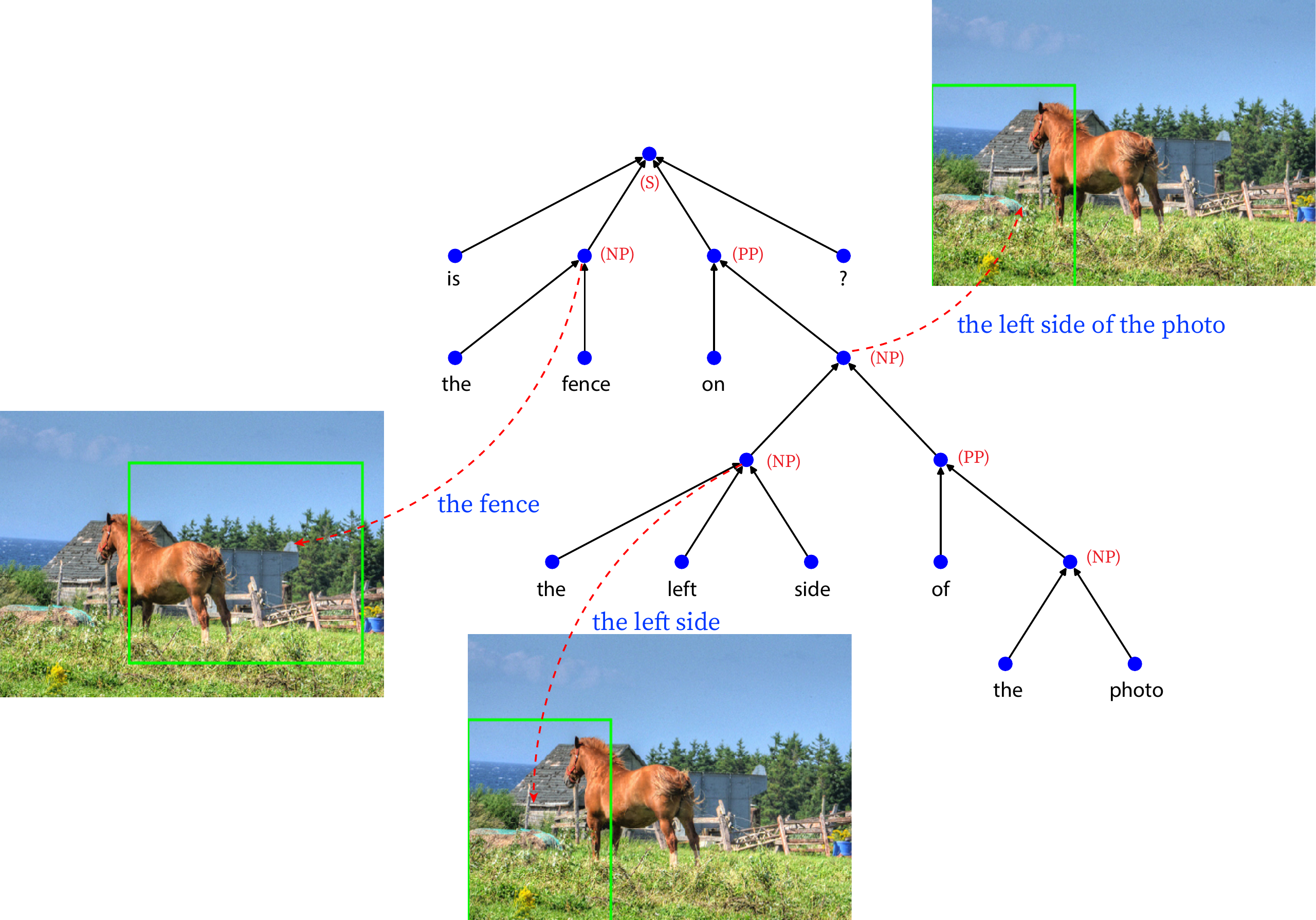}
\par\end{center}
\begin{center}
{\small{}(Constituency parse tree)}{\small\par}
\par\end{center}%
\end{minipage}\quad{}{\color{blue}\vrule}\quad{}%
\begin{minipage}[t]{0.4\textwidth}%
\begin{center}
\includegraphics[width=0.98\columnwidth]{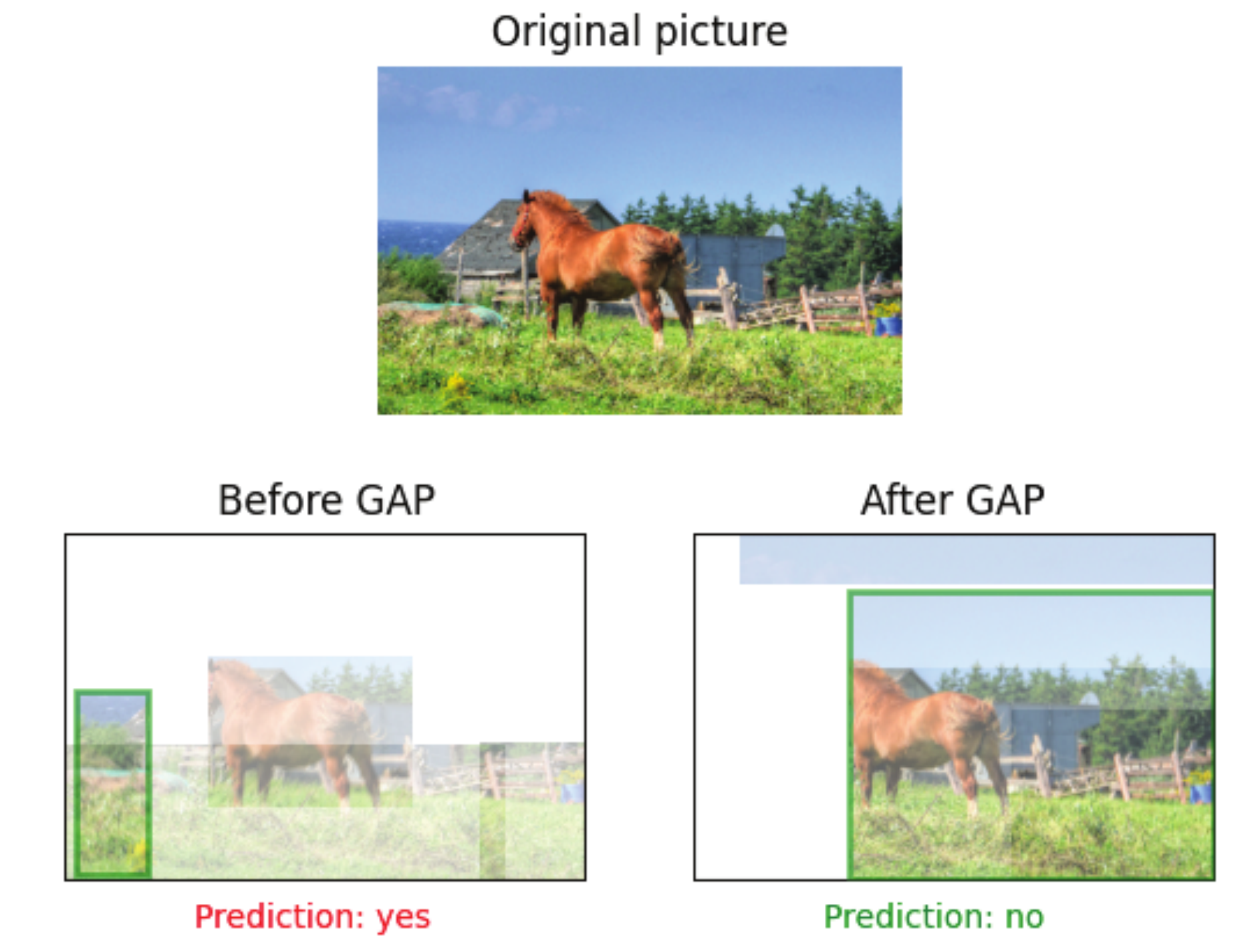}
\par\end{center}
\begin{center}
{\small{}(Visual attentions and predictions)}{\small\par}
\par\end{center}%
\end{minipage}
\par\end{center}%
\end{minipage}\vspace{1cm}
\par\end{centering}
\begin{centering}
\noindent\begin{minipage}[t]{1\textwidth}%
\begin{center}
\noindent\begin{minipage}[t]{1\textwidth}%
\begin{center}
\textbf{\small{}Question:}{\small{} is the small plate to the right
or to the left of the silver fork?}\textbf{\small{} }{\small{}\quad{}-\quad{}}\textbf{\small{}
GT answer:}{\small{} right}
\par\end{center}%
\end{minipage}
\par\end{center}
\begin{center}
\begin{minipage}[t]{0.45\textwidth}%
\begin{center}
\includegraphics[width=0.9\columnwidth]{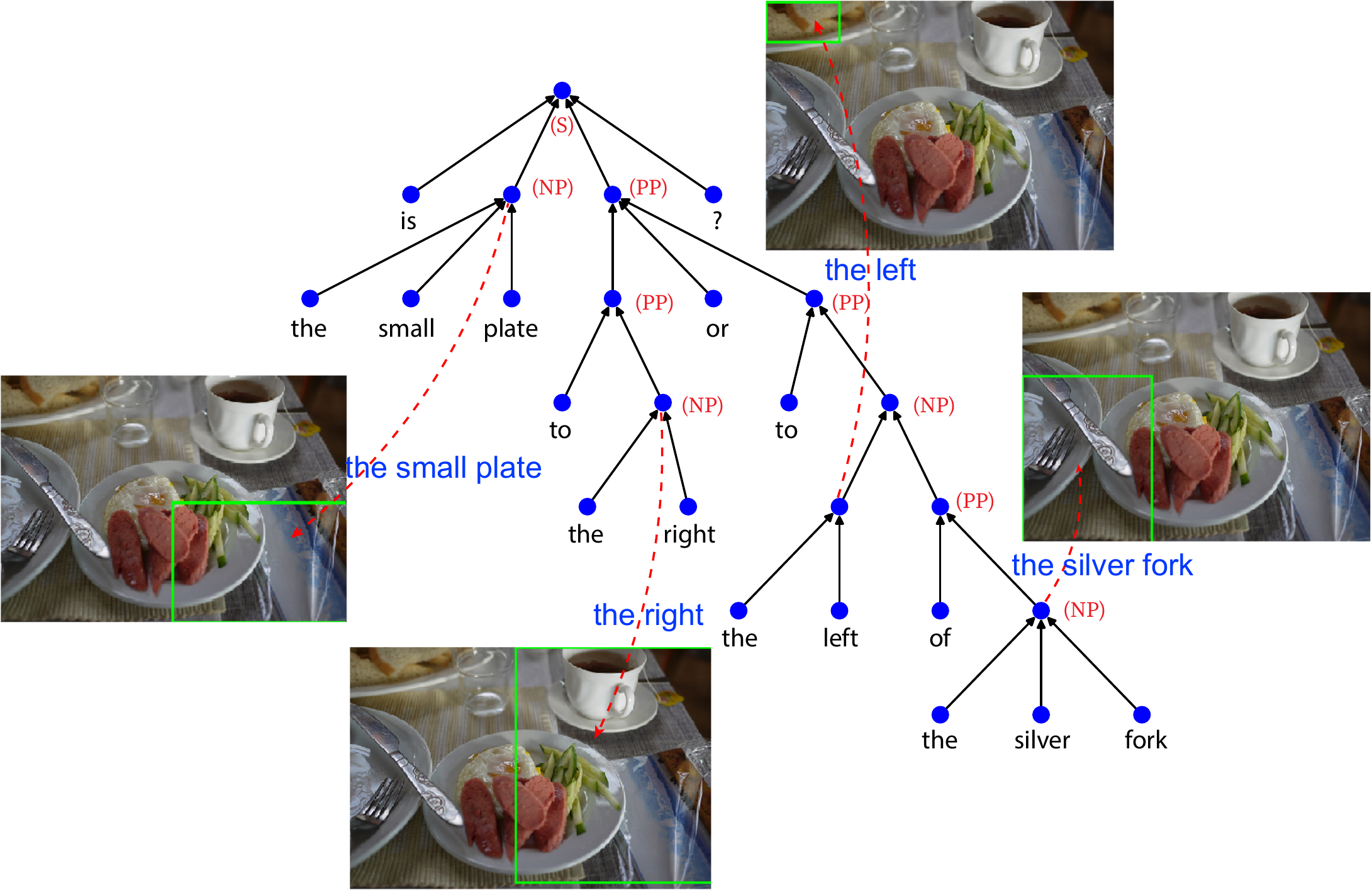}
\par\end{center}
\begin{center}
{\small{}(Constituency parse tree)}{\small\par}
\par\end{center}%
\end{minipage}\quad{}{\color{blue}\vrule}\quad{}%
\begin{minipage}[t]{0.4\textwidth}%
\begin{center}
\includegraphics[width=1\columnwidth]{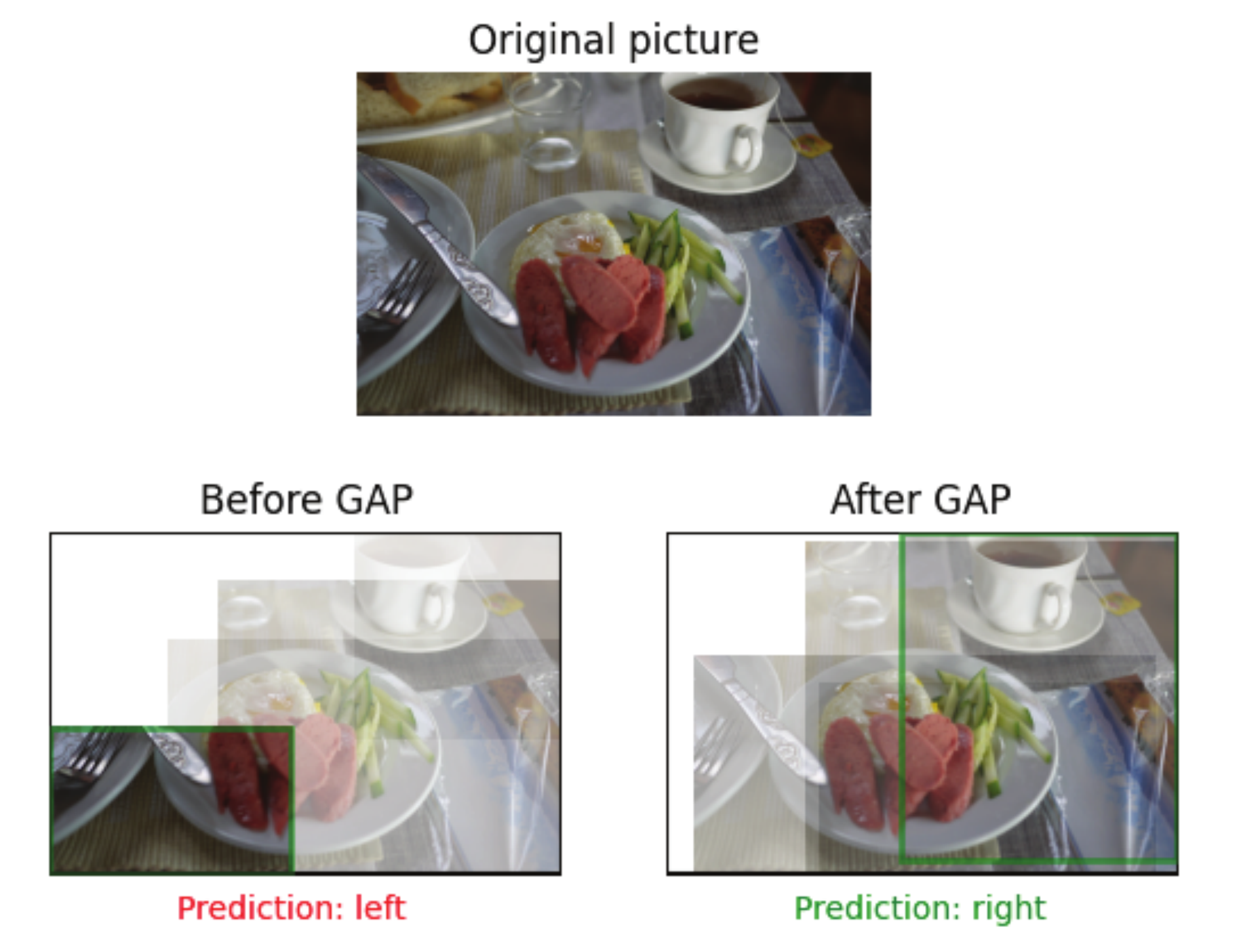}
\par\end{center}
\begin{center}
{\small{}(Visual attentions and predictions)}{\small\par}
\par\end{center}%
\end{minipage}
\par\end{center}%
\end{minipage}\medskip{}
\par\end{centering}
\caption{Qualitative open-box analysis of GAP\textquoteright s operation and
effects on LOGNet backbone. The attentions are obtained at the last
reasoning step of LOGNet. See Fig.\ref{fig:appendix_Qualitative-results-butd}
caption for conventions and legends.\label{fig:appendix_Qualitative-results-lognet}}
\end{figure*}

\subsection{Qualitative Analysis}

Fig. \ref{fig:chap7_Qualitative-results} provides only one case of
visualisation on the internal operation of our proposed method GAP
and its effect on the performance of Image QA models. We provide more
examples here for UpDn backbone (Fig. \ref{fig:appendix_Qualitative-results-butd})
and LOGNet backbone (Fig. \ref{fig:appendix_Qualitative-results-lognet})
with the same convention and legends. 

In each figure, left subfigures present the tree structures built
from respective questions and the linguistic-visual grounding priors
produced by the pre-trained MAttNet model. Right subfigures compare
the visual attentions before and after applying GAP. In all cases
across two different backbones (UpDn and LOGNet), GAP clearly helps
direct the models to pay attention to more appropriate visual regions,
partly explaining their answer predictions.
\clearpage{}

\bibliographystyle{plainnat}
\bibliography{thaole_local}

\begin{center}
\textemdash \textemdash \textemdash \textemdash \textemdash \textemdash \textemdash \textemdash \textemdash \textemdash \textemdash \textemdash \textemdash \textemdash \textemdash \textemdash \textemdash \textemdash \textemdash \textemdash \textemdash \textemdash \textemdash \textemdash \textemdash \textemdash \textendash \textemdash \textemdash \textemdash \textemdash \textemdash \textemdash \textemdash \textemdash \textemdash \textemdash{}
\par\end{center}

\begin{center}
\textbf{Copyright Information}
\par\end{center}

\textbf{Every reasonable effort has been made to acknowledge the owners
of copyright material. I would be pleased to hear from any copyright
owner who has been omitted or incorrectly acknowledged.}
\end{document}